\documentclass[a4paper,12pt,authoryear]{Classes/PhDThesisPSnPDF}  %

\input{Preamble/preamble}

\title{Deep Learning on Attributed Graphs}
\subtitle{A Journey from Graphs to Their Embeddings and Back}

\author{Martin Simonovsky}

\dept{LIGM / IMAGINE}

\university{Université Paris-Est / Ecole des Ponts}
\degree{Doctor of Philosophy}

\ifdefineAbstract
\fi

\ifdefineChapter
\fi

\begin{document}

\frontmatter

\begin{titlepage}
%
{\setlength{\parindent}{0cm}
\begin{center}
\begin{tabularx}{\linewidth}{c X c}
\includegraphics[height=3cm]{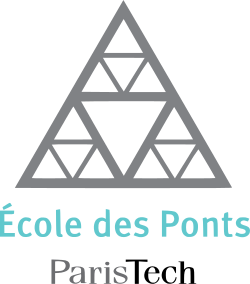} 
& &
\raisebox{0.6cm}{\includegraphics[height=1.1cm]{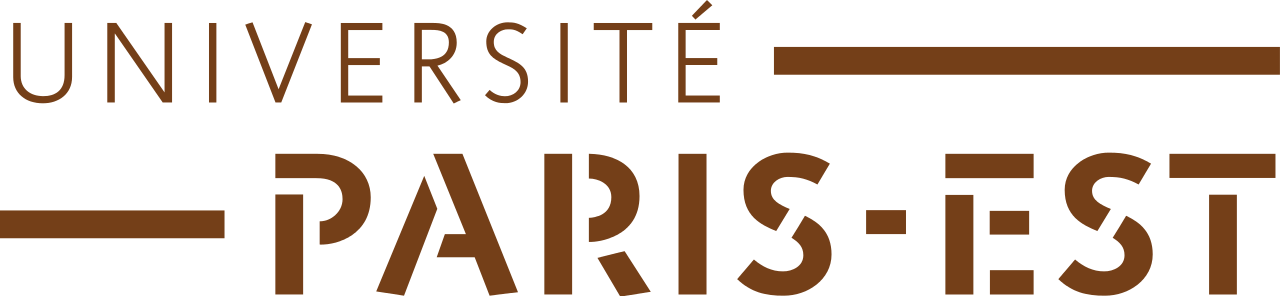}}
\end{tabularx}
\\[1cm]
\rule{\linewidth}{1pt}
\\[0.8cm]
{\LARGE \textbf{Deep Learning on Attributed Graphs}}
\\[0.4cm]
{\Large A Journey from Graphs to Their Embeddings and Back} 
\\[0.8cm]
{\Large \textbf{Martin SIMONOVSKY}}
\\[0.5cm]
\rule{\linewidth}{1pt}
\\[1.5cm]
{\large A doctoral thesis in the domain of automated signal and\\image processing supervised by \textbf{Nikos KOMODAKIS}}
\\[0.8cm]
{\large Submitted to \textbf{{\'E}cole Doctorale Paris-Est \\ Mathématiques et Sciences et Technologies \\de l'Information et de la Communication}.}
\\[1.5cm]
{\large Presented on 14 December 2018 to a committee consisting of:}
\\[0.3cm]
\begin{tabular}{lll}
  Nikos \textsc{Komodakis} & École des Ponts ParisTech & Supervisor\\
  Matthew B. \textsc{Blaschko} & KU Leuven & Reviewer\\
  Stephen \textsc{Gould} & Australian National University & Reviewer\\
  Renaud \textsc{Marlet} & École des Ponts ParisTech & Examiner\\
  Josiane \textsc{Zerubia} & INRIA Sophia Antipolis & Committee chair
\end{tabular}
\end{center}

\newpage
\thispagestyle{empty}
\vspace*{\fill}
École des Ponts ParisTech\\
LIGM-IMAGINE\\
6, Av Blaise Pascal - Cité Descartes\\
Champs-sur-Marne\\
77455 Marne-la-Vallée cedex 2\\
France
\\[1.5cm]
Université Paris-Est Marne-la-Vallée\\
École Doctorale Paris-Est MSTIC\\
Département Études Doctorales\\
6, Av Blaise Pascal - Cité Descartes\\
Champs-sur-Marne\\
77454 Marne-la-Vallée cedex 2\\
France

}

\end{titlepage}

\cleardoublepage
\begingroup
\let\cleardoublepage\clearpage
\setsinglecolumn
\thispagestyle{empty}

\vspace{4cm}
\section*{\centering \Large Abstract}
\vspace{0.8cm}
A graph is a powerful concept for representation of relations between pairs of entities. Data with underlying graph structure can be found across many disciplines, describing chemical compounds, surfaces of three-dimensional models, social interactions, or knowledge bases, to name only a few. There is a natural desire for understanding such data better. Deep learning (DL) has achieved significant breakthroughs in a variety of machine learning tasks in recent years, especially where data is structured on a grid, such as in text, speech, or image understanding. However, surprisingly little has been done to explore the applicability of DL on arbitrary graph-structured data directly. 

The goal of this thesis is to investigate architectures for DL on graphs and study how to transfer, adapt or generalize concepts that work well on sequential and image data to this domain. We concentrate on two important primitives: embedding graphs or their nodes into a continuous vector space representation (encoding) and, conversely, generating graphs from such vectors back (decoding). To that end, we make the following contributions.

First, we introduce Edge-Conditioned Convolutions (ECC), a convolution-like operation on graphs performed in the spatial domain where filters are dynamically generated based on edge attributes. The method is used to encode graphs with arbitrary and varying structure.

Second, we propose SuperPoint Graph, an intermediate point cloud representation with rich edge attributes encoding the contextual relationship between object parts. Based on this representation, ECC is employed to segment large-scale point clouds without major sacrifice in fine details.

Third, we present GraphVAE, a graph generator allowing us to decode graphs with variable but upper-bounded number of nodes making use of approximate graph matching for aligning the predictions of an autoencoder with its inputs. The method is applied to the task of molecule generation.

\vspace{1cm}
\noindent\textit{Keywords:} deep learning, graph convolutions, graph embedding, graph generation, point cloud segmentation
\vfill

\newpage
\thispagestyle{empty}

\vspace{4cm}
\section*{\centering \Large Résumé}
\vspace{0.8cm}
Le graphe est un concept puissant pour la représentation des relations entre des paires d'entités. Les données ayant une structure de graphes sous-jacente peuvent être trouvées dans de nombreuses disciplines, décrivant des composés chimiques, des surfaces des modèles tridimensionnels, des interactions sociales ou des bases de connaissance, pour n'en nommer que quelques-unes. L'apprentissage profond (DL) a accompli des avancées significatives dans une variété de tâches d'apprentissage automatique au cours des dernières années, particulièrement lorsque les données sont structurées sur une grille, comme dans la compréhension du texte, de la parole ou des images. Cependant, étonnamment peu de choses ont été faites pour explorer l'applicabilité de DL directement sur des données structurées sous forme des graphes. 

L'objectif de cette thèse est d'étudier des architectures de DL sur des graphes et de rechercher comment transférer, adapter ou généraliser à ce domaine des concepts qui fonctionnent bien sur des données séquentielles et des images. Nous nous concentrons sur deux primitives importantes : le plongement de graphes ou leurs nœuds dans une représentation de l'espace vectorielle continue (codage) et, inversement, la génération des graphes à partir de ces vecteurs (décodage). Nous faisons les contributions suivantes.

Tout d'abord, nous introduisons Edge-Conditioned Convolutions (ECC), une opération de type convolution sur les graphes réalisés dans le domaine spatial où les filtres sont générés dynamiquement en fonction des attributs des arêtes. La méthode est utilisée pour coder des graphes avec une structure arbitraire et variable.

Deuxièmement, nous proposons SuperPoint Graph, une représentation intermédiaire de nuages de points avec de riches attributs des arêtes codant la relation contextuelle entre des parties des objets. Sur la base de cette représentation, l'ECC est utilisé pour segmenter les nuages de points à grande échelle sans sacrifier les détails les plus fins.

Troisièmement, nous présentons GraphVAE, un générateur de graphes permettant de décoder des graphes avec un nombre de nœuds variable mais limité en haut, en utilisant la correspondance approximative des graphes pour aligner les prédictions d'un auto-encodeur avec ses entrées. La méthode est appliquée à génération de molécules.

\vspace{1cm}
\noindent\textit{Mots clés:} apprentissage profond, convolution sur de graphes, plongement de graphes, génération de graphes, segmentation des nuage de points
\vfill

\newpage
\thispagestyle{empty}

\vspace{4cm}
\section*{\centering \Large Publication List}
\vspace{0.8cm}
{\setlength{\parindent}{0cm}

Simonovsky, M. and Komodakis, N. (2016). OnionNet: Sharing Features in Cascaded Deep Classifiers. In \textit{Proceedings of the British Machine Vision Conference (BMVC)}.
\\[15pt]
Simonovsky, M., Gutiérrez-Becker, B., Mateus, D., Navab, N., and Komodakis, N. (2016). A Deep Metric for Multimodal Registration. In \textit{International Conference on Medical Image Computing and Computer-Assisted Intervention (MICCAI)}.
\\[15pt]
Simonovsky, M. and Komodakis, N. (2017). Dynamic Edge-conditioned Filters in Convolutional Neural Networks on Graphs. In \textit{IEEE Conference on Computer Vision and Pattern Recognition (CVPR)}.
\\[15pt]
Landrieu*, L. and Simonovsky*, M. (2018). Large-scale Point Cloud Semantic Segmentation with Superpoint Graphs. In \textit{IEEE Conference on Computer Vision and Pattern Recognition (CVPR)}.
\\[15pt]
Simonovsky, M. and Komodakis, N. (2018). Towards Variational Generation of Small Graphs. In \textit{Sixth International Conference on Learning Representations (ICLR), workshop track}.
\\[15pt]
Simonovsky, M. and Komodakis, N. (2018). GraphVAE: Towards Generation of Small Graphs Using Variational Autoencoders. In \textit{27th International Conference on Artificial Neural Networks (ICANN)}.

\blfootnote{* Joint first authorship}
}

\newpage
\thispagestyle{empty}

\vspace{4cm}
\section*{\centering \Large Acknowledgements}
\vspace{0.8cm}
{\setlength{\parindent}{0cm}

Foremost, I would like to express gratitude to my supervisor Nikos Komodakis for accepting me as his student and securing funding for the whole period, for useful discourses and insights, and especially for giving me the trust and freedom to pursue the research directions I personally found the most interesting and promising. I'm also grateful to the rest of my committee for devoting their precious time to reviewing this thesis and for coming up with many great, thought-provoking questions.

I have experienced very friendly and relaxed atmosphere in my research group, Imagine. In particular, I'm obliged to Renaud Marlet for his warm attitude, negotiating difficult issues and, with the strong support of Brigitte Mondou, for shielding the lab from the outside world. I thank Guillaume Obozinski for his research attitude and didactics, as well as the organization of reading groups, together with Mathieu Aubry. I was also fortunate enough to collaborate on interesting problems with external researchers who taught me a lot along the way. Thank you, Lo{\"{\i}}c Landrieu, Benjam{\'{\i}}n Guti{\'{e}}rrez{-}Becker, Martin Čadík, Jan Brejcha, and Shinjae Yoo.

Uprooted and replanted to France, I was blessed with amazing labmates, many of whom have become good friends of mine. Thank you for all the fun, chats, help, ideas, as well as putting up with GPU exploitation, Benjamin Dubois, Francisco Massa, Laura Fernández Julià, Marina Vinyes, Martin De La Gorce, Mateusz Koziński, Pierre-Alain Langlois, Praveer Singh, Raghudeep Gadde, Sergey Zagoruyko, Shell Hu, Spyros Gidaris, Thibault Groueix, Xuchong Qiu, and Yohann Salaun, as well as Maria Vakalopoulou and Norbert Bus. The past three years would have been lonely and boring without you! Professionally, I'm especially grateful to Sergey, who bootstrapped me into the engineering side of deep learning and taught me lots of good practice, to Francisco for stimulating discussions on everything, and to Shell for keeping me in touch with the theory.

Furthermore, I'm indebted to the community behind Torch, PyTorch, and the whole Python universe. Without all of you, I would still be implementing my first paper now.

A very special thanks goes to my dear parents, Dana and Vašek, for their strong  belief in education and for their unconditional support through my whole studies; to you I dedicate this thesis. Last but certainly not least, I would like to thank my partner Beata for her patience and tolerance, for enduring our long long-distance relationship, as well as for all the love, moral support, and fun.

}

\endgroup

\tableofcontents

\mainmatter

\newtheorem{definition}{Definition}[section]

\chapter{Introduction} \label{chap:intro}

\section{Motivation}

A graph is a powerful representation of relations between pairs of entities. Its versatility and strong theoretical understanding has resulted in a plethora of use cases across a variety of science disciplines and engineering problems. For example, graphs can naturally be used to describe the structure of chemical compounds, the interactions between regions in the brain, social interactions between people, the topology of three-dimensional models and transportation plans, dependencies of linked web pages, program flows, or knowledge bases. The development of the theory and algorithms to handle graphs has therefore been of major interest. 

Formally, a (directed) graph is an ordered pair $G=(\mathcal{V},\mathcal{E})$ such that $\mathcal{V}$ is a non-empty set of vertices (also called nodes) and $\mathcal{E}\subseteq \mathcal{V}\times \mathcal{V}$ is a set of edges. Additional information can be attached to both vertices and edges in the form of attributes. A vertex-attributed graph assumes function $\mathcal{V} \to\mathbb{R}^{d_v}$ assigning attributes to each vertex. An edge-attributed graph assumes function $\mathcal{E}\to\mathbb{R}^{d_e}$ assigning attributes to each edge.

The classical graph theory, foundations of which date back to the 18th century, focuses on understanding and analyzing the graph structure and addressing combinatorial problems such as finding mappings, paths, flows or colorings. 
On the other hand, there has been a lot of research on graph-structured data in the past decades, especially in the signal processing and machine learning communities. In the former, the research has notably brought Fourier transform to irregular domains, giving rise to spectral graph theory and adapting fundamental signal processing operations such as filtering, translation, dilation, or downsampling~\citep{shuman2013emerging}. Usually, a fixed graph is considered with changing data on its nodes. 
In machine learning, researchers have explicitly made use of graph structures in the data in areas such as graph partitioning for clustering, label propagation for semi-supervised learning, manifold methods for dimensionality reduction, graph kernels to describe similarity between different graphs, or graphical models for capturing probabilistic interpretation of data.

In recent years, deep learning (DL) has achieved significant breakthroughs in quantitative performance over traditional machine learning approaches in a broad variety of tasks and the field has matured into technologies useful for commercial applications. Prominently, a specific class of architecture, called Convolutional Neural Networks (CNNs), has gained massive popularity in tasks where the underlying data representation has a grid structure, such as in speech processing and natural language understanding (1D, temporal convolutions), in image classification and segmentation (2D, spatial convolutions), and in video and medical image understanding (3D, volumetric convolutions)~\citep{lecun2015deep}. On the other hand, the case of data lying on irregular or generally non-Euclidean domains has received comparatively much less attention in the community but has become a hot topic during the preparation of this thesis~\citep{bronstein2017geometric}.

Encouraged by the success of DL on grids, in this thesis we study DL architectures for graph-structured data. Specifically, we concentrate on two important primitives: embedding graphs or their nodes into a continuous vector space representation (encoding) and, conversely, generating graphs from such vectors back (decoding). We introduce both problems, their applications and challenges in the following two subsections. Following the DL philosophy of abandoning hand-crafted features for their learned counterparts, our contributions pragmatically build more on accomplished DL methods rather than on past approaches from the machine learning and signal processing community amendable to differentiable re-formulation. The main research question we ask is: \textit{How can we transfer, adapt or generalize DL concepts working well on sequential and image data to graphs?}

\subsection{Embedding Graphs and Their Nodes}

One of the central tasks addressed by DL is that of representation learning~\citep{bengio2013representation}, \ie finding a mapping from raw input objects to a continuous vector space of fixed dimensionality $\mathbb{R}^d$. Both the mapping and the vector space are also interchangeably called embedding or encoding in the literature. The goal is to optimize this mapping so that task-relevant object properties are preserved and reflected in the geometric relationships within the learned embedding space. As an example, in the task of hand-written digit classification, the mapping should capture properties regarding the type of digit, be invariant to those concerning writing style, and strive to make embeddings of images belonging to different classes separable by hyperplanes. %

In the case of graphs, we are primarily interested in finding representation of individual nodes as well as of the entire graphs. Usually, information from the local neighborhood of a node is considered in its embedding, while graph embeddings are further computed as some form of aggregation (pooling) of node embeddings. The analogy in deep learning on images is computing pixel-wise features (\eg for semantic image segmentation) and image-wise features (\eg for image recognition), respectively. In the case of graphs, there are two kinds of information available. The first is the graph structure as defined by the nodes and edges. The second is the data associated to the graph, \ie its node and edge attributes. Arguably, a powerful embedding network should be able to exploit as much information as possible.

The ability to compute node embeddings has given rise to many application, such as link prediction~\citep{schlichtkrull2017modeling}, (semi-supervised) node classification in citation networks~\citep{kipf}, performing logical reasoning tasks~\citep{yujia16}, finding correspondences across meshes~\citep{MontiBMRSB17} or point cloud segmentation~(Chapter \ref{chap:cvpr18}). Graph embeddings has also been useful in many tasks, for example measuring similarity of brain networks by metric learning~\citep{ktena2017distance}, suggesting heuristic moves for approximating NP-hard problems~\citep{dai2017learning}, predicting satisfaction of formal properties in computer programs~\citep{yujia16}, classifying chemical effects of molecules~(Chapter \ref{chap:cvpr17}) and regressing their physical properties~\citep{GilmerSRVD17}, or point cloud classification~(Chapter \ref{chap:cvpr17}).

\subsubsection{Challenges} \label{in_emb_challenges}

The cornerstone of popular network architectures for processing natural images, video or speech is the convolutional layer~\citep{mnist}. The translation invariance of the operation, the use of filters with compact support and the application over multiple resolutions fit very well to the statistical properties of such data, namely the stationarity, locality and compositionality.

Besides the fact that the architecture imposes a prior especially suitable for natural images~\citep{UlyanovVL17}, another major benefit is the induced strong tying of parameters (weight sharing) across the grid, which greatly reduces the number of free parameters in the network compared to a fully-connected layer (perceptron) while still being able to capture the statistics of real data and enables processing of variable-sized inputs in so-called fully-convolutional style~\citep{Long2015fnc}.

Assuming that the stationarity, locality and possibly compositionality principles of representation hold to at least some level in graph data as well, it is meaningful to consider a hierarchical CNN-like architecture for processing it in order to benefit from advantages described above. However, the graph domain poses several challenges which make extensions of CNNs to graphs not straightforward. We detail the main challenges below.

\paragraph*{Irregular Neighborhoods}
Unlike on grids, where each node has the same amount of neighbors except for boundaries, nodes in general graphs can have arbitrary number of adjacent nodes. This means that there is no trivial analogy of translation invariance and thus the research objective is to come up with strategies to share parameters of the convolution operator among its applications to different nodes.

\paragraph*{Unordered Neighborhoods}
There is no specific ordering of neighbors of a node, as both nodes and edges are defined as sets. In order to do better that isotropic smoothing, the ability to assign different convolutional weights to different edges is desired. This requires making assumptions or exploiting additional information such as the global structure, degrees of nodes, or edge attributes. On contrary, grids imply a natural ordering of neighborhoods.

\paragraph*{Structural Variability}
In practical scenarios, the graph structure may vary throughout the dataset or may remain fixed, with only the data on nodes changing. The latter is the usual case for applications stemming from the signal processing community, centered around spectral filtering methods~\citep{shuman2013emerging}, and data-mining community, centered around matrix factorization methods~\citep{hamilton2017representation}. However, as such methods cannot naturally handle datasets with varying graph structure (for instance meshes, molecules, or frequently updated large-scale graphs), spatial filtering methods have gained on popularity recently. The research question here is to devise explicit local propagation rules and build links to established past work.

\paragraph*{Computational Complexity}
The current DL architectures on grids benefit from heavy parallelism on Graphics Processing Units (GPUs) enabled by well-engineering implementations in popular frameworks such as PyTorch~\citep{pytorch} or TensorFlow~\citep{tensorflow}. Unfortunately, the irregular structure of graphs is less amendable to parallelism. The research investigates implementation details such as the use of sparse matrices or other data structures. In addition, for methods based on spectral processing, the naive use of Graph Fourier transform has quadratic complexity in the number of nodes.

\subsection{Generating Graphs}

Deep learning-based generative models have gained massive popularity during several past years, especially in the case of images and text. The principal idea is to collect a large amount of (unannotated) data in some domain and then train a model to generate similar data, \ie draw samples from a learned distribution closely approximating the true data distribution. As the model has much less parameters than the size of the dataset, it is forced to discover the "essence" of the data rather than to memorize it. Usually, the generative process (also called decoding) is conditioned on a random vector and/or a point from a defined vector space, such as the embedding space of an encoder.

In the case of graphs, we are interested in generating both the structure and associated attributes. There are many applications of such a graph generator, for example creating similar graphs~\citep{netgan2018} or maintaining intermediate representations in reasoning tasks~\citep{johnson2017learning}. In an encoder-decoder setup, any kind of input can be encoded to a latent embedding space and then translated to a graph. Possible applications include drug discovery by continuous optimization of certain chemical properties in the latent space~\citep{Bombarelli16}, sampling molecules with certain properties~(Chapter \ref{chap:iclr18}) or corresponding to given laboratory measurements (either by conditioning or translation), or simply encoder pre-training in cases where labeled data is expensive.

\subsubsection{Challenges}

In addition to the challenges usually demonstrated in generative models on grids, such as tricky training~\citep{radford2015unsupervised,BowmanVVDJB16}, problematic capturing of all distribution modes~\citep{radford2015unsupervised} or difficult quantitative evaluation~\citep{TheisOB15}, generative models for graph data face several more challenges, detailed below.

\paragraph*{Unordered Nodes} 
Data on regular grids provide an implicit ordering of nodes (that is of pixels, characters or words) that is amendable both to step by step generation with autoregressive processes~\citep{oord2016pixelrnn,BowmanVVDJB16} and generation of entire content in a feed-forward network~\citep{radford2015unsupervised}. However, nodes in graphs are defined as a set and so there is often no clear way how to linearize the construction in a sequence of steps for training autoregressive generators or order nodes in a graph adjacency matrix for generation by a feed-forward network. While there are practical algorithms for approximate graph canonization, \ie finding consistent node ordering, available~\citep{McKay2014nauty}, the empirical result of \citet{vinyals2015order} that the linearization order matters when learning on sets suggest that it is a priori unclear that enforcing a specific canonical ordering would lead to the best results.

\paragraph*{Discrete Nature} 
Similar to text and unlike images, graphs are discrete structures. Incremental construction with autoregressive methods involves discrete decisions, which are not differentiable and thus problematic for gradient-based optimization methods common in DL. In sequence generation, this is typically circumvented by a maximum likelihood objective with so-called teacher forcing~\citep{WilliamsZ89}, which allows to decompose the generation into a sequence of conditional decisions and maximize the likelihood of each of them. However, the choice of such a decomposition for graphs is not clear, as we argue above, and teacher forcing is prone the inability to fix its mistakes, resulting in possibly poor prediction performance if the conditioning context diverges from sequences seen during training~\citep{bengio2015scheduled}. On the other hand, generation of graphs in a probabilistic formulation can only postpone dealing with non-differentiablity issues to the final step, when the discretized graph has to be output.

\paragraph*{Large Graphs}
The level to which the locality and compositionality principles apply to a particular graph can strongly vary due to possibly arbitrary connectivity - for example, a generator should be able to output both a path and a complete graph. This seems problematic for scaling up to large graphs, as one cannot trivially enforce some form of hierarchical decomposition or coarse-to-fine construction, conceptually similar to using up-sampling operators in images~\citep{radford2015unsupervised}. This may bring about the necessity to use more parameters while having less regularization implied by the architecture design in the case a feed-forward network. For autoregressive methods, large graphs may make training difficult due to very long chains of construction steps, especially for denser graphs.

\section{Summary of Contributions}

Having charted the two main directions of this thesis and their challenges, we summarize our contributions in this section. Most of these contributions were published as "Dynamic Edge-Conditioned Filters in Convolutional Neural Networks on Graphs" at CVPR~\citep{simonovsky2017dynamic}, as "Large-scale Point Cloud Semantic Segmentation with Superpoint Graphs" at CVPR~\citep{superpointgraphs}, as "Towards Variational Generation of Small Graphs" at ICLR workshop track~\citep{simonovsky2018towards}, and as "GraphVAE: Towards Generation of Small Graphs Using Variational Autoencoders" at ICANN~\citep{simonovsky2018graphvae}. Martin Simonovsky is the leading author in all publications with the exception of~\citet{superpointgraphs}, where the contribution is equally shared with Lo{\"{\i}}c Landrieu.

\begin{itemize}
\item We propose Edge-Conditioned Convolution (ECC) in Chapter~\ref{chap:cvpr17}, a novel convolution-like operation on graphs performed in the spatial domain where filters are conditioned on edge attributes (discrete or continuous) and dynamically generated for each specific input graph. This allows the algorithm to exploit enough structural information in local neighborhoods so that our formulation can be shown to generalize discrete convolution on grids. Due the application in spatial domain, the method can work on graphs with arbitrary and varying structure. In Chapter~\ref{chap:cvpr18}, we integrate ECC into a recurrent network and reduce its memory and computational requirements, reformulating it as ECC-VV. 

\item We introduce SuperPoint Graph (SPG) in Chapter~\ref{chap:cvpr18}, a novel point cloud representation with rich edge attributes encoding the contextual relationship between object parts. Based on this representation, we are able to apply deep learning in the form of ECC/ECC-VV on large-scale point clouds without major sacrifice in fine details.

\item We demonstrate multiple applications of ECC/ECC-VV. Chapter~\ref{chap:cvpr17} investigates graph classification applications, in particular for graphs representing chemical compounds and for neighborhood graphs of point clouds. Further, we evaluate the performance on node classification in the context of semantic segmentation of large-scale scenes in Chapter~\ref{chap:cvpr18}. ECC is also used as encoder in the molecular autoencoder presented in Chapter~\ref{chap:iclr18}. Besides having obtained the state of the art performance on several datasets among DL methods at the respective times of publication (NCI1~\citep{nci1db}, Sydney Urban Objects~\citep{trianglesvm}, Semantic3D~\citep{hackel2017semantic3d} and S3DIS~\citep{armeni_cvpr16}), we were the first to apply graph convolutions to point cloud processing, with the motivation of preserving sparsity and presumably fine details.

\item We propose a graph decoder formulated in the framework of variational autoencoders~\citep{vae} in Chapter~\ref{chap:iclr18}. The decoder outputs a probabilistic fully-connected graph of a predefined maximum size directly at once, which somewhat sidesteps the issues of discretization, and uses graph matching during training in order to attempt to overcome the challenge of undefined node ordering. We evaluate on the difficult task of molecule generation.

\end{itemize}

Besides the central topic of deep learning on graphs, we have conducted research along several other lines within the duration of the thesis. These activities have resulted in two publications, which are not part of this manuscript and which we therefore briefly summarize below:

\begin{itemize} 

\item In the domain of medical image registration, we proposed a patch similarity metric based on CNNs and applied it to registering volumetric images of different modalities. The key insight was to exploit the differentiability of CNNs and directly backpropagate through the trained network to update transformation parameters within a continuous optimization framework. The training was formulated as a classification task, where the goal was to discriminate between aligned and misaligned patches. The metric was validated on intersubject deformable registration on a dataset different from the one used for training, demonstrating good generalization. In this task, we outperformed mutual information by a significant margin. The work was done in collaboration with Benjam{\'{\i}}n Guti{\'{e}}rrez{-}Becker from TU Munich as a second author and was published as "A Deep Metric for Multimodal Registration" at MICCAI 2016~\citep{simonovsky2016deep}.

\item For retrieval scenarios in computer vision, we proposed a way of speeding up evaluation of deep neural networks by replacing a monolithic network with a cascade of feature-sharing classifiers. The key feature was to allow subsequent stages to add both new layers as well as new feature channels to the previous ones. Intermediate feature maps were thus shared among classifiers, preventing them from the necessity of being recomputed. As a result demonstrated in three applications (patch matching, object detection, and image retrieval), the cascade could operate significantly faster than both monolithic networks and traditional cascades without sharing at the cost of marginal decrease in precision. The work was published as "OnionNet: Sharing Features in Cascaded Deep Classifiers" at BMVC 2016~\citep{simonovsky2016onionnet}.

\end{itemize}

\section{Outline}

In this section we summarize the chapters of the thesis and relate them to each other.

\begin{description}

\item[Chapter~\ref{chap:background}: Background and Related Work] This chapter provides the background and overviews related and prior work on graph-structured data analysis. Both classic and deep learning-based methods are covered. Further discussions of related and contemporary work are also provided within each core chapter.

\item[Chapter~\ref{chap:cvpr17}: Edge-Conditioned Convolutions] This chapter introduces the main \\workhorse of this thesis, graph convolutions able to process continuous edge attributes. The method is applied to small-scale point cloud classification and biological graph datasets.

\item[Chapter~\ref{chap:cvpr18}: Large-scale Point Cloud Segmentation] This chapter applies ECC to the the problem of segmentation of large areas, such as streets or office blocks, by working on homogeneous segments instead of individual points.

\item[Chapter~\ref{chap:iclr18}: Generation of Small Graphs] This chapter introduces a graph autoencoder based on ECC and graph matching with an application to drug discovery, where novel molecules need to be sampled.

\item[Chapter~\ref{chap:conclusion}: Conclusion] This chapter reflects on the contributions of the thesis and hypothesizes on directions of future work.

\end{description}

\chapter{Background and Related Work}   \label{chap:background}

This chapter provides background and overview of related and prior work on (sub)graph embedding techniques. After introducing our notation in Section~\ref{sec:graph_notation}, we dive into classical (non-deep) signal processing on graphs in Section~\ref{sub:bg_sig_proc}, which is mostly formulated in the spectral domain. This knowledge is then useful for the overview of spectral deep learning-based convolution methods, introduced in Section \ref{sub:bg_spectralgc}. We mention their shortcoming and move on to review the progress in spatial convolution methods in Section \ref{sub:bg_spatialgc}, also in relation to the contributions of this thesis. To make the picture complete, we briefly mention non-convolutional approaches for embedding, namely direct embedding and graph kernels, in Section \ref{bg_nonconv}. 

While this thesis focuses exclusively on graphs, it is worth to mention the progress made in (deep) learning on manifolds. Whereas the fields of differential geometry and graph theory are relatively distant, both graphs and manifolds are examples of non-Euclidean domains and the approaches for (spectral) signal processing and for convolutions in particular are closely related, especially when considering discretized manifolds - meshes. We refer the interested reader to the nice survey paper of \citet{bronstein2017geometric} for concrete details.

\section{Attributed Graphs} \label{sec:graph_notation}

Let us repeat and extend graph-related definitions from the introduction. A directed graph is an ordered pair $G=(\mathcal{V},\mathcal{E})$ such that $\mathcal{V}$ is a non-empty set of vertices (also called nodes) and $\mathcal{E}\subseteq \mathcal{V}\times \mathcal{V}$ is a set of edges (also called links). Undirected graph can be defined by constraining the relation $\mathcal{E}$ to be symmetric. 

It is frequently very useful to be able to attach additional information to elements of graphs in the form of attributes. For example, in graph representation of molecules, vertices may be assigned atomic numbers while edges may be associated with information about the chemical bond between two atoms. Concretely, a vertex-attributed graph assumes function $F: \mathcal{V} \to\mathbb{R}^{d_v}$ assigning attributes to each vertex. An edge-attributed graph assumes function $E: \mathcal{E}\to\mathbb{R}^{d_e}$ assigning attributes to each edge. In the special case of non-negative one dimensional attributes $\mathcal{E}\to\mathbb{R}^+$, we call the graph weighted. Discrete attributes $\mathcal{E}\to\mathbb{N}$ can be also called types. %

In addition to attributes, which are deemed being a fixed part of the input, we introduce mapping $H: \mathcal{V} \to\mathbb{R}^{d}$ to denote vertex representation due to signal processing or graph embedding algorithms operating on the graph. This mapping is called signal (especially if $d=1$) or simply data. Often, vertex attributes $F$ may be used to initialize $H$. The methods for computation of $H$ will constitute a major part of discussion in this thesis. 

Finally, assuming a particular node ordering, the graph can be conveniently described in matrix, resp. tensor form using its (weighed) adjacency matrix $A$, signal matrix $H$, node attribute matrix $F$ and edge attribute tensor $E$ of rank 3.

\section{Signal Processing on Graphs} \label{sub:bg_sig_proc}

\def\laplacian{L}
\newcommand{\DEF}[1]{{\emph{#1}}}

Here we briefly introduce several concepts regarding multiscale signal filtering on graphs, which are the basis of spectral graph convolution methods presented in Section~\ref{sub:bg_spectralgc} and also used for pooling in Chapter~\ref{chap:cvpr17}.

Mathematically, both a one-dimensional discrete-time signal with $N$ samples as well as a signal on a graph with $N$ nodes can be regarded as vectors $\mathbf{f} \in \mathbb{R}^{N}$. Nevertheless, applying classical signal processing techniques on the graph domain would yield suboptimal results due to disregarding (in)dependencies arising from the irregularity present in the domain. 

However, the efforts to generalize basic signal processing primitives such as filtering or downsampling faces two major challenges, related to those mentioned in Section~\ref{in_emb_challenges}: 
\begin{description}
\item[Translation] While translating time signal $f(t)$ to the right by a unit can be expressed as $f(t-1)$, there is no analogy to shift-invariant translation in graphs, making the classical definition of convolution operation $(f*w)(t) := \int_{\tau} g(\tau) w(t-\tau) d\tau$ not directly applicable.
\item[Coarsening] While time signal decimation may amount to simply removing every other point, it is not immediately clear which nodes in a graph to remove and how to aggregate signal on the remaining ones.
\end{description}

In the following subsections, we describe three core concepts developed by the signal processing community: graph Fourier transform, graph filtering operations and graph coarsening operations. We refer to the excellent overview papers of~\citet{shuman2013emerging, shumanFV16} for a broader perspective as well as details.

\subsection{Graph Fourier Transform}

Assuming undirected finite weighted graph $G$ on $N$ nodes, let $A$ be its weighted adjacency matrix and $D$ its diagonal degree matrix, \ie $D_{i,i}=\sum_j{A_{i,j}}$. The central concept in spectral graph analysis is the family of Laplacians, in particular the unnormalized graph Laplacian $\laplacian:=D-A$ and the normalized graph Laplacian $\laplacian:=D^{-1/2}(D-A)D^{-1/2}=I-D^{-1/2}AD^{-1/2}$. Intuitively, Laplacians capture the difference between the local average of a function around a node and the value of the function at the node itself. We denote $\{\mathbf{u}_l\}$ the set of real-valued orthonormal eigenvectors of a Laplacian and $\{\lambda_l\}$ the set of corresponding eigenvalues. Interestingly, Laplacian eigenvalues provide a notion of frequency in the sense that eigenvectors associated with small eigenvalues are smooth and vary slowly across the graph (\ie, the values of an eigenvector at two nodes connected by an edge with large weight are likely to be similar) and those associated with larger eigenvalues oscillate more rapidly. %

The analogy between graph Laplacian spectra and the set of complex exponentials, which are eigenfunctions of the classical Laplacian operator on Euclidean space and provide the basis for the classical Fourier transform, is the motivation for introducing an analogy to Fourier transform on graphs. The \DEF{graph Fourier transform} $\mathbf{\hat{f}}$ of signal $\mathbf{f} \in \mathbb{R}^{N}$ on the nodes of $G$ is defined as the expansion in terms of Laplacian eigenvectors $\hat{f}(\lambda_l) := <\mathbf{f},\mathbf{u}_l> = \sum_{i=0}^{N-1}{f(i)u_l(i)}$ and the \DEF{inverse graph Fourier transform} is then computed as $f(i) = \sum_{l=0}^{N-1}{\hat{f}(\lambda_l)u_l(i)}$.

We remark that extensions to directed graphs are complicated~\citep{SardellittiBL17}. Also, handling of discrete or multidimensional edge attributes is not supported by the framework, to the best of our knowledge, although one way of circumventing the latter might be to simultaneously consider multiple graphs with the same connectivity but different edge weights, each graph corresponding to a single real, non-negative attribute dimension.

\subsection{Signal Filtering} \label{sub:bg_spectr_filtering}

Using graph Fourier transform, a signal can be equivalently represented in the spatial (node) domain and in the graph spectral domain. Thus, as in classical signal processing, the signal can be filtered in either of the domains. 

\DEF{Graph spatial filtering} amounts to expressing the filtered signal $g$ at node $i$ as a linear combination of the input signal $\mathbf{f}$ at nodes within its $K$-hop local neighborhood $\mathcal{N}_K$: $g(i) = B_{i,i}f(i) + \sum_{j \in \mathcal{N}_K(i)}{B_{i,j}f(j)}$ with filter coefficients matrix $B$. While this is a (spatially) localized operation by design, it requires specifying $\mathcal{O}(N^2)$ coefficients unless some form of regularity is assumed.

\DEF{Graph spectral filtering} follows the classical case, where filtering corresponds to multiplication in the spectral domain, \ie the amplification or attenuation of the set of basis functions. It is defined as $\hat{g}(\lambda_l) = \hat{f}(\lambda_l) \hat{w}(\lambda_l)$. This suggests a way of generalizing the convolution operation on graphs by the means of spectral filtering as 

\begin{equation} \label{eq:bg_spectr_conv}
g(i) = (f *_G w)(i) := \sum_{l=0}^{N-1}\sum_{j=0}^{N-1}{f(j) \hat{w}(\lambda_l) u_l(i) u_l(j)}
\end{equation}

It is important to remark that spectral filter coefficients $\mathbf{\hat{w}}$ are specific to the given graph, resp. the particular choice of its Laplacian eigenvectors. 

Using the full range of $N$ coefficients makes spectral filtering perfectly localized in the Fourier domain but not localized in the spatial domain due to uncertainty principle, its extension to graphs first developed by~\citet{AgaskarL13} but still considered an open problem. However, this property is usually undesired in practice. The popular remedy is to make the filter spectrum "smooth" by ordering the eigenvectors according to their eigenvalues and expressing coefficients as continuous function of only a few free parameters. To make the link between spectral and spatial filtering, it is interesting to parameterize the spectral filter as order $K$ polynomial $\hat{w}(\lambda_l) = \sum_{k=0}^{K-1}{a_k \lambda_l^k}$ with coefficients $\mathbf{a}$. It can be shown that this construction corresponds to a spatial filter on $K$-hop neighborhood for a particular choice of $B_{i,j} := \sum_{k=0}^{K-1}{a_k \laplacian_{i,j}^k}$~\citep{shuman2013emerging}. Thus, polynomial spectral filters are exactly localized within a certain spatial neighborhood and have only a limited number of free parameters, independent of $N$, which is beneficial for any learning task, as shown by \citet{defferrard16}.

Nevertheless, even in the localized case the cost of filtering a graph signal in the spectral domain is an $\mathcal{O}(N^2)$ operation due to the multiplications with Fourier basis in Equation~\ref{eq:bg_spectr_conv}. Fortunately, approximation methods have been developed for fast filtering in the case of sparse graphs, where the filter is formulated directly in the spatial domain as a polynomial in $\laplacian$ that can be evaluated recursively, leading to computational complexity linear with the number of edges. A popular choice for the polynomial is truncated Chebyshev expansion~\citep{ShumanVF11}. Note that this does not require an explicit computation of the Laplacian eigenbasis.

Building banks of filters localized both in space and frequency has been intensively studied in the context of graph wavelets and we refer to the review in~\citep{shuman2013emerging} for more details. Wavelet coefficients can serve as descriptors for classification or matching tasks in non-deep machine learning \citet{MasoumiH17}.

\subsection{Graph Coarsening} \label{sub:bg_coarsening}

In the Euclidean world, multiresolution representation and processing of signals and images with techniques such as pyramids, wavelets or CNNs has been very successful. Finding a similar concept for graphs is therefore of interest. The task of \DEF{graph coarsening} is to form a series of successively coarser graphs $G^{(s)}=(\mathcal{V}^{(s)}, \mathcal{E}^{(s)})$ from the original graph $G=G^{(0)}$. Coarsening typically consists of three steps: subsampling or merging nodes, creating the new edge structure $E^{(s)}$ and weights (so-called \DEF{reduction}), and mapping the nodes in the original graph to those in the coarsened one with $C^{(s)}: \mathcal{V}^{(s-1)}\to \mathcal{V}^{(s)}$. In many practical cases, a desirable property of coarsening is to halve the number nodes at each level while clustering nodes connected with large weights together. For unweighted bipartite graphs, there is a natural way of coarsening by sampling "every other node". But the task is more complex for other cases, which has led to creation of various algorithms.

One line of work is based on partitioning the set of nodes into clusters and representing these with a single node in the
coarsened graph. For example, Graclus~\citep{dhillon2007weighted} greedily merges pairs of nodes based on clustering heuristics such the normalized cut $A_{i,j}(1/D_{i,i}+1/D_{j,j})$, which is efficient but not optimal. The union of the neighborhoods of the two original nodes becomes the neighborhood of the merged node, leading to decreased sparsity of the coarsened graphs. This algorithm is used for pooling in~\citet{defferrard16}. 

Another line of work keeps a strict subset of original nodes during the coarsening steps. For example, \citet{shumanFV16} select the nodes to keep based on the sign of entries in the eigenvector associated with the largest Laplacian eigenvalue (frequency), which can be computed efficiently by the power method. The sign is shown to behave intuitively on bipartite graphs, such as grids or trees. Kron reduction~\citep{kron} is suggested as a method of choice to compute the coarsened Laplacian, defned as the Schur complement of the original Laplacian with respect to the removed node indices. It offers many nice properties but leads to a decrease in sparsity, which can be amended by postprocessing with spectral sparsification~\citep{spielman2011graph}, among others. This randomized algorithm samples edges according to the probability of the edge appearing in a random spanning tree of the graph, thus mostly removing those structurally unimportant edges. We use this pipeline for pooling operation later in Chapter~\ref{chap:cvpr17}.

Finally, let us remark that graph resolutions generated in this fashion are completely independent of any signals residing on the graph nodes.

\section{Spectral Graph Convolutions}   \label{sub:bg_spectralgc}

In this section, we review how the deep learning community have applied spectral signal processing methods, introduced above, to learning tasks. The characteristic feature of these methods is the explicit use of graph Laplacian in the convolution operation.

In their pioneering work, \citet{bruna13} learned filters using the formulation of spectral convolution in Equation~\ref{eq:bg_spectr_conv}. This can be conveniently expressed in matrix form as $\mathbf{h}^{t+1}=U \Theta U^T \mathbf{h}^t$, where $U$ is the column-wise concatenation of Laplacian eigenvectors $\{\mathbf{u}_l\}$, $\Theta$ is a diagonal matrix of filter coefficients and $\mathbf{h}^t$ and $\mathbf{h}^{t+1}$ is the input and output signal vector, respectively. Each spectral filter $\Theta$ is parameterized as B-spline with the number of free parameters independent of graph size $N$ to reduce the risk of overfitting and to make filters smooth and therefore localized in space. In addition, high frequency eigenvectors may be discarded for better computational efficiency of Fourier transform, which nevertheless remains quadratic in the number of eigenvectors. By using multiple filters per layer, stacking multiple convolutional layers and considering coarsened graphs created by agglomerative clustering, Bruna \etal were the first to present a graph analogue of classic CNNs.

\citet{defferrard16} later proposed an efficient spectrum-free method by formulating spectral filters as Chebyshev polynomials, as briefly introduced in Section~\ref{sub:bg_spectr_filtering} above. The filtering is evaluated directly in the spatial domain on a $K$-hop neighborhood and can be written in matrix notation as $\mathbf{h}^{t+1} = \sum_{k=0}^{K-1}{\mathbf{\theta}_k T_k(\tilde{L}) \mathbf{h}^t}$ where $\mathbf{\theta}$ is a vector of $K$ parameters and $T_k$ is Chebyshev polynomial of order $k$ evaluated at the scaled Laplacian $\tilde{L} := 2L/\max{\lambda} - I$. Coarsening for pooling purposes is done by Graclus~\citep{dhillon2007weighted}, which does not require computation of the spectrum either.

This approach was further simplified by \citet{kipf}, who proposed its first-order approximation in spatial domain as $\mathbf{h}^{t+1} = \theta (I - \hat{L}) \mathbf{h}^t$ where $\theta$ is a scalar parameter and $\hat{L}$ is the Laplacian of the adjacency matrix with added self-connections $\hat{A} = A+I$. In effect, this model corresponds to simply taking average of neighboring nodes' signal weighted proportionally to the Laplacian.

Note that despite the spatial evaluation of spectral methods, model parameters are still tied to the particular Laplacian and the trained networks may not generalize to other graph structures; this has been empirically demonstrated by \citet{MontiBMRSB17} for the model of \citet{defferrard16}. \citet{SyncSpecCNN} embarked on addressing this for the case of 3D shapes represented as meshes by mapping the spectral representation of signal on an input graph to a predefined canonical graph, in the context of which filters are learned. The mapping is a linear transformation predicted by a special network from the voxelized eigenbasis of the input mesh. The authors leverage duality in functional maps between spatial and spectral correspondences and use ground truth correspondences between shapes  to initialize the training, which they described as extremely challenging.

Finally, let us recap that spectral methods have been shown to work on undirected graphs only, as the adaptation of Laplacian to directed graphs is not straightforward~\citep{SardellittiBL17}. An interesting way of circumventing this was proposed by \citet{MotifNet}, who exploit the concept of motifs (small subgraphs) to construct a symmetric motif-induced adjacency matrix based on a non-symmetric adjacency matrix by counting the number of times an edge takes part in a set of canonical motifs.

\section{Spatial Graph Convolutions}   \label{sub:bg_spatialgc}

We have seen above that while the spectral construction is theoretically well motivated, there are obstacles in going beyond a single undirected graph structure. The spatial construction, on the other hand, may not have these limitations, though its designs are perhaps less principled. This section overviews the development in spatial construction methods for graph convolution and is closely related to the contributions presented in Chapters~\ref{chap:cvpr17} and~\ref{chap:cvpr18}.

Let $H^t \in \mathbb{R}^{N \times d^t}$ be $d^t$-dimensional signal matrix at $N$ nodes. Spatial graph convolution amounts to expressing the output signal $H_i^{t+1} \in \mathbb{R}^{d^{t+1}}$ at node $i$ as a function of the input signal $H_j^t \in \mathbb{R}^{d^t}$ at nodes $j$ within its $K$-hop local neighborhood.
\citet{scarselli09} proposed one of the first propagation models $M$, which (somewhat ironically) is arguably also the most general one: it may depend in an arbitrary way on the signals $H$, as well as edge and node attributes $E$ and $F$ within an arbitrary neighborhood $\mathcal{N}(i)$, formally written as 
\begin{equation} \label{eq:scarselli_eq_gen}
H_i^{t+1} = m(\{H_a^t | a \in \mathcal{N}(i) \cup \{i\}\}, \{F_a | a \in \mathcal{N}(i) \cup \{i\}\}, \{E_{a,b} | a,b \in \mathcal{N}(i) \cup \{i\}\}) .
\end{equation}
To account for the fact that neighboring nodes are typically unordered, the general form is then restricted to pairwise potentials with an aggregation over direct neighbors as
\begin{equation} \label{eq:scarselli_eq_rest}
H_i^{t+1} = u(F_i, \sum_{j \in \mathcal{N}(i)}{m(H_j^t, F_i, F_j, E_{j,i})}.
\end{equation}
The functions $m$ and $u$ are designed to be contractive operators, the whole network is treated as a dynamical system and solved for fixed point solution for $H$ from any initial state in its forward pass\footnote{It would be a very interesting exercise to implement this approach in a current deep learning framework and benchmark it against contemporary RNN-based graph convolution methods. \citet{yujia16} provide a theoretical insight that contraction is expected to attenuate the effect of long-range dependencies, though.}. No concept of graph-level output or graph coarsening was mentioned.

Recently, many graph convolution variants have appeared in the community. To make their comparison easier, \citet{GilmerSRVD17} suggested Message Passing Neural Network (MPNN) family fitting the vast majority of models of that time. It is a modified version of Scarselli's Equation~\ref{eq:scarselli_eq_rest}, where a) node labels are used for initializing node state $H^0$ rather than for propagation and b) states of both the source and the destination node can be used for propagation. Concretely, \citet{GilmerSRVD17} use the message function $m$ and the node update function $u$ in the following way: 
\begin{align} \label{eq:mpnn}
\begin{split}
M_i^{t+1} &= \sum_{j \in \mathcal{N}(i)}{m_t(H_i^t, H_j^t, E_{j,i})} \\
H_i^{t+1} &= u_t(H_i^t, M_i^{t+1})
\end{split}
\end{align}
These function define a recurrent neural network (RNN) where the state is processed over several iterations (time steps) $t = 1\dots T$. At each iteration $t$, the update function takes its hidden state $H_i^t$ and updates it with $M_i^{t+1}$, which is the aggregation of messages $m_t$ incoming from each neighbor. Finally, a readout function is defined to aggregate final node states into a graph-level state. We will use this framework (without the readout function, for simplicity) to chronologically review interesting related works in the following. Note that the list is far from being exhaustive, as there has been an explosion of many very closely related or sometimes even identical models in the community recently.

\begin{description}
\item[\citet{bruna13}] suggested a spatial method, in addition to his spectral model described in Section~\ref{sub:bg_spectralgc}. Here, $m_t: W^{t,j,i} H_j^t$ and $u_t: \mathrm{ReLU}(M_i^{t+1})$ with an edge-specific matrices of parameters $W$. The disadvantage of no weight sharing is a large number of parameters and no generalization over different graphs.

\item[\citet{duvenaud}] proposed message passing function $m_t: H_j^t$ and update function $u_t: \mathrm{tanh}(W^{t,\mathrm{deg}(i)} M_i^{t+1})$ with node-degree specific matrices of parameters $W$. The signal is passed over edges indiscriminately (nevertheless, there are weak constraints for possible edge types and node degree in their specific context of molecular fingerprints).

\item[\citet{MouLZWJ16}] proposed convolution operation on ordered trees, where the weights linearly depend on the position of a node within a triangular  window containing the parent node and children, $m_t: (\eta_j^1 W^1 + \eta_j^2 W^2 +  \eta_j^3 W^3) H_j^t + \mathbf{b}$ and $u_t: \mathrm{tanh}(M_i^{t+1})$, where $\eta$ are barycentric coordinates and $W$, $\mathbf{b}$ are learnable parameters.

\item[\citet{yujia16}] introduced support for discrete edge types and successful RNN layers, such as the Gated Recurrent Unit~\citep{cho-gru14}, by having $m_t: W^{E_{j,i}} H_j^t$ and $u_t: \mathrm{GRU}(H_i^t, M_i^{t+1})$, where $W$ are edge-type and edge-direction specific weight matrices.

\item[\citet{KearnesMBPR16}] developed a model where both nodes and edges hold hidden representations updated in an alternating fashion. Formally $m_t: E_{j,i}^t$, $u_t: \mathrm{ReLU}(W^2[\mathrm{ReLU}(W^1 H_i^t), M_i^{t+1}])$ and $E_{j,i}^{t+1} = \mathrm{ReLU}(W^5[\mathrm{ReLU}(W^3 E_{j,i}^t),$\\$\mathrm{ReLU}(W^4 [H_i^t, H_j^t])])$ where $[\cdot]$ denotes concatenation and $W$ weight matrices.

\item[\citet{BattagliaPLRK16}] implemented the general version of MPNN in Equation~\ref{eq:mpnn} with $m_t$ and $u_t$ as multi-layer perceptrons. This stands out from most other approaches by not explicitly decomposing message passing into a form of multiplication of $H_j^t$ with some weights. While the formulation has been proposed before \citet{GilmerSRVD17}, it has been demonstrated only as a single convolutional layer on fully connected graphs with discrete edge attributes.

\item[\citet{Schtt2017Quantum}] replaced the usual matrix-vector multiplication with element-wise multiplication in lower-dimensional space, leading to message function $m_t: \tanh(W^1((W^2 H_j^t + \mathbf{b}^1) \odot (W^3 E_{j,i} + \mathbf{b}^2))$ and $u_t: H_i^t + M_i^{t+1}$ where $W$ and $\mathbf{b}$ are learnable parameters. Later in \citep{SchuttKFCTM17}, the authors allowed for handling continuous edge attributes, leading to basically the same message function as in \citet{GilmerSRVD17}.

\item[\citet{MontiBMRSB17}], building on their previous work on learning on manifolds and meshes, proposed a Gaussian mixture model for message passing conditioned on continuous edge attributes. The model with $K$ mixture components can be written as $m_t: w(E_{j,i}) H_j^t$ with edge weighting function $w$, \\$w(\mathbf{x}) = \sum_{k=0}^K{\mathbf{a}_k \exp(-0.5(\mathbf{x}-\bm{\mu}^k)^T \mathrm{diag}(\bm{\sigma}^k)^{-1} (\mathbf{x}-\bm{\mu}^k))}$ where $\mathbf{a}$, $\bm{\mu}$, $\bm{\sigma}$ are learnable parameters and $\mathrm{diag(\cdot)}$ builds a diagonal matrix from a vector.

\item[\citet{GilmerSRVD17}], the authors of MPNN, proposed message passing conditioned on continuous edge attributes using a neural network $w$ outputting a edge-specific weight matrix, which is then used in $m_t: w(E_{j,i}) H_j^t$. A GRU is used for maintaining node state, $u_t: \mathrm{GRU}(H_i^t, M_i^{t+1})$.

\item[\citet{HamiltonYL17}] applied graph convolutions to unsupervised node embedding using a loss encouraging close nodes to have a similar embedding and distant ones otherwise. The concrete functions are $m_t: \mathrm{ReLU}(W^1 H_j^t + \mathbf{b})$ and $u_t: \mathrm{ReLU}(W^{t,2}[H_i^t, M_i^{t+1}])$ where $W$ and $\mathbf{b}$ are learnable parameters. In addition, max is used instead of sum in Equation~\ref{eq:mpnn} The neighborhoods are subsampled to support large node degrees and edges are untyped.

\item[\citet{GraphAtt}] used self-attention mechanism to weigh contributions of neighbors using coefficients $\alpha_{i} = \mathrm{softmax}_j(\mathbf{a}[W H_i^t, W H_j^t])$ in functions $m_t: \alpha_{ij} W H_j^t$ and $u_t: \mathrm{ReLU}(M_i^{t+1})$, where $W$ and $\mathbf{a}$ are learnable parameters and the neighborhood $\mathcal{N}(i)$ includes the central node $i$ itself. Independently, \citet{wang2018non} proposed further formulations of $\alpha_{ij}$ in the context of non-local image processing.

\item[\citet{dgcnn}] proposed a message passing function formulated as $m_t: w(H_i^t, H_j^t - H_i^t)$ with a neural network $w$. While being less general than \citet{BattagliaPLRK16}, it forces the network to perceive $H_j^t$ relatively to $H_i^t$, which is arguably advantageous if the embeddings also encode global information, such as position. Max aggregation is used instead of sum in Equation~\ref{eq:mpnn} and the update function is simply $u_t: M_i^{t+1}$.

\item[\citet{verma2018feastnet}] used a variant of message passing $m_t: w(H_i^t, H_j^t) H_j^t$ with weights defined as a mixture of $K$ matrices, $w(\mathbf{x}, \mathbf{y}) = \sum_{k=0}^K{W^k \mathrm{softmax}_j(\mathbf{x}^T\mathbf{a}^j + \mathbf{y}^T\mathbf{b}^j + \mathbf{c}^j)_k}$, where $W$ and $\mathbf{a}, \mathbf{b}, \mathbf{c}$ are learnable parameters.

\end{description}

Our work also fits into MPNN framework. Specifically, Edge-Conditioned Convolutions (ECC), presented in Chapter~\ref{chap:cvpr17}, can be defined as $m_t: w^t(E_{j,i}) H_j^t + \mathbf{b}^t$ and $u_t: \mathrm{ReLU}(M_i^{t+1})$, where $w^t$ is a neural network\footnote{The message function is equivalent to that in \citet{GilmerSRVD17}, which was a concurrent work.} and $\mathbf{b}^t$ a learnable bias. In addition, mean is used instead of sum in Equation~\ref{eq:mpnn} and the neighborhood $\mathcal{N}(i)$ includes the central node $i$ itself. Several models listed above are put into relation to our formulation in more detail in Subsection~\ref{subsec:relgridconv}. In Chapter~\ref{chap:cvpr18}, we further extend ECC with element-wise multiplication and GRU input gating, defined as $m_t: w(E_{j,i}) \odot H_j^t$ and $u_t: \mathrm{GRU}(H_i^t, (W H_i^t + \mathbf{b}) \odot M_i^{t+1})$, where $W$ and $\mathbf{a}$ are learnable parameters.

Interestingly, the spatial representation of spectral methods can also be interpreted within MPNN framework, see Appendix in \citet{GilmerSRVD17} for details. For example, the equivalent formulation of \citet{kipf} is $m_t: \hat{A}_{i,j}(\mathrm{deg}(i)\mathrm{deg}(j))^{-1/2} H_j^t$ and $u_t: \mathrm{ReLU}(W^t M_i^{t+1})$.

Finally, let us note that there are also several works which do not easily fit the MPNN framework. Diffusion-convolutional NN \citep{dcnn} produces features by applying diffusion of different length, \ie computes the average of features at all nodes weighted proportionally to transition probabilities given by powers of graph stochastic matrix $P=D^{-1}A$. Patchy-SAN \citep{niepert} linearizes selected graph neighborhoods by choosing a lexicographically maximal adjacency matrix so that a conventional 1D CNN can be used.

\section{Non-convolutional Embedding Methods} \label{bg_nonconv}

Besides convolutional methods, there have been other approaches of learning the embeddings of nodes or parts of graphs. While these methods are not directly related to the contributions of this thesis, we find it interesting to provide a brief overview for completeness. In particular, we review direct node embedding and graph kernels in the following.

\subsection{Direct Node Embedding}

Direct methods define the $i$-th node embedding $\mathbf{z}_i$ to be a function of only the node's own attributes $F_i$ or even its unique identity if no attributes are given. Note that this is different from the convolutional approach, where the embedding of a node is defined as the function of its surroundings as well. The main idea of the majority of direct embedding approaches is to use unsupervised objective functions encouraging the embedding $\mathbf{z}$ of "close" nodes to be similar, usually measured as their inner-product or Euclidean distance. 

Early methods, \eg Laplacian eigenmaps~\citep{BelkinN01}, often approached the task as a dimensionality reduction problem. Given proximity matrix $S$ describing the "closeness" between nodes, such as the (weighted) adjacency matrix, the optimization problem has the form of matrix factorization $||Z^T Z - S||$, where $Z$ is the sought matrix of embeddings. In effect, the obtained embedding vectors just approximates some prescribed measure.

The advent of deep learning in natural language processing opened new ways for representation learning of discrete objects, such as words. Multiple successful techniques for unsupervised learning of word embeddings were based on co-occurence statistics, \ie the assumption that similar words tend to appear in similar word neighborhoods (context). In particular, Skip-gram~\citep{mikolov2013efficient} aims to predict a word's embedding based on the embedding of its contextual words and vice versa. 

This idea has been adapted to graphs in various ways of defining the context of a node. A popular way of establishing context is by using random walks on graph, pioneered by DeepWalk~\citep{PerozziAS14} and later improved on by node2vec with biased walks~\citep{GroverL16}, allowing for smooth transition between depth-first and breath-first exploration. The similarity of embeddings approximates the stochastic transition matrix $P$ by $e^{\mathbf{z}_i^T \mathbf{z}_j} / \sum_{v_k \in V} e^{\mathbf{z}_i^T \mathbf{z}_k} \approx P_{i,j}$, $P_{i,j}$ being the probability of visiting node $j$ from $i$. The unsupervised loss function can combined with a supervised task objective, such as node classification.

Nevertheless, the vast majority of direct encoding approaches learn a unique embedding vector for each node individually, which leads to two major drawbacks: the number of parameters grows linearly with the size of the graph and it is not possible to generate embeddings for nodes not seen during training or generalize across graphs. In semi-supervised learning, this is called transductive learning, as opposed to inductive learning, which can deal with novel test nodes. One way of addressing this deficiency is to learn a network encoding node attributes instead of identities, as in \citet{BojchevskiG17}. The other way is to use spatial graph convolutions introduced in Section~\ref{sub:bg_spatialgc} above (\citet{HamiltonYL17} in particular), which may not be suitable for link prediction of yet unconnected new nodes, though.

\subsection{Graph Kernels}

Kernel methods~\citep{scholkopf2002kernels} are a family of established approaches based on measuring the similarity between two objects using an explicit kernel function $\mathcal{K}$ corresponding to the inner product in reproducing kernel Hilbert space $\mathcal{H}$. Such kernel is then used in place of inner products in various kernelized algorithms, notably Support Vector Machines.

A kernel between two graphs $G$ and $G'$ is given by $\mathcal{K}(G,G') := <\Phi(G), \Phi(G')>_\mathcal{H}$, where $\Phi(G)$ denotes a feature vector containing counts of a certain class of substructures in graph $G$ and $\mathcal{H}$ is typically a Euclidean space. The challenge is to find substructure decompositions that captures the semantics of the graph while being computationally tractable to enumerate and evaluate, such as decompositions into graphlets (subgraphs of limited size), subtrees, or random walks. The Weisfeiler-Lehman framework increased the efficiency of previous kernels by using a relabeling procedure and scaled to graphs with thousands of nodes~\citep{shervashidze}. Recently, \citet{lei2017deriving} cast the computation process of Weisfeiler-Lehman kernel as a (differentible) recurrent network.

Similarly to direct node embedding introduced above, there has been works exploiting Skip-gram for substructure embedding. Such embeddings can be used to improve graph kernels; \citet{deepkern} proposed not treating substructures as independent and introduced bilinear form $\Phi(G)^T M \Phi(G')$ called Deep graph kernels, where learnable matrix $M$ captures the relationship between substructures and is computed from inner products of their individual embeddings. Subgraph2vec~\citep{narayanan2016subgraph2vec} later improved the way how the context of subgraphs in a graph should be modeled.

\chapter{Edge-Conditioned Convolutions} \label{chap:cvpr17}

\section{Introduction}

Convolutional Neural Networks (CNNs) have gained massive popularity in tasks where the underlying data representation has a grid structure, such as in speech processing and natural language understanding (1D, temporal convolutions), in image classification and segmentation (2D, spatial convolutions), or in video parsing (3D, volumetric convolutions)~\citep{lecun2015deep}. 

On the other hand, in many other tasks the data naturally lie on irregular or generally non-Euclidean domains, which can be structured as graphs in many cases. These include problems in 3D modeling, computational chemistry and biology, geospatial analysis, social networks, or natural language semantics and knowledge bases, to name a few. Assuming that the locality, stationarity, and composionality principles of representation hold to at least some level in the data, it is meaningful to consider a hierarchical CNN-like architecture for processing it.

However, a generalization of CNNs from grids to general graphs is not straightforward and has recently become a topic of increased interest. We identify that the formulations of graph convolution predating this work do not exploit continuous edge attributes, which results in an overly homogeneous view of local graph neighborhoods, with an effect similar to enforcing rotational invariance of filters in regular convolutions on images. Hence, in this work we propose a convolution operation which can make use of this information channel and show that it leads to an improved graph classification performance.

This novel formulation also opens up a broader range of applications; we concentrate here on point clouds specifically. Point clouds have been mostly ignored by deep learning before publication of the work presented in this chapter, their voxelization being the only trend  \citep{voxnet,pclabeling16,qi16}. To offer a competitive alternative with a different set of advantages and disadvantages, we construct graphs in Euclidean space from point clouds in this work and demonstrate state of the art performance on Sydney dataset of LiDAR scans~\citep{trianglesvm}.

This chapter is largely based on our CVPR 2017 publication~\citep{simonovsky2017dynamic}. Its contributions to the field at the time of submission are as follows:

\begin{itemize}
\item We formulate a convolution-like operation on graph signals `performed in the spatial domain where filter weights are conditioned on edge attributes (discrete or continuous) and dynamically generated for each specific input sample. Our networks work on graphs with arbitrary varying structure throughout a dataset.
\item We are the first to apply graph convolutions to point cloud processing. Our method outperforms volumetric approaches and attains the new state of the art performance on Sydney dataset, with the benefit of preserving sparsity and presumably fine details.
\item We reach a competitive level of performance on graph classification benchmark NCI1~\citep{nci1db}, outperforming other approaches based on deep learning there.
\end{itemize}

\section{Related Work} \label{sec:ecc-related}

\subsection{Graph Convolutions}

Here we put the contributions of this chapter in the context of related work and refer the reader to Chapter \ref{chap:background} for the review of the topic in general. 
Spectral methods \citep{bruna13, defferrard16} offer a mathematically sound definition of convolution operator and learn filters in relation to the spectrum of graph Laplacian, which therefore has to be the same for all graphs in a dataset. This means that the graph structure is fixed and only the signal defined on the nodes may differ, precluding applications on problems where the graph structure varies in the dataset, such as meshes, point clouds, or diverse biochemical data. 

To cover these important cases, we formulate our filtering approach in the spatial domain, where the limited complexity of evaluation and the localization property is provided by construction. The main challenge here is dealing with weight sharing among local neighborhoods, as the number of nodes adjacent to a particular vertex varies and their ordering is often not well definable.

\citet{bruna13} assumed fixed graph structure and did not share any weights among neighborhoods. Several works have independently dealt with this problem. \citet{duvenaud} sum the signal over neighboring nodes followed by a weight matrix multiplication given by the degree of central node. \citet{dcnn} share weights based on the number of hops between two nodes. \citet{kipf} further approximate the spectral method of \citet{defferrard16} and weaken the dependency on the Laplacian, but ultimately arrive at center-surround weighting of neighborhoods. None of these methods captures finer structure of the neighborhood and thus does not generalize regular convolution on grids. In contrast, here we show that methods using additional information in the form of edge attributes do offer such a convenient property. These includes the work of \citet{yujia16} and \citet{MontiBMRSB17}. Section~\ref{subsec:relgridconv} demonstrates that our method generalizes the above mentioned models. In a concurrent work, \citet{GilmerSRVD17} presented an equivalent formulation to our method. Finally, the approach of \citet{niepert} introduces a heuristic for linearizing selected graph neighborhoods so that a conventional 1D CNN can be used. We share their goal of capturing structure in neighborhoods but approach it in a different way. 

\subsection{Deep Learning on Sparse 3D Data}

Three-dimensional data can be captured and processed in various forms. Here we restrict ourselves to point clouds and polygonal meshes, which can be treated as volumes, sets or graphs for their processing.

\paragraph*{Volumetric representation.} Volumetric representation, which can be processed by regular convolutions on grid, was the first one to be used in deep learning-based 3D understanding methods and has enjoyed popularity since then. Medical images \citep{shen2017deep}, distance fields \citep{LiPSQG16} or uncertainties in output space \citep{WuSKYZTX15} are just a few examples of dense spatial data. Nevertheless, for simplicity and efficient processing in hardware, also sparse data may be voxelized into a dense volumes. In fact, prior to the submission of our work (2016), this was the only way of processing point clouds using deep learning, be it for classification \citep{voxnet,qi16} or segmentation \citep{pclabeling16} purposes. However, voxel grid representation of sparse data tends to be much more expensive in terms of memory and brings about discretization artifacts due to limited resolution or the need to resort to a sliding window approach.

To address this, \citet{Graham15} proposed a way of computing sparse convolution on lattices where the input is gathered into a dense matrix to allow for standard matrix multiplication. Unfortunately, the bookkeeping is performed on the CPU in their implementation and the evaluation has been fairly limited.  %
In a work concurrent to ours, OctNet~\citep{Riegler2017OctNet} and O-CNN~\citep{Wang-2017-ocnn} have expanded on the topic and replaced uniform grids by non-uniform octrees of a predefined depth. An octree is a spatial subdivision structure with adaptive cell size that represents empty space efficiently while allowing to store data in a linear array. However, unlike 
in \citet{Graham15} where the sparsity is decreased at each layer due to convolution dilating the number of active (non-zero) locations, octree structures are kept unmodified. While preserving sparsity, such a formulation of convolution has different behavior from the one on dense grids in how it propagates information across empty space~\footnote{Interestingly, this is also true for all methods operating directly on point clouds and meshes mentioned further below: their convolutions add no new points or nodes to the domain. In fact, the computation of convolution on octree cells can be seen as a particular instance of a graph, which could be processed by our graph convolution method presented in this Chapter to give rise to the same output.}. Nevertheless, this does not seem to be a problem in practice and has not prevented octree-based methods from reaching state of the art results. This formulation of spatial convolution has also been named submanifold sparse convolutions by~\citet{SubmanifoldSparseConvNet}. There, the authors revisit the algorithm of \citet{Graham15} and rely on strided convolutions to allow information to flow between disconnected components in the input, claiming better efficiency than OctNet. The necessity to know the grid subdivision structure or the submanifold ahead is inconvenient in reconstruction and shape generation applications. However, in later works~\citep{ogn2017,Riegler2017OctNetFusion}, octrees were successfully used also in such applications by refining spatial subdivisions in a multi-resolution pipeline. Independently, \citet{KlokovL17KdNet} proposed hierarchical processing of point clouds according to the partitioning given by balanced kd-trees, where cells are merged by multilayer perceptrons. However, the distance metric induced by axis-aligned kd-trees is often rather different to the Euclidean one, demonstrated in the method's sensitivity to global rotations.

\paragraph*{Set representation.} A set is an intuitive representation for point clouds. Concurrently to our work, \citet{qi2016pointnet} was the first to propose a network operating directly on sets of points. Their key idea is to embed each point individually into high-dimensional latent space and then apply a permutation-invariant aggregation function (channel-wise maximum) to arrive at global descriptors, which can be again concatenated with point embeddings for point-wise prediction tasks. As the learned mapping is sensitive to global transformations, a spatial transformer network~\citep{JaderbergSTN} in the form of a smaller PointNet is used to roughly normalize the global coordinate system of input point clouds. While the architecture is remarkably simple, efficient and achieves great performance on a variety of tasks, \citet{QiYSG17PointNetPP} observed its limitations in fine-grained segmentation tasks due to its obliviousness to any concept of spatial closeness. Their follow-up work PointNet++~\citep{QiYSG17PointNetPP} applies PointNet hierarchically on a nested partitioning of the input set, in reminiscence of strided convolution. While improving over PointNet, points in local point sets are still treated independently. \citet{Engelmann17_3dsemseg} later investigated incorporation of spatial context into PointNet by several ways of fusing embeddings of neighboring point cloud parts. 

\paragraph*{Graph representation.} Unlike sets, graphs can model the context of spatial data explicitly. While neighborhood graphs constructed from point clouds express only geometric information, polygonal meshes additionally capture topological constraints (faces) and represent discretized manifolds (surfaces). We regard point cloud as graphs in Euclidean space in this work. \citet{masci15} are motivated by connections between meshes and manifolds and define convolution on spatially binned neighborhoods around every node in a mesh using geodesic distances. Their architecture contains only a single convolutional layer operating on combination of intrinsic spectral shape features. Follow-up work proposed a different local binning technique using anisotropic diffusion~\citep{BoscainiMRB16} and, concurrently to our work, using Gaussian mixture model~\citep{MontiBMRSB17}. The latter, already mentioned above, can be regarded as a general graph convolution approach, though subsumed by our proposed model. Later, \citet{dgcnn} introduced an architecture where edge attributes are defined as the difference of current node embeddings and the neighborhood graph is repeatedly dynamically rebuilt.

\begin{figure*}[ht]
\centering
\includegraphics[width=\linewidth]{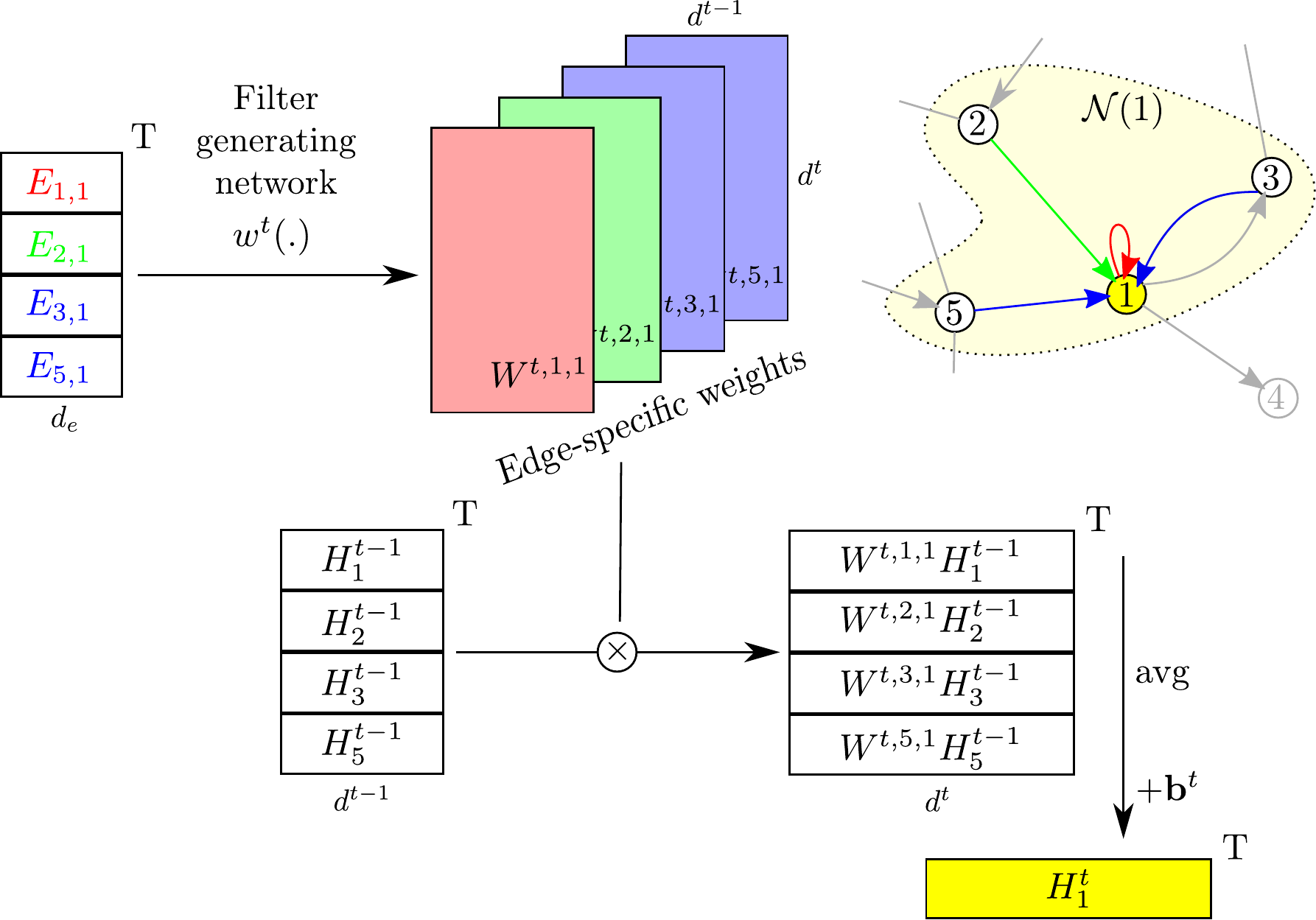}
\vspace{1.5ex}
\caption{\label{fig:gconv} Illustration of edge-conditioned convolution on a directed subgraph. Signal $H_1^t$ on vertex 1 in the $t$-th network layer is computed as a weighted sum of signals $H^{t-1}$ on the set of its predecessor nodes, assuming self-loops. The particular weight matrices $W$ are dynamically generated by filter-generating network $w^t$ based on the corresponding edge attributes $E$, visualized as colors.}
\end{figure*}

\section{Method} \label{sec:method}

We propose a method for performing convolutions over local graph neighborhoods exploiting edge attributes (Section~\ref{subsec:ecc}) and show it to generalize regular convolutions (Section~\ref{subsec:relgridconv}). Afterwards, we present deep networks with our convolution operator (Section~\ref{subsec:eccnet}) in the case of point clouds (Section~\ref{subsec:applclouds}) and general graphs (Section~\ref{subsec:applgraphs}).

\subsection{Edge-Conditioned Convolution} \label{subsec:ecc}

Let us consider a directed graph $G=(\mathcal{V},\mathcal{E})$ with edge attributes $E: \mathcal{E}\to\mathbb{R}^{d_e}$. Building on top of graph notation introduced in Section~\ref{sec:graph_notation}, let $H^t:\mathcal{V} \to\mathbb{R}^{d^t}$ denote the node representation (signal) computed at convolutional layer $t\in\{1,\ldots,T\}$ in a neural network. We make the input signal equal to node attributes $H^0=F$ if these are available and set $H^0=0$ otherwise. The neighborhood $\mathcal{N}(i)=\{j;(j,i)\in \mathcal{E}\}$ of node $i$ is defined to contain all its direct predecessors.

\def\dimo{{\mathbb{R}^{d^t}}}
\def\dimi{{\mathbb{R}^{d^{t-1}}}}
\def\dimjoint{{\mathbb{R}^{d^t\times d^{t-1}}}}

Our approach computes the filtered signal $H_i^t\in\dimo$ at node $i$ as a weighted sum of its own signal $H_i^{t-1}\in\dimi$ and signals $H_j^{t-1}\in\dimi$ in its neighborhood, $j\in \mathcal{N}(i)$. While such a commutative aggregation solves the problem of undefined node ordering and varying neighborhood sizes, it also smooths out any structural information. To retain it, we propose to condition each filtering weight on the respective edge attribute. To this end, we borrow the idea from Dynamic Filter Networks~\citep{dfn16} and define filter-generating network $w^t: \mathbb{R}^{d_e} \to \dimjoint$ with parameters $\theta^t$ which outputs an edge-specific weight matrix $W^{t,j,i} \in\dimjoint$ given edge attributes $E_{j,i}$. See Figure~\ref{fig:gconv} for an illustration. 

The convolution operation, coined Edge-Conditioned Convolution (ECC), is formalized as follows:
\begin{equation}
\begin{split}
H_i^{t+1} &= \mathrm{ReLU}\left(\frac{1}{|\mathcal{N}(i)\cup\{i\}|} \sum_{j\in \mathcal{N}(i)\cup\{i\}} w^t(E_{j,i};\theta^t) H_j^{t} + \mathbf{b}^t\right) \label{eq:C1} \\
&= \mathrm{ReLU}\left(\frac{1}{|\mathcal{N}(i)\cup\{i\}|} \sum_{j\in \mathcal{N}(i)\cup\{i\}} W^{t,j,i} H_j^{t} + \mathbf{b}^t\right)
\end{split}
\end{equation}

Equivalently, in the framework of Message Passing Neural Networks (see Section~\ref{sub:bg_spatialgc}), ECC can be defined with message function $m_t: w^t(E_{j,i};\theta^t) H_j^t + \mathbf{b}^t$, update function $u_t: \mathrm{ReLU}(M_i^{t+1})$ and modified Equation~\ref{eq:mpnn} where mean is used instead of sum and the neighborhood includes the central node $i$ itself.

It is important to understand that we do not learn convolution weights directly but rather a network which predicts them. In fact, it would be intractable to learn convolution weights for individual values of continuous attributes. Therefore, $\theta^t$ and $\mathbf{b}^t$ are learned model parameters updated with gradient descent during training and $W^{t,j,i}$ are dynamically predicted parameters for attributes of a particular edge in a particular input graph. 

Parameters $\theta^t$ can be untied in each network layer $t$ (such as in this chapter and Chapter~\ref{chap:iclr18}) or shared across layers as in Chapter~\ref{chap:cvpr18}, where a more involved update function $u_t$ is used to compensate for that.

The filter-generating networks $w$ can be implemented using any differentiable architecture. As attributes are simple vectors in all our applications, we use multi-layer perceptrons (MLPs). In the case of structured attributes such as images or text (\eg a free-form description of relations in a knowledge graph), CNNs or RNNs would be a more appropriate choice for $w$.

Note that ECC is defined on directed graphs. However, directed edges $(i,j)$ without corresponding opposite edges $(j,i)$ cause asymmetries in message passing. In the extreme case, start nodes in a directed acyclic graph would obtain no information about any other node. Therefore, we suggest making sure that both edge directions are present in input graphs and that their attributes differ so that message passing depends on the direction of that edge. For instance, it is reasonable to set $E_{i,j}=-E_{j,i}$ if attributes represent offsets between points in spaces. Having $E_{i,j}=E_{j,i}$ is sufficient in the case of undirected graphs, which are supported by replacing each undirected edge $\{i,j\}$ with two directed ones $(i,j)$ and $(j,i)$.

\paragraph*{Identity connections (ECC-id).} Finally, we discuss the treatment of central node $i$. The formulation of ECC in Equation~\ref{eq:C1} effectively adds self-loops to the graph and does not treat $i$ in any special way (other than consistent weights $W^{t,i,i}$ for all $i$ if self-loops have a distinct attribute $E_{i,i}$), which is consistent with the definition of convolution on grids. However, the success of Residual Networks \citep{residuals} is a strong motivation to consider adding identity skip-connections to the model and encourage ECC in learning residuals. In addition, moving the central node out of the aggregation makes its influence independent of the size of the neighborhood. We formulate ECC-id as follows:
\begin{equation}
H_i^{t+1} = \mathrm{ReLU}\left(\mathrm{id}(H_i^{t}) + \frac{1}{|\mathcal{N}(i)|} \sum_{j\in \mathcal{N}(i)} w(E_{j,i};\theta^t) H_j^{t} + \mathbf{b}^t\right) \label{eq:Crn}
\end{equation}
where $\mathrm{id}()$ is an identity mapping if $d_t=d_{t-1}$ and a linear mapping otherwise. The equivalent update function in Message Passing framework is $u_t: \mathrm{ReLU}(\mathrm{id}(H_i^{t}) + M_i^{t+1})$.

\paragraph*{Complexity.} Computing $H^t$ for all nodes requires at most\footnote{If edge attributes are represented by $d_e$ discrete values in a particular graph and $d_e<|\mathcal{E}|$, $w$ can be evaluated only $d_e$-times.} $|\mathcal{E}|$ evaluations of $w$ and $|\mathcal{E}|+|\mathcal{V}|$ or $2|\mathcal{E}|+|\mathcal{V}|$ matrix-vector multiplications for directed, resp. undirected graphs. Both operations can be carried out efficiently on the GPU in batch-mode. Aggregation over neighborhood can be implemented as sparse-dense matrix multiplication, available in both PyTorch and TensorFlow frameworks. Therefore, the method can scale well to large sparse graphs both theoretically and in practice.

\subsection{Relationship to Existing Formulations} \label{subsec:relgridconv}

Our formulation of convolution on graph neighborhoods retains the key properties of the standard convolution on regular grids that are useful in the context of CNNs: weight sharing and locality. 

\begin{figure}[bt]
\centering
\includegraphics[width=0.5\linewidth]{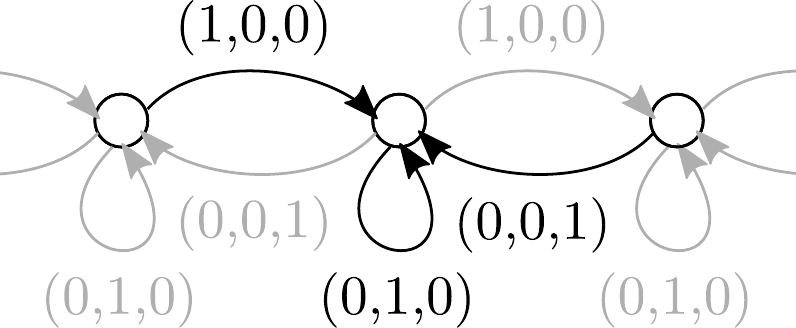}
\vspace{1.5ex}
\caption{\label{fig:regularequiv} Construction of a directed graph with one-hot edge attributes where the proposed edge-conditioned convolution is equivalent to the regular 1D convolution with a centered filter of size $d_e=3$.}
\end{figure}

Importantly, the standard discrete convolution on grids is a special case of ECC, which we demonstrate in 1D for clarity. Consider an ordered set of nodes $\mathcal{V}$ forming a path graph (chain). To obtain convolution with a centered kernel of size $d_e$, we form $\mathcal{E}$ so that each node is connected to its $d_e$ spatially nearest neighbors including self by a directed edge labeled with one-hot coding of the neighbor's discrete offset $\delta$, see Figure~\ref{fig:regularequiv}. Taking $w^t$ as a single linear layer without bias, we have $w(E_{j,i};\theta^t) = \theta_{\delta}^t$, where $\theta_{\delta}^t$ denotes the respective reshaped column of the parameter matrix $\theta^t\in \mathbb{R}^{(d_t\times d_{t-1})\times d_e}$. With a slight abuse of notation, we arrive at the equivalence to the standard convolution: $H_i^{t+1} = \mathrm{ReLU}(\sum_{j\in \mathcal{N}(i)} W^{t,j,i} H_j^{t}) =  \mathrm{ReLU}(\sum_{\delta} \theta_{\delta}^t H_{i-\delta}^{t}$), ignoring the normalization factor $1/|\mathcal{N}(i)\cup\{i\}|$ playing a role only at grid boundaries.

This shows that ECC can retain the same number of parameters and computational complexity of the regular convolution in the case of grids. Note that such reduction is not possible with any of related work oblivious to edge attributes due to their way of weight tying. The weights in ECC are tied by edge attribute, which is in contrast to tying them by hop distance from a node~\citep{dcnn}, according to a neighborhood linearization heuristic~\citep{niepert}, by being the central node or not~\citep{kipf}, by node degree~\citep{duvenaud}, or not at all~\citep{bruna13}.

In fact, our definition of message passing function can be shown to generalize a number of prior graph convolution methods introduced in Section~\ref{sub:bg_spatialgc} by using a single linear layer without bias as $w^t$ and defining attributes appropriately. The model of \citet{bruna13} can be recovered by assigning a unique one-hot code to every edge. \citet{duvenaud} requires a one-hot code indicating the degree of target node $\mathrm{deg}(i)$ to every edge $(j,i)$, while the model of \citep{kipf} is obtained by setting $E_{i,j} =(\mathrm{deg}(i)\mathrm{deg}(j))^{-1/2}$. The support for discrete attributes in \citet{yujia16,schlichtkrull2017modeling} can be obtained by using them directly in one-hot coding. The Gaussian mixture model of \citet{MontiBMRSB17} (concurrent work) is subsumed by our general neural network $w^t$ by the universal approximation theorem~\citep{Hornik91}. Finally, the message passing function of \citet{GilmerSRVD17} (also concurrent work) is equivalent to our method.

Last, let us remark that the idea of exchanging messages among nodes is also the algorithmic basis of many inference techniques within the context of graphical models, such as belief propagation \citep{Pearl:1988}. We postpone further discussion to Section~\ref{subsec:contexseg} in the following Chapter, where we contrast message passing networks and conditional random fields.

\subsection{Deep Networks with ECC}  \label{subsec:eccnet}

\begin{figure*}[bt]
\centering
\includegraphics[width=\linewidth]{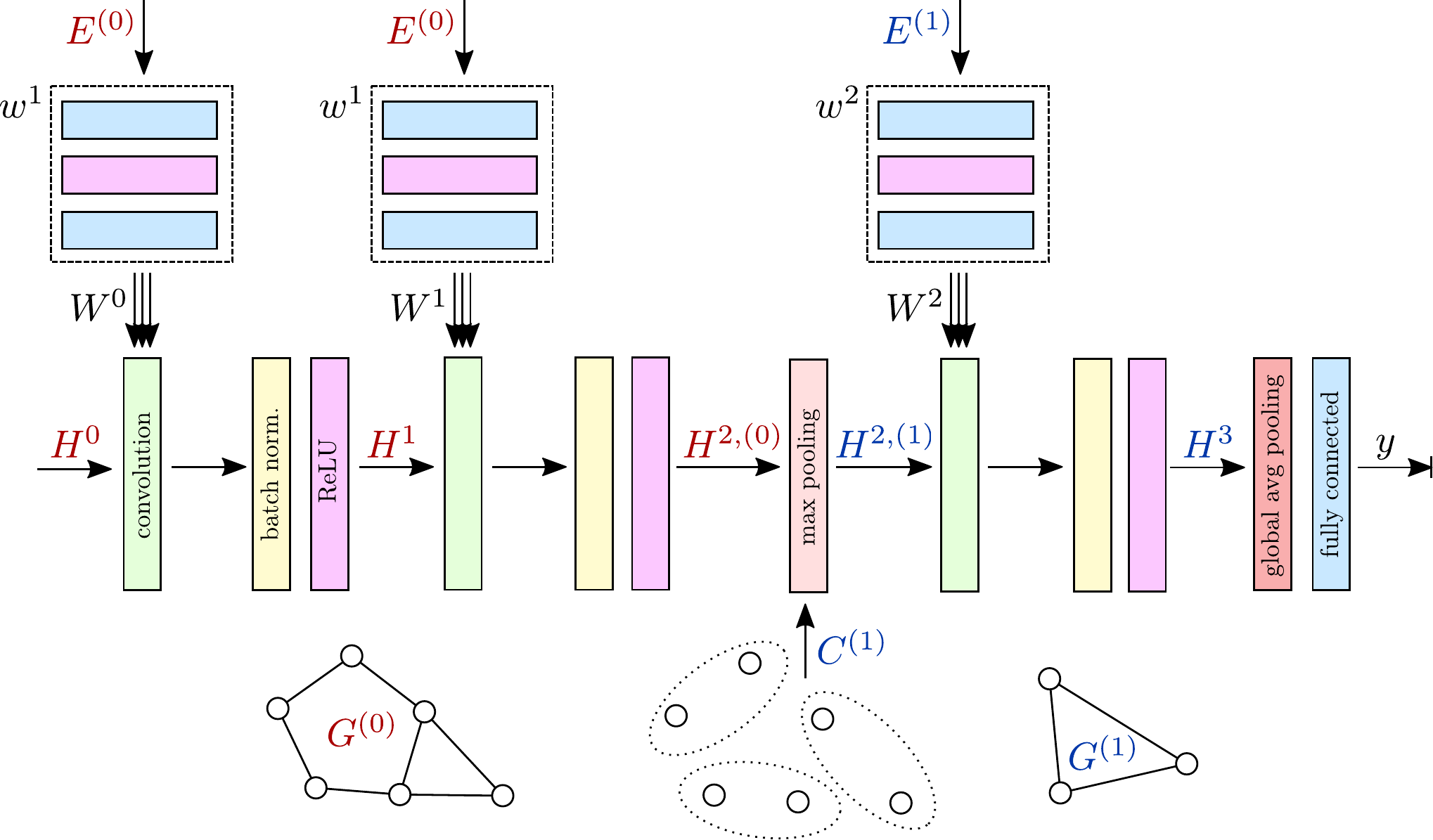}
\vspace{1.5ex}
\caption{\label{fig:netexample} Illustration of a deep network with three edge-conditioned convolutions and one pooling. The last convolution is executed on a structurally different graph $G^{(1)}$, which is related to the input graph $G^{(0)}$ by coarsening and signal aggregation in the max pooling step according to mapping $C^{(1)}$. See Section~\ref{subsec:eccnet} for more details.}
\end{figure*}

While ECC is in principle applicable to both node classification and graph classification tasks, in this chapter we restrict ourselves only to the latter one, \ie predicting a class for the whole input graph. Hence, we follow the common architectural pattern for feed-forward networks of interlaced convolutions and poolings topped by global pooling and fully-connected layers, see Figure~\ref{fig:netexample} for an illustration. This way, information from local neighborhoods gets combined over successive layers to gain context due to enlarged receptive field. While edge attributes are fixed for a particular graph, their (learned) interpretation by the means of filter generating networks $w^t$ may change from layer to layer if the weights are untied. Therefore, the restriction of ECC to 1-hop neighborhoods $\mathcal{N}(i)$ is not a constraint, akin to using small 3$\times$3 filters in exchange for deeper networks in CNNs on grids, which is known to be beneficial~\citep{hesun14}.

We use batch normalization~\citep{batchnorm} after each convolution, which was necessary for the learning to converge. Interestingly, we had no success with other feature normalization techniques such as data-dependent initialization~\citep{goodinit} or layer normalization~\citep{layernorm}. On the other hand, explicitly assuring the convolution operation being a non-expansive operator \citep{scarselli09} by applying activation function $\nu\mathrm{tanh}(w(\cdot))$ with $0<\nu<1$ can also make the network converge; the experimental performance was worse than with batch normalization, though.

\paragraph*{Pooling.} While (non-strided) convolutional layers and all point-wise layers do not change the underlying graph and only evolve the signal on nodes, pooling layers are defined to output aggregated signal on the nodes of a new, coarsened graph. Therefore, a pyramid of $S$ progressively coarser graphs has to be constructed for each input graph. Let us extend here our notation with an additional scale superscript $s\in\{0,\ldots,S\}$ to distinguish among different graphs $G^{(s)}=(\mathcal{V}^{(s)}, \mathcal{E}^{(s)})$ in the pyramid when necessary. Each $G^{(s)}$ has also its associated attributes $E^{(s)}$ and signal $H^{t,(s)}$.

As introduced in Section~\ref{sub:bg_coarsening}, a coarsening of $G^{(s-1)}$ typically consists of three steps: subsampling or merging nodes, creating the new edge structure $\mathcal{E}^{(s)}$ and labeling $E^{(s)}$ (so-called reduction), and mapping the nodes in the original graph to those in the coarsened one with $C^{(s)}: \mathcal{V}^{(s-1)}\to \mathcal{V}^{(s)}$. We use a different algorithm depending on whether we work with general graphs or graphs in Euclidean space, therefore we postpone discussing the details to the applications. Finally, the $s$-th pooling layer aggregates $H^{t,(s-1)}$ into a lower dimensional $H^{t,(s)}$ based on $C^{(s)}$. See Figure~\ref{fig:netexample} for an example of using the introduced notation. 

During coarsening, a small graph may be reduced to several disconnected nodes in its lower resolutions without problems as our formulation of graph convolution always assumes a virtual self-edge. Since the architecture is designed to process graphs with variable number of edges and nodes, we deal with varying node count $|\mathcal{V}^{(S)}|$ in the lowest graph resolution by global average/max pooling.

\subsection{Application in Point Clouds} \label{subsec:applclouds}

Point clouds are an important 3D data modality arising from many acquisition techniques, such as laser scanning (LiDAR) or multi-view reconstruction. However, their natural irregularity and sparsity present a challenge to the massive parallelism in GPU-based processing. In fact, at the submission time of our paper~\citep{simonovsky2017dynamic}, the only way of processing point clouds using deep learning has been to first voxelize them before feeding them to a 3D CNN, be it for classification~\citep{voxnet} or segmentation~\citep{pclabeling16} purposes. Such a dense representation is very hardware friendly and simple to handle with the current deep learning frameworks.

On the other hand, there are several disadvantages too. First, voxel representation tends to be much more expensive in terms of memory than usually sparse point clouds. Second, the necessity to fit them into a fixed size 3D grid brings about discretization artifacts and the loss of details and possibly metric scale. With the work presented in this chapter, we attempted at offering a competitive alternative to the mainstream by performing deep learning on point clouds directly. While we concentrate on point clouds in this thesis, we believe this work can also directly apply to meshes, as the graph structure is given and mesh downsampling is a well studied problem.

\paragraph*{Graph Construction.} Given a point cloud $\mathcal{P}$ with point positions $P$ and features $F$ (such as laser return intensity or color) we build a directed graph $G=(\mathcal{P},\mathcal{E})$ by connecting each node $i$ to all nodes $j$ in its spatial neighborhood by a directed edge $(j,i)$. In our experiments with neighborhoods, fixed metric radius $\rho$ worked better than a fixed number of neighbors, likely due to more symmetric message passing and more homogeneous speed of information propagation. The node signal is set as $H^0=F$ (or 0 if there are no features $F$). 
The offset $\delta=P_j-P_i$ between the points corresponding to nodes $j$, $i$ is represented in Cartesian and spherical coordinates as 6D edge attribute vector $E_{j,i} = (\delta_x, \delta_y, \delta_z, ||\delta||, \arccos \delta_z/||\delta||, \arctan \delta_y/\delta_x)$.

\paragraph*{Graph Coarsening.} For a single input point cloud $\mathcal{P}$, a pyramid of downsampled point clouds $\mathcal{P}^{(s)}$ is obtained by the VoxelGrid algorithm~\citep{pclrusu}, which overlays a grid of resolution $r^{(s)}$ over the point cloud and replaces all points within a voxel with their centroid (and thus maintains subvoxel accuracy). Each of the resulting point clouds $\mathcal{P}^{(s)}$ is then independently converted into a graph $G^{(s)}$ and attributes $E^{(s)}$ with neighborhood radius $\rho^{(s)}$ as described above. The pooling map $C^{(s)}$ is defined so that each point in $\mathcal{P}^{(s-1)}$ is assigned to its spatially nearest point in the subsampled point cloud $\mathcal{P}^{(s)}$.

\paragraph*{Data Augmentation.} In order to reduce overfitting on small datasets, we perform online data augmentation. In particular, we randomly rotate point clouds about their up-axis, jitter their scale, perform mirroring, or delete random points.

\subsection{Application in General Graphs} \label{subsec:applgraphs}

Many problems can be modeled directly as graphs. In such cases the graph dataset is already given and only the appropriate graph coarsening scheme needs to be chosen. Without any concept of spatial localization of nodes in general graphs, this is by no means trivial and there exists a large body of literature on this problem, briefly reviewed in Section~\ref{sub:bg_coarsening}. 

Here, we resort to established graph coarsening algorithms and utilize the multiresolution framework of Shuman \etal~\citep{shumanFV16,gspbox}, which works by repeated downsampling and graph reduction of the input graph. The downsampling step is based on splitting the graph into two components by the sign of the largest eigenvector of the Laplacian. This is followed by Kron reduction~\citep{kron}, which also defines new scalar edge attributes, enhanced with spectral sparsification of edges~\citep{spielman2011graph}. Note that the algorithm regards graphs as unweighted for the purpose of coarsening.

This method is attractive for us because of two reasons. Each downsampling step removes approximately half of the nodes, guaranteeing a certain level of pooling strength, and the sparsification step is randomized. The latter property is exploited as a useful data augmentation technique since several different graph pyramids can be generated from a single input graph. This is in spirit similar to the effect of fractional max-pooling~\citep{fractpool}. We do not perform any other data augmentation, as this would need to be domain specific.

\section{Experiments}

The proposed method is evaluated in point cloud classification (real-world data in Section~\ref{subsec:evalpc} and synthetic in \ref{subsec:evalmn}) and on a standard graph classification benchmark (Section~\ref{subsec:evalgg}). In addition, we validate our method and study its properties on MNIST (Section~\ref{subsec:evalmnist}). Finally, we perform several ablation studies in Section~\ref{subsec:eccablation}.

\subsection{Sydney Urban Objects} \label{subsec:evalpc}

\begin{figure}[bt]
\centering
\includegraphics[width=\linewidth]{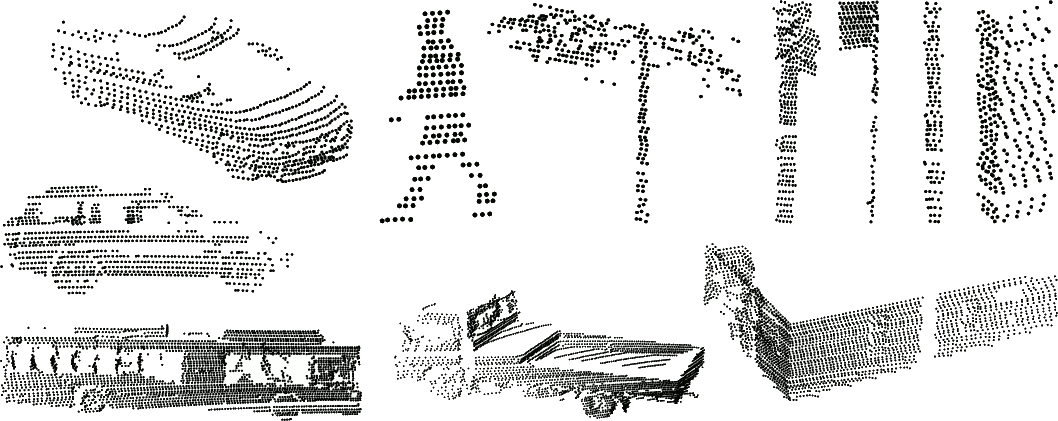}
\vspace{1.5ex}
\caption{\label{fig:sydney} Illustrative samples of the majority of classes in Sydney Urban Objects dataset, reproduced from \citet{trianglesvm}.}
\end{figure}

This point cloud dataset \citep{trianglesvm} consists of 588 objects in 14 categories (vehicles, pedestrians, signs, and trees) manually extracted from 360$^{\circ}$ LiDAR scans, see Figure~\ref{fig:sydney}. It demonstrates non-ideal sensing conditions with occlusions (holes) and a large variability in viewpoint (single viewpoint). This makes object classification a challenging task. 

Following the protocol employed by the dataset authors, we report the mean F1 score weighted by class frequency, as the dataset is imbalanced. This score is further aggregated over four standard training/testing splits.

\paragraph*{Network Configuration.}
Our ECC-network has 7 parametric layers and 4 levels of graph resolution. Its configuration can be described as C(16)-C(32)-MP(0.25,0.5)-C(32)-C(32)-MP(0.75,1.5)-C(64)-MP(1.5,1.5)-GAP-FC(64)-D(0.2)-FC(14), where C($c$) denotes ECC with $c$ output channels followed by affine batch normalization and ReLU activation, MP($r$,$\rho$) stands for max-pooling down to grid resolution of $r$ meters and neighborhood radius of $\rho$ meters, GAP is global average pooling, FC($c$) is fully-connected layer with $c$ output channels, and D($p$) is dropout with probability $p$. The filter-generating networks $w^t$ have configuration FC(16)-FC(32)-FC($d_t d_{t-1}$) with orthogonal weight initialization~\citep{orthoinit} and ReLUs in between. Input graphs are created with $r^0=0.1$ and $\rho^0=0.2$ meters to break overly dense point clusters. Networks are trained with SGD and cross-entropy loss for 250 epochs with batch size 32 and learning rate 0.1 step-wise decreasing after 200 and 245 epochs. Node signal $H^0$ is scalar laser return intensity (0-255), representing depth.

\paragraph*{Results.} 
Table~\ref{tab:respc} compares our result (ECC, 78.4) against two methods based on volumetric CNNs evaluated on voxelized occupancy grids of size $32^3$ (VoxNet \citep{voxnet} 73.0 and ORION \citep{orion} 77.8), which we outperform by a small margin and set the new state of the art result on this dataset.

\begin{table}[bt]
\centering
\begin{tabular}{ccc}
\toprule
Model & Representation & Mean F1\tabularnewline
\midrule
Triangle+SVM \citep{trianglesvm}& image & 67.1 \tabularnewline
GFH+SVM \citep{gfhsvm} & histogram & 71.0 \tabularnewline
VoxNet \citep{voxnet} & volumetric & 73.0 \tabularnewline
ORION \citep{orion} & volumetric & 77.8\tabularnewline
\midrule
ECC & graph & 78.4\tabularnewline
\bottomrule
\end{tabular}
\vspace{1.5ex}
\caption{\label{tab:respc}
Mean F1 score weighted by class frequency on Sydney Urban Objects dataset \cite{trianglesvm}. Only the best-performing models of each baseline are listed.}
\end{table}

\subsection{ModelNet} \label{subsec:evalmn}

ModelNet~\citep{modelnet} is a large scale collection of object meshes. We evaluate classification performance on its subsets ModelNet10 (3991/908 train/test examples in 10 categories) and ModelNet40 (9843/2468 train/test examples in 40 categories). Synthetic point clouds are created from meshes by uniformly sampling 1000 points on mesh faces according to face area (a simulation of acquisition from multiple viewpoints) and rescaled into a unit sphere.

\paragraph*{Network Configuration.}
Our ECC-network for ModelNet10 has 7 parametric layers and 3 levels of graph resolution with configuration C(16)-C(32)-MP(2.5/32,7.5/32)-C(32)-C(32)-MP(7.5/32,22.5/32)-C(64)-GMP-FC(64)-D(0.2)-FC(10), GMP being global max pooling. Other definitions and filter-generating networks $w^t$ are as in Section~\ref{subsec:evalpc}. Input graphs are created with $r^0=1/32$ and $\rho^0=2/32$ units, mimicking the typical grid resolution of $32^3$ in voxel-based methods. The network is trained with SGD and cross-entropy loss for 175 epochs with batch size 64 and learning rate 0.1 step-wise decreasing after every 50 epochs. There is no node signal, \ie $H^0$ are zero. For ModelNet40, the network is wider (C(24), C(48), C(48), C(48), C(96), FC(64), FC(40)) and is trained for 100 epochs with learning rate decreasing after each 30 epochs.

\paragraph*{Results.} 
Table~\ref{tab:resmn} compares our result to several previous works, based either on volumetric \citep{modelnet, voxnet, orion, qi16} or rendered image representation \citep{Su15}, as well as to contemporary methods. Test sets were expanded to include 12 orientations (ECC). We also evaluate voting over orientations (ECC 12 votes), which slightly improves the results likely due to the rotational variance of VoxelGrid algorithm. While not having fully reached the state of the art at the time of submission, our method remained very competitive (90.8\%, resp. 87.4\% mean instance accuracy). 

\begin{table}[bt]
\centering
\resizebox{1\linewidth}{!}{
\begin{tabular}{cccccc}
\toprule
Model & Representation & \multicolumn{2}{c}{ModelNet10} & \multicolumn{2}{c}{ModelNet40} \tabularnewline
\midrule
3DShapeNets \citep{modelnet} & volumetric & 83.5 & & 77.3 & \tabularnewline
MVCNN \citep{Su15} & images & & & 90.1 &\tabularnewline
VoxNet \citep{voxnet} & volumetric & 92 & & 83 &\tabularnewline
ORION \citep{orion} & volumetric & 93.8 & & & \tabularnewline
SubvolumeSup \citep{qi16} & volumetric & & & 86.0 & (89.2) \tabularnewline
** O-CNN \citep{Wang-2017-ocnn} & sparse vol. & & & & (90.6) \tabularnewline
** KdNet \citep{KlokovL17KdNet} & sparse vol. & & (94.0) & & (91.8) \tabularnewline
** PointNet \citep{qi2016pointnet} & point set & & & 86.2 & (89.2) \tabularnewline
** PointNet++ \citep{QiYSG17PointNetPP} & point set & & & & (90.7) \tabularnewline
** DynamicGraph \citep{dgcnn} & graph & & & 90.2 & (92.2) \tabularnewline
\midrule
ECC & graph & 89.3 & (90.0) & 82.4 & (87.0) \tabularnewline
ECC (12 votes) & graph & 90.0 & (90.8) & 83.2 & (87.4)\tabularnewline
\bottomrule
\end{tabular}}
\vspace{1.5ex}
\caption{\label{tab:resmn}
Mean class accuracy (resp. mean instance accuracy) on ModelNets~\citep{modelnet}. Only the best models of each baseline are listed. Concurrent baselines are denoted with a star.}
\end{table}

\subsection{Graph Classification} \label{subsec:evalgg}

We evaluate on a graph classification benchmark frequently used in the community, consisting of five datasets: NCI1, NCI109, MUTAG, ENZYMES, and D{\&}D. Their properties can be found in Table~\ref{tab:stats}, indicating the variability in dataset sizes, in graph sizes, and in the availability of attributes. Following \citet{shervashidze}, we perform 10-fold cross-validation with 9 folds for training and 1 for testing and report the average prediction accuracy.

NCI1 and NCI109~\citep{nci1db} consist of graph representations of chemical compounds screened for activity against non-small cell lung cancer and ovarian cancer cell lines, respectively. MUTAG~\citep{debnath1991structure} is a dataset of nitro compounds labeled according to whether or not they have a mutagenic effect on a bacterium. ENZYMES~\citep{borgwardtK05} contains representations of tertiary structure of 6 classes of enzymes. D{\&}D~\citep{dobson2003} is a database of protein structures (nodes are amino acids, edges indicate spatial closeness) classified as enzymes and non-enzymes. 

\paragraph*{Network Configuration.}
Our ECC-network for NCI1 and NCI109 has 8 parametric layers and 3 levels of graph resolution. Its configuration can be described as C(48)-C(48)-C(48)-MP-C(48)-C(64)-MP-C(64)-GAP-FC(64)-D(0.1)-FC(2), where C($c$) denotes ECC with $c$ output channels followed by affine batch normalization, ReLU activation and dropout (probability 0.05), MP stands for max-pooling onto a coarser graph, GAP is global average pooling, FC($c$) is fully-connected layer with $c$ output channels, and D($p$) is dropout with probability $p$. The filter-generating networks $w^t$ have configuration FC(64)-FC($d_t d_{t-1}$) with orthogonal weight initialization~\cite{orthoinit} and ReLU in between. attributes are encoded as one-hot vectors ($d_0=37$ and $s=4$ due to an extra attribute for self-connections). Networks are trained with SGD and cross-entropy loss for 50 epochs with batch size 64 and learning rate 0.1 step-wise decreasing after 25, 35, and 45 epochs. The dataset is expanded five times by randomized sparsification (Section~\ref{subsec:applgraphs}). 

We made small deviations from this description for the other three datasets  as follows. As MUTAG is a tiny dataset of small graphs, we trained a downsized ECC-network to combat overfitting with configuration C(16)-C(32)-C(48)-MP-C(64)-MP-GAP-FC(64)-D(0.2)-FC(2). Due to higher complexity of ENZYMES we use a wider ECC-network configured as C(64)-C(64)-C(96)-MP-C(96)-C(128)-MP-C(128)-C(160)-MP-C(160)-GAP-FC(192)-D(0.2)-FC(6). Finally, due to large graphs in D{\&}D dataset we designed a ECC-network with more pooling configured as C(48)-C(48)-C(48)-MP-C(48)-MP-C(64)-MP-C(64)-MP-C(64)-MP-C(64)-MP-GAP-FC(64)-D(0.2)-FC(2).

\paragraph*{Baselines.}
We compare our method (ECC) to the state of the art Weisfeiler-Lehman graph kernel \citep{shervashidze} and to four approaches using deep learning as at least one of their components \citep{dcnn, niepert, deepkern, struct2vec, narayanan2016subgraph2vec}. Randomized sparsification used during training time can also be exploited at test time, when the network prediction scores (ECC-5-scores) or votes (ECC-5-votes) are averaged over 5 runs. To judge the influence of edge attributes, we run our method with uniform attributes and $w^t$ being a single layer FC($d_t d_{t-1}$) without bias\footnote{Also possible for unlabeled ENZYMES and D{\&}D, since our method uses attributes from Kron reduction for all coarsened graphs by default.} (ECC no edge attributes). 

\paragraph*{Results.}
Table~\ref{tab:resgraph} conveys that while there is no clear winning algorithm, our method performs at the level of state of the art for edge-labeled datasets (NCI1, NCI109, MUTAG). The results demonstrate the importance of exploiting edge attributes for convolution-based methods, as the performance of DCNN \citep{dcnn} and ECC without edge attributes is distinctly worse, justifying the motivation behind this chapter. Averaging over random sparsifications at test time improves accuracy by a small amount. Our results on datasets without edge attributes (ENZYMES, D{\&}D) are somewhat below the state of the art but still at a reasonable level, though improvement in this case was not the aim of this work. This indicates that further research is needed into the adaptation of CNNs to general graphs. In the following, we discuss the results for each dataset in more detail.

\textit{NCI1.} ECC (83.80\%) performs distinctly better than convolution methods that are not able to use edge attributes (DCNN \citep{dcnn} 62.61\%, PSCN \citep{niepert} 78.59\%). Methods not approaching the problem as convolutions on graphs but rather combining deep learning with other techniques are stronger (subgraph2vec \citep{narayanan2016subgraph2vec} 78.05\%, Deep WL \citep{deepkern} 80.31\%, structure2vec \citep{struct2vec} 83.72\%) but are still outperformed by ECC. While the Weisfeiler-Lehman graph kernel remains the strongest method (WL \citep{shervashidze} 84.55\%), it is fair to conclude that ECC, structure2vec, and WL perform at the same level.

\textit{NCI109.} ECC (82.14\%) performs distinctly better than DCNN \citep{dcnn} (62.86\%), which is not able to use edge attributes, and is on par with non-convolutional approaches (subgraph2vec \citep{narayanan2016subgraph2vec} 78.39\%, Deep WL \citep{deepkern} 80.32\%, structure2vec \citep{struct2vec} 82.16\%, WL \citep{shervashidze} 84.49\%).

\textit{MUTAG.} While by numbers ECC (89.44\%) outperforms all other approaches except of PSCN \citep{niepert} (92.63\%), we note that all four leading methods (subgraph2vec \citep{narayanan2016subgraph2vec} 87.17\%, Deep WL \citep{deepkern} 87.44\%, structure2vec \citep{struct2vec} 88.28\%, ECC, PSCN) can be seen to perform equally well due to fluctuations caused by the dataset size. We account the tiny decrease in performance with test-time randomization (88.33\%) to the same reason.

\textit{ENZYMES.} As this dataset is not edge-labeled, we do not expect to obtain the best performance. Indeed, our method (53.50\%) performs at the level of Deep WL \citep{deepkern} (53.43\%) and is overperformed by WL \citep{shervashidze} (59.05\%) and structure2vec \citep{struct2vec} (61.10\%). Note that the gap to the other convolution-based method DCNN \citep{dcnn} (18.10\%) is huge and there is an improvement of more than 4 percentage points due to edge attributes in coarser graph resolutions from Kron reduction.

\textit{D{\&}D.} As this dataset is also not edge-labeled, we do not expect to obtain the best performance. Our method (74.10\%) is overperformed by the others who evaluated on this dataset (PSCN \citep{niepert} 77.12\%, WL \citep{shervashidze} 79.78\%, structure2vec \citep{struct2vec} 82.22\%), though the margin is not very large.

\begin{table}[bt]
\centering
\addtolength{\tabcolsep}{-3pt}
\begin{tabular}{cccccc}
\toprule
 & NCI1 & NCI109 & MUTAG & ENZYMES & D{\&}D \tabularnewline
\midrule
\# graphs & 4110 & 4127 & 188 & 600 & 1178\tabularnewline
mean $|V|$ & 29.87 & 29.68 & 17.93 & 32.63 & 284.32 \tabularnewline
mean $|E|$ & 32.3 & 32.13 & 19.79 & 62.14 & 715.66 \tabularnewline
\# classes & 2 & 2 & 2 & 6 & 2\tabularnewline
\# node attributes & 37 & 38 & 7 & 3 & 82\tabularnewline
\# edge attributes & 3 & 3 & 11 & {\textemdash} & {\textemdash}\tabularnewline
\bottomrule
\end{tabular}
\addtolength{\tabcolsep}{3pt}
\vspace{1.5ex}
\caption{\label{tab:stats}
Characteristics of the graph benchmark datasets, extended from \citep{struct2vec}. Both edge and node attributes are categorical.}
\end{table}

\begin{table}[bt]
\centering
\addtolength{\tabcolsep}{-3pt}
\resizebox{1\linewidth}{!}{
\begin{tabular}{cSSSSS}
\toprule
Model & {NCI1} & {NCI109} & {MUTAG} & {ENZYMES} & {D{\&}D} \tabularnewline
\midrule
DCNN \citep{dcnn} & 62.61 & 62.86 & 66.98 & 18.10 & {\textemdash} \tabularnewline
subgraph2vec \citep{narayanan2016subgraph2vec} & 78.05 & 78.39 & 87.17 & {\textemdash} & {\textemdash}\tabularnewline
PSCN \citep{niepert} & 78.59 & {\textemdash} & 92.63 & {\textemdash} & 77.12 \tabularnewline
Deep WL \citep{deepkern} & 80.31 & 80.32 & 87.44 & 53.43 & {\textemdash} \tabularnewline
structure2vec \citep{struct2vec} & 83.72 & 82.16 & 88.28 & 61.10 & 82.22\tabularnewline
WL \citep{shervashidze} & 84.55 & 84.49 & 83.78 & 59.05 & 79.78\tabularnewline
** CCN \citep{kondorCCN} & 76.27 & 75.54 & 91.64 & {\textemdash} & {\textemdash}\tabularnewline
** DGCNN \citep{ZhangCNC18} & 74.44 & {\textemdash} & 85.83 & {\textemdash} & 79.37 \tabularnewline
\midrule
ECC (no edge attributes)& 76.82 & 75.03 & 76.11 & 45.67 & 72.54 \tabularnewline
ECC 				& 83.80 & 81.87 & 89.44 & 50.00 & 73.65 \tabularnewline
ECC (5 votes) 		& 83.63 & 82.04 & 88.33 & 53.50 & 73.68 \tabularnewline
ECC (5 scores)		& 83.80 & 82.14 & 88.33 & 52.67 & 74.10 \tabularnewline
\bottomrule
\end{tabular}}
\addtolength{\tabcolsep}{3pt}
\vspace{1.5ex}
\caption{\label{tab:resgraph}
Mean accuracy (10 folds) on graph classification datasets. Only the best-performing models of each baseline are listed. Baselines newer than our work are denoted with a star.}
\end{table}

\subsection{MNIST} \label{subsec:evalmnist}

To further validate our method, we applied it to the MNIST classification problem \citep{mnist}, a dataset of 70k greyscale images of handwritten digits represented on a 2D grid of size 28$\times$28. We regard each image $I$ as point cloud $\mathcal{P}$ with points $p=(x,y,0)$ and signal $F_P(p)=I(x,y)$ representing each pixel, $x,y\in\{0,..,27\}$. Edge labeling and graph coarsening is performed as explained in Section~\ref{subsec:applclouds}. 
We are mainly interested in two questions: Is ECC able to reach the standard performance on this classic baseline? What kind of representation does it learn?

\paragraph*{Network Configuration.}
Our ECC-network has 5 parametric layers with configuration C(16)-MP(2,3.4)-C(32)-MP(4,6.8)-C(64)-MP(8,30)-C(128)-D(0.5)-FC(10); the notation and filter-generating network being as in Section~\ref{subsec:evalpc}. The last convolution has a stride of 30 and thus maps all $4\times 4$ points to only a single point. Input graphs are created with $r^0=1$ and $\rho^0=2.9$. This model exactly corresponds to a regular CNN with three convolutions with filters of size 5$\times$5, 3$\times$3, and 3$\times$3 interlaced with max-poolings of size 2$\times$2, finished with two fully connected layers. Networks are trained with SGD and cross-entropy loss for 20 epochs with batch size 64 and learning rate 0.01 step-wise decreasing after 10 and 15 epochs.

\paragraph*{Results.} Table~\ref{tab:resmnist} proves that our ECC network can achieve the level of quality comparable to the good standard in the community (99.14). This is exactly the same accuracy as reported by \citet{defferrard16} and better than what is offered by other spectral-based approaches (98.2 \citep{bruna13}, 94.96 \citep{edwards16}). Note that we are not aiming at becoming the state of the art on MNIST by this work. 

Next, we investigate the effect of regular grid and irregular mesh. To this end, we discard all black points $i: F_i=0$) from the point clouds, corresponding to 80.9\% of data, and retrain the network (ECC sparse input). Exactly the same test performance is obtained (99.14), indicating that our method is very stable with respect to graph structure changing from sample to sample.

Furthermore, we check the quality of the learned filter generating networks $w^t$. We compare with ECC configured to mimic regular convolution using single-layer filter networks and one-hot encoding of offsets (ECC one-hot), as described in Section~\ref{subsec:relgridconv}. This configuration reaches 99.37 accuracy, or 0.23 more than ECC, implying that $w^t$ are not perfect but still perform very well in learning the proper partitioning of edge attributes.

\begin{figure}[bt]
\centering
\includegraphics[width=0.4\linewidth]{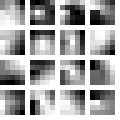}
\includegraphics[width=0.4\linewidth]{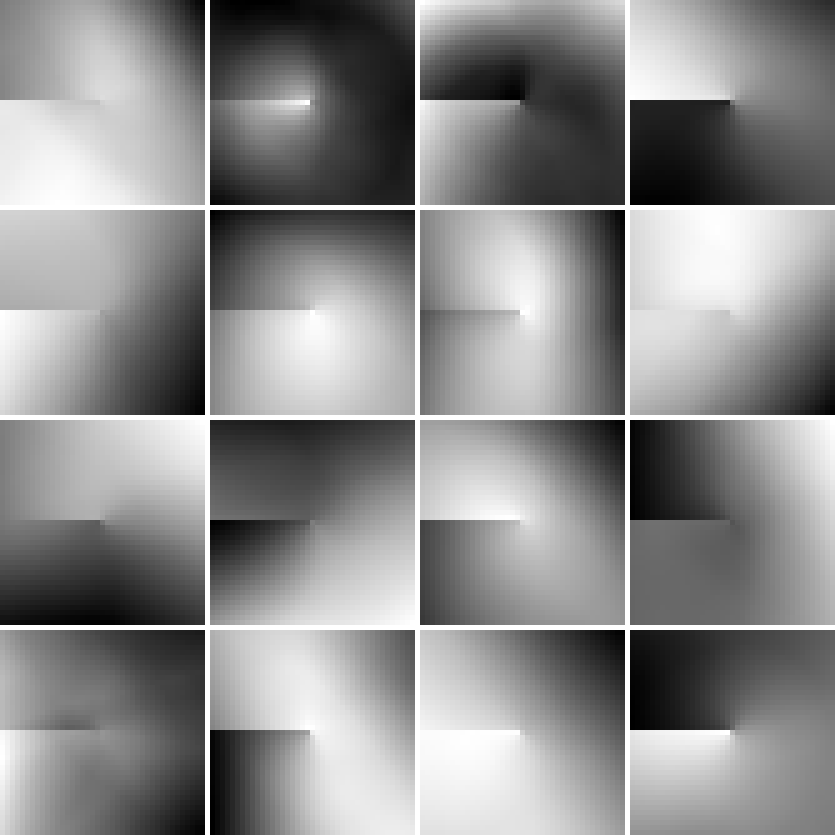}
\vspace{1.5ex}
\caption{\label{fig:filters} Convolutional filters learned on MNIST in the first layer for sparse input ECC, sampled in two different resolutions. See Section~\ref{subsec:evalmnist} for details.}
\end{figure}

Last, we explore the generated filters visually for the case of the sparse input ECC. As filters $W^0 \in \mathbb{R}^{16\times 1}$ are a continuous function of an edge attribute, we can visualize the change of values in each dimension in 16 images by sampling attributes over grids of two resolutions. The coarser one in Figure~\ref{fig:filters} has integer steps corresponding to the offsets $\delta_x,\delta_y\in\{-2,..,2\}$. It shows filters exhibiting the structured patterns typically found in the first layer of CNNs. The finer resolution in Figure~\ref{fig:filters} (sub-pixel steps of 0.1) reveals that the filters are in fact smooth and do not contain any discontinuities apart from the angular artifact due to the $2\pi$ periodicity of azimuth. Interestingly, the artifact is not distinct in all filters, suggesting the network may learn to overcome it if necessary.

\begin{table}[bt]
\centering
\begin{tabular}{ccc}
\toprule
Model & Train accuracy & Test accuracy\tabularnewline
\midrule
ECC & 99.12 & 99.14 \tabularnewline
ECC (sparse input) & 99.36 & 99.14 \tabularnewline
ECC (one-hot) & 99.53 & 99.37\tabularnewline
\bottomrule
\end{tabular}
\vspace{1.5ex}
\caption{\label{tab:resmnist} Accuracy on MNIST dataset \citep{mnist}.}
\end{table}

\subsection{Detailed Analyses and Ablations} \label{subsec:eccablation}

This section provides analysis of several design choices and investigates robustness of point cloud classification to noise. In the second part, we explore two extensions of our ECC formulation, specifically with degree attributes and with a learned normalization factor.

\paragraph*{Neighborhood Radius.}
In Table~\ref{tab:res_radius} we study the dependence on convolution radii $\rho$: increasing them $1.5\times$ or $2\times$ in all convolutional layers leads to a drop in performance in Sydney dataset, which would correspond to a preference of using smaller filters in regular CNNs. The average neighborhood size is roughly 10 nodes for our best-performing network.
We hypothesized that larger radii smooth out the information in the central node. 
To investigate this, we increased the importance of the self-loop by adding an identity skip-connection (ECC-id) and retrained the networks. Indeed, stronger identity connection allowed for successful integration of a larger context, up to some limit, which suggests that information should be aggregated neither too much nor too little.

\begin{table}[bt]
\centering
\begin{tabular}{cc}
\toprule
Model & Mean F1\tabularnewline
\midrule
ECC $2\rho$ & 74.4 \tabularnewline
ECC $1.5\rho$ & 76.9 \tabularnewline
ECC & 78.4\tabularnewline
ECC-id $2\rho$ & 77.4 \tabularnewline
ECC-id $1.5\rho$ & 79.5 \tabularnewline
ECC-id & 77.0\tabularnewline
\bottomrule
\end{tabular}
\vspace{1.5ex}
\caption{\label{tab:res_radius}
Influence of varied neighborhood radius on Sydney Urban Objects dataset \cite{trianglesvm}.}
\end{table}

\paragraph*{Identity Connections} \label{sec:idconn}
We compare the performance of ECC (quation~\ref{eq:C1}) and ECC-id (Equation~\ref{eq:C1}) in Table~\ref{tab:idconntable}. With two exceptions (NCI109 and ENZYMES), ECC does not benefit from identity connections in the specific network configurations. The trend may be different for other configurations, \eg ECC $1.5\rho$ improved from 76.9 to 79.5 mean F1 score on Sydney due to identity connections as mentioned above.

\begin{table}[bt]
\centering
\resizebox{1\linewidth}{!}{%
\begin{tabular}{cccccccc}
\toprule
 & NCI1 & NCI109 & MUTAG & ENZYMES & D{\&}D & Sydney & ModelNet10\tabularnewline
\midrule
ECC-id & 83.24  & \emph{81.97 } & 85.56  & \emph{51.83 } & 70.48 & 77.0 & 88.5 (89.3) \tabularnewline
ECC & 83.80  & 81.87  & 89.44  & 50.00  & 73.65  & 78.4  & 89.3 (90.0) \tabularnewline
\bottomrule
\end{tabular}%
}
\vspace{1ex}
\caption{\label{tab:idconntable}
The effect of adding identity connections (improvements in italics). Performance metrics vary and are specific to each dataset, as introduced in the respective sections.}
\end{table}

\paragraph*{Edge Attributes for Point Clouds} \label{sec:pclabels}
In Section \ref{subsec:applclouds} we defined edge attributes $E_{j,i}$ as the offset $\delta=P_j-P_i$ expressed in Cartesian and spherical coordinates. Here, we explore the importance of individual elements in the proposed edge labeling and further evaluate attributes invariant to rotation about objects' vertical axis $z$ (IRz). Table~\ref{tab:edgelimp} conveys that models with isotropic (60.7) or no attributes (38.9) perform poorly as expected, while either of the coordinate systems is important. IRz labeling performs comparably or even slightly better than our proposed one. However, we believe this is a property of the specific dataset and may not necessarily generalize, an example being MNIST, where IRz is equivalent to full isotropy and decreases accuracy to 89.9\%.

\begin{table}[bt]
\centering
\begin{tabular}{cc}
\toprule
attribute $E_{j,i}$ & Mean F1\tabularnewline
\midrule
$(\delta_x, \delta_y, \delta_z, ||\delta||, \arccos \delta_z/||\delta||, \arctan \delta_y/\delta_x)$ & 78.4 \tabularnewline
$(\delta_x, \delta_y, \delta_z)$ & 76.1 \tabularnewline
$(||\delta||, \arccos \delta_z/||\delta||, \arctan \delta_y/\delta_x)$ & 77.3 \tabularnewline
\midrule 
$(||\delta_{xy}||, \delta_z, ||\delta||, \arccos \delta_z/||\delta||)$ & 75.8 \tabularnewline
$(||\delta_{xy}||, \delta_z)$ & 78.2 \tabularnewline
$(||\delta||, \arccos \delta_z/||\delta||)$ & 78.7 \tabularnewline
\midrule
$(||\delta||)$ & 60.7 \tabularnewline
$(0)$ & 38.9 \tabularnewline
\bottomrule
\end{tabular}
\vspace{1ex}
\caption{\label{tab:edgelimp}
ECC on Sydney dataset with varied edge attribute definition.}
\end{table}

\begin{figure}[bt]
\centering
\includegraphics[width=0.48\linewidth]{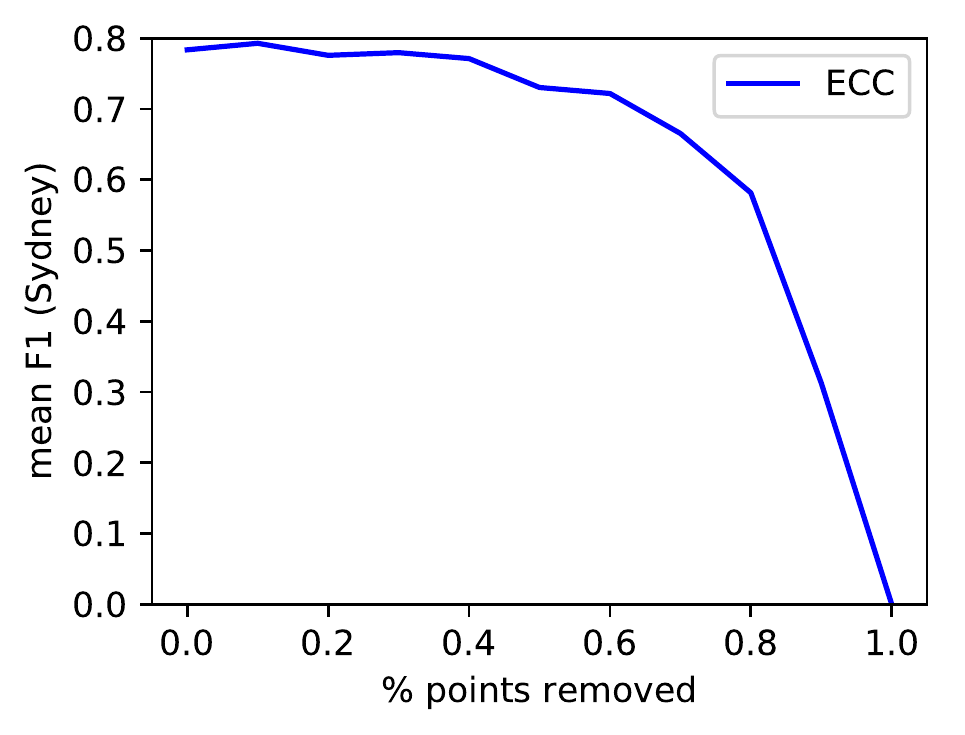}
\includegraphics[width=0.48\linewidth]{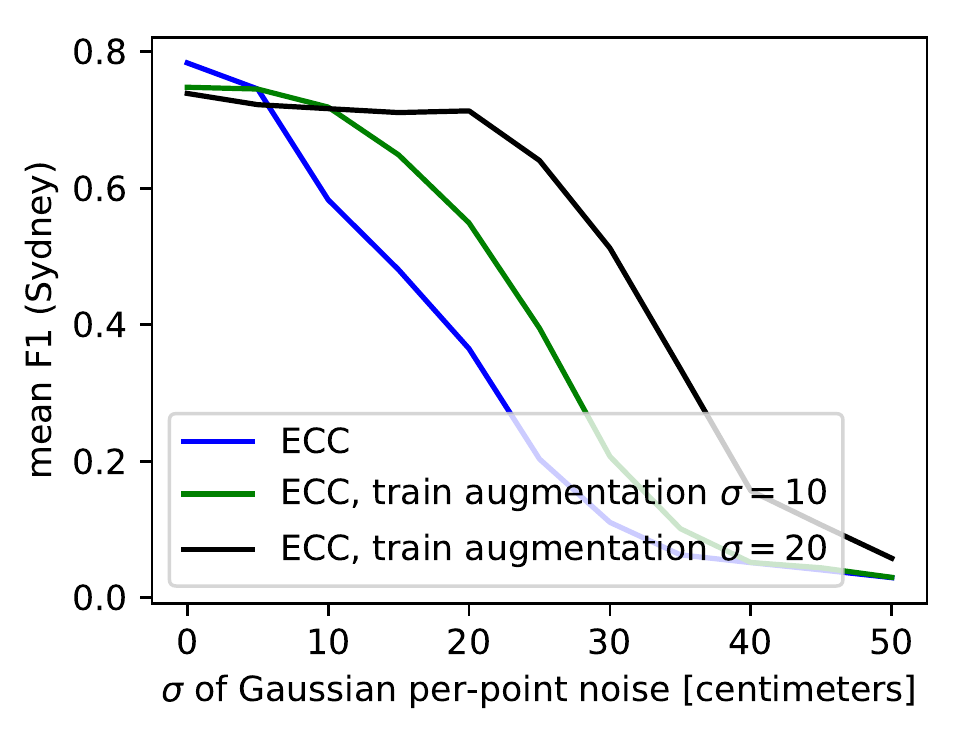}
\caption{\label{fig:robutness}
Robustness to point removal and Gaussian noise on Sydney dataset.}
\end{figure}

\paragraph*{Robustness to Noise} \label{sec:pcnoise}
Real-world point clouds contain several kinds of artifacts, such as holes due to occlusions and Gaussian noise due to measurement uncertainty. Figure~\ref{fig:robutness} shows that ECC is highly robust to point removal and can be made robust to additive Gaussian noise by a proper training data augmentation.

\paragraph*{Node Degrees in Edge Attributes} \label{sec:degrees}
In the task of graph classification, we used categorical attributes (if present) encoded as one-hot vectors for edges in the input graph and scalars computed by Kron reduction for edges in all coarsened graphs. Here we investigate making the edge attributes more informative by including the degrees of the pair of nodes forming an edge. The degree information is implicitly used by spectral convolution methods, as the degree information is contained in the graph Laplacian, and also appears in the explicit propagation rules \citep{kipf, dcnn}. 

Our model can be easily extended to make use of this information by simply appending it to the existing edge attribute vectors. We consider four variants of providing additional degree attributes $L_{deg}(j)$ and  $L_{deg}(i)$ about a directed edge $(j,i)$: $L_{deg}(i) = 1/\sqrt{\mathrm{deg}(i)}$, $L_{deg}(i) = 1/\mathrm{deg}(i)$, $L_{deg}(i) = \sqrt{\mathrm{deg}(i)}$, and $L_{deg}(i) = \mathrm{deg}(i)$, where $\mathrm{deg}(i) = |\mathcal{N}(i)|$ is the degree of node $i\in \mathcal{V}$. We use these additional attributes in all graph resolutions.

Table~\ref{tab:degrees} reveals that degree information can improve the results considerably, especially for datasets without given edge attributes (by up to 5 percentage points for ENZYMES and up to 2.14 percentage points for D{\&}D). However, no variant of $L_{deg}(i)$ can guarantee consistent improvement over all datasets.

\begin{table}[bt]
\centering
\begin{tabular}{cccccc}
\toprule
 & NCI1 & NCI109 & MUTAG & ENZYMES & D{\&}D \tabularnewline
\midrule
$L_{deg}(i) = 1/\sqrt{\mathrm{deg}(i)}$ & 82.99  & 81.94  & 87.78  & \emph{53.67 } & 73.65 \tabularnewline
$L_{deg}(i) = 1/\mathrm{deg}(i)$ & 83.60  & \emph{82.40 } & 88.89  & \emph{52.67 } & 71.77 \tabularnewline
$L_{deg}(i) = \sqrt{\mathrm{deg}(i)}$ & 83.58  & \emph{82.28 } & 86.67  & \emph{55.00 } & \emph{75.79}\tabularnewline
$L_{deg}(i) = \mathrm{deg}(i)$ & 83.16  & \emph{83.03 } & 86.67  & \emph{52.83 } & \emph{73.74 }\tabularnewline
\midrule
ECC without $L_{deg}(i)$ & 83.80  & 81.87  & 89.44  & 50.00  & 73.65 \tabularnewline
\bottomrule
\end{tabular}
\vspace{1ex}
\caption{\label{tab:degrees}
The effect in mean classification accuracy of adding node degrees to edge attributes (improvements in italics).}
\end{table}

\paragraph*{Node Degrees in Normalization} \label{sec:normnet}
The formulation of ECC in Equation~\ref{eq:C1} performs normalization by the neighborhood size. Here we explore learning an additional multiplicative factor, conditioned on the neighborhood size $1/|\mathcal{N}(i)|$. The motivation is to investigate whether modeling a relation between neighborhood size and feature magnitude (such as a smooth transition from a sum to an average) could improve predictive performance. To this end, we again make use of Dynamic filter networks~\citep{dfn16} and design a factor-generating network $f: \mathbb{R} \to \mathbb{R}$ which given node degree $\mathrm{deg}(i) = |\mathcal{N}(i)|$ outputs a node-specific normalization factor. We formulate ECC-f as follows:

\begin{equation}
H_i^{t+1} = \mathrm{ReLU}\left(\frac{f^t(|\mathcal{N}(i)|;\theta^t)}{|\mathcal{N}(i)\cup\{i\}|} \sum_{j\in \mathcal{N}(i)\cup\{i\}} w^t(E_{j,i};\theta^t) H_j^{t} + \mathbf{b}^t\right) \label{eq:CZ}
\end{equation}

In our experiments, the factor-generating networks $f^t$ have configuration FC(32)-FC(1) with orthogonal weight initialization~\citep{orthoinit} and ReLUs in between. 

The results in Table~\ref{tab:normnet} show that while being helpful on some datasets (NCI109, ENZYMES, ModelNet10), ECC-f harms the performance on the other ones. Embedding node information in attributes instead seems to achieve higher performance, see above.

\begin{table}[bt]
\centering
\resizebox{1\linewidth}{!}{%
\begin{tabular}{cccccccc}
\toprule
 & NCI1 & NCI109 & MUTAG & ENZYMES & D{\&}D & Sydney & ModelNet10\tabularnewline
\midrule
ECC-f & 83.48  & \emph{82.57} & 86.67  & \emph{52.50} & 72.03 & 75.5 & \emph{89.9 (90.6)} \tabularnewline %
ECC & 83.80  & 81.87  & 89.44  & 50.00  & 73.65  & 78.4  & 89.3 (90.0) \tabularnewline
\bottomrule
\end{tabular}%
}
\vspace{1ex}
\caption{\label{tab:normnet}
The effect of adding a learned normalization factor (improvements in italics). Performance metrics vary and are specific to each dataset, as introduced in the respective sections.}
\end{table}

\section{Discussion}

In this section, we take a critical viewpoint and discus several limitations of our work as well. 

\paragraph*{Memory requirements.} While the possibility to work with continuous attributes in ECC is very convenient in many problems, the model may become quite GPU memory demanding, especially during training when all intermediate activations need to be stored (\eg around 9 GB for our ModelNet10 network). This is particularly valid for point clouds, where nearly all edges have their unique edge attributes in practice and thus also their unique filter weights $W$. 

There are at least two options for addressing this problem. From a modeling perspective, one can decrease the number of parameters in generated weights $W$. One way is to restrict $W$ to low rank $q$ by generating $W_a \in \mathbb{R}^{q\times d^{t-1}}$ and $W_b \in \mathbb{R}^{d^t\times q}$ and computing $W H_j^{t} = W_b(W_a H_j^{t})$, which is more efficient as long as $q(d^t+d^{t-1}) < d^t d^{t-1}$. Unfortunately, we have seen considerably worse performance in practice. It turned out that a better way in terms of both performance and memory is to make $W$ diagonal, effectively replacing matrix-vector multiplication with element-wise multiplication; we discuss this extension in more detail farther in Chapter~\ref{chap:cvpr18}.

From an engineering perspective, we experimented with clustering of edge attributes during training with K-means. This can be done independently for each training sample offline or even online in each iteration. Unlike simple predefined quantization, this approach can maintain the original empirical distribution of attributes over the dataset and also doubles as data augmentation. In spite of that, we observed a small decrease in performance.

In general, the overhead in terms of memory and computation together with no distinct benefit in predictive performance is likely the main reason why concurrently introduced set-based approaches such as PointNet~\citep{qi2016pointnet} have become more popular in the community for simple, small-scale point clouds. In such cases, modeling nearest neighbor structures using graphs may indeed be a bit of an overkill. However, in Chapter~\ref{chap:cvpr18} we demonstrate a major advantage of using explicit spatial relationships for modeling large-scale point clouds.

\paragraph*{Feed-forward update function.} In this Chapter, we were using simple update functions $u_t: \mathrm{ReLU}(M_i^{t+1})$ and $u_t: \mathrm{ReLU}(\mathrm{id}(H_i^{t}) + M_i^{t+1})$. While being consistent with many popular feed-forward architectures for images, the community has drawn inspiration from recurrent networks and often adopted gated update functions \citep{yujia16,GilmerSRVD17,SchuttKFCTM17}, which should provide increased protection against oversmoothing information within neighborhoods. In Chapter~\ref{chap:cvpr18}, we also adopt this view.

\paragraph*{Rotation variance.} Our definitions of edge attributes for point clouds bind together the degree of (in)variance to local and global transformations, in particular to rotations. Instead, very often, the ideal would be to remain equivariant for reasoning about local spatial relationships and obtain invariance only at the global scale. Popular remedies include data augmentation, as e.g. in this thesis\footnote{We also performed preliminary experiment finding a local coordinate system in each neighborhood with principal component analysis of point coordinates but were not able to achieve reasonable results.}, or spatial transformer networks~\citep{JaderbergSTN}, as e.g. in PointNet~\citep{qi2016pointnet}. Solving the problem of equivariance  efficiently and in a principled way is an active research topic both in sparse and dense representations and in 2D and 3D - see \citet{nBodyNets,tensorFieldNets} for very recent methods targeting the point cloud domain.
\section{Conclusion}

We have introduced edge-conditioned convolution (ECC), an operation on graph node signal performed in the spatial domain where filter weights are conditioned on edge attributes and dynamically generated for each specific input sample. We have shown that our formulation generalizes the standard convolution on graphs if edge attributes are chosen properly and experimentally validated this assertion on MNIST. We applied our approach to point cloud classification in a novel way, setting a new state of the art performance on Sydney dataset. Furthermore, we have outperformed other deep learning-based approaches on graph classification dataset NCI1. The source code has been published at \url{https://github.com/mys007/ecc}.

In the next chapter, we integrate ECC into a recurrent network, investigate a reduction of its memory and computational requirements, and apply it to node-wise prediction task rather than graph classification tasks as in this chapter.

\chapter{Large-scale Point Cloud Segmentation} \label{chap:cvpr18}

\newcommand*{\bbR}{\mathbb{R}}
\newcommand*{\cE}{\mathcal{E}}
\newcommand*{\cS}{\mathcal{S}}
\newcommand*{\cG}{\mathcal{G}}
\newcommand*{\cL}{\mathcal{L}}
\newcommand*{\cN}{\mathcal{N}}

\renewcommand*{\Pa}[1]{\left(#1\right)}
\newcommand*{\Cur}[1]{\left\{#1\right\}}
\newcommand*{\Bra}[1]{\left[#1\right]}
\newcommand*{\Norm}[1]{\left\lVert#1\right\rVert}
\newcommand*{\Abs}[1]{\left\lvert#1\right\rvert}
\newcommand*{\upp}[2]{#1^{\Pa{#2}}}
\newcommand*{\argmin}{\mathrm{arg}\,\mathrm{min}}

\renewcommand*{\eqref}[1]{Equation~\hyperref[#1]{\ref*{#1}}}
\newcommand*{\figref}[1]{Figure~\hyperref[#1]{\ref*{#1}}}
\newcommand*{\tabref}[1]{Table~\hyperref[#1]{\ref*{#1}}}
\newcommand*{\Subref}[1]{\hyperref[#1]{(\subref*{#1})}}
\newcommand*{\secref}[1]{Section~\hyperref[#1]{\ref*{#1}}}

\newcommand*{\Gvor}{G_{\text{vor}}}
\newcommand*{\Evor}{\mathcal{E}_{\text{vor}}}
\newcommand*{\Gnn}{G_{\text{nn}}}
\newcommand*{\Enn}{\mathcal{E}_{\text{nn}}}
\newcommand*{\dmea}{\delta_\text{mean}}
\newcommand*{\dstd}{\delta_\text{std}}
\newcommand*{\dnorm}{\delta_\text{norm}}
\newcommand*{\dacos}{\delta_\text{acos}}
\newcommand*{\datan}{\delta_\text{atan}}
\newcommand*{\Length}[1]{\text{length}\Pa{#1}}
\newcommand*{\Surface}[1]{\text{surface}\Pa{#1}}
\newcommand*{\Volume}[1]{\text{volume}\Pa{#1}}
\newcommand*{\Lin}{\text{Lin}}
\newcommand*{\Pla}{\text{Pla}}
\newcommand*{\Sca}{\text{Sca}}
\newcommand*{\Ver}{\text{Ver}}
\newcommand*{\Ele}{\text{Ele}}

\newcommand*{\Gruh}[2]{H_{#2}^{#1}}
\newcommand*{\Gruhn}[1]{H_{i}^{t,#1}}
\newcommand*{\Gruu}{U_i^{t}}
\newcommand*{\Grur}{R_i^{t}}
\newcommand*{\Grux}{X_i^{t}}
\newcommand*{\Gruxn}[1]{X_{i}^{t,#1}}
\newcommand*{\Gruq}{Q_i^{t}}
\newcommand*{\Grum}{M_i^{t}}
\newcommand*{\Gruy}{Y_i}
\newcommand*{\Emb}{Z_i}

\def\etal{{et al.}~}
\def\eg{\textit{e.g.}~}
\def\Eg{\textit{E.g.}~}
\def\ie{\textit{i.e.}~}
\def\cf{\textit{c.f.}~}
\def\wrt{{w.r.t.}~}

\newcolumntype{C}[1]{>{\centering\arraybackslash}p{#1}}

\newbox\mybox
\def\centerfigure#1{%
    \setbox\mybox\hbox{#1}%
    \raisebox{-0.5\dimexpr\ht\mybox+\dp\mybox}{\copy\mybox}%
}

\section{Introduction}

Visual understanding of 3D environment is a fundamental requirement for agents acting in the real world, allowing for applications in autonomous driving and robotics, perception assistance tools or augmented reality. Point cloud representation frequently arises in such systems, usually originating in dedicated sensors, such as LiDAR and other active remote sensing devices, or coming from multi-view / structure-from-motion reconstruction methods. However, analysis of large 3D point clouds presents numerous challenges, the most obvious one being the scale of the data. Another hurdle is the lack of clear structure akin to the regular grid arrangement in images, especially in cases when the data has been fused from multiple sensors or the position of the sensor is uncertain.

Previous attempts at using deep learning for large 3D data were trying to replicate successful CNN architectures used for image segmentation. For example, SnapNet \citep{boulch2017unstructured} converts a 3D point cloud into a set of virtual 2D color and depth (RGB-D) snapshots, the semantic segmentation of which can then be projected on the original data. SegCloud \citep{tchapmi2017segcloud} uses 3D convolutions on a regular voxel grid. However, we argue that such methods do not capture the inherent structure of 3D point clouds, which results in limited discrimination performance. Indeed, converting point clouds to 2D format comes with loss of information and requires to perform surface reconstruction, a problem arguably as hard as semantic segmentation. Volumetric representation of point clouds is inefficient and tends to discard small details, as discussed in Section~\ref{sec:ecc-related} before.

Deep learning architectures specifically designed for 3D point clouds \citep{qi2016pointnet, Riegler2017OctNet,QiYSG17PointNetPP,Engelmann17_3dsemseg}, including our work presented in Chapter~\ref{chap:cvpr17}, display good results but are limited by the size of inputs they can handle at once.

\begin{figure*}[t!]
\begin{tabular}{cc}
	\begin{subfigure}[b]{0.48\textwidth}
		\includegraphics[width=1\textwidth]{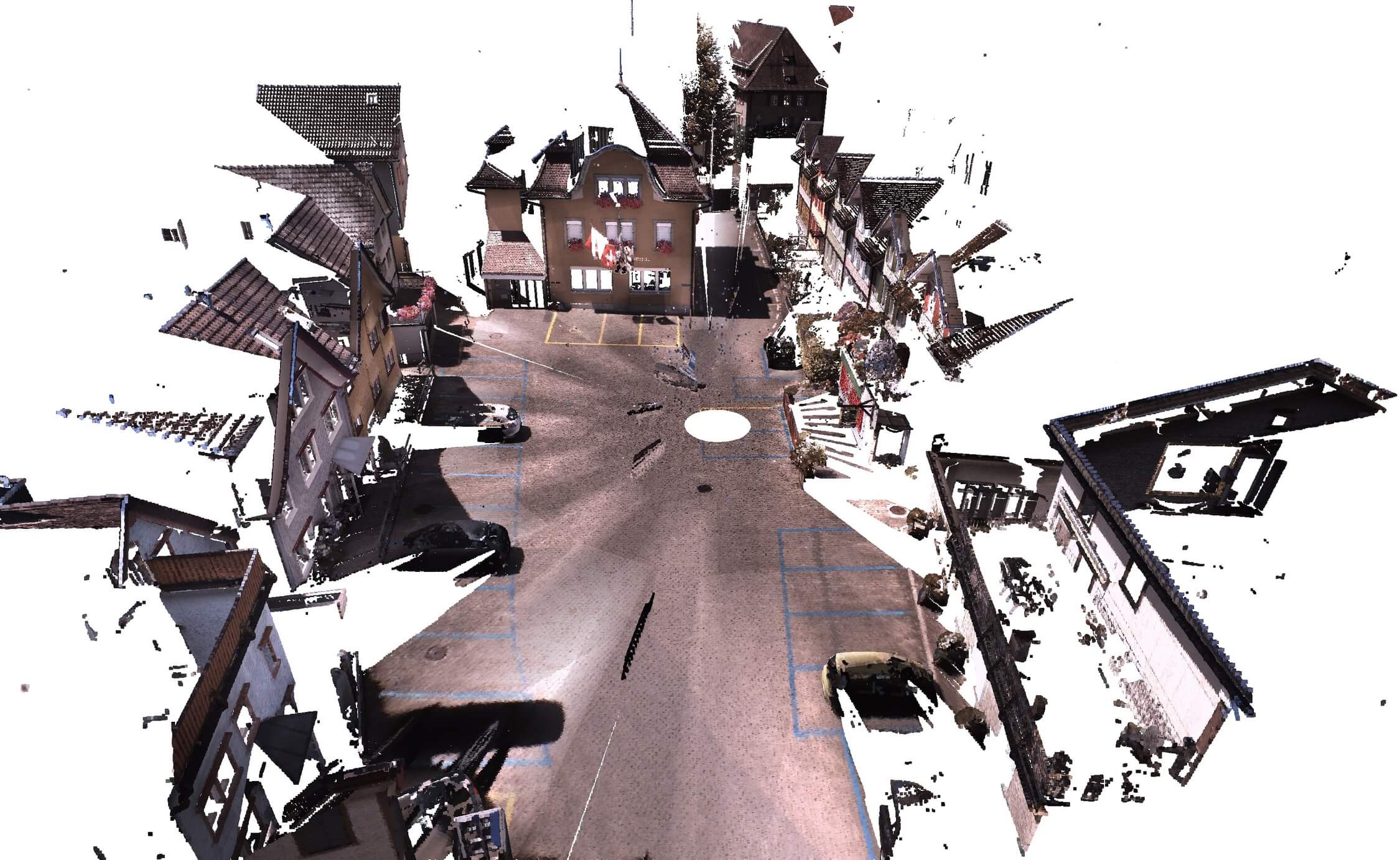}
        \caption{RGB point cloud}
        \label{fig:illu_rgb}
	\end{subfigure}&
    \begin{subfigure}[b]{0.48\textwidth}
		\includegraphics[width=1\textwidth]{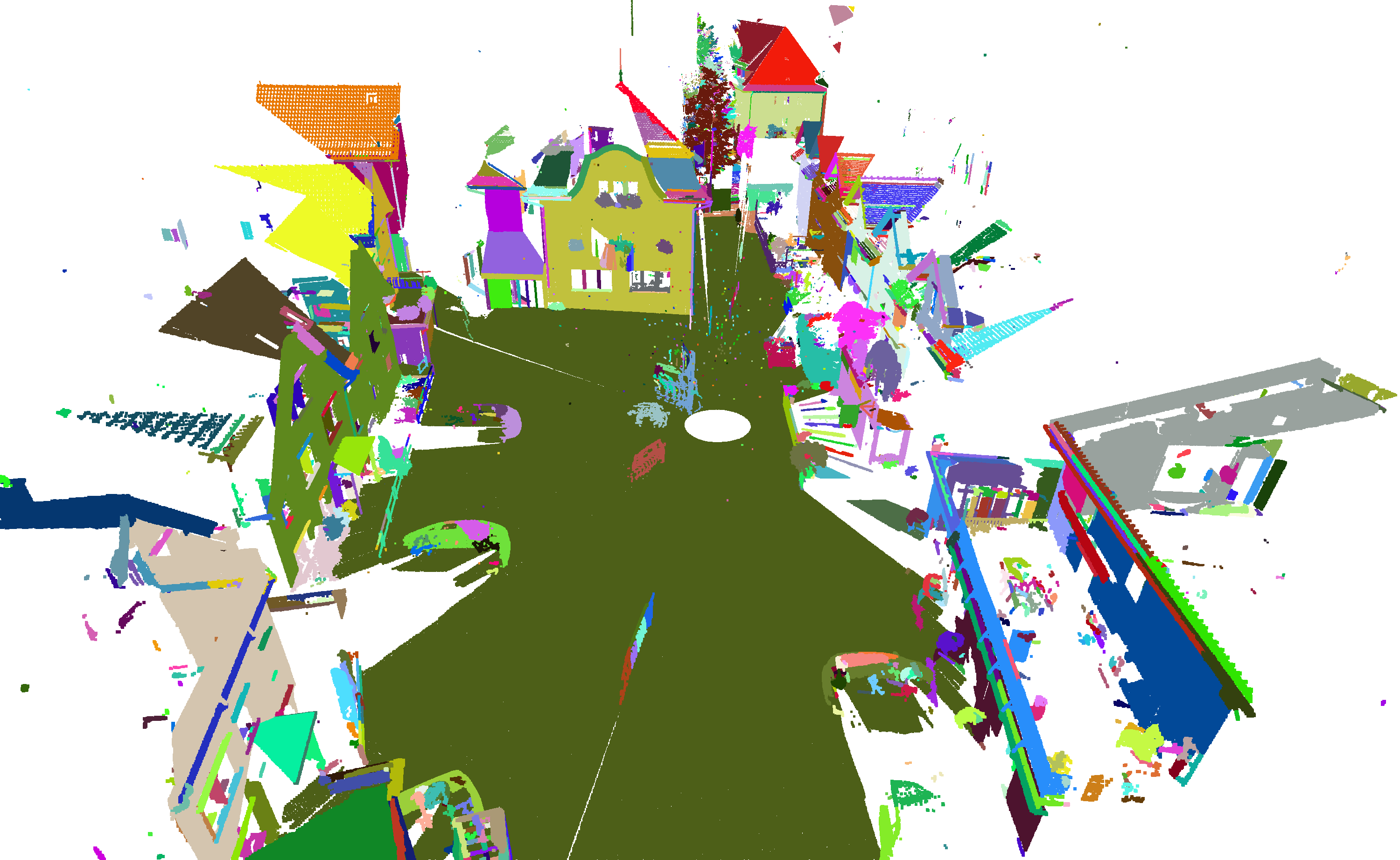}
        \caption{Geometric partition}
        \label{fig:illu_seg}
	\end{subfigure}\\
     \begin{subfigure}[b]{0.48\textwidth}   
		\includegraphics[width=1\textwidth]{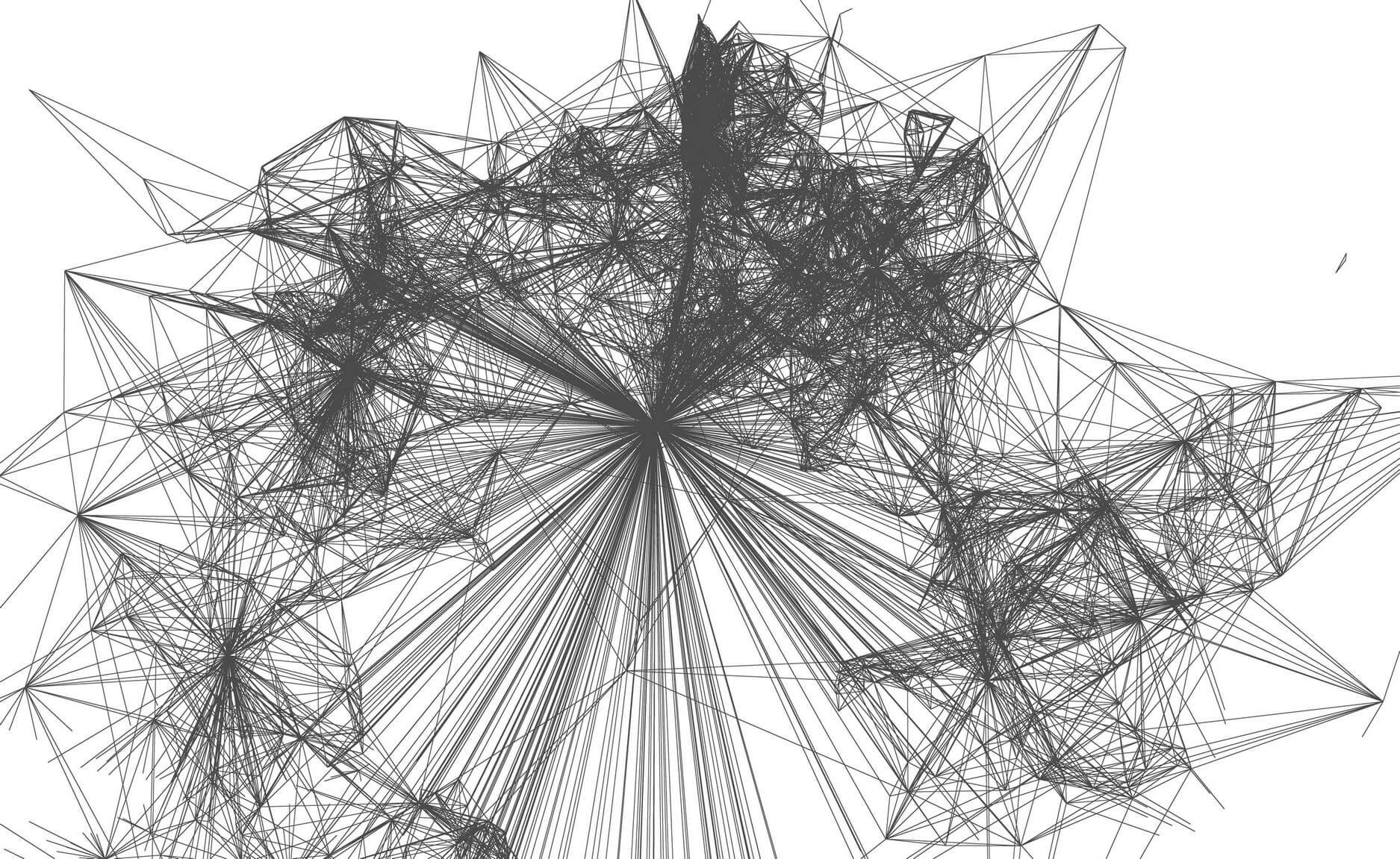}
          \caption{Superpoint graph}
           \label{fig:illu_graph} 
	\end{subfigure}&
    \begin{subfigure}[b]{0.48\textwidth}
		\includegraphics[width=1\textwidth]{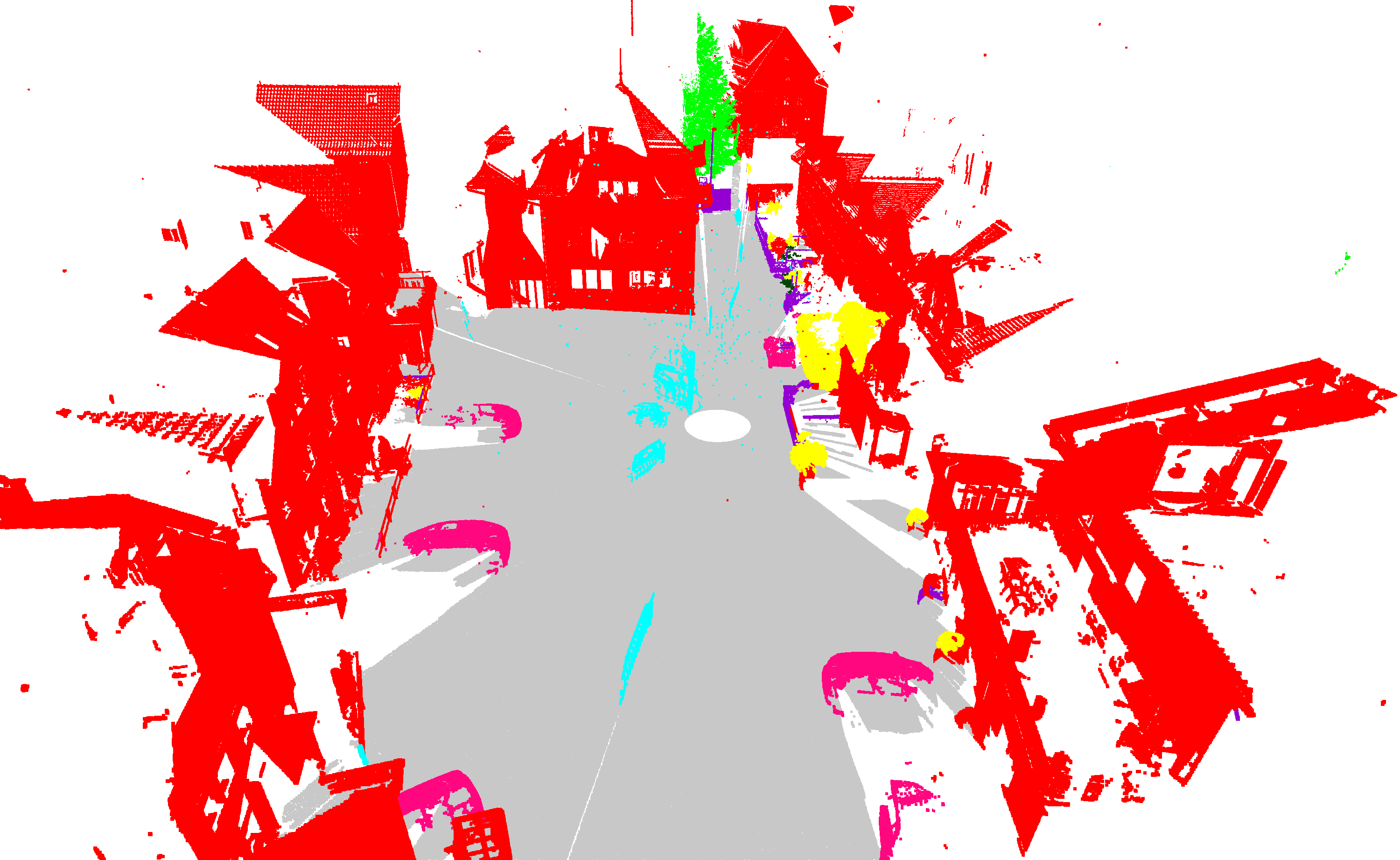}       
        \caption{Semantic segmentation}
         \label{fig:illu_pred}
	\end{subfigure}
\end{tabular}
\caption{Visualization of individual steps in our pipeline. An input point cloud \Subref{fig:illu_rgb} is partitioned into geometrically simple shapes, called superpoints \Subref{fig:illu_seg}. Based on this preprocessing, a superpoint graph (SPG) is constructed by linking nearby superpoints by superedges with rich attributes \Subref{fig:illu_graph}. Finally, superpoints are transformed into compact embeddings, processed with graph convolutions to make use of contextual information, and classified into semantic labels.}
\label{fig:teaser}
\end{figure*}

In this chapter we propose a representation of large 3D point clouds as a collection of interconnected simple shapes, coined superpoints, in spirit similar to superpixel methods for image segmentation \citep{achanta2012slic}. As illustrated in \figref{fig:teaser}, this structure can be captured by an attributed directed graph called the superpoint graph (SPG). Its nodes represent simple shapes while edges describe their adjacency relationship characterized by rich edge attributes.

The SPG representation has several compelling advantages. First, instead of classifying individual points or voxels, it considers entire object parts as whole, which are easier to identify. Second, it is able to describe in detail the relationship between adjacent objects, which is crucial for contextual classification: cars are generally above roads, ceilings are surrounded by walls, etc. Third, the size of the SPG is defined by the number of simple structures in a scene rather than the total number of points, which is typically several order of magnitude smaller. This allows us to model long-range interaction which would be intractable otherwise without strong assumptions on the nature of the pairwise connections.

This chapter is largely based on our CVPR 2018 publication~\citep{superpointgraphs}, where the contribution was equally shared with Lo{\"{\i}}c Landrieu. Its contributions to the field at the time of publication are as follows:
\begin{itemize}
\item We introduce superpoint graphs, a novel point cloud representation with rich edge attributes encoding the contextual relationship between object parts in 3D point clouds.
\item Based on this representation, we are able to apply deep learning on large-scale point clouds without major sacrifice in fine details. Our architecture consists of PointNets~\citep{qi2016pointnet} for superpoint embedding and graph convolutions for contextual segmentation. For the latter, we introduce a novel, more efficient version of Edge-Conditioned Convolutions from previous chapter as well as a new form of input gating in Gated Recurrent Units~\citep{cho-gru14}.
\item We set a new state of the art on two publicly available datasets: Semantic3D \citep{hackel2017semantic3d} and S3DIS \citep{armeni_cvpr16}. In particular, we improve mean per-class intersection over union (mIoU) by $11.9$ points for the Semantic3D reduced test set, by $8.8$ points for the Semantic3D full test set, and by up to $12.4$ points for the S3DIS dataset.
\end{itemize}

\section{Related Work}
The classic approach to large-scale point cloud segmentation is to classify each point or voxel independently using handcrafted features derived from their local neighborhood \citep{weinmann_contextual_2015}.
The solution is then spatially regularized using graphical models \citep{munoz2009contextual, koppula2011semantic, lu2012simplified, shapovalov2013spatial, kim20133d, anand2013contextually, niemeyer2014contextual,martinovic20153d,wolf2015fast} or structured optimization \citep{LANDRIEU2017102}.
Clustering as  preprocessing \citep{hu2013efficient,guinard_weakly_2017} or postprocessing \citep{weinmann2017hybrid} have been used by several frameworks to improve the accuracy of the classification.

\textbf{Deep Learning on Point Clouds.} Several different approaches going beyond naive volumetric processing of point clouds have been proposed recently and briefly reviewed in Section~\ref{sec:ecc-related} before. However, very few methods with deep learning components have been demonstrated to be able to segment large-scale point clouds. PointNet~\citep{qi2016pointnet} can segment large clouds with a sliding window approach, therefore constraining contextual information within a small area only. \citet{Engelmann17_3dsemseg} improves on this by increasing the context scope with multi-scale windows or by considering directly neighboring window positions on a voxel grid. SEGCloud~\citep{tchapmi2017segcloud} handles large clouds by voxelizing followed by interpolation back to the original resolution and post-processing with a conditional random field (CRF). None of these approaches is able to consider fine details and long-range contextual information simultaneously. In contrast, our pipeline partitions point clouds in an adaptive way according to their geometric complexity and allows deep learning architecture to use both fine detail and interactions over long distance.

\textbf{Graph Convolutions.} A key step of our approach is using graph convolutions to spread contextual information. Formulations that are able to deal with graphs of variable sizes can be seen as a form of message passing over graph edges~\citep{GilmerSRVD17}. Of particular interest are models supporting continuous edge attributes, which we use to represent interactions. In image segmentation, convolutions on graphs built over superpixels have been used for post-processing: \citet{LiangSFLY16,LiangLSFYX17} traverses such graphs in a sequential node order based on unary confidences to improve the final labels. We update graph nodes in parallel and exploit edge attributes for informative context modeling. \citet{Xu17GraphGen} convolves information over graphs of object detections to infer their contextual relationships. Our work infers relationships implicitly to improve segmentation results. \citet{QiLJFU17} also relies on graph convolutions on 3D point clouds. However, we process large point clouds instead of small RGB-D images with nodes embedded in 3D instead of 2D in a novel, rich-attributed graph. Finally, we note that graph convolutions also bear functional similarity to deep learning formulations of CRFs~\citep{Zheng15crf}, which we discuss more in \secref{subsec:contexseg}.

\section{Method}
\begin{figure*}[t]
\input{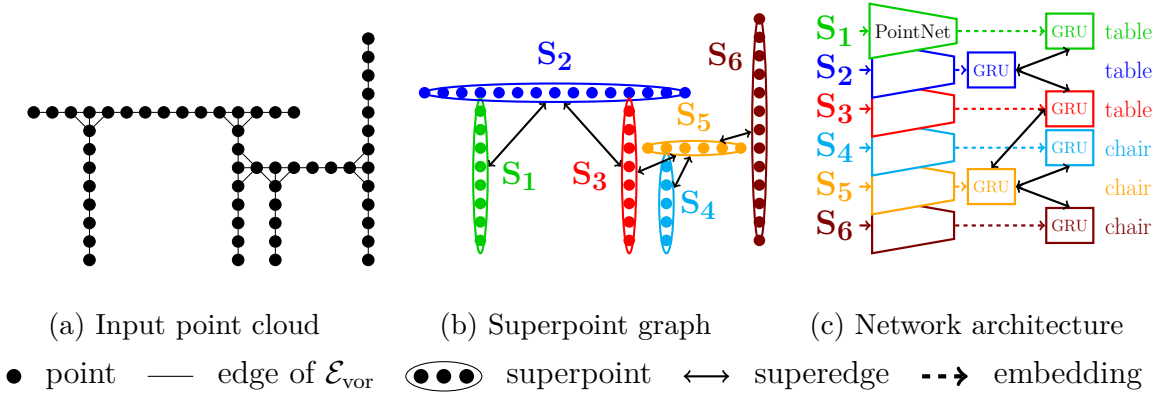}
\caption{Illustration of our framework on a toy scan of a table and a chair. We perform geometric partitioning on the  point cloud \Subref{fig:pipea}, which allows us to build the superpoint graph \Subref{fig:pipeb}.
 Each superpoint is embedded by a PointNet network. The embeddings are then refined in GRUs by message passing along superedges to produce the final labeling \Subref{fig:pipec}.}
\label{fig:pipeline}
\end{figure*}
The main obstacle that our framework tries to overcome is the size of typical LiDAR scans.
Indeed, they can reach hundreds of millions of points, making direct deep learning approaches intractable. The proposed superpoint graph (SPG) representation allows us to split the semantic segmentation problem into three distinct problems of different scales, shown in \figref{fig:pipeline}, which can in turn be solved by methods of corresponding complexity:
\begin{itemize}
\item[1] \textbf{Geometrically homogeneous partition:} The first step of our algorithm is to partition the point cloud into geometrically simple yet meaningful shapes, called superpoints. This unsupervised step takes the whole point cloud as input, and therefore must be computationally very efficient. The SPG can be easily computed from this partition.
\item[2] \textbf{Superpoint embedding:} Each node of the SPG corresponds to a small part of the point cloud corresponding to a geometrically simple primitive, which we assume to be semantically homogeneous. Such primitives can be reliably represented by downsampling small point clouds to at most hundreds of points. This small size allows us to utilize recent point cloud embedding methods such as PointNet~\citep{qi2016pointnet}.
\item[3] \textbf{Contextual segmentation:} The graph of superpoints is by orders of magnitude smaller than any graph built on the original point cloud. Deep learning algorithms based on graph convolutions can then be used to classify its nodes using rich edge attributes facilitating long-range interactions.
\end{itemize}
The SPG representation allows us to perform end-to-end learning of the trainable two last steps. We will describe each step of our pipeline in the following subsections.
\subsection{Geometric Partition with a Global Energy}
\label{sec:energy}
In this subsection, we describe our method for partitioning the input point cloud into parts of simple shape.
Our objective is not to retrieve individual objects such as cars or chairs, but rather to break down the objects into simple parts, as seen in \figref{fig:illustration}. However, the clusters being geometrically simple, one can expect them to be semantically homogeneous as well, \ie not to cover objects of different classes. Note that this step of the pipeline is purely unsupervised and makes no use of class labels beyond validation.

We follow the global energy model described by \citet{guinard_weakly_2017} for its computational efficiency. Another advantage is that the segmentation is adaptive to the local geometric complexity. In other words, the segments obtained can be large simple shapes such as roads or walls, as well as much smaller components such as parts of a car or a chair.

Let us consider the input point cloud $\mathcal{P}$ as a set of $n$ 3D points. Each point $i\in \mathcal{P}$ is defined by its 3D position $P_i$, and, if available, other observations $O_i$ such as color or intensity. For each point, we compute a set of $d_g$ geometric features $F_i\in \bbR^{d_g}$ characterizing the shape of its local neighborhood. In this paper, we use three dimensionality values proposed by \citet{demantke_dimensionality_2011}: linearity, planarity and scattering, as well as the verticality feature introduced by \citet{guinard_weakly_2017}.
We also compute the elevation of each point, defined as the $z$ coordinate of $P_i$ normalized over the whole input cloud.

The global energy proposed by \citet{guinard_weakly_2017} is defined with respect to the $10$-nearest neighbor adjacency graph $\Gnn=\Pa{\mathcal{P},\Enn}$ of the point cloud (note that this is not the SPG). The geometrically homogeneous partition is defined as the constant connected components of the solution of the following optimization problem:
\begin{equation}
\label{eq:minimal_partition}
\argmin_{G \in \bbR^{d_g \times n}}
\sum_{i \in \mathcal{P}}\Norm{G_i - F_i}^2
+ \mu
\sum_{(i,j) \in \Enn}\bm{\alpha}_{(i,j)}\Bra{G_i \neq G_j},
\end{equation}
where $\Bra{\cdot}$ is the Iverson bracket. The edge weight $\bm{\alpha} \in \bbR^{\Abs{E}}_+$ is chosen to be inversely proportional to the edge length. The factor $\mu$ is the regularization strength and determines the coarseness of the resulting partition.

The problem defined in \eqref{eq:minimal_partition}
is known as generalized minimal partition problem, and can be seen as a continuous-space version of the Potts energy model, or an $\ell_0$ variant of the graph total variation.
The minimized functional being nonconvex and noncontinuous implies that the problem cannot realistically be solved exactly for large point clouds. However, the $\ell_0$-cut pursuit algorithm introduced by \citet{landrieu2017cut} is able to quickly find an approximate solution with a few graph-cut iterations. In contrast to other optimization methods such as $\alpha$-expansion \citep{boykov2001fast}, the $\ell_0$-cut pursuit algorithm does not require selecting  the size of the partition in advance.
The constant connected components $\cS=\Cur{S_1, \cdots, S_k}$ of the solution of \eqref{eq:minimal_partition} define our geometrically simple elements, and are referred  as \emph{superpoints} (\ie set of points) in the rest of this chapter.
\subsection{Superpoint Graph Construction}
\label{sec:SPGconstruction}
In this subsection, we describe how we compute the SPG as well as its key features. The SPG is a structured representation of the point cloud, defined as an oriented attributed graph $\cG=\Pa{\cS, \cE, E}$ whose nodes are the set of superpoints $\cS$ and edges $\cE$ (referred to as \emph{superedges}) represent the adjacency between superpoints. The superedges are annotated by a set of $d_f$ attributes: $E \in \bbR^{|\cE| \times d_f}$ characterizing the adjacency relationship between superpoints.

We define $\Gvor=\Pa{\mathcal{P},\Evor}$ as the symmetric Voronoi adjacency graph of the complete input point cloud as defined by \citet{jaromczyk1992relative}.
Two superpoints $S$ and $R$ are adjacent if there is at least one edge in $\Evor$ with one end in $S$ and one end in $R$:
\begin{equation}
\label{eq:superedges}
\cE = \Cur{\Pa{S,R} \in \cS^2 \mid \exists \Pa{i,j} \in \Evor \cap \Pa{S \times R}}.
\end{equation}
Important spatial attributes associated with a superedge $\Pa{S,R}$ are obtained from the set of offsets $\delta(S,R)$ for edges in $\Evor$ linking both superpoints:
\begin{equation} 
\label{eq:delta}
\delta\Pa{S,R}=\Cur{\Pa{p_i-p_j} \mid \Pa{i,j} \in \Evor \cap \Pa{S \times R}}.
\end{equation}
Superedge attributes can also be derived by comparing the shape and size of the adjacent superpoints. To this end, we compute $\Abs{S}$ as the number of points comprised in a superpoint $S$, as well as shape attribute $\Length{S}=\lambda_1$, $\Surface{S}=\lambda_1\lambda_2$, $\Volume{S}=\lambda_1\lambda_2\lambda_3$ derived from the eigenvalues $\lambda_1, \lambda_2, \lambda_3$ of the covariance of the positions of the points comprised in each superpoint, sorted by decreasing value. 
In \tabref{table:superedge_features}, we describe a list of the different superedge attributes used in this paper. Note that the break of symmetry in the edge attributes makes the SPG a directed graph.
\begin{table}\begin{center}
  \begin{tabular}{ccc} \hline
  	\small
    Attribute name & Size & Description \\\hline
    mean offset & $3$ & $\mathrm{mean}_{m \in \delta\Pa{S,R}}\; \delta_m$\\ 
    offset deviation & $3$ & $\mathrm{std}_{m \in \delta\Pa{S,R}}\; \delta_m$\\ %
    centroid offset & $3$ & $\mathrm{mean}_{i \in S}\; P_i - \mathrm{mean}_{j \in R}\; P_j$\\
    length ratio & $1$ & $\log\Length{S} / \Length{R}$\\
    surface ratio & $1$ & $\log\Surface{S} / \Surface{R}$\\
    volume ratio & $1$ & $\log\Volume{S} / \Volume{R}$\\
    point count ratio & $1$ & $\log |S| / |R|$\\
    \hline 
  \end{tabular}
  \end{center}
  \caption{List of $d_f=13$ superedge attributes characterizing the adjacency between two superpoints $S$ and $R$.}
\label{table:superedge_features}
\end{table}

\subsection{Superpoint Embedding} \label{sec:embedding}
The goal of this stage is to compute a descriptor for every superpoint $S_i$ by embedding it into a vector $\Emb$ of fixed-size dimensionality $d_z$. Note that each superpoint is embedded in isolation; contextual information required for its reliable classification is provided only in the following stage by the means of graph convolutions. 

Several deep learning-based methods have been proposed for this purpose recently. We choose PointNet~\citep{qi2016pointnet} for its remarkable simplicity, efficiency, and robustness. In PointNet, input points are first aligned by a Spatial Transformer Network~\citep{JaderbergSTN}, independently processed by multi-layer perceptrons (MLPs), and finally max-pooled to summarize the shape. 

In our case, input shapes are geometrically simple objects, which can be reliably represented by a small amount of points and embedded by a rather compact PointNet. This is important to limit the memory needed when evaluating many superpoints on current GPUs. In particular, we subsample superpoints on-the-fly down to $n_p=128$ points to maintain efficient computation in batches and facilitate data augmentation. Superpoints of less than $n_p$ points are sampled with replacement, which in principle does not affect the evaluation of PointNet due to its max-pooling. However, we observed that including very small superpoints of less than $n_\mathrm{minp}=40$ points in training harms the overall performance. Thus, embedding of such superpoints is set to zero so that their classification relies solely on contextual information.

In order for PointNet to learn spatial distribution of different shapes, each superpoint is rescaled to unit sphere before embedding. Points are represented by their normalized position $P'_i$, observations $O_i$, and geometric features $F_i$ (since these are already available precomputed from the partitioning step). Furthermore, the original metric diameter of the superpoint is concatenated as an additional feature after PointNet max-pooling in order to stay covariant with shape sizes, see \secref{sec:model_details} for further details.

\subsection{Contextual Segmentation}
\label{subsec:contexseg}
The final stage of the pipeline is to classify each superpoint $S_i$ based on its embedding $\Emb$ and its local surroundings within the SPG. Graph convolutions are naturally suited to this task. In this section, we explain the propagation model of our system.

Our approach builds on the ideas from Gated Graph Neural Networks~\citep{yujia16} and Edge-Conditioned Convolutions (ECC) introduced in the previous chapter. The general idea is that superpoints refine their embedding according to pieces of information passed along superedges. Concretely, each superpoint $S_i$ maintains its state hidden in a Gated Recurrent Unit (GRU)~\citep{cho-gru14}. The hidden state is initialized with embedding $\Emb$ and is then processed over several iterations (time steps) $t=1\ldots T$. At each iteration $t$, a GRU takes its hidden state $\Gruh{t}{i}$ and an incoming message $\Grum$ as input, and computes its new hidden state $\Gruh{t+1}{i}$. The incoming message $\Grum$ to superpoint $i$ is computed as a weighted sum of hidden states $\Gruh{t}{j}$ of neighboring superpoints $j$. The actual weighting for a superedge $(j,i)$ depends on its attributes $E_{j,i}$, listed in \tabref{table:superedge_features}. In particular, it is computed from the attributes by a multi-layer perceptron $w$, so-called Filter Generating Network. Formally:
\begin{equation}
\begin{aligned}
\label{eq:gru}
\begin{split}
\Gruh{t+1}{i} &= (1-\Gruu) \odot \Gruq + \Gruu \odot \Gruh{t}{i}\\
\Gruq &= \tanh(\Gruxn{1} + \Grur \odot \Gruhn{1})\\
\Gruu &= \sigma(\Gruxn{2} + \Gruhn{2})\\
\Grur &= \sigma(\Gruxn{3} + \Gruhn{3})\\
\end{split}
\end{aligned}
\end{equation}
\begin{equation}
\begin{aligned}
\label{eq:gru_norm}
(\Gruhn{1}, \Gruhn{2}, \Gruhn{3})^T &= \rho(W_h \Gruh{t}{i} + \mathbf{b}_h) \\
(\Gruxn{1}, \Gruxn{2}, \Gruxn{3})^T &= \rho(W_x \Grux + \mathbf{b}_x)
\end{aligned}
\end{equation}
\vspace{-5pt}
\begin{align}
\Grux &= \sigma(W_g \Gruh{t}{i} + \mathbf{b}_g) \odot \Grum \label{eq:gru_ig} \\ 
\Grum &= \mathrm{mean}_{j \mid (j,i) \in \cE}\; w(E_{j,i,\cdot}; W_e) \odot \Gruh{t}{j} \label{eq:gru_ecc}
\end{align}
\vspace{-24pt}
\begin{align}
\Gruh{1}{i} &= \Emb\\
\Gruy &= W_o (\Gruh{1}{i}, \ldots, \Gruh{T+1}{i})^T \label{eq:gru_io},
\end{align}
where $\odot$ is element-wise multiplication, $\sigma(\cdot)$ sigmoid function, and $W_\cdot$ and $\mathbf{b}_\cdot$ are trainable parameters shared among all GRUs. \eqref{eq:gru} lists the standard GRU rules~\citep{cho-gru14} with its update gate $\Gruu$ and reset gate $\Grur$. To improve stability during training, in \eqref{eq:gru_norm} we apply Layer Normalization~\citep{layernorm} defined as $\rho(\mathbf{a}) := (\mathbf{a} - \mathrm{mean}(\mathbf{a})) / (\mathrm{std}(\mathbf{a}) + \epsilon)$ separately to linearly transformed input $\Grux$ and transformed hidden state $\Gruh{t}{i}$, with $\epsilon$ being a small constant. Finally, the model includes three interesting extensions in Equations~\ref{eq:gru_ig}--\ref{eq:gru_io}, which we detail below.

\paragraph*{Input Gating.} We argue that GRU should possess the ability to down-weight (parts of) an input vector based on its hidden state. For example, GRU might learn to ignore its context if its class state is highly certain or to direct its attention to only specific feature channels. \eqref{eq:gru_ig} achieves this by gating message $\Grum$ by the hidden state before using it as input $\Grux$.

\paragraph*{Edge-Conditioned Convolution.} ECC plays a crucial role in our model as it can dynamically generate filtering weights for any value of continuous attributes $E_{j,i}$ by processing them with a multi-layer perceptron $w$. 
In the original formulation, $w$ regresses a weight matrix to perform matrix-vector multiplication $w(E_{j,i}; W_e) \Gruh{t}{j}$ for each edge. In this chapter, we propose a lightweight variant with lower memory requirements and fewer parameters, which is beneficial for datasets with few but large point clouds. Specifically, we regress only an edge-specific weight vector and perform element-wise multiplication as in \eqref{eq:gru_ecc} (ECC-VV). Channel mixing, albeit in an edge-unspecific fashion, is postponed to \eqref{eq:gru_norm}. Finally, let us remark that $w$ is shared over time iterations and that self-loops are not necessary due to the existence of hidden states in GRUs.

\paragraph*{State Concatenation.} Inspired by DenseNet~\citep{densenet17}, we concatenate hidden states over all time steps and linearly transform them to produce segmentation logits $\Gruy$ in \eqref{eq:gru_io}. This allows to exploit the dynamics of hidden states due to increasing receptive field for the final classification.

\subsection{Implementation Details}
\label{sec:model_details}

\paragraph*{Training.} While the geometric partitioning step is unsupervised, superpoint embedding and contextual segmentation are trained jointly in a supervised way with cross entropy loss. Superpoints are assumed to be semantically homogeneous and, consequently, assigned a hard ground truth label corresponding to the majority label among their contained points. We also considered using soft labels corresponding to normalized histograms of point labels and training with Kullback-Leibler~\citep{Kullback_and_Leibler_1951} divergence loss. It performed slightly worse in our initial experiments, though.

Naive training on large SPGs may approach memory limits of current GPUs. We circumvent this issue by randomly subsampling the sets of superpoints at each iteration and training on induced subgraphs, \ie graphs composed of subsets of nodes and the original edges connecting them. Specifically, graph neighborhoods of order $3$ are sampled to select at most $512$ superpoints per SPG with more than $n_\mathrm{minp}$ points (smaller superpoints are not embedded). Note that as the induced graph is a union of small neighborhoods, relationships over many hops may still be formed and learned. This strategy also doubles as data augmentation and a strong regularization, together with randomized sampling of point clouds described in \secref{sec:embedding}. Additional data augmentation is performed by randomly rotating superpoints around the vertical axis and jittering point features by Gaussian noise $N(0,0.01)$ truncated to $[-0.05,0.05]$.

We train using Adam~\citep{KingmaB14_adam} with initial learning rate 0.01 and batch size 2, \ie effectively up to 1024 superpoints per batch. For Semantic3D, we train for 500 epochs with stepwise learning rate decay of 0.7 at epochs 350, 400, and 450. For S3DIS, we train for 250 epochs with steps at 200 and 230. We clip gradients within $[-1,1]$.

\paragraph*{Testing.} In modern deep learning frameworks, testing can be made very memory-efficient by discarding layer activations as soon as the follow-up layers have been computed. In practice, we were able to label full SPGs at once. To compensate for randomness due to subsampling of point clouds in PointNets, we average logits obtained over $10$ runs with different seeds.

\paragraph*{Voxelization.} We pre-process input point clouds with voxelization subsampling by computing per-voxel mean positions and observations over a regular 3D grid ($5$ cm bins for Semantic3D and $3$ cm bins for S3DIS dataset).
The resulting semantic segmentation is interpolated back to the original point cloud in a nearest neighbor fashion.  Voxelization helps decreasing the computation time and memory requirement, and improves the accuracy of the semantic segmentation by acting as a form of geometric and radiometric denoising as well. The quality of further steps is practically not affected, as superpoints are usually strongly subsampled for embedding during learning and inference anyway (\secref{sec:embedding}).

\paragraph*{Geometric Partition.} We set regularization strength $\mu=0.8$ for Semantic3D and $\mu=0.03$ for S3DIS, which strikes a balance between semantic homogeneity of superpoints and the potential for their successful discrimination (S3DIS is composed of smaller semantic parts than Semantic3D). In addition to five geometric features $f$ (linearity, planarity, scattering, verticality, elevation), we use color information $O$ for clustering in S3DIS due to some classes being geometrically indistinguishable, such as boards or doors.

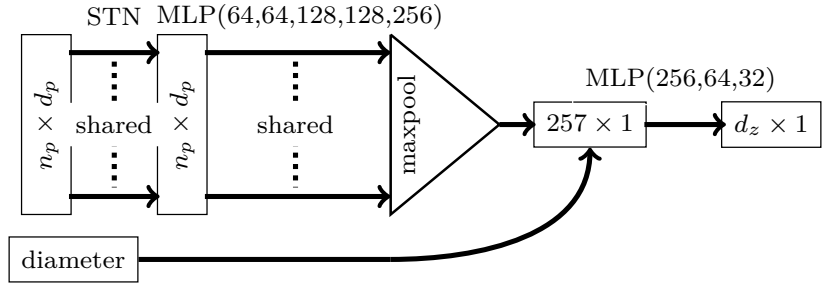
\begin{figure*}[t]
\begin{center}
\resizebox{0.7\textwidth}{!}{ 
    \begin{tikzpicture}[
 triangle/.style = {fill=white, regular polygon, regular polygon sides=3 }]
      \node [rectangle, draw = white, minimum width=2cm,  minimum height=5mm] at (0.75,1.2)   (transformNet)  {\scriptsize STN};
\node [rectangle, draw = black, rotate=90,  minimum width=2cm,  minimum height=5mm] at (0,0)   (input)  {\scriptsize $n_p \times d_p$};
   \node [rectangle, draw = black,  minimum width=.6cm,  minimum height=5mm] at (0.3,-1.5)   (scale)  {\scriptsize diameter};
\node [rectangle, draw = black, rotate=90,  minimum width=2cm,  minimum height=5mm] at (1.5,0)   (transformed)  {\scriptsize $n_p \times d_p$};
  
  \node [rectangle, draw = none, minimum width=2cm,  minimum height=5mm] at (2.8,1.2)   (MLP1)  {\scriptsize MLP(64,64,128,128,256)};
\draw[-, ultra thick, dotted] (0.75, 0.7) -- (0.75, -0.7);
\draw[-, ultra thick, dotted] (2.75, 0.7) -- (2.75, -0.7);
  \node [rectangle, draw = white, minimum width=.5cm,  minimum height=2mm, fill = white] at (0.75,0)   (MLP11)  {\scriptsize shared};
  \node [rectangle, draw = white, minimum width=.5cm,  minimum height=2mm, fill = white] at (2.75,0)   (MLP11)  {\scriptsize shared};
  \node [rectangle, draw = black,  minimum width=1cm,  minimum height=5mm] at (6,0)   (global)  {\scriptsize $257 \times 1$};

  \node [draw = white, rotate=90] at (4,0)   (maxpool)  {\scriptsize maxpool};

  \node [rectangle, draw = white , minimum width=1cm,  minimum height=2mm] at (7,0.5)   (MLP2)  {\scriptsize MLP(256,64,32)};

  \node [rectangle, draw = black,  minimum width=1cm,  minimum height=5mm] at (8,0)   (output)  {\scriptsize $d_z \times 1$};

  \draw[->, ultra thick] (0.25, 0.8) -- (1.25, 0.8);
  \draw[->, ultra thick] (0.25, -0.8) -- (1.25, -0.8);

  \draw[-, ultra thick] (scale) to (3.8, -1.5);
  \draw[->, ultra thick] (3.8, -1.5) to[out = 0, in = 270] (global);

  \draw[->, ultra thick] (1.75, 0.8 ) -- (3.8, 0.8);
  \draw[->, ultra thick] (1.75, -0.8) -- (3.8, -0.8);

  \draw[thick] (3.8, 1) -- (5, 0) -- (3.8, -1) -- (3.8, 1);

 \draw[->, ultra thick] (5, 0) -- (5.4, 0);

 \draw[->, ultra thick] (6.6, 0) -- (7.5, 0);

    \end{tikzpicture}
}
\end{center}
\caption{The PointNet embedding $n_p$ $d_p$-dimensional samples of a superpoint to a $d_z$-dimensional vector.}
\label{ref:fig_pointnet}
\end{figure*}

\paragraph*{PointNet.} We use a simplified shallow and narrow PointNet architecture with just a single Spatial Transformer Network (STN), see \figref{ref:fig_pointnet}. Input points are processed by a sequence of MLPs (widths 64, 64, 128, 128, 256) and max pooled to a single vector of 256 features. The scalar metric diameter is appended and the result further processed by a sequence of MLPs (widths 256, 64, $d_z$=32). A residual matrix $\Phi \in \mathbb{R}^{2\times 2}$ is regressed by STN and $(I+\Phi)$ is used to transform XY coordinates of input points as the first step. The architecture of STN is a "small PointNet" with 3 MLPs (widths 64, 64, 128) before max pooling and 3 MLPs after (widths 128, 64, 4). Batch Normalization~\citep{batchnorm} and ReLUs are used everywhere. Input points have $d_p$=11 dimensional features for Semantic3D (position $P$, color $O$, geometric features $F$), with 3 additional ones for S3DIS (room-normalized spatial coordinates, as in past work \citep{qi2016pointnet}).

\paragraph*{Segmentation Network.} We use embedding dimensionality $d_z=32$ and $T=10$ iterations. ECC-VV is used for Semantic3D (there are only $15$ point clouds even though the amount of points is large), while full ECC is used for S3DIS (large number of point clouds).
Filter-generating network $w$ is a MLP with 4 layers (widths 32, 128, 64, and 32 or $32^2$ for ECC-VV or ECC) with ReLUs. Batch Normalization is used only after the third parametric layer. No bias is used in the last layer. Superedges have $d_f=13$ dimensional attributes, normalized by mean subtraction and scaling to unit variance based on the whole training set.

\section{Experiments}
We evaluate our pipeline on two large point cloud segmentation benchmarks, Semantic3D \citep{hackel2017semantic3d} and Stanford Large-Scale 3D Indoor Spaces (S3DIS) \citep{armeni_cvpr16}, on both of which we set the new state of the art. Furthermore, we perform a thorough ablation study of our pipeline in \secref{subsec:ablation} and \secref{subsec:ext_ablation}.

Even though the two data sets are quite different in nature (large outdoor scenes for Semantic3D, smaller indoor scanning for S3DIS), we use nearly the same model for both, described above. The deep model is rather compact and $6$ GB of GPU memory is enough for both testing and training.

Performance is evaluated using three metrics: per-class intersection over union (IoU), per-class accuracy (Acc), and overall accuracy (OA), defined as the proportion of correctly classified points. We stress that the metrics are computed on the original point clouds, not on superpoints. 
\subsection{Semantic3D}
Semantic3D \citep{hackel2017semantic3d} is currently the largest available LiDAR dataset with over 3 billion points from a variety of urban and rural scenes. Each point has RGB and intensity values (the latter of which we do not use). The dataset consists of 15 training scans and 15 test scans with withheld labels. We also evaluate on the reduced set of 4 subsampled scans, as common in past work.

\begin{figure*}
\begin{center}
\begin{tabular}{cc}
 \begin{subfigure}{0.47\textwidth}
  \parbox{1\textwidth}{
 \begin{tabular}{c}
 	\includegraphics[width=1\textwidth]{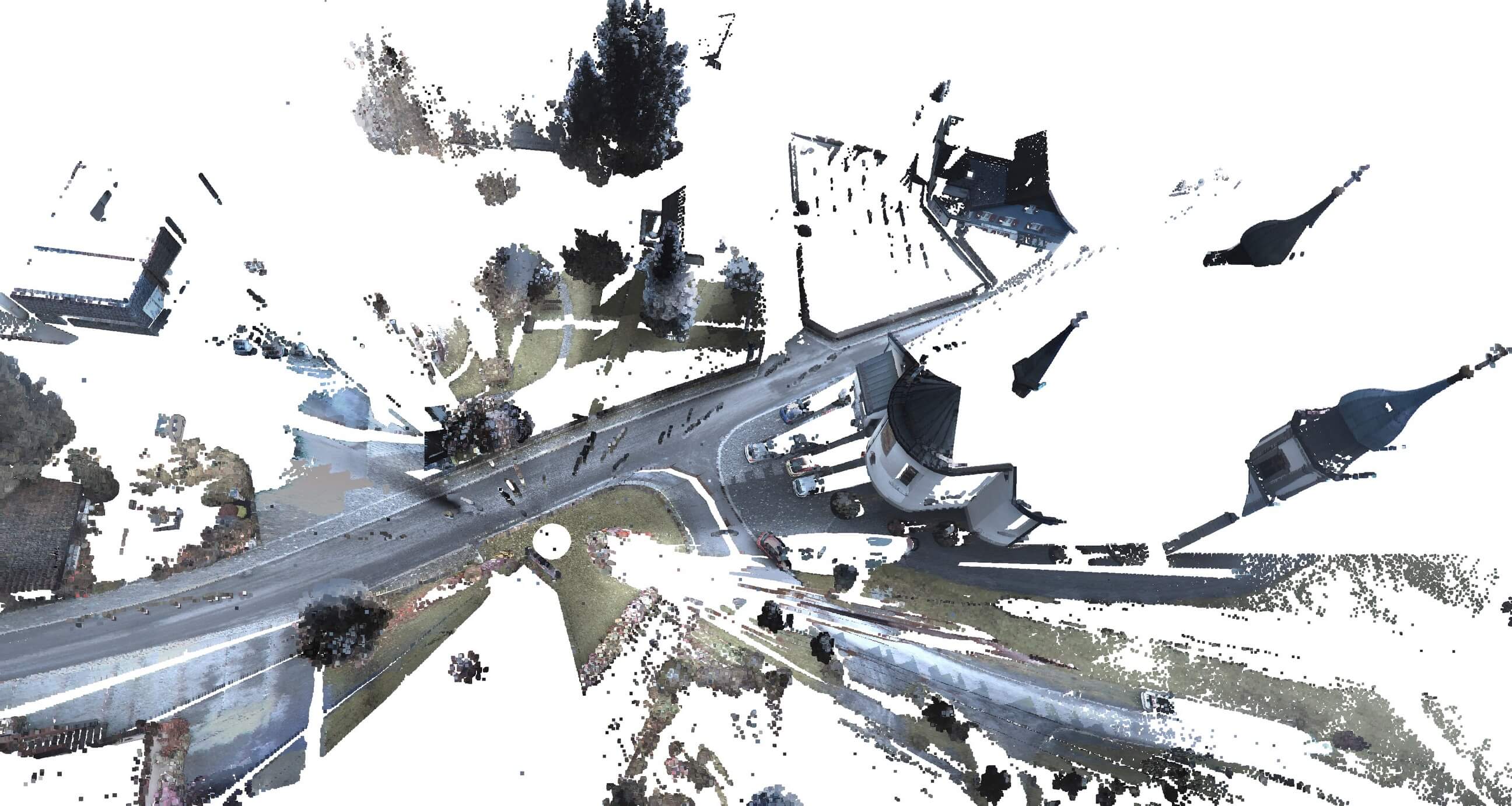}\\
    \includegraphics[width=1\textwidth] {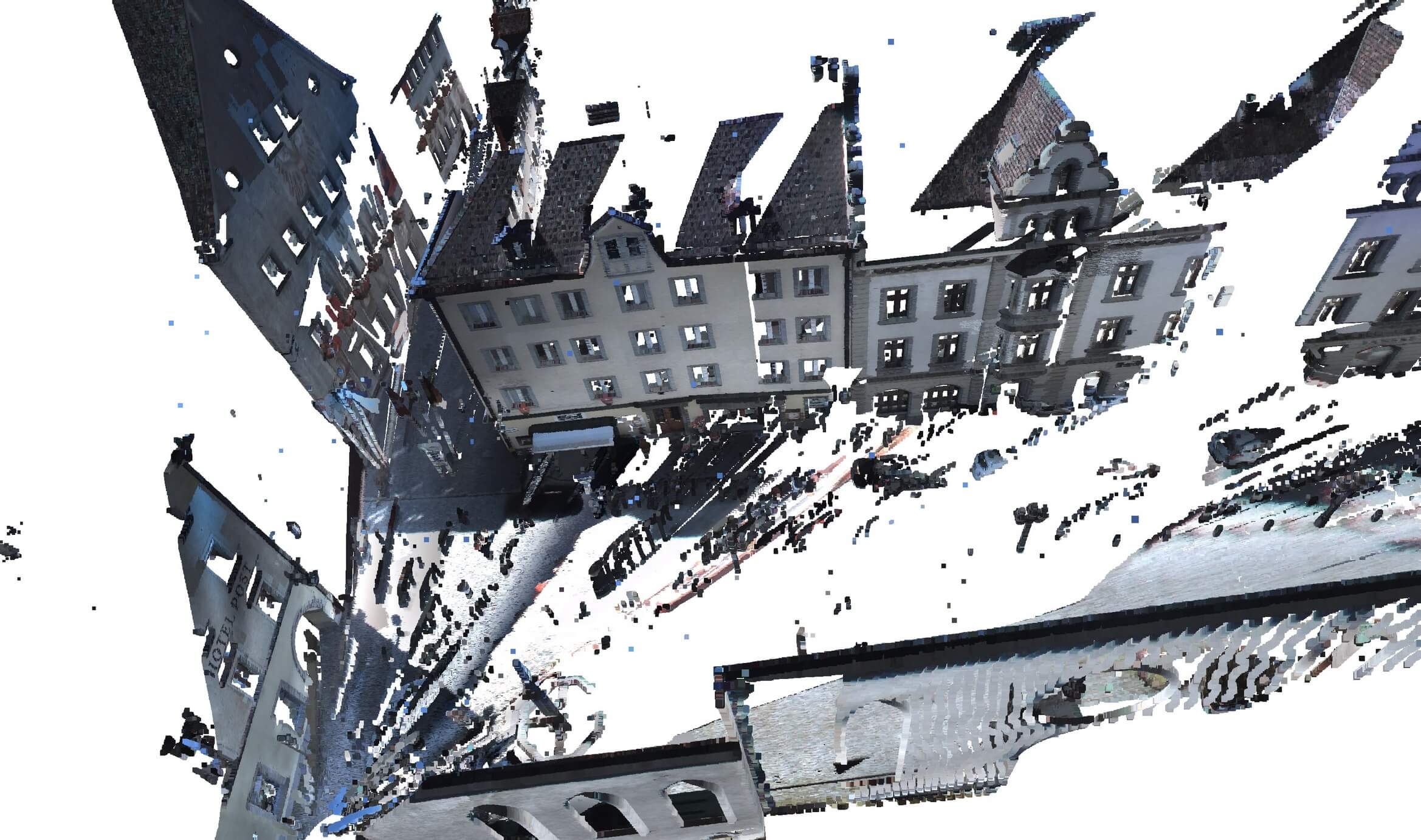}
 \end{tabular}
 \caption{RGB point cloud}
\label{fig:illustration_RGB}
 }\end{subfigure}
 &
  \begin{subfigure}{0.47\textwidth}
  \parbox{1\textwidth}{
 \begin{tabular}{c}
 	\includegraphics[width=1\textwidth]{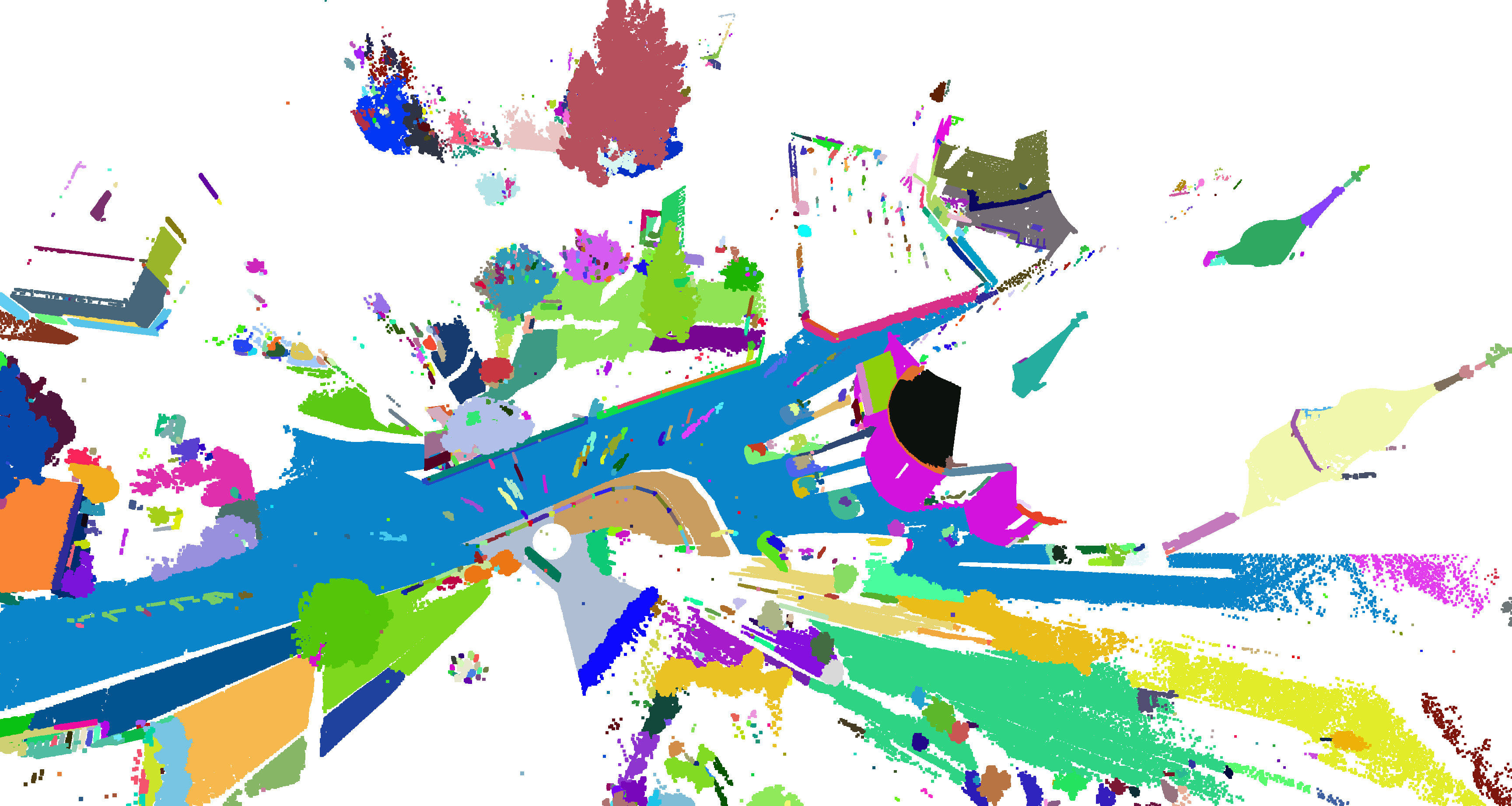}\\
    \includegraphics[width=1\textwidth]{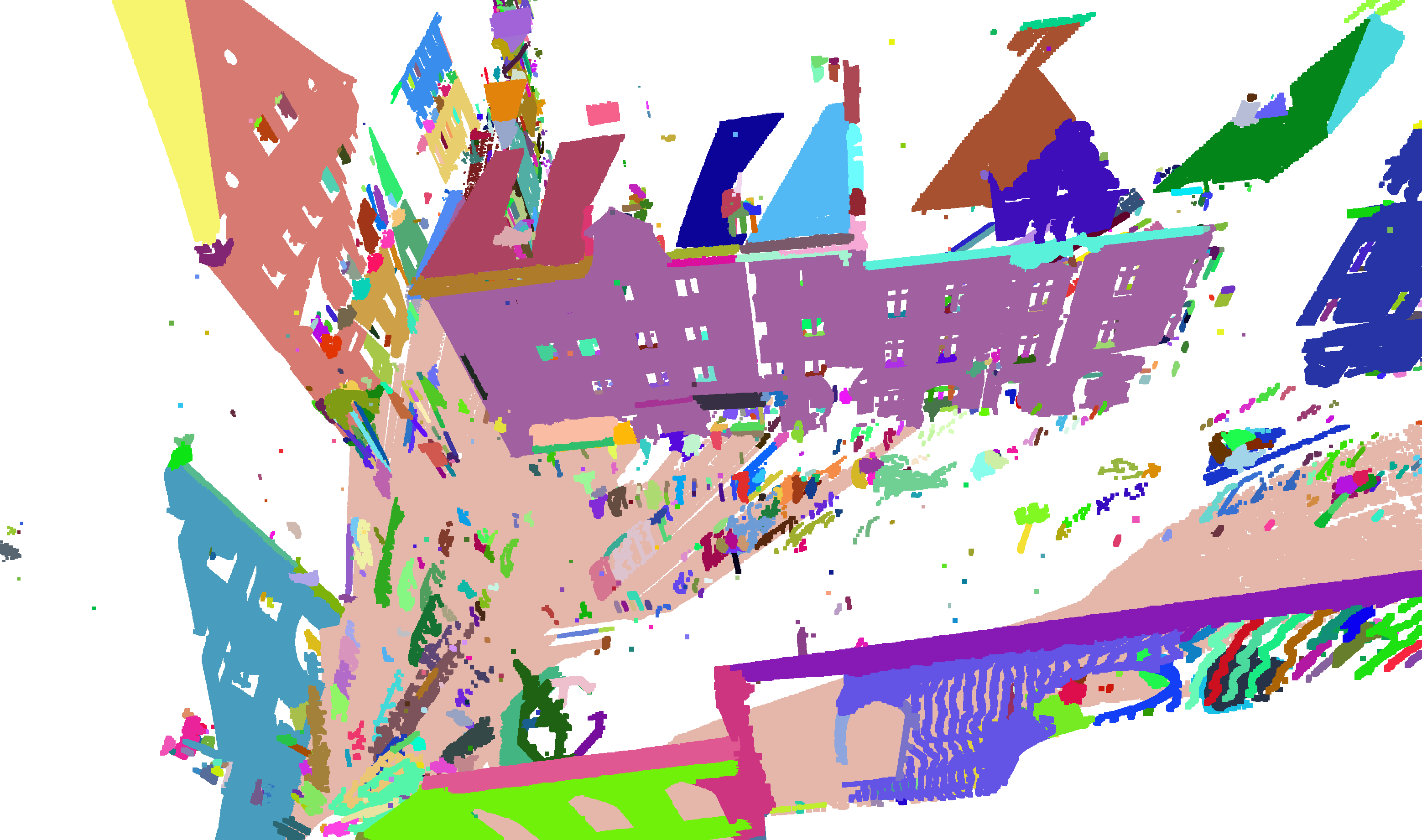}
 \end{tabular}
 \caption{Geometric partitioning}
\label{fig:illustration_seg}
 }\end{subfigure}
 \\
  \begin{subfigure}{0.47\textwidth}
  \parbox{1\textwidth}{
 \begin{tabular}{c}
 	\includegraphics[width=1\textwidth]{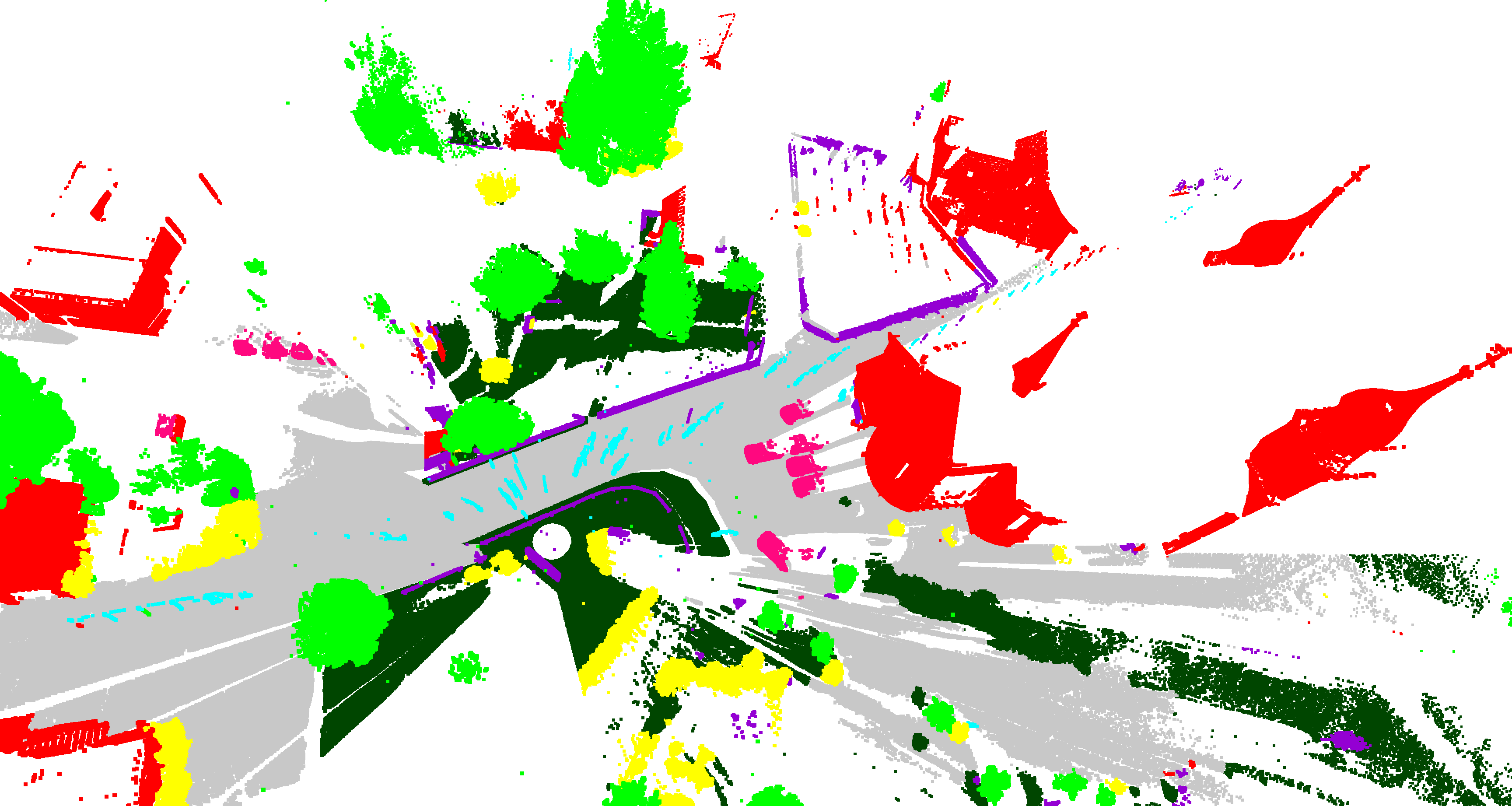}\\
    \includegraphics[width=1\textwidth]{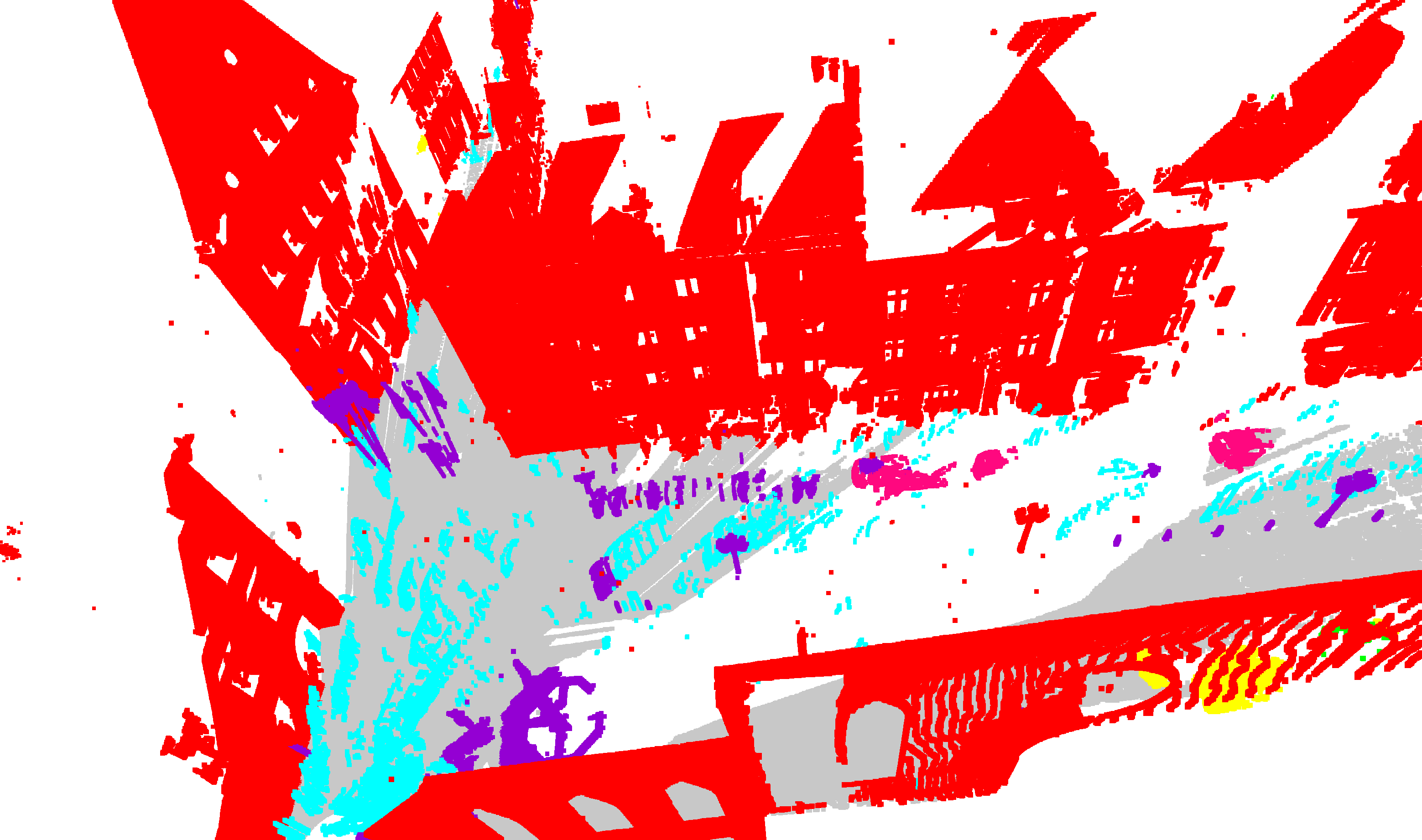}
 \end{tabular}
 \caption{Prediction}
\label{fig:illustration_pred}
 }\end{subfigure}   
 &
  \begin{subfigure}{0.47\textwidth}
  \parbox{1\textwidth}{
 \begin{tabular}{c}
 	\includegraphics[width=1\textwidth]{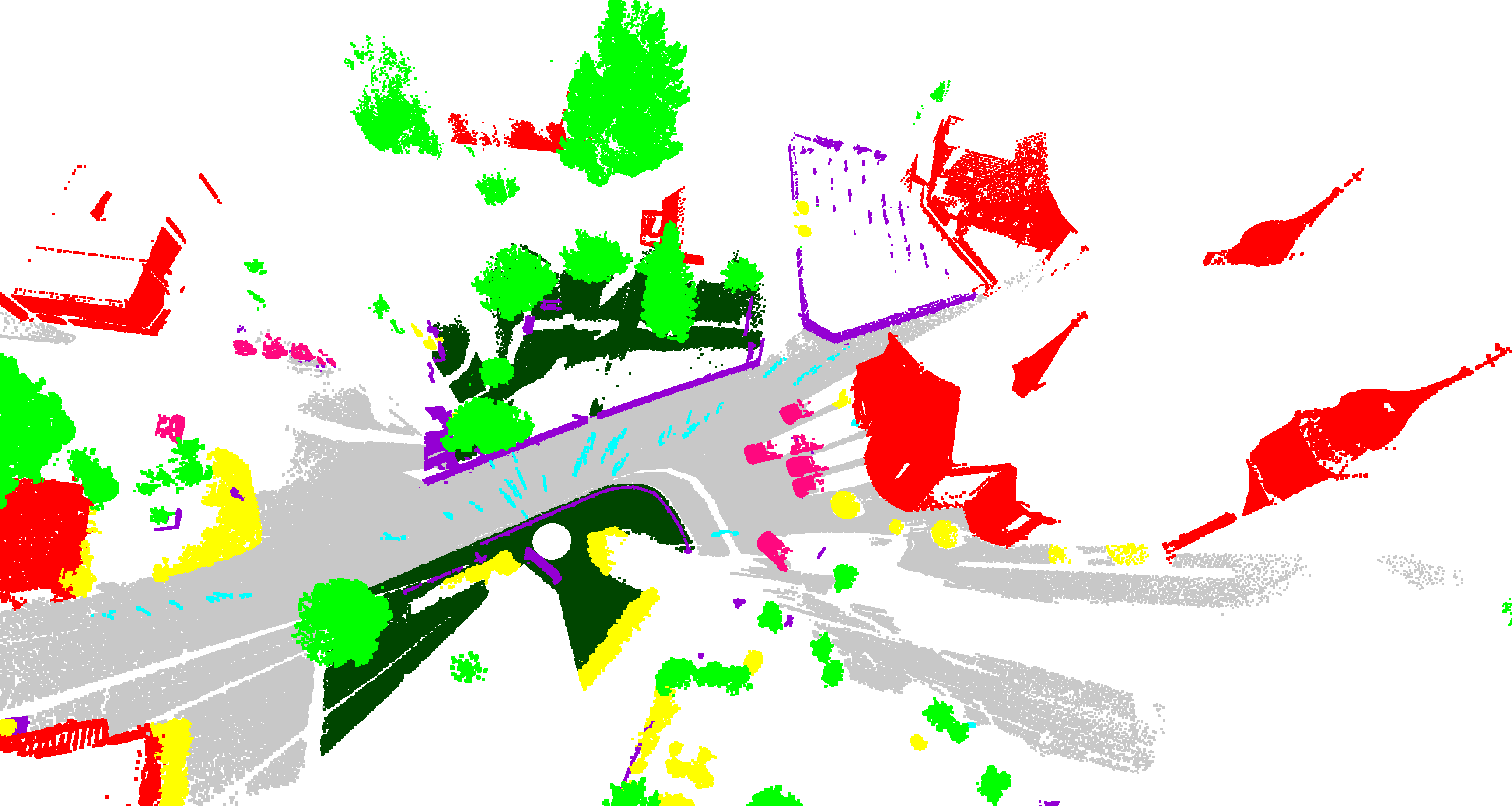}\\
    \includegraphics[width=1\textwidth]{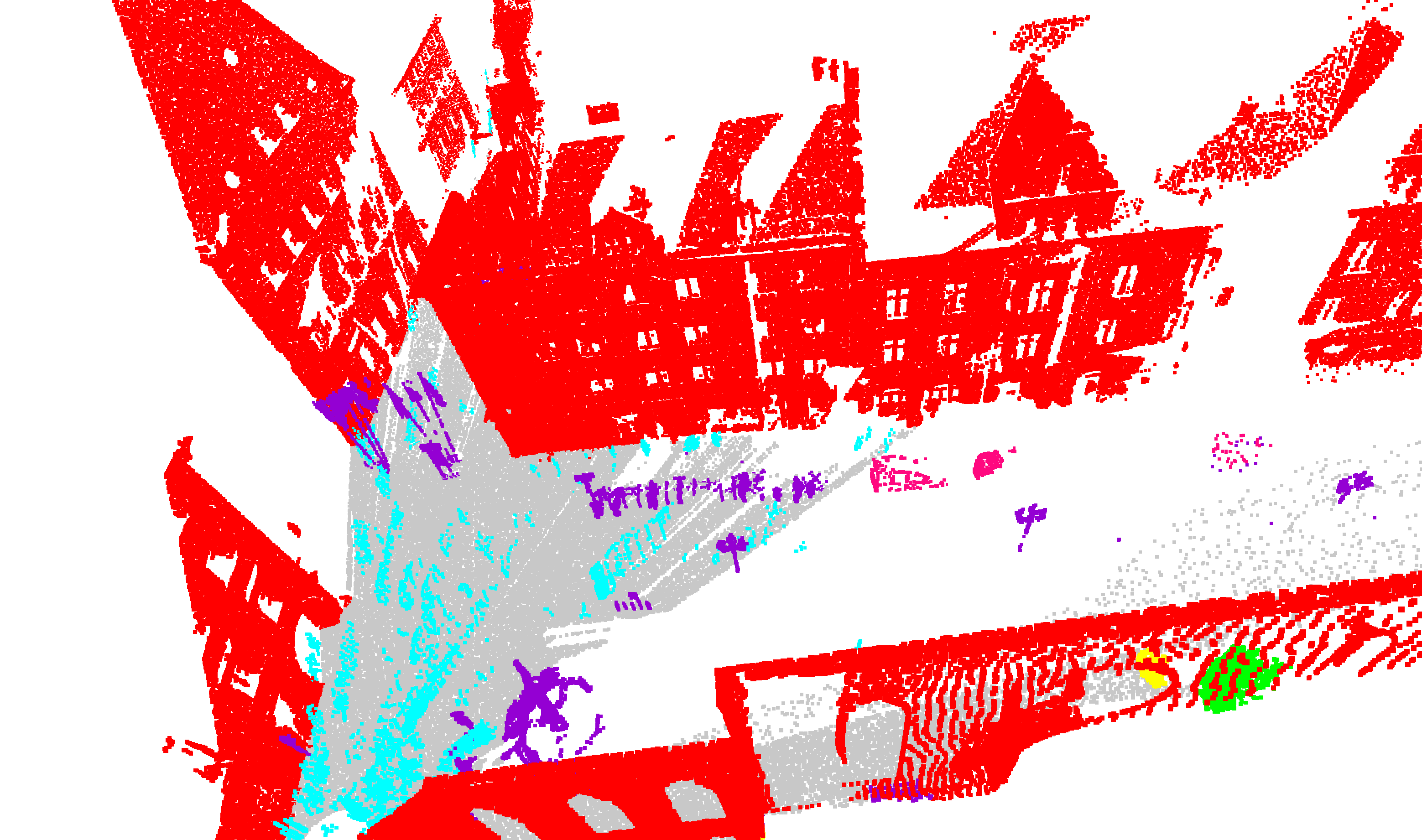}
 \end{tabular}
 \caption{Ground truth}
\label{fig:illustration_GT}
 }\end{subfigure}  
 \\
 \addlinespace[1.5ex]
 \multicolumn{2}{c}{
	 \definecolor{tempcolor1}{RGB}{200,200,200}
	 \tikz\node at (0,0) [rectangle, minimum width = 3mm, draw = none, fill = tempcolor1] (n){};
	 \hspace{0.2em} man-made terrain \hspace{0.5em}
	 \definecolor{tempcolor2}{RGB}{0,70,0}
	 \tikz\node at (0,0) [rectangle, minimum width = 3mm, draw = none, fill = tempcolor2] (n){};
	 \hspace{0.2em} natural terrain \hspace{0.5em}
	 \definecolor{tempcolor3}{RGB}{0,255,0}
	 \tikz\node at (0,0) [rectangle, minimum width = 3mm, draw = none, fill = tempcolor3] (n){};
	 \hspace{0.2em} high vegetation \hspace{0.5em}
	 \definecolor{tempcolor4}{RGB}{255,255,0}
	 \tikz\node at (0,0) [rectangle, minimum width = 3mm, draw = none, fill = tempcolor4] (n){};
	 \hspace{0.2em} low vegetation } \\
\multicolumn{2}{c}{
	 \definecolor{tempcolor5}{RGB}{255,0,0}
	 \tikz\node at (0,0) [rectangle, minimum width = 3mm, draw = none, fill = tempcolor5] (n){};
	 \hspace{0.2em} buildings \hspace{0.5em}
	 \definecolor{tempcolor6}{RGB}{148,0,211}
	 \tikz\node at (0,0) [rectangle, minimum width = 3mm, draw = none, fill = tempcolor6] (n){};
	 \hspace{0.2em} hardscape \hspace{0.5em}
	 \definecolor{tempcolor7}{RGB}{0,255,255}
	 \tikz\node at (0,0) [rectangle, minimum width = 3mm, draw = none, fill = tempcolor7] (n){};
	 \hspace{0.2em} scanning artefacts \hspace{0.5em}
	 \definecolor{tempcolor8}{RGB}{255,8,127}
	 \tikz\node at (0,0) [rectangle, minimum width = 3mm, draw = none, fill = tempcolor8] (n){};
	 \hspace{0.2em} cars 
 }
\end{tabular}
\end{center}
\caption{Example visualizations on Semantic3D. The colors  are chosen randomly for each element of the partition.}
\label{fig:illustration}
\end{figure*}

In \tabref{tab:results_semantic3D}, we provide the results of our algorithm compared to other recent works and in \figref{fig:illustration}, we provide qualitative results of our framework.
Our framework improves significantly on the state of the art of semantic segmentation for this data set, \ie by nearly 12 mIoU points on the reduced set and by nearly 9 mIoU points on the full set. In particular, we observe a steep gain on the "artefact" class. This can be explained by the ability of the partitioning algorithm to detect artifacts due to their singular shape, while they are hard to capture using snapshots, as suggested by \citep{boulch2017unstructured}. Furthermore, these small object are often merged with the road when performing spatial regularization. 
\begin{table*}
\begin{center}
\resizebox{\textwidth}{!}{ 
\small
\begin{tabular}{ccC{2em}C{2em}|C{2em}C{2em}C{2em}C{2em}C{2em}C{2em}C{2em}C{2em}}\hline
&Method& OA & mIoU&MaTe&NaTe&HiVe&LoVe&Bu&Ha&ScAr&Ca
\\\hline
\multirow{5}{*}{\rotatebox{90}{reduced set}}
&\citet{hackel2016fast}&86.2&54.2&89.8&74.5&53.7&26.8&88.8&18.9&36.4&44.7\\
&\citet{lawin2017deep}&88.9&58.5&85.6&83.2&74.2&32.4&89.7&18.5&25.1&59.2\\ 
&\citet{boulch2017unstructured}&88.6&59.1&82.0&77.3&79.7&22.9&91.1&18.4&37.3&64.4\\
&\citet{tchapmi2017segcloud}&88.1&61.3&83.9&66.0&86.0&40.5&91.1&30.9&27.5&64.3\\
&SPG (Ours)&\bf 94.0&\bf73.2&\bf97.4&\bf92.6&\bf87.9&\bf44.0&\bf93.2&\bf31.0&\bf63.5&\bf76.2
\\\hline 
\multirow{3}{*}{\rotatebox{90}{full set}}
&\citet{hackel2016fast}&85.0&49.4&91.1&69.5&32.8&21.6&87.6&25.9&11.3&55.3\\
&\citet{boulch2017unstructured}&91.0&67.4&89.6&\bf79.5&74.8&56.1&90.9&36.5&34.3&77.2\\
&SPG (Ours)&\bf92.9&\bf76.2&\bf91.5&75.6&\bf78.3&\bf71.7&\bf94.4&\bf56.8 &\bf52.9&\bf88.4\\\hline
\end{tabular}}
\end{center}
\caption{Intersection over union metric for the different classes of the Semantic3D dataset: man-made terrain (MaTe), natural terrain (NaTe), high vegetation (HiVe), low vegetation , (LoVe), buildings (Bu), hardscape (Ha), scanning artefact (ScAr), cars (Ca). OA is the global accuracy, while mIoU refers to the unweighted average of IoU of each class. Full test set has $\num{2 091 952 018}$ points, reduced test set $\num{78 699 329}$ points.
}
\label{tab:results_semantic3D}
\end{table*}

\begin{figure*}
\begin{center}
\begin{tabular}{cc}
 \begin{subfigure}{0.47\textwidth}
  \parbox{1\textwidth}{
 \begin{tabular}{c}
 	\includegraphics[width=1\textwidth]{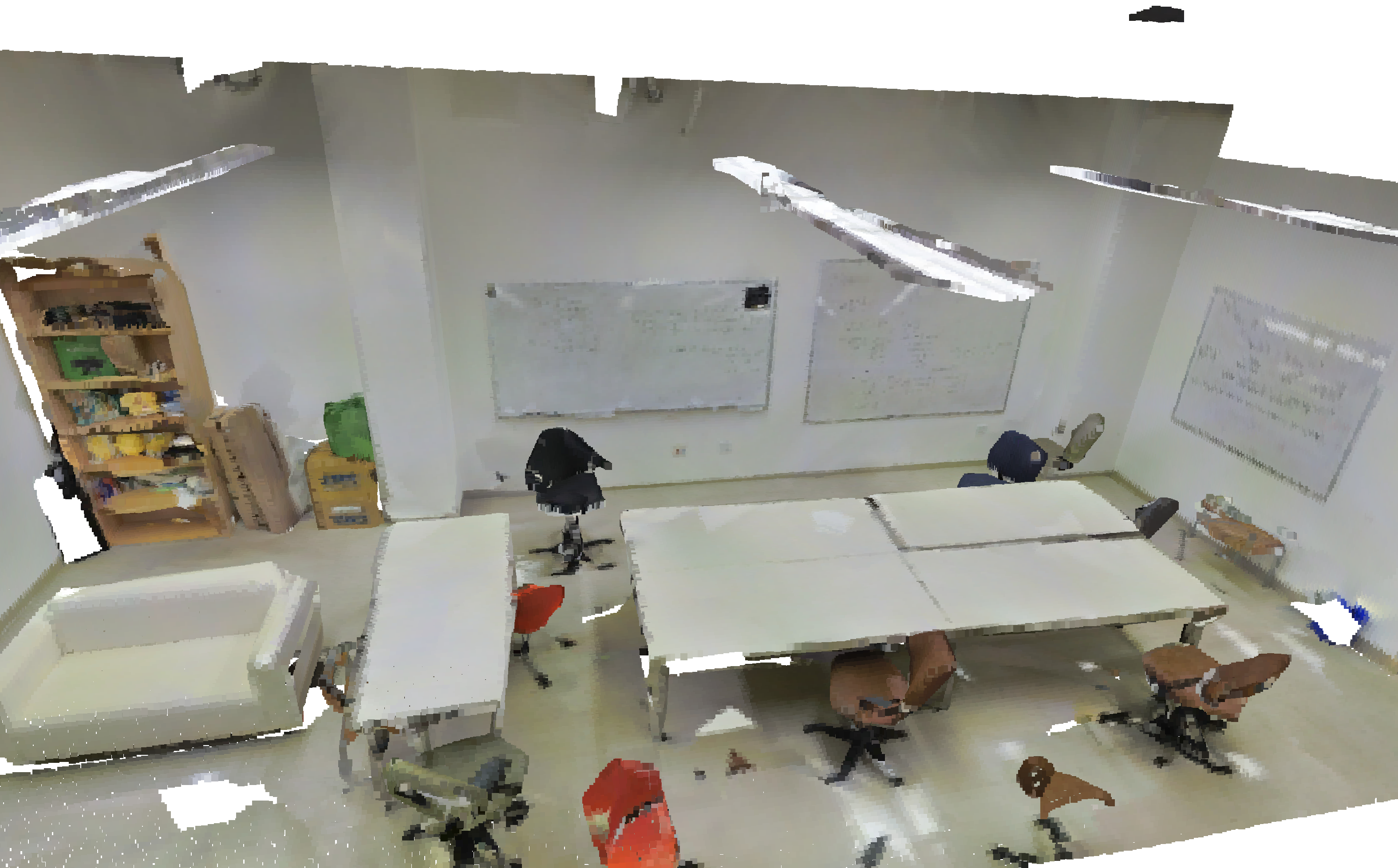}\\ 	  
 	\includegraphics[width=1\textwidth]{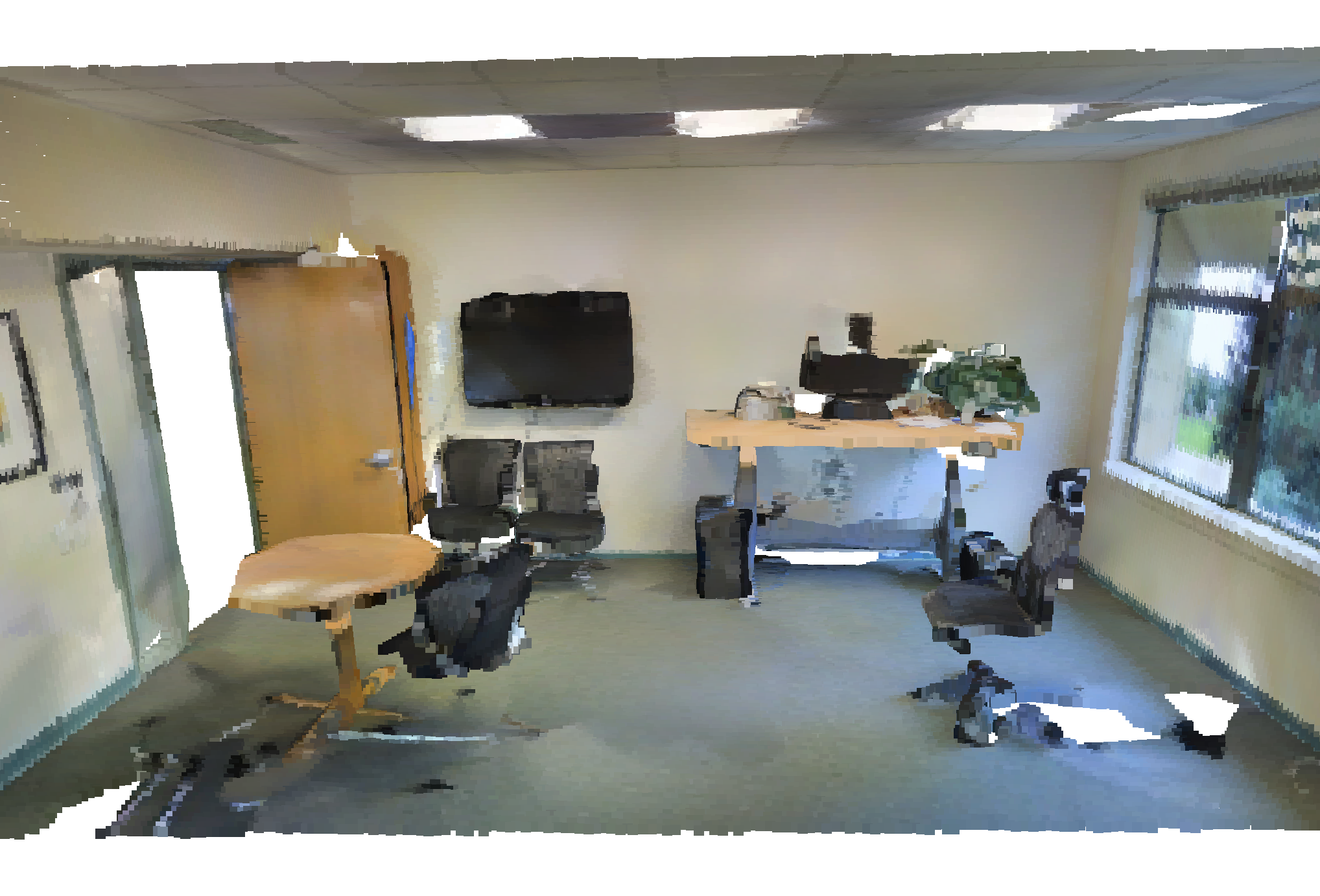}
 \end{tabular}
 \caption{RGB point cloud}
\label{fig:illustration2_RGB}
 }\end{subfigure}
 &
  \begin{subfigure}{0.47\textwidth}
  \parbox{1\textwidth}{
 \begin{tabular}{c}
 	\includegraphics[width=1\textwidth]{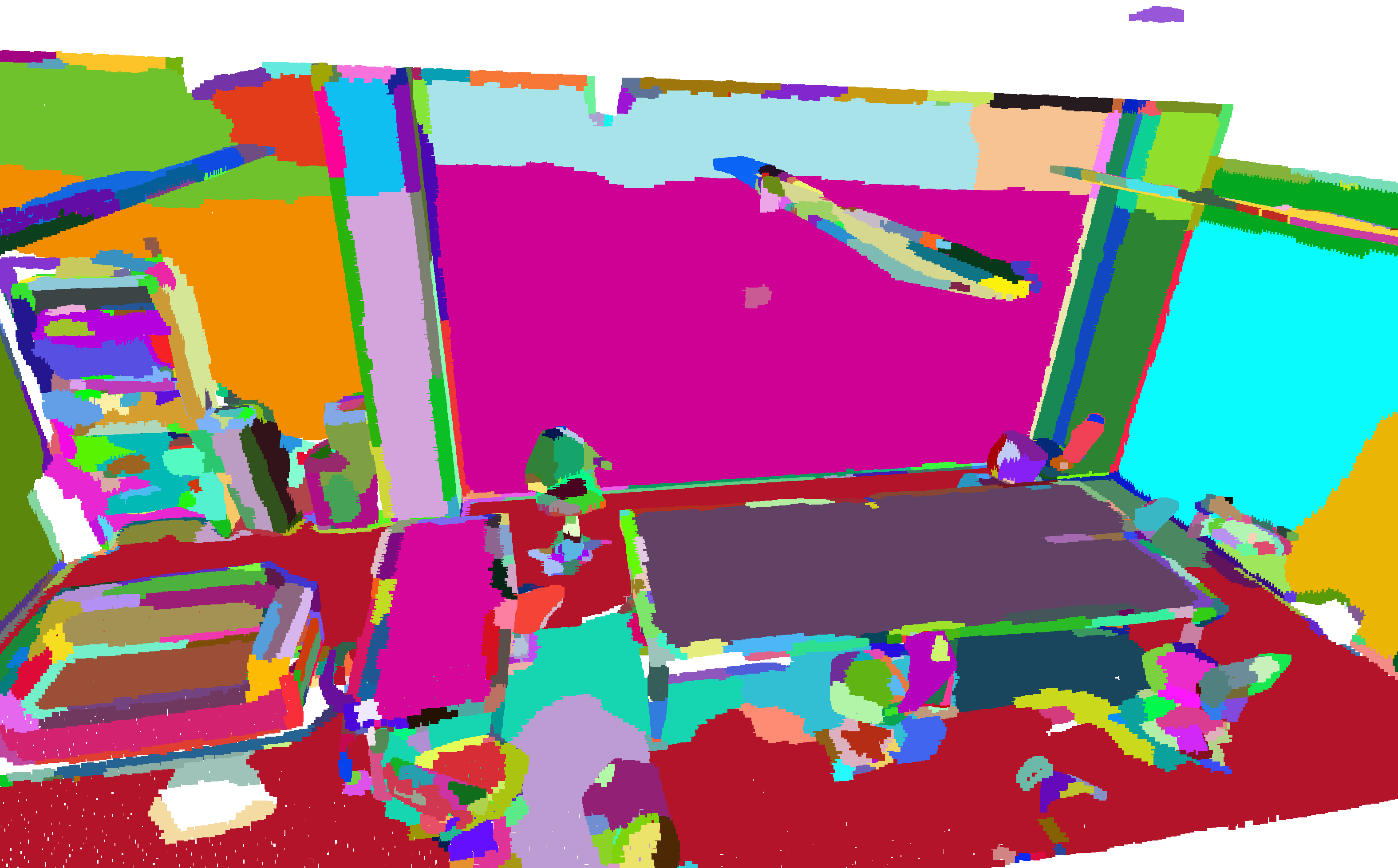}\\ 	  
 	\includegraphics[width=1\textwidth]{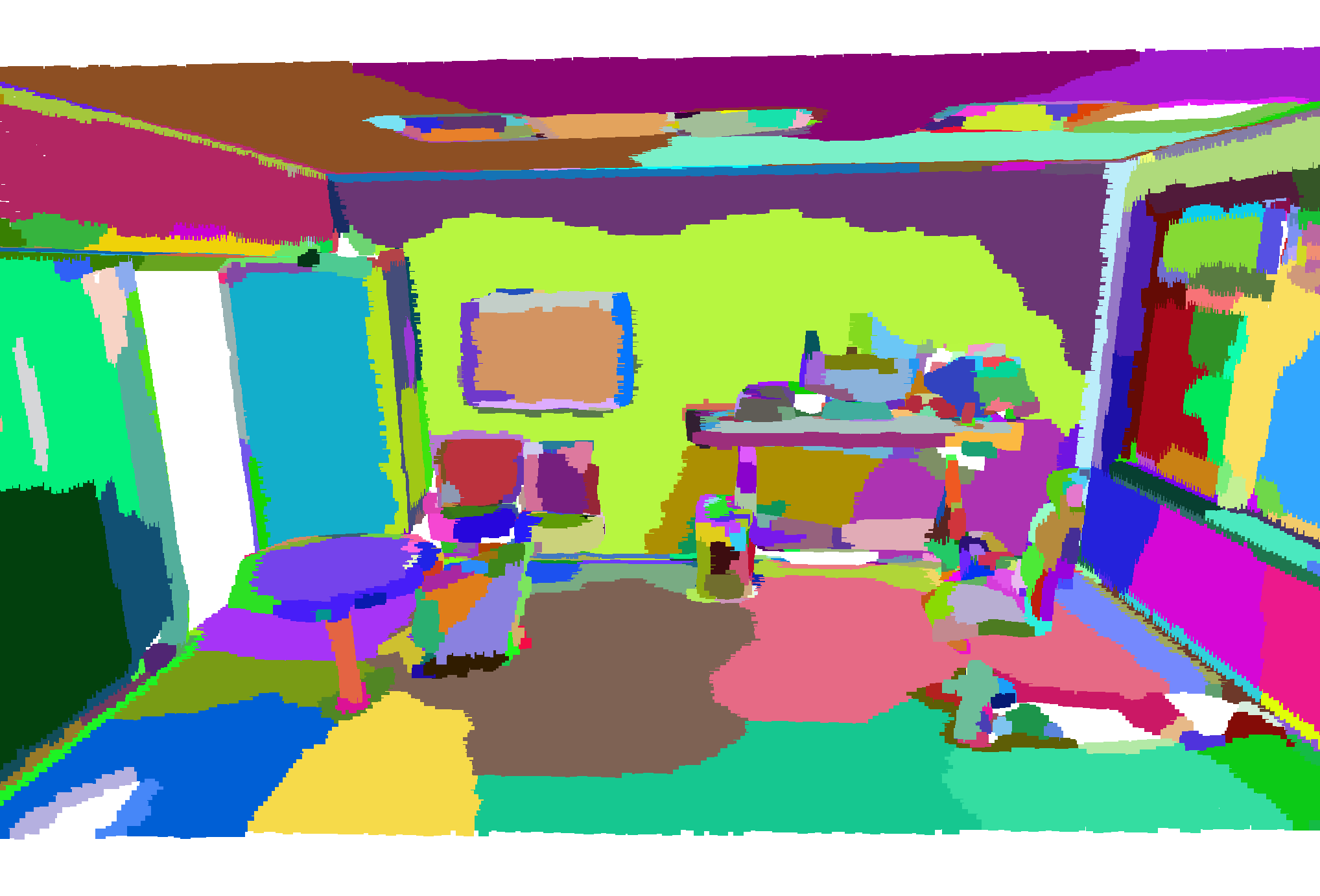}
 \end{tabular}
 \caption{Geometric partitioning}
\label{fig:illustration2_seg}
 }\end{subfigure}
 \\
  \begin{subfigure}{0.47\textwidth}
  \parbox{1\textwidth}{
 \begin{tabular}{c}
 	\includegraphics[width=1\textwidth]{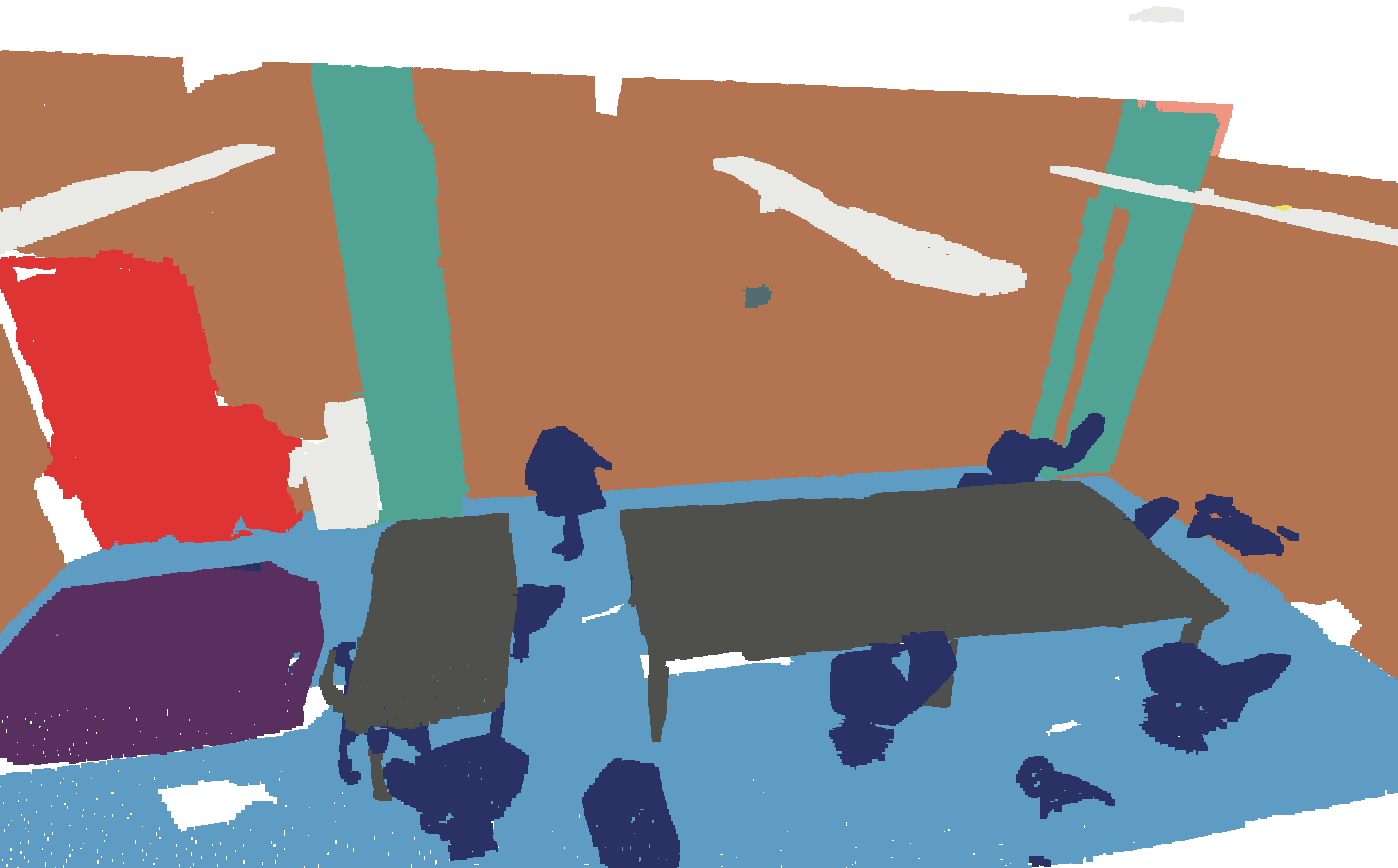}\\ 	  
 	\includegraphics[width=1\textwidth]{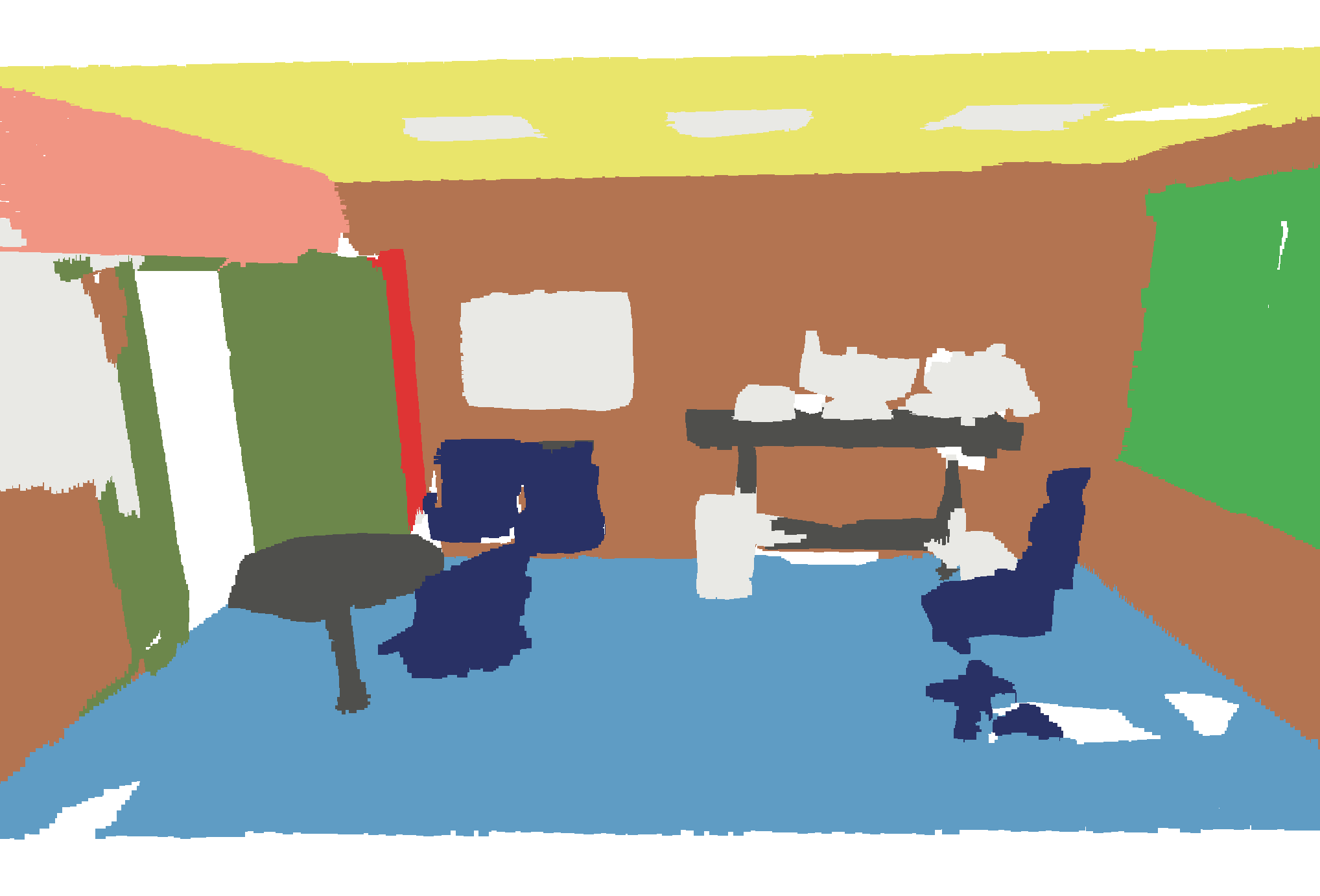}
 \end{tabular}
 \caption{Prediction}
\label{fig:illustration2_pred}
 }\end{subfigure}   
 &
  \begin{subfigure}{0.47\textwidth}
  \parbox{1\textwidth}{
 \begin{tabular}{c}
 	\includegraphics[width=1\textwidth]{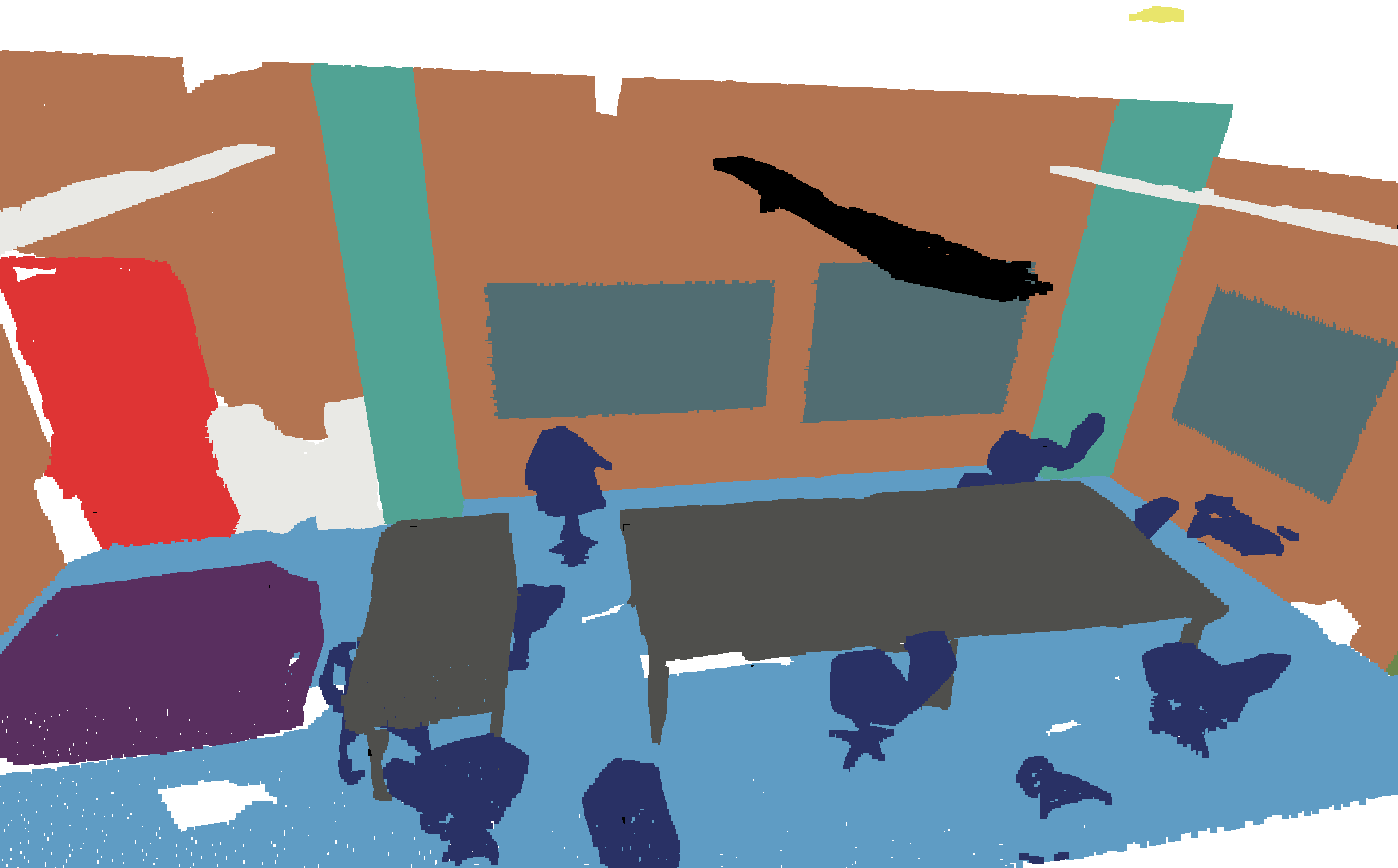}\\ 	  
 	\includegraphics[width=1\textwidth]{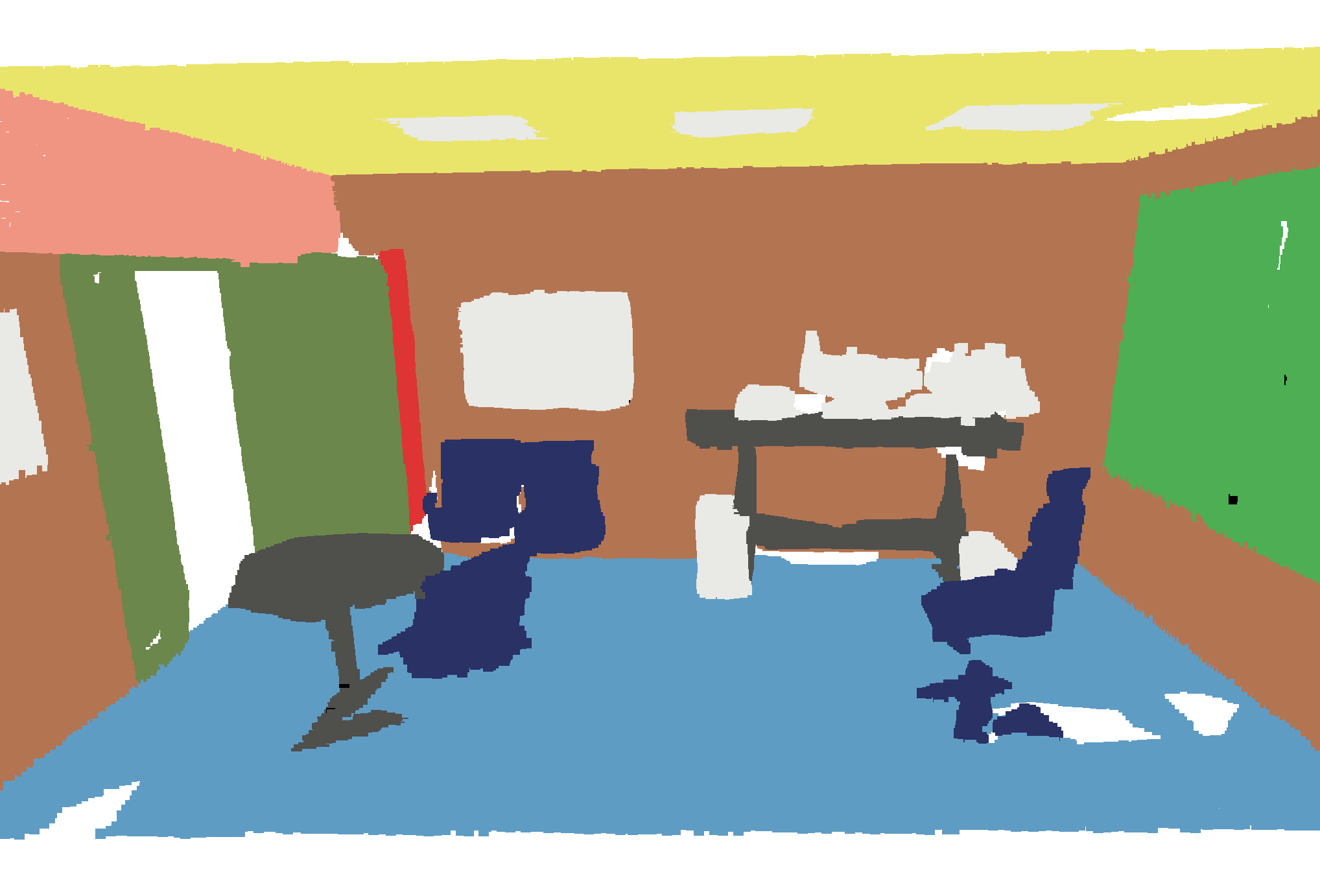}
 \end{tabular}
 \caption{Ground truth}
\label{fig:illustration2_GT}
 }\end{subfigure}  
 \\
 \addlinespace[1.5ex]
 \multicolumn{2}{c}{
	 \definecolor{tempcolor9}{RGB}{233,229,107}
	 \tikz\node at (0,0) [rectangle, minimum width = 3mm, draw = none, fill = tempcolor9] (n){};
	 \hspace{0.2em} ceiling \hspace{0.5em}
	 \definecolor{tempcolor10}{RGB}{95,156,196}
	 \tikz\node at (0,0) [rectangle, minimum width = 3mm, draw = none, fill = tempcolor10] (n){};
	 \hspace{0.2em} floor \hspace{0.5em}
	 \definecolor{tempcolor11}{RGB}{179,116,81}
	 \tikz\node at (0,0) [rectangle, minimum width = 3mm, draw = none, fill = tempcolor11] (n){};
	 \hspace{0.2em} wall \hspace{0.5em}
	 \definecolor{tempcolor12}{RGB}{81,163,148}
	 \tikz\node at (0,0) [rectangle, minimum width = 3mm, draw = none, fill = tempcolor12] (n){};
	 \hspace{0.2em} column \hspace{0.5em}
	 \definecolor{tempcolor13}{RGB}{241,149,131}
	 \tikz\node at (0,0) [rectangle, minimum width = 3mm, draw = none, fill = tempcolor13] (n){};
	 \hspace{0.2em} beam  } \\
\multicolumn{2}{c}{
	 \definecolor{tempcolor14}{RGB}{77,174,84}
	 \tikz\node at (0,0) [rectangle, minimum width = 3mm, draw = none, fill = tempcolor14] (n){};
	 \hspace{0.2em} window \hspace{0.5em}
	 \definecolor{tempcolor15}{RGB}{108,135,75}
	 \tikz\node at (0,0) [rectangle, minimum width = 3mm, draw = none, fill = tempcolor15] (n){};
	 \hspace{0.2em} door \hspace{0.5em}
	 \definecolor{tempcolor16}{RGB}{79,79,76}
	 \tikz\node at (0,0) [rectangle, minimum width = 3mm, draw = none, fill = tempcolor16] (n){};
	 \hspace{0.2em} table \hspace{0.5em}
	 \definecolor{tempcolor17}{RGB}{41,49,101}
	 \tikz\node at (0,0) [rectangle, minimum width = 3mm, draw = none, fill = tempcolor17] (n){};
	 \hspace{0.2em} chair \hspace{0.5em}
	 \definecolor{tempcolor18}{RGB}{223,52,52}
	 \tikz\node at (0,0) [rectangle, minimum width = 3mm, draw = none, fill = tempcolor18] (n){};
	 \hspace{0.2em} bookcase  } \\
\multicolumn{2}{c}{
	 \definecolor{tempcolor19}{RGB}{100,20,100}
	 \tikz\node at (0,0) [rectangle, minimum width = 3mm, draw = none, fill = tempcolor19] (n){};
	 \hspace{0.2em} sofa \hspace{0.5em}
	 \definecolor{tempcolor20}{RGB}{81,109,114}
	 \tikz\node at (0,0) [rectangle, minimum width = 3mm, draw = none, fill = tempcolor20] (n){};
	 \hspace{0.2em} board \hspace{0.5em}
	 \definecolor{tempcolor21}{RGB}{220,220,220}
	  \tikz\node at (0,0) [rectangle, minimum width = 3mm, draw = none, fill = tempcolor21] (n){};
	 \hspace{0.2em} clutter \hspace{0.5em}
	 \definecolor{tempcolor22}{RGB}{0,0,0}
	\tikz\node at (0,0) [rectangle, minimum width = 3mm, draw = none, fill = tempcolor22] (n){};
	 \hspace{0.2em} unlabelled  
 }
\end{tabular}
\end{center}

\caption{Example visualizations on S3DIS. The colors  are chosen randomly for each element of the partition.}
\label{fig:illustration2}
\end{figure*}

\subsection{Stanford Large-Scale 3D Indoor Spaces}
The S3DIS dataset \citep{armeni_cvpr16} consists of 3D RGB point clouds of six floors from three different buildings split into individual rooms.
We evaluate our framework following two dominant strategies found in previous works. As advocated by~\citet{qi2016pointnet}, we perform $6$-fold cross validation with micro-averaging, \ie computing metrics once over the merged predictions of all test folds. Following \citet{tchapmi2017segcloud}, we also report the performance on the fifth fold only (Area 5), corresponding to a building not present in the other folds.
Since some classes in this data set cannot be partitioned purely using geometric features (such as boards or paintings on walls), we concatenate the color information $O$ to the geometric features $F$ for the partitioning step.

The quantitative results are displayed in \tabref{tab:results_S3DIS}, with qualitative results in \figref{fig:illustration2}.
S3DIS is a difficult dataset with hard to retrieve classes such as white boards on white walls and columns within walls. From the quantitative results we can see that our framework performs better than other methods on average. Notably, doors are able to be correctly classified at a higher rate than other approaches, as long as they are open. Indeed, doors are geometrically similar to walls, but their position with respect to the door frame allows our network to retrieve them correctly. On the other hand, the partition merges white boards with walls, depriving the network from the opportunity to even learn to classify them: the IoU of boards for theoretical perfect classification of superpoints (\secref{subsec:ablation}) is only $51.3$.

\textbf{Computation Time.} In \tabref{tab:computation_time}, we report computation time over the different steps of our pipeline for the inference on Area 5 measured on a 4 GHz CPU and GTX 1080 Ti GPU. While the bulk of time is spent on the CPU for partitioning and SPG computation, we show that voxelization as pre-processing leads to a significant speed-up as well as improved accuracy.

\begin{table*}\begin{center}
\begin{tabular}{c}
\begin{subfigure}{1\textwidth} %
\begin{center}
\begin{tabular}{ccccc}\hline
& Method & OA & mAcc & mIoU\\\hline
\multirow{6}{*}{\rotatebox{90}{Area 5}}
& \citet{qi2016pointnet}&--&48.98&41.09\\ 
& \citet{tchapmi2017segcloud}&--&57.35&48.92\\
& ** \citet{tanconv18}& 82.5 & 62.2 & 52.8\\
& ** \citet{huangslice18}&--& 59.42& 51.93\\
& ** \citet{pointcnn18}& 85.91 & 63.86 & 57.26\\
& SPG (Ours)
&\bf86.38&\bf66.50&\bf58.04\\
\hline
\multirow{4}{*}{\rotatebox{90}{6-fold}}
& \citet{qi2016pointnet} \# &78.5&66.2&47.6\\
& \citet{Engelmann17_3dsemseg}&81.1&66.4&49.7\\
& ** \citet{huangslice18}&--& 66.45& 56.47\\
& ** \citet{pointcnn18}& \bf88.14 &\bf75.61& \bf65.39\\
& SPG (Ours)&85.5&73.0&62.1\\\hline
\end{tabular}
\end{center}
\end{subfigure}  
\\\addlinespace[2ex]
\begin{subfigure}{1\textwidth}
\resizebox{\textwidth}{!}{ 
\footnotesize
\begin{tabular}{ccC{0.04\textwidth}C{0.04\textwidth}C{0.04\textwidth}C{0.04\textwidth}C{0.05\textwidth}C{0.05\textwidth}C{0.04\textwidth}C{0.04\textwidth}C{0.04\textwidth}C{0.07\textwidth}C{0.04\textwidth}C{0.04\textwidth}C{0.05\textwidth}}\hline
& Method & ceiling & floor & wall & beam & column & window & door & chair & table & bookcase & sofa & board & clutter\\\hline
\multirow{4}{*}{\rotatebox{90}{Area 5}}
& \citet{qi2016pointnet}&88.80&97.33&69.80&\bf0.05 &3.92&46.26&10.76&52.61&58.93&40.28&5.85&26.38&33.22\\ 
& \citet{tchapmi2017segcloud}&90.06 &96.05 &69.86&0.00&18.37&38.35&23.12&75.89&70.40&58.42&40.88 &12.96&41.60\\
& ** \citet{huangslice18}& \bf93.34 & \bf98.36 & \bf79.18 &0.00 & 15.75 &45.37 &50.10 & 65.52 & 67.87 & 22.45 & 52.45 &\bf41.02 &43.64\\
& SPG (Ours)& 89.35&96.87&78.12&0.0&\bf42.81 &\bf48.93&\bf61.58&\bf84.66&\bf75.41& \bf69.84&\bf52.60&2.10&\bf52.22\\
\hline
\multirow{4}{*}{\rotatebox{90}{6-fold}}
& \citet{qi2016pointnet}  \#&88.0&88.7&69.3&42.4&23.1&47.5&51.6&42.0&54.1&38.2&9.6&29.4&35.2\\
& \citet{Engelmann17_3dsemseg}&90.3&92.1&67.9&44.7&24.2&52.3&51.2&47.4&58.1&39.0&6.9&30.0&41.9\\
& ** \citet{huangslice18}&\bf92.48 &92.83 & \bf78.56&32.75 &34.37&51.62 &68.11& 59.72& 60.13&16.42&\bf50.22 &\bf44.85& 52.03\\
& SPG (Ours)&89.9&\bf95.1&76.4&\bf62.8&\bf47.1&\bf55.3&\bf68.4&\bf73.5&\bf69.2&\bf63.2&45.9&8.7&\bf52.9\\\hline
\end{tabular}
}\end{subfigure} 
\end{tabular}
\end{center}
\caption{Results on the S3DIS dataset on fold ``Area 5'' and micro-averaged over all 6 folds. Intersection over union is shown split per class. Concurrent baselines are denoted with a star. Hash denotes results obtained by \citet{Engelmann17_3dsemseg}.}
\label{tab:results_S3DIS}
\end{table*}
\begin{table}\begin{center}
\begin{tabular}{ccccc}\hline
Step & Full cloud & $2$ cm & $3$ cm & $4$ cm\\\hline
Voxelization & $\num{0}$ &$\num{40}$&$\num{24}$&$\num{16}$\\
Feature computation & $\num{439}$&$\num{194}$&$\num{88}$&$\num{43}$\\
Geometric partition & $\num{3 428}$&$\num{1013}$&$\num{447}$&$\num{238}$\\
SPG computation & $\num{3 800}$&$\num{958}$&$\num{436}$&$\num{252}$\\
Inference & $10 \times 24$ & $10 \times 11$ & $10 \times 6$ & $10 \times 5$\\\hline
Total &$\num{7 907}$&$\num{2315}$&$\num{1055}$&$\num{599}$\\
mIoU $6$-fold&54.1&60.2&62.1&57.1\\\hline
\end{tabular}
\end{center}
\caption{Computation time in seconds for the inference on S3DIS Area 5 ($68$ rooms, $\num{78 649 682}$ points) for different voxel sizes.}
\label{tab:computation_time}
\end{table}
\subsection{Segmentation Baselines} \label{subsec:ablation}
To demonstrate the advantage of our overall pipeline design, we compare it to several baselines. Due to the lack of public ground truth for test sets of Semantic3D, we evaluate on S3DIS with 6-fold cross validation and show comparison of different models to our $\mathrm{Best}$ model in \tabref{tab:results_ablation}.

\textbf{Performance Limits.} The contribution of contextual segmentation can be bounded both from below and above. The lower bound ($\mathrm{Unary}$) is estimated by training PointNet with $d_z=13$ but otherwise the same architecture, denoted as PointNet13, to directly predict class logits, without SPG and GRUs. The upper bound ($\mathrm{Perfect}$) corresponds to assigning each superpoint its ground truth label, and thus sets the limit of performance due to the geometric partition. We can see that contextual segmentation is able to win roughly $22$ mIoU points over unaries, confirming its importance. Nevertheless, the learned model still has room of up to $26$ mIoU points for improvement, while about $12$ mIoU points are forfeited to the semantic inhomogeneity of superpoints.

\textbf{CRFs.} We compare the effect of our GRU+ECC-based network to CRF-based regularization. As a baseline ($\mathrm{iCRF}$), we post-process $\mathrm{Unary}$ outputs by CRF inference over SPG connectivity with scalar transition matrix, as described by \citep{guinard_weakly_2017}. Next ($\mathrm{CRF-ECC}$), we adapt CRF-RNN framework of \citet{Zheng15crf} to general graphs with edge-conditioned convolutions (see \ref{sec:crfecc} for details) and train it with PointNet13 end-to-end. Finally ($\mathrm{GRU13}$), we modify $\mathrm{Best}$ to use PointNet13. We observe that $\mathrm{iCRF}$ barely improves accuracy (+1 mIoU), which is to be expected, since the partitioning step already encourages spatial regularity. $\mathrm{CRF-ECC}$ does better (+15 mIoU) due to end-to-end learning and use of edge attributes, though it is still below $\mathrm{GRU13}$ (+18 mIoU), which performs more complex operations and does not enforce normalization of the embedding. Nevertheless, the 32 channels used in $\mathrm{Best}$ instead of the 13 used in $\mathrm{GRU13}$ provide even more freedom for feature representation (+22 mIoU).

\begin{table}\begin{center}\small
\begin{tabular}{ccc}\hline
Model & mAcc & mIoU \\\hline
$\mathrm{Best}$ & 73.0 & 62.1  \\
$\mathrm{Perfect}$ & 92.7 & 88.2\\
$\mathrm{Unary}$ & 50.8 & 40.0 \\
$\mathrm{iCRF}$ & 51.5 &  40.7 \\
$\mathrm{CRF-ECC}$ & 65.6 & 55.3\\
$\mathrm{GRU13}$ & 69.1 & 58.5 \\
\hline
\end{tabular}
\end{center}
\caption{Comparison to various segmentation baselines on S3DIS (6-fold cross validation).}
\label{tab:results_ablation}
\end{table}

\subsection{Ablation Studies} \label{subsec:ext_ablation}

To better understand the influence of made design decisions and chosen hyperparameters, we perform ablation studies and present their results in \tabref{tab:results_ablation2}. Concretely, we explore the advantages of design choices by individually removing them from $\mathrm{Best}$ in order to compare the framework's performance with and without them.

\textbf{a) Spatial Transformer Network.} While STN makes superpoint embedding orientation invariant, the relationship with surrounding objects are still captured by superedges, which are orientation variant. In practice, STN helps by $4$ mIoU points.

\textbf{b) Geometric Features.} Geometric features $f_i$ are computed in the geometric partition step and can therefore be used in the following learning step for free. While PointNets could be expected to learn similar features from the data, this is hampered by superpoint subsampling, and therefore their explicit use helps (+4 mIoU).

\textbf{c) State Concatenation.} The advantage of state concatenation is compared to considering only the last hidden state in GRU for output ($\Gruy = W_o \Gruh{T+1}{i}$). This accounts for about $5$ mIoU points.

\textbf{d) ECC Variant.} $\mathrm{ECC-VV}$ decreases the performance on the S3DIS dataset by $3$ mIoU points. Nevertheless, it has improved the performance on Semantic3D by $2$ mIoU. 

\textbf{e) Sampling Superpoints.} The main effect of subsampling SPG is regularization by data augmentation. Too small a sample size leads to disregarding contextual information (-4 mIoU) while too large a size leads to overfitting (-2 mIoU). Lower memory requirements at training is an extra benefit. There is no subsampling at test time.

\textbf{f) Long-range Context.} We observe that limiting the range of context information in SPG harms the performance. Specifically, capping distances in $G_{\text{vor}}$ to $1$ m (as used in PointNet~\citep{qi2016pointnet}) or $5$ m (as used in SegCloud\footnote{Furthermore, SegCloud divides the inference into cubes without overlap, possibly causing inconsistencies across boundaries.}~\citep{tchapmi2017segcloud}) worsens the performance of our method (even more on our Semantic 3D validation set).

\textbf{g) Input Gate.} We evaluate the effect of input gating (IG) for GRUs as well as LSTM units. While a LSTM unit achieves higher score than a GRU (-3 mIoU), the proposed IG reverses this situation in favor of GRU (+1 mIoU). Unlike the standard input gate of LSTM, which controls the information flow from the hidden state and input to the cell, our IG controls the input even before it is used to compute all other gates.

\textbf{h) Regularization Strength $\mu$.} We investigate the balance between superpoints' discriminative potential and their homogeneity controlled by parameter $\mu$ . We observe that the system is able to perform reasonably over a range of SPG sizes. 

\textbf{Superedge Attributes.} Finally, in \tabref{tab:edge_ablation} we evaluate empirical importance of individual superedge attributes by removing them from $\mathrm{Best}$. Although no single attribute is crucial, the most being offset deviation (+3 mIoU), without any superedge attributes\footnote{We perform homogeneous regularization by setting all superedge attributes to scalar $1$.} the network performs distinctly worse (-22 mIoU), falling back even below $\mathrm{iCRF}$ to the level of $\mathrm{Unary}$, which validates their design and the overall motivation for SPG.

\begin{table}\small\begin{center}
\begin{tabular}{ccccc}
\emph{a) Spatial transf.} & no &\bf yes & &\\
mIoU & 58.1 & 62.1 & & \\\hline
\emph{b) Geometric features} & no &\bf yes & &\\
mIoU & 58.4  & 62.1 & & \\\hline
\emph{c) State concatenation} & no &\bf yes & &\\
mIoU & 57.7 & 62.1 & & \\\hline
\emph{d) ECC variant} & ECC-VV &\bf ECC & &\\
mIoU & 59.4 & 62.1 & & \\\hline
\emph{e) Max superpoints} & 256 & \bf 512 & 1024 &\\
mIoU & 57.9 & 62.1 & 60.4 &\\\hline
\emph{f) Superedge limit} & 1 m & 5 m & \boldmath$\infty$ & \\
mIoU & 61.0 & 61.3 & 62.1 &\\\hline
\emph{g) Input gate} & LSTM & LSTM+IG & GRU & \bf GRU+IG \\
mIoU & 61.0 & 61.0 & 57.5 & 62.1 \\\hline
\emph{h) Regularization $\mu$} & 0.01 & 0.02 & \bf 0.03 & 0.04 \\
\# superpoints & 785 010& 385 091 & 251 266& 186 108 \\
perfect mIoU & 90.6& 88.2  & 86.6 & 85.2\\
mIoU & 59.1 & 59.2 &   62.1&  58.8 \\
\end{tabular}\end{center}
\caption{Ablation study of design decisions on S3DIS (6-fold cross validation). Our choices in bold.}
\label{tab:results_ablation2}
\vspace*{-0.3cm}
\end{table} 

\begin{table}\begin{center}
\begin{tabular}{ccc}\hline
Attribute set & mAcc & mIoU\\\hline
$\mathrm{Best}$ & 73.0 & 62.1\\
no superedge attributes & 50.1 & 39.9\\
no mean offset & 72.5 & 61.8\\
no offset deviation & 71.7 & 59.3\\
no centroid offset & 74.5 & 61.2\\
no len/surf/vol ratios & 71.2 & 60.7\\
no point count ratio & 72.7 & 61.7 \\
\hline
\end{tabular}
\end{center}
\caption{Ablation study of superedge attributes on S3DIS (6-fold cross validation).}
\label{tab:edge_ablation}
\end{table}

\section{Discussion} \label{sec:crfecc}
\paragraph*{Relation to CRFs.}
In image segmentation, post-processing of convolutional outputs using Conditional Random Fields (CRFs) is widely popular. Several inference algorithms can be formulated as (recurrent) network layers amendable to end-to-end learning~\citep{Zheng15crf,SchwingU15}, possibly with general pairwise potentials \citep{LinSHR16,ChandraK16,LarssonK0ATH17}. While  our method of information propagation shares both these characteristics, our GRUs operate on $d_z$-dimensional intermediate feature space, which is richer and less constrained than low-dimensional vectors representing beliefs over classes, as also discussed in~\citet{gadde16bi}. Such enhanced access to information is motivated by the desire to learn a powerful representation of context, which goes beyond belief compatibilities, as well as the desire to be able to discriminate our often relatively weak unaries (superpixel embeddings). 

We have empirically evaluated these claims in our experiments above. To provide a fair comparison, we have adapted CRF-RNN mean field inference introduced by \citet{Zheng15crf} to use the same pairwise information as our model and denoted it CRF-ECC. Here we describe this adaptation in more detail. The original work proposed a dense CRF with unary potentials $U_i$ ($=\Emb$) and pairwise potentials $\Psi$ defined to be a mixture of $m$ Gaussian kernels as $\Psi_{ij} = \gamma \sum_m \mathbf{w}_m K_m(E_{i,j})$, where $\gamma$ is label compatibility matrix, $\mathbf{w}$ are parameters, and $K$ are fixed Gaussian kernels applied on edge features. 

We replaced this definition of the pairwise term with a Filter generating network $w$ parameterized with weights $W_e$, which generalizes the message passing and compatibility transform steps of Zheng \etal. Furthermore, we use superedge connectivity $\cE$ instead of assuming a complete graph. The pseudo-code is listed in Algorithm~\ref{alg:crfecc}. Its output are marginal probability distributions $Q$. In practice we run the inference for $T=10$ iterations.

\begin{algorithm}
\caption{CRF-ECC}\label{alg:crfecc}
\begin{algorithmic}
\State $Q_i \gets \mathrm{softmax}(U_i)$
\While{not converged}
  \State $\hat{Q}_i \gets \sum_{j \mid (j,i) \in \cE} w(E_{j,i}; W_e) Q_j$
  \State $\breve{Q}_i \gets U_i - \hat{Q}_i$
  \State $Q_i \gets \mathrm{softmax}(\breve{Q}_i)$
\EndWhile
\end{algorithmic}
\end{algorithm}

\paragraph*{Adjacency Graphs.}
In this paper, we use two different adjacency graphs on points of the input clouds: $\Gnn$ in \secref{sec:energy} and $\Gvor$ in \secref{sec:SPGconstruction}.
Indeed, different definitions of adjacency have different advantages. Voronoi adjacency is more suited to capture long-range relationships between points, which is beneficial for the SPG.
Nearest neighbors adjacency tends not to connect objects separated by a small gap. This is desirable for the global energy but tends to produce a SPG with many small connected components, decreasing embedding quality.
Fixed radius adjacency should be avoided in general as it handles the variable density of LiDAR scans poorly.

\paragraph*{Limitations.}
Despite the strong experimental results, it is worth discussing limitations of our method as well. Foremost, the assumed semantic homogeneity of the obtained partitions does not necessarily hold up once applied to real world data. The framework has been evaluated on high-end scanner acquisitions and may not immediately work on point clouds coming from structure-from-motion reconstructions and especially cheap depth cameras, such as Kinect.  %

While the regularization strength $\mu$ provides a powerful and an intuitive handle for the user to adapt the method for a particular dataset, we believe a trainable partitioning step or at least partitioning of learned features might go even further and allow the model to learn \eg SLAM-specific features and provide a better partition. 
Very recently, several attempts at learning features for clustering have been presented: \citet{WolfSKH17} predict altitudes for watershed algorithm, \citet{Tu-CVPR-2018} predict affinities for computing superpixels, and \citet{Verelst18} looks into backpropagation through SLIC, a popular superpixel algorithm. This makes us speculate that features for global energy term in Equation~\ref{eq:minimal_partition} could be learned in a similar spirit. Nevertheless, end-to-end training may remain challenging from the engineering point of view on the current hardware simply due to the scale of the full problem.

\section{Conclusion}
We presented a deep learning framework for performing semantic segmentation of large point clouds based on a partition into simple shapes. We showed that SPGs allow us to use effective deep learning tools, which would not be able to handle the data volume otherwise. Our method significantly improves on the state of the art on two publicly available datasets. Our experimental analysis suggested that future improvements can be made in both partitioning and learning deep contextual classifiers. The source code has been published at \url{https://github.com/loicland/superpoint_graph}.

\chapter{Generation of Small Graphs} \label{chap:iclr18}

\section{Introduction} \label{sec:genintro}

Generative models capture data distribution in an unsupervised way and may allow us to draw samples from it. In the recent years, deep generative models have gone through fast-paced advances leading to massive rise in the realism of generated samples. They have found use in a myriad of problems and applications, including image super-resolution~\citep{LedigTHCCAATTWS17}, domain or modality transfer~\citep{IsolaZZE17, ReedAYLSL16}, anomaly detection~\citep{SchleglSWSL17}, or proposing new molecules in drug discovery as in this chapter. The most popular model families include generative adversarial networks (GANs)~\citep{GoodfellowGAN14}, posing the training process as a game between a generator and a discriminator, variational autoencoders (VAEs)~\citep{vae}, approximating data likelihood by its variational lower bound, and auto-regressive networks~\citep{SutskeverMH11}, learning the conditional distribution of each next sampling step.

Progress in deep learning on graphs, on the other hand, has been nearly exclusively concentrated on learning graph embedding tasks before the submission of our work, \ie encoding an input graph into a vector representation. Hence, it is an intriguing question how one can transfer the achievements in generative models for images and text to the domain of graphs, \ie their decoding from a vector representation.

However, learning to generate graphs is a difficult problem for methods based on gradient optimization, as graphs are discrete structures.  Unlike sequence (text) generation, graphs can have arbitrary connectivity and it is not trivial to choose how to linearize their construction in a sequence of steps. On the other hand, learning the order for incremental construction involves discrete decisions, which are not differentiable.
 
In this work, we propose to sidestep these hurdles by having the decoder output a probabilistic fully-connected graph of a predefined maximum size directly at once. In a probabilistic graph, the existence of nodes and edges, as well as their attributes, are modeled as independent random variables. The method is formulated in the framework of variational autoencoders by \citet{vae}.

We demonstrate our method, coined GraphVAE, in cheminformatics on the task of molecule generation. Molecular datasets are a challenging but convenient testbed for our generative model, as they easily allow for both qualitative and quantitative tests of decoded samples. Past generative models have operated on textual input data only. An advantage of a graph-based decoder compared to text-based decoder is the possibility to predict detailed attributes of atoms and bonds in addition to the base structure.

This chapter is largely based on our ICLR workshop paper~\citep{simonovsky2018towards} and follow-up ICANN 2018 publication~\citep{simonovsky2018graphvae}. While our method is applicable for generating smaller graphs only and its practical performance leaves space for improvement, we believe our work at the time of its submission was an important initial step towards powerful and efficient graph decoders with the following contributions:

\begin{itemize}
\item We present one of the first methods for graph generation using deep learning. Our method can generate graphs of variable though limited sizes without any step-wise supervision.
\item We evaluate on two molecular datasets, offering a better selection of valid but at the same time diverse samples on QM9 dataset \citep{qm9} than previous text generation-based methods. We also suggest a conditional setting of our model as an alternative way of controlling the molecule generation process. 
\end{itemize}

\section{Related work} \label{sec:genrelated}

\paragraph*{De Novo Molecular Design.} In the pharmaceutical industry, generative models may become promising for computational design of molecules fulfilling certain criteria (such as solubility) and optimizing certain desirable properties (such as binding strength to a protein). The hypothesis is that optimization in the continuous space of models or embeddings is easier than in the original discrete chemical space. In the context of deep generative models, the field has explored mainly two strategies for navigating the chemical space, either by fine-tuning a generative model, \eg with simple hill-climbing~\citep{segler17} or with reinforcement learning~\citep{OlivecronaBEC17}, or by optimization in the continuous embedding space~\citep{Bombarelli16}. Here, we suggest to directly drive certain properties by introducing a conditional version of the decoder, also concurrently proposed by \citet{LiDGM18}.
 
While molecules have an intuitive and richer representation as graphs, the field has had to resort to textual representations with fixed syntax, \eg so-called SMILES strings \citep{Weininger88}, to exploit recent progress made in text generation with RNNs \citep{OlivecronaBEC17,segler17,Bombarelli16}. As their syntax is brittle, many invalid strings tend to be generated, which has been recently addressed by \citet{KusnerPH17} by incorporating context-free grammar rules into decoding to mask out invalid steps, effectively generating a parse tree. While encouraging, their approach does not guarantee semantic (chemical) validity, similarly to our method. In a concurrent work, \citet{sd-vae} propose to structure the string decoding space even more tightly by using attribute grammars, which allow to capture semantic rules as well and thus generate higher proportions of valid outputs.

\paragraph*{Discrete Data Decoders.} Text is the most common discrete representation. Generative models there are usually trained in a maximum likelihood fashion by teacher forcing \citep{WilliamsZ89}, which avoids the need to backpropagate through output discretization by feeding the ground truth instead of the past sample at each step. \citet{bengio2015scheduled} argued this may lead to expose bias, \ie possibly reduced ability to recover from own mistakes. Recently, efforts have been made to overcome this problem. Notably, computing a differentiable approximation using Gumbel distribution \citep{KusnerH16} or bypassing the problem by learning a stochastic policy in reinforcement learning \citep{SeqGAN}. Our work deals with the non-differentiability problem by formulating the loss on a probabilistic graph.

\paragraph*{Graph Decoders.} Related work pre-dating deep learning includes random graphs \citep{erdos1960evolution,barabasi1999emergence}, stochastic blockmodels \citep{Snijders1997}, or state transition matrix learning \citep{GongX03}.

Graph generation was largely unexplored in deep learning before 2018. The closest work to ours is by \citet{johnson2017learning}, who incrementally constructs a probabilistic (multi)graph as a world representation  according to a sequence of input sentences to answer a query. While our model also outputs a probabilistic graph, we do not assume having a prescribed order of construction transformations available and we formulate the learning problem as an autoencoder. 

\citet{Xu17GraphGen} learns to produce a scene graph from an input image. They construct a graph from a set of object proposals, provide initial embeddings to each node and edge, and use message passing to obtain a consistent prediction. In contrast, our method is a generative model which produces a probabilistic graph from a single vector, without specifying the number of nodes or the structure explicitly. 

Multiple concurrent and follow up methods have then appeared, which we briefly review here. These works do not sample edges in an independent way and outperform the method presented here.

\citet{LiDGM18} relax several assumptions made by \citet{johnson2017learning} and present the first self-contained auto-regressive graph decoder. The graphs are incrementally constructed by a sequence of decisions on adding nodes and edges, predicted by RNN running on embeddings continuously updated throughout the construction. The network is trained with random or fixed node orderings and the authors report difficulties due to extensive length of sequences. \citet{graphrnn} improve on this by proposing a more concise graph building strategy with graph-level and edge-level RNNs and introduce a breadth-first-search node-ordering scheme, which limits the number of possible permutations and steps, to significantly improve scalability allowing to handle hundreds of nodes and edges. \citet{LiZL18} propose even more lightweight construction strategy (graph-level RNN or even casting it as Markov process) and use depth-first ordering with random noise. \citet{YouPolicy18} then casted the construction as a Markov process in a reinforcement learning framework with a GAN loss.

\citet{samanta18} also train to generate probabilistic graphs in a VAE framework. However, they model graph embedding as a set of node embeddings rather than a single opaque vector, which allows them to avoid graph matching for loss computation but also means that node embeddings must be sampled independently during inference and the number of them is not part of the latent code. Edge probabilities are modeled as functions of node embeddings, similarly to  \citet{kipf2016variational}, and invalid steps are masked out during edge decoding to improve the ratio of semantically valid graphs, similarly to \citet{KusnerPH17}. \citet{liu18} further builds up on \citet{samanta18} and, among other minor improvements, adds conditioning on the embedding of the current state of the graph to the edge sampling function, making it a Markov process unlike in \citet{LiDGM18, graphrnn}, where decisions depend on the generation history. \citet{molgan18} generate graphs at once as in our work but use GAN framework for training, which alleviates the need to use graph matching but makes the model more susceptible to mode collapse.

For the specific task of molecular graph generation, \citet{JinBJ18} builds on the insight that molecules can be decomposed into trees of fragments. From a combined tree and graph embedding, a molecule is decoded by reconstructing the tree, followed by a series of greedy choices of compatible fragments with a scoring function involving the latent embedding as a context. The masking of invalid combinations prevents creation of invalid molecules.

\section{Method} \label{sec:genmethod}

\def\enc{{q_{\phi}(\mathbf{z}|G)}}
\def\dec{{p_{\theta}(G|\mathbf{z})}}
\def\prior{{p(\mathbf{z})}}

\begin{figure*}[bt]
\centering
\includegraphics[width=1\linewidth]{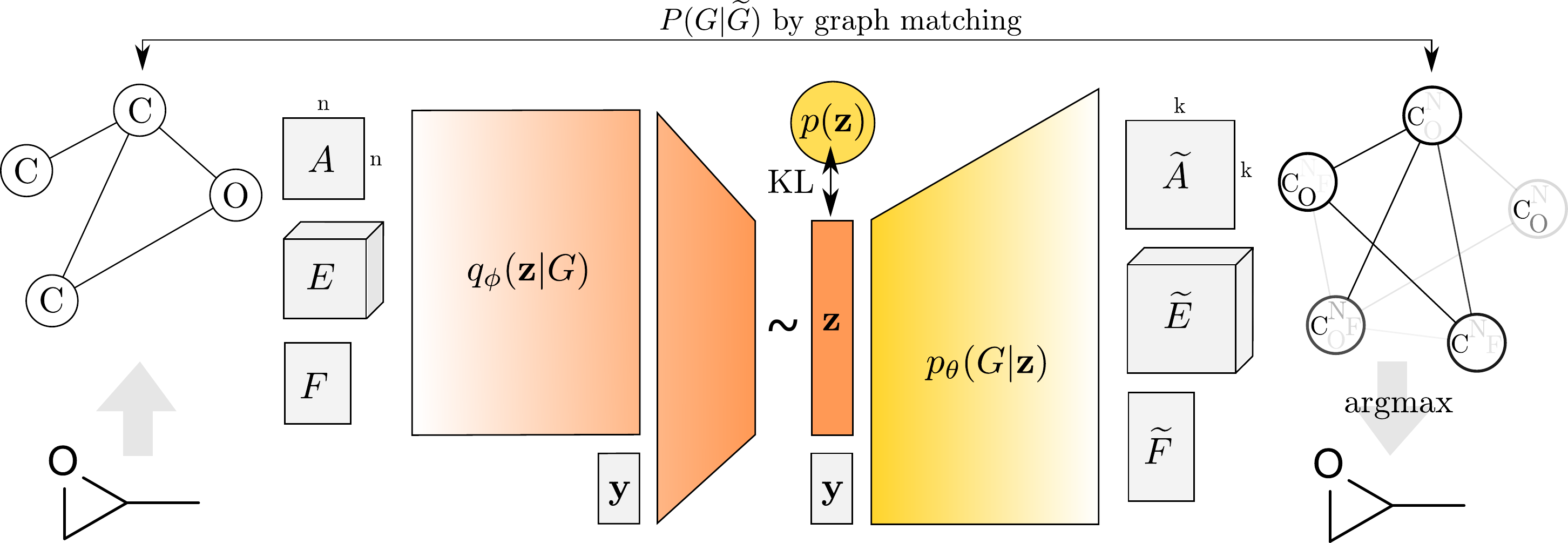}
\caption{\label{fig:net} 
Illustration of the proposed variational graph autoencoder. Starting from a discrete attributed graph $G=(A,E,F)$ on $n$ nodes (\eg a representation of propylene oxide), stochastic graph encoder $\enc$ embeds the graph into continuous representation $\mathbf{z}$. Given a point in the latent space, our novel graph decoder $\dec$ outputs a probabilistic fully-connected graph $\widetilde{G}=(\widetilde{A},\widetilde{E},\widetilde{F})$ on predefined $k \geq n$ nodes, from which discrete samples may be drawn. The process can be conditioned on label $\textbf{y}$ for controlled sampling at test time. Reconstruction ability of the autoencoder is facilitated by approximate graph matching for aligning $G$ with $\widetilde{G}$.
}
\end{figure*}

We approach the task of graph generation by devising a neural network able to translate vectors in a continuous code space to graphs. Our main idea is to output a probabilistic fully-connected graph and use a standard graph matching algorithm to align it to the ground truth. The proposed method is formulated in the framework of variational autoencoders (VAE) by \citet{vae}, although other forms of regularized autoencoders would be equally suitable \citep{aae,gmmn}. We briefly recapitulate VAE below and continue with introducing our novel graph decoder together with an appropriate objective.
 
\subsection{Variational Autoencoder} \label{subsec:genmethod-vae}

Let $G=(A,E,F)$ be a graph specified with its adjacency matrix $A$, edge attribute tensor $E$, and node attribute matrix $F$. We wish to learn an encoder and a decoder to map between the space of graphs $G$ and their continuous embedding $\mathbf{z} \in \mathbb{R}^c$, see Figure~\ref{fig:net}. In the probabilistic setting of a VAE, the encoder is defined by a variational posterior $\enc$ and the decoder by a generative distribution $\dec$, where $\phi$ and $\theta$ are learned parameters. Furthermore, there is a prior distribution $\prior$ imposed on the latent code representation as a regularization; we use a simplistic isotropic Gaussian prior $\prior = N(0,I)$. The whole model is trained by minimizing the upper bound on negative log-likelihood $-\log p_{\theta}(G)$ \citep{vae}:   %
\begin{equation}
\mathcal{L}(\phi, \theta; G) = \mathbb{E}_{\enc}[-\log \dec] + \mathrm{KL}[\enc || \prior] \label{eq:elbo}
\end{equation}
The first term of $\mathcal{L}$, the reconstruction loss, enforces high similarity of sampled generated graphs to the input graph $G$. The second term, KL-divergence, regularizes the code space to allow for sampling of $\mathbf{z}$ directly from $\prior$ instead from $\enc$ later. The dimensionality of $\mathbf{z}$ is usually fairly small so that the autoencoder is encouraged to learn a high-level compression of the input instead of learning to simply copy any given input. While the regularization is independent on the input space, the reconstruction loss must be specifically designed for each input modality. In the following, we introduce our graph decoder together with an appropriate reconstruction loss. %

\subsection{Probabilistic Graph Decoder} \label{subsec:genmethod-dec}

Graphs are discrete objects, ultimately. While this does not pose a challenge for encoding, demonstrated by the recent developments in graph convolution networks, graph generation was an open problem until early 2018. In a related task of text sequence generation, the currently dominant approach is character-wise or word-wise prediction \citep{BowmanVVDJB16}. However, graphs can have arbitrary connectivity and there is no clear way how to linearize their construction in a sequence of steps\footnote{While algorithms for canonical graph orderings are available \citep{McKay2014nauty}, \citet{vinyals2015order} empirically found out that the linearization order matters when learning on sets. However, see also the evolution of node ordering strategies used by follow up works discussed in Section~\ref{sec:genrelated}}. On the other hand, iterative construction of discrete structures during training without step-wise supervision involves discrete decisions, which are not differentiable and therefore problematic for back-propagation.

Fortunately, the task can become much simpler if we restrict the domain to the set of all graphs on maximum $k$ nodes, where $k$ is fairly small (in practice up to the order of tens). Under this assumption, handling dense graph representations is still computationally tractable. We propose to make the decoder output a probabilistic fully-connected graph $\widetilde{G}=(\widetilde{A},\widetilde{E},\widetilde{F})$ on $k$ nodes at once. This effectively sidesteps both problems mentioned above. 

In probabilistic graphs, the existence of nodes and edges is modeled as Bernoulli variables, whereas node and edge attributes are multinomial variables. While not discussed in this work, continuous attributes could be easily modeled as Gaussian variables represented by their mean and variance. We assume all variables to be independent.

Each tensor of the representation of $\widetilde{G}$ has thus a probabilistic interpretation.  Specifically, the predicted adjacency matrix $\widetilde{A} \in [0,1]^{k \times k}$ contains both node probabilities $\widetilde{A}_{a,a}$ and edge probabilities $\widetilde{A}_{a,b}$ for nodes $a \neq b$. The edge attribute tensor $\widetilde{E} \in \mathbb{R}^{k \times k \times d_e}$ indicates class probabilities for edges and, similarly, the node attribute matrix $\widetilde{F} \in \mathbb{R}^{k \times d_n}$ contains class probabilities for nodes.  %

The decoder itself is deterministic. Its architecture is a simple multi-layer perceptron (MLP) with three outputs in its last layer. Sigmoid activation function is used to compute $\widetilde{A}$, whereas edge- and node-wise softmax is applied to obtain $\widetilde{E}$ and $\widetilde{F}$, respectively. At test time, we are often interested in a (discrete) point estimate of $\widetilde{G}$, which can be obtained by taking edge- and node-wise argmax in $\widetilde{A},\widetilde{E}$, and $\widetilde{F}$. Note that this can result in a discrete graph on less than $k$ nodes.

\subsection{Reconstruction Loss} \label{subsec:genmethod-loss}

\def\Gst{{G}} 
\def\Gsp{{\widetilde{G}}} 

Given a particular instance of a discrete input graph $\Gst$ on $n \leq k$ nodes and its probabilistic reconstruction $\Gsp$ on $k$ nodes, evaluation of Equation~\ref{eq:elbo} requires computation of likelihood $p_{\theta}(\Gst | \mathbf{z}) = P(\Gst | \Gsp)$. 

Since no particular ordering of nodes is imposed in either $\Gsp$ or $\Gst$ and matrix representation of graphs is not invariant to permutations of nodes, comparison of two graphs is hard. However, approximate graph matching described further in Subsection~\ref{subsec:genmethod-match} can obtain a binary assignment matrix $X \in \{0,1\}^{k \times n}$, where $X_{a,i}=1$ only if node $a \in \Gsp$ is assigned to $i \in \Gst$ and $X_{a,i}=0$ otherwise.

Knowledge of $X$ allows to map information between both graphs. Specifically, input adjacency matrix is mapped to the predicted graph as $A' = X A X^T$, whereas the predicted node attribute matrix and slices of edge attribute matrix are transferred to the input graph as $\widetilde{F}' = X^T \widetilde{F}$  and $\widetilde{E}'_{\cdot,\cdot,l} = X^T \widetilde{E}_{\cdot,\cdot,l} X$. The maximum likelihood estimates, \ie cross-entropy, of respective variables are as follows:
\begin{equation}
\begin{aligned}
\log p(A'|\mathbf{z}) &= 1/k \sum_{a} A'_{a,a} \log \widetilde{A}_{a,a} + (1-A'_{a,a}) \log (1-\widetilde{A}_{a,a}) + \\
& + 1/{k(k-1)} \sum_{a \neq b} A'_{a,b} \log \widetilde{A}_{a,b} + (1-A'_{a,b}) \log (1-\widetilde{A}_{a,b}) \\
\log p(F|\mathbf{z}) &= 1/n \sum_{i} \log F^T_{i,\cdot} \widetilde{F}'_{i,\cdot} \\
\log p(E|\mathbf{z}) &= 1/(||A||_1 - n) \sum_{i \neq j} \log E^T_{i,j,\cdot} \widetilde{E}'_{i,j,\cdot}
\end{aligned}
\end{equation}
where we assumed that $F$ and $E$ are encoded in one-hot notation. The formulation considers existence of both matched and unmatched nodes and edges but attributes of only the matched ones. Furthermore, averaging over nodes and edges separately has shown beneficial in training as otherwise the edges dominate the likelihood. The overall reconstruction loss is a weighed sum of the previous terms:
\begin{equation}
-\log p(G|\mathbf{z}) = - \lambda_A \log p(A'|\mathbf{z}) - \lambda_F \log p(F|\mathbf{z}) - \lambda_E \log p(E|\mathbf{z}) \label{eq:logli}
\end{equation}

\subsection{Graph Matching} \label{subsec:genmethod-match}

The goal of (second-order) graph matching is to find correspondences $X \in \{0,1\}^{k \times n}$ between nodes of graphs $G$ and $\widetilde{G}$ based on the similarities of their node pairs $S: (i,j)\times(a,b) \to  \mathbb{R}^+$ for $i,j \in G$ and $a,b \in \widetilde{G}$. It can be expressed as integer quadratic programming problem of similarity maximization over $X$ and is typically approximated by relaxation of $X$ into continuous domain: $X^* \in [0,1]^{k \times n}$ \citep{Cho}. For our use case, the similarity function is defined as follows:
\begin{equation}
\begin{aligned}
S((i,j),(a,b)) &= (E^T_{i,j,\cdot} \widetilde{E}_{a,b,\cdot})  A_{i,j} \widetilde{A}_{a,b} \widetilde{A}_{a,a} \widetilde{A}_{b,b} [i \neq j \wedge a \neq b] + \\
& + (F^T_{i,\cdot} \widetilde{F}_{a,\cdot})  \widetilde{A}_{a,a} [i=j \wedge a=b]
\end{aligned}
\label{eq:affin}
\end{equation}
The first term evaluates similarity between edge pairs and the second term between node pairs, $[\cdot]$ being the Iverson bracket. Note that the scores consider both feature compatibility ($\widetilde{F}$ and $\widetilde{E}$) and existential compatibility ($\widetilde{A}$), which has empirically led to more stable assignments during training. To summarize the motivation behind both Equations~\ref{eq:logli} and \ref{eq:affin}, our method aims to find the best graph matching and then further improve on it by gradient descent on the loss. Given the stochastic way of training deep networks, we argue that solving the matching step only approximately is sufficient. This is conceptually similar to the approach for learning to output unordered sets by \citep{vinyals2015order}, where the closest ordering of the training data is sought.

In practice, we are looking for a graph matching algorithm robust to noisy correspondences which can be easily implemented on GPU in batch mode. Max-pooling matching (MPM) by \citet{Cho} is a simple but effective algorithm following the iterative scheme of power methods. It can be used in batch mode if similarity tensors are zero-padded, \ie $S((i,j),(a,b))=0$ for $n < i,j \leq k$, and the amount of iterations is fixed.

Max-pooling matching outputs continuous assignment matrix $X^*$. Unfortunately, attempts to directly use $X^*$ instead of $X$ in Equation~\ref{eq:logli} performed badly, as did experiments with direct maximization of $X^*$ or soft discretization with softmax or straight-through Gumbel softmax \citep{JangGP16} or using Sinkhorn iterations to arrive at doubly-stochastic matrices \citep{sinkhorn1967concerning}. We therefore discretize $X^*$ to $X$ using Hungarian algorithm to obtain a strict one-on-one mapping\footnote{Some predicted nodes are not assigned for $n<k$. Our current implementation performs this step on CPU although a GPU version has been published \citep{DateN16}.}. While this operation is non-differentiable, gradient can still flow to the decoder directly through the loss function and training convergence proceeds without problems. Note that this approach is often taken in works on object detection, \eg \citet{stewart2016end}, and more recently point cloud generation \citet{FanSG17}, where a set of detections needs to be matched to a set of ground truth entities and treated as fixed before computing a differentiable loss.

\subsection{Further Details} \label{subsec:genmethod-det}

\paragraph*{Encoder.} A feed forward network with edge-conditioned graph convolutions (ECC) as presented in Chapter~\ref{chap:cvpr17} is used as encoder, although any other graph embedding method is applicable. As our edge attributes are categorical, a single linear layer for the filter generating network in ECC is sufficient. Due to smaller graph sizes no pooling is used in encoder except for the global one, for which we employ gated pooling by \citet{yujia16}. As usual in VAE, we formulate the encoder as probabilistic and enforce Gaussian distribution of $\enc$ by having the last encoder layer outputs $2c$ features interpreted as mean and variance, allowing to sample $\mathbf{z}_l \sim N(\mathbf{\mu}_l(G), \mathbf{\sigma}_l(G))$ for $l \in {1,..,c}$ using the re-parameterization trick \citep{vae}.

\paragraph*{Disentangled Embedding.} In practice, rather than random drawing of graphs, one often desires more control over the properties of generated graphs. In such case, we follow \citet{cvae} and condition both encoder and decoder on label vector $\mathbf{y}$ associated with each input graph $G$. Decoder $p_{\theta}(G|\mathbf{z},\mathbf{y})$ is fed a concatenation of $\mathbf{z}$ and $\mathbf{y}$, while in encoder $q_{\phi}(\mathbf{z}|G,\mathbf{y})$, $\mathbf{y}$ is concatenated to every node's features just before the graph pooling layer. If the size of latent space $c$ is small, the decoder is encouraged to exploit information in the label. If the latent space is large, however, one might better explicitly minimize mutual information between the condition and the embedding, an important detail which we nevertheless leave for future work.

\def\xx{{\mathbf{x}}}

\paragraph*{Max-Pooling Matching} We briefly review max-pooling matching algorithm of \citet{Cho} for clarity. In its relaxed form, a continuous correspondence matrix $X^* \in [0,1]^{k \times n}$ between nodes of graphs $G$ and $\widetilde{G}$ is determined based on similarities of node pairs $i,j \in G$ and $a,b \in \widetilde{G}$ represented as matrix elements $S_{ia;jb} \in \mathbb{R}^+$. 

Let $\xx^*$ denote the column-wise replica of $X^*$ (with overloaded indexing notation $\xx^*_{ia}=X^*_{ia}$). The relaxed graph matching problem is expressed as quadratic programming task $\mathbf{x}^* = \arg \max_{\xx} \xx^T S \xx$ such that $\sum_{i=1}^n \xx_{ia} \leq 1$, $\sum_{a=1}^k \xx_{ia} \leq 1$, and $\xx \in [0,1]^{kn}$. The optimization strategy of choice is derived to be equivalent to the power method with iterative update rule $\xx^{t+1} = S \xx^{(t)}/||S \xx^{t}||_2$. The starting correspondences $\xx^{0}$ are initialized as uniform and the rule is iterated until convergence; in our use case we run for a fixed amount of iterations.

In the context of graph matching, the matrix-vector product $S \xx$ can be interpreted as sum-pooling over match candidates: $\xx_{ia} \leftarrow \xx_{ia} S_{ia;ia} + \sum_{j \in N_i} \sum_{b \in N_a} \xx_{jb}S_{ia;jb}$, where $N_i$ and $N_a$ denote the set of neighbors of node $i$ and $a$.
The authors argue that this formulation is strongly influenced by uninformative or irrelevant elements and propose a more robust max-pooling version, which considers only the best pairwise similarity from each neighbor: $\xx_{ia} \leftarrow \xx_{ia} S_{ia;ia} + \sum_{j \in N_i} \max_{b \in N_a} \xx_{jb}S_{ia;jb}$.

\section{Evaluation} \label{sec:exper}

We demonstrate our method for the task of molecule generation by evaluating on two large public datasets of organic molecules, QM9 and ZINC.

\subsection{Application in Cheminformatics}

Quantitative evaluation of generative models of images and texts has been troublesome \citep{TheisOB15}, as it very difficult to measure realness of generated samples in an automated and objective way. Thus, researchers frequently resort there to qualitative evaluation and embedding plots. However, qualitative evaluation of graphs can be very unintuitive for humans to judge unless the graphs are planar and fairly simple. 

\defcitealias{rdkit}{RDKit} 	
Fortunately, we found graph representation of molecules, as undirected graphs with atoms as nodes and bonds as edges, to be a convenient testbed for generative models. On one hand, generated graphs can be easily visualized in standardized structural diagrams. On the other hand, chemical validity of graphs, as well as many further properties a molecule can fulfill, can be checked using software packages (\texttt{SanitizeMol} in \citetalias{rdkit}) or simulations. This makes both qualitative and quantitative tests possible.

Chemical constraints on compatible types of bonds and atom valences make the space of valid graphs complicated and molecule generation challenging. In fact, a single addition or removal of edge or change in atom or bond type can make a molecule chemically invalid. Comparably, flipping a single pixel in MNIST-like number generation problem is of no issue.

To help the network in this application, we introduce three remedies. First, we make the decoder output symmetric $\widetilde{A}$ and $\widetilde{E}$ by predicting their (upper) triangular parts only, as undirected graphs are sufficient representation for molecules.
Second, we use prior knowledge that molecules are connected and, at test time only, construct maximum spanning tree on the set of probable nodes $\{a: \widetilde{A}_{a,a} \geq 0.5\}$ in order to include its edges $(a,b)$ in the discrete pointwise estimate of the graph even if $\widetilde{A}_{a,b}<0.5$ originally. Third, we do not generate Hydrogen explicitly and let it be added as "padding" during chemical validity check~\footnote{The implicit Hydrogen count is calculated as the difference between the valence of a particular atom and the sum of its bonds to other heavy (non-Hydrogen) atoms. Typically, the number of heavy atoms is approximately half the total number of atoms in organic molecules. Thus, implicit Hydrogen makes molecule representations considerably smaller, saving memory and running time. In addition, such a padding is forgiving to decoders generating less bonds than required by the respective valences. Note that this does not affect the reconstruction error, though.}.

\subsection{QM9 Dataset} \label{subsec:qm9}

QM9 dataset \citep{qm9} contains about 134k organic molecules of up to 9 heavy (non Hydrogen) atoms with 4 distinct atomic numbers and 4 bond types, we set $k=9$, $d_e=4$ and $d_n=4$. We set aside 10k samples for testing and 10k for validation (model selection).

We compare our unconditional model to the character-based generator of \citet{Bombarelli16} (CVAE) and the grammar-based generator of \citet{KusnerPH17} (GVAE). We used the code and architecture in \citet{KusnerPH17} for both baselines, adapting the maximum input length to the smallest possible. In addition, we demonstrate a conditional generative model for an artificial task of generating molecules given a histogram of heavy atoms as 4-dimensional label $\mathbf{y}$, the success of which can be easily validated.

\paragraph*{Setup.} The encoder has two graph convolutional layers (32 and 64 channels) with identity connection, batchnorm, and ReLU; followed by the graph-level output formulation in Equation~7 of \citet{yujia16} with auxiliary networks being a single fully connected layer (FCL) with 128 output channels; finalized by a FCL outputting $(\mathbf{\mu}, \mathbf{\sigma})$. The decoder has 3 FCLs (128, 256, and 512 channels) with batchnorm and ReLU; followed by parallel triplet of FCLs to output graph tensors. We set $c=40$, $\lambda_A=\lambda_F=\lambda_E=1$, batch size 32, 75 MPM iterations and train for 25 epochs with Adam with learning rate 1e-3 and $\beta_1$=0.5.

\paragraph*{Embedding Visualization.} To visually judge the quality and smoothness of the learned embedding $\mathbf{z}$ of our model, we may traverse it in two ways: along a slice and along a line. For the former, we randomly choose two $c$-dimensional orthonormal vectors and sample $\mathbf{z}$ in regular grid pattern over the induced 2D plane. For the latter, we randomly choose two molecules $G^{(1)}, G^{(2)}$ of the same label from test set and interpolate between their embeddings $\mathbf{\mu}(G^{(1)}), \mathbf{\mu}(G^{(2)})$. This also evaluates the encoder, and therefore benefits from low reconstruction error.

\begin{figure*}[bt]
\centering
\includegraphics[width=0.49\linewidth]{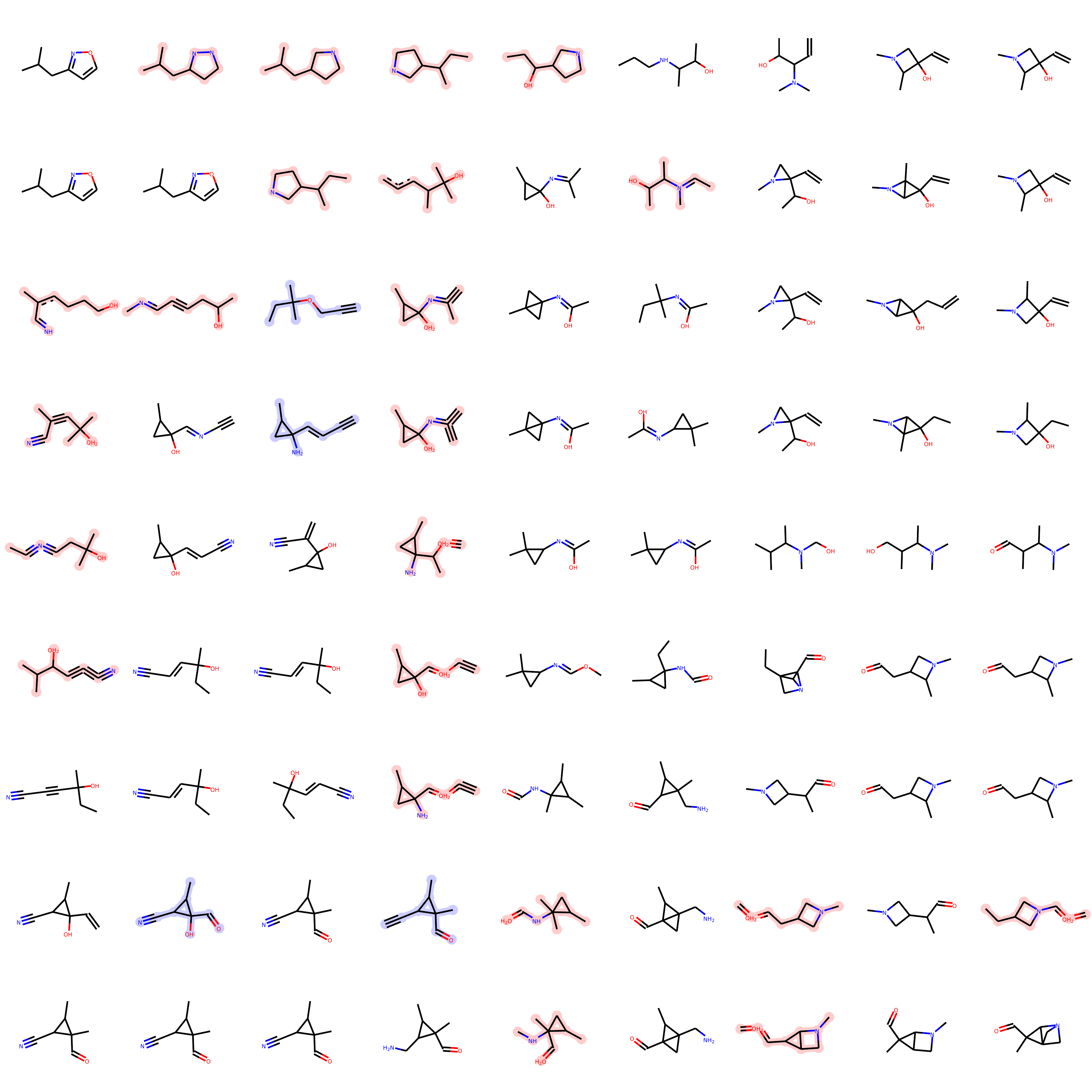}
\hfill\vline\hfill
\includegraphics[width=0.49\linewidth]{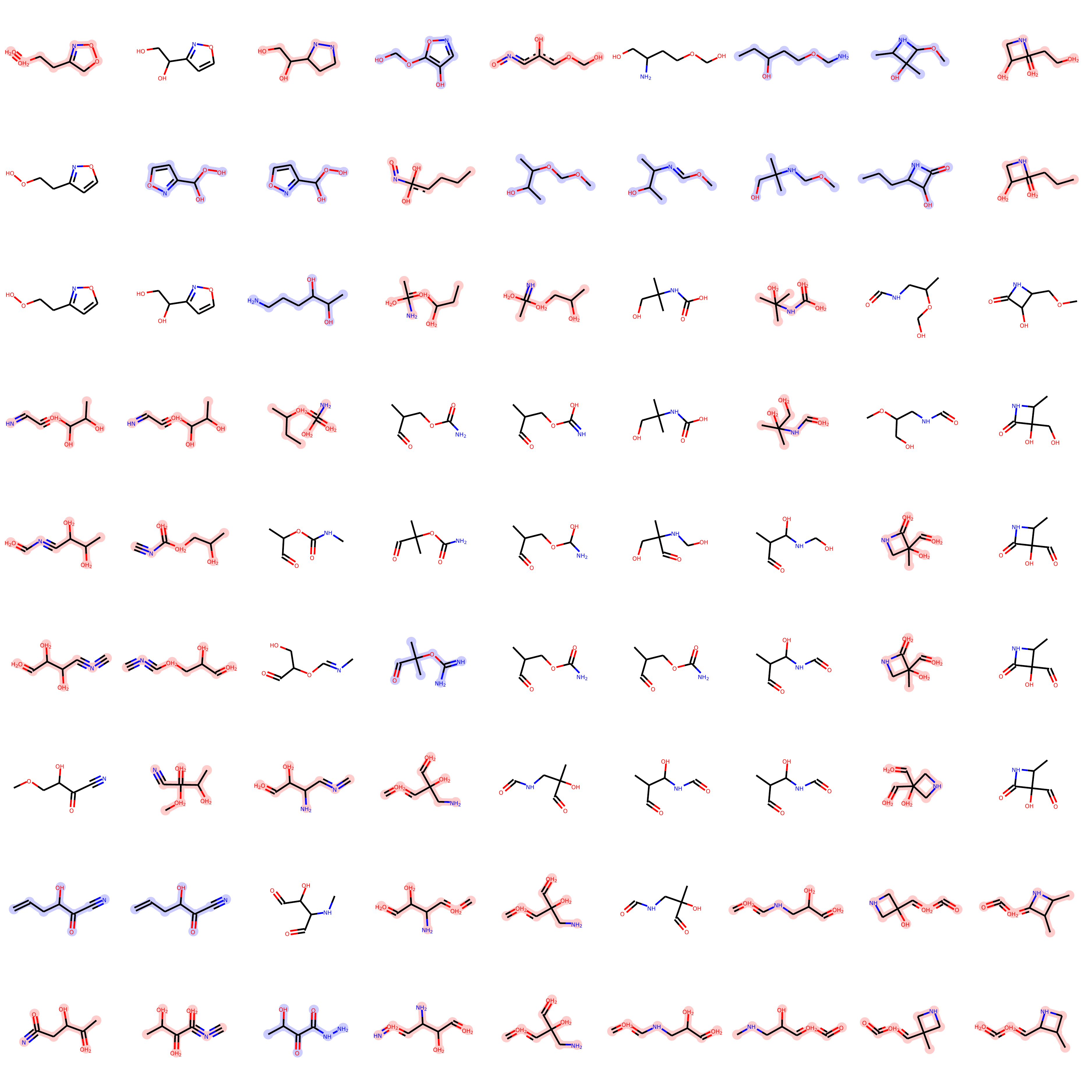}
\vspace{1.5ex}
\caption{\label{fig:planes}
Decodings of latent space points of a conditional model sampled over a random 2D plane in $\mathbf{z}$-space of $c=40$ (within 5 units from center of coordinates). Left: Samples conditioned on 7x Carbon, 1x Nitrogen, 1x Oxygen (12\% QM9). Right: Samples conditioned on 5x Carbon, 1x Nitrogen, 3x Oxygen (2.6\% QM9). Color legend as in Figure \ref{fig:interp}.
}
\end{figure*}

We plot two planes in Figure~\ref{fig:planes}, for a frequent label (left) and a less frequent label in QM9 (right). Both images show a varied and fairly smooth mix of molecules. The left image has many valid samples broadly distributed across the plane, as presumably the autoencoder had to fit a large portion of database into this space. The right exhibits stronger effect of regularization, as valid molecules tend to be only around center. 

\begin{figure*}[bt]
\centering
\includegraphics[width=\linewidth]{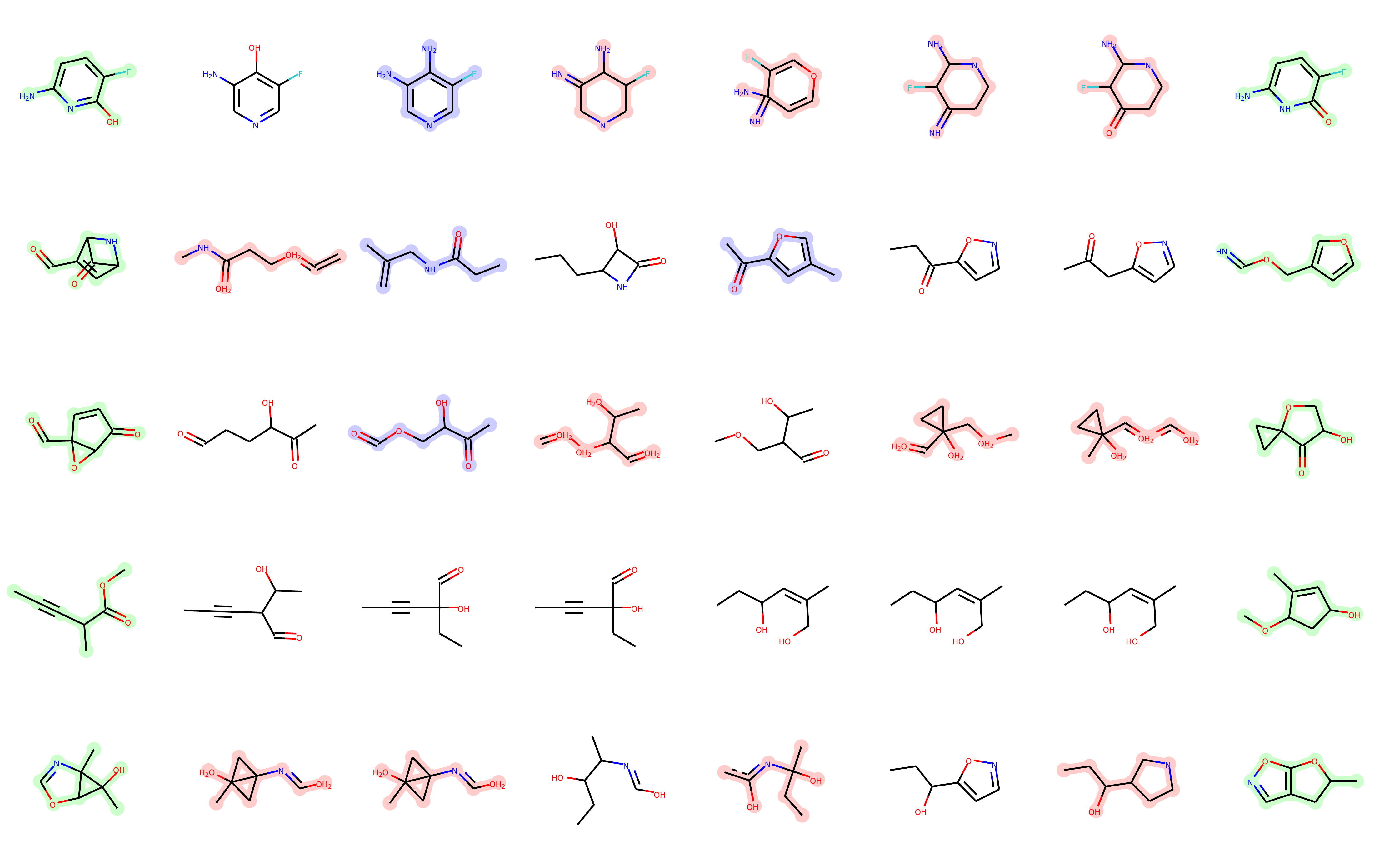}
\vspace{1.5ex}
\caption{\label{fig:interp}
Linear interpolation between row-wise pairs of randomly chosen molecules in $\mathbf{z}$-space of $c=40$ in a conditional model. Color legend: encoder inputs (green), chemically invalid graphs (red), valid graphs with wrong label (blue), valid and correct (white).
}
\end{figure*}

An example of several interpolations is shown in Figure~\ref{fig:interp}. We can find both meaningful (1st, 2nd and 4th row) and less meaningful transitions, though many samples on the lines do not form chemically valid compounds.

\paragraph*{Decoder Quality Metrics.} The quality of a conditional decoder can be evaluated by the validity and variety of generated graphs. For a given label $\mathbf{y}^{(l)}$, we draw $n_s=10^4$ samples $\mathbf{z}^{(l,s)} \sim \prior$ and compute the discrete point estimate of their decodings $\hat{G}^{(l,s)} = \arg \max p_{\theta}(G|\mathbf{z}^{(l,s)},\mathbf{y}^{(l)})$. 

Let $V^{(l)}$ be the list of chemically valid molecules from $\hat{G}^{(l,s)}$ and $C^{(l)}$ be the list of chemically valid molecules with atom histograms equal to $\mathbf{y}^{(l)}$. We are interested in ratios $\mathrm{Valid}^{(l)} = |V^{(l)}|/n_s$ and $\mathrm{Accurate}^{(l)} = |C^{(l)}|/n_s$. Furthermore, let $\mathrm{Unique}^{(l)} = |\mathrm{set}(C^{(l)})|/|C^{(l)}|$ be the fraction of unique correct graphs and $\mathrm{Novel}^{(l)} = 1 -  |\mathrm{set}(C^{(l)}) \cap \mathrm{QM9}| /|\mathrm{set}(C^{(l)})|$ the fraction of novel out-of-dataset graphs; we define $\mathrm{Unique}^{(l)}=0$ and $\mathrm{Novel}^{(l)}=0$ if $|C^{(l)}|=0$. Finally, the introduced metrics are aggregated by frequencies of labels in QM9, \eg $\mathrm{Valid} = \sum_{l} \mathrm{Valid}^{(l)} \mathrm{freq}(\mathbf{y}^{(l)})$. Unconditional decoders are evaluated by assuming there is just a single label, therefore $\mathrm{Valid}=\mathrm{Accurate}$.

In Table~\ref{tab:qm9}, we can see that on average 50\% of generated molecules are chemically valid and, in the case of conditional models, about 40\% have the correct label which the decoder was conditioned on. Larger embedding sizes $c$ are less regularized, demonstrated by a higher number of $\mathrm{Unique}$ samples and by lower accuracy of the conditional model, as the decoder is forced less to rely on actual labels. The ratio of $\mathrm{Valid}$ samples shows less clear behavior, likely because the discrete performance is not directly optimized for. For all models, it is remarkable that about 60\% of generated molecules are out of the dataset, \ie the network has never seen them during training. %

Looking at the baselines, CVAE can output only very few valid samples as expected, while GVAE generates the highest number of valid samples (60\%) but of very low variance (less than 10\%). Additionally, we investigate the importance of graph matching by using identity assignment $X$ instead and thus learning to reproduce particular node permutations in the training set, which correspond to the canonical ordering of SMILES strings from RDKit. This ablated model (denoted as NoGM in Table~\ref{tab:qm9}) produces many valid samples of lower variety and, surprisingly, outperforms GVAE in this regard. In comparison, our model can achieve good performance in both metrics at the same time.

\begin{table*}[bt]
\centering
\begin{tabular}{cccccccc}
\toprule
& & $\log \dec$ & ELBO & $\mathrm{Valid}$ & $\mathrm{Accurate}$ & $\mathrm{Unique}$ & $\mathrm{Novel}$\tabularnewline
\midrule

\multirow{4}{*}{\rotatebox{90}{Cond.}}
& Ours $c=20$ & -0.578 & -0.722 & 0.565 & 0.467 & 0.314 & 0.598 \tabularnewline
& Ours $c=40$ & -0.504 & -0.617 & 0.511 & 0.416 & 0.484 & 0.635 \tabularnewline
& Ours $c=60$ & -0.492 & -0.585 & 0.520 & 0.406 & 0.583 & 0.613 \tabularnewline
& Ours $c=80$ & -0.475 & -0.557 & 0.458 & 0.353 & 0.666 & 0.661 \tabularnewline

\midrule
\multirow{7}{*}{\rotatebox{90}{Unconditional}}
& Ours $c=20$ & -0.660 & -0.916 & 0.485 & 0.485 &	0.457 &	0.575 \tabularnewline
& Ours $c=40$ & -0.537 & -0.744 & 0.542 & 0.542 & 0.618 & 0.617 \tabularnewline
& Ours $c=60$ & -0.486 & -0.656 & 0.517 & 0.517 & 0.695 & 0.570 \tabularnewline
& Ours $c=80$ & -0.482 & -0.628 & 0.557 & 0.557 & 0.760 & 0.616 \tabularnewline
\cline{2-8}
\rule{0pt}{2.5ex} & NoGM $c=80$ & -2.388 & -2.553 & 0.810 & 0.810 & 0.241 & 0.610	 \tabularnewline
& CVAE $c=60$ & -- & -- & 0.103 & 0.103 & 0.675 & 0.900 \tabularnewline
& GVAE $c=20$ & -- & -- & 0.602 & 0.602 & 0.093 & 0.809 \tabularnewline

\bottomrule
\end{tabular}
\vspace{1.5ex}
\caption{\label{tab:qm9}
Performance on conditional and unconditional QM9 models evaluated by mean test-time reconstruction log-likelihood $(\log \dec$), mean test-time evidence lower bound (ELBO), and decoding quality metrics (Section~\ref{subsec:qm9}). Baselines CVAE \citep{Bombarelli16} and GVAE \citep{KusnerPH17} are listed only for the embedding size with the highest $\mathrm{Valid}$.}
\end{table*}

\paragraph*{Likelihood.} Besides the application-specific metric introduced above, we also report evidence lower bound (ELBO) commonly used in VAE literature, which corresponds to $-\mathcal{L}(\phi, \theta; G)$ in our notation. In Table~\ref{tab:qm9}, we state mean bounds over test set, using a single $\mathbf{z}$ sample per graph. We observe both reconstruction loss and KL-divergence decrease due to larger $c$ providing more freedom. However, there seems to be no strong correlation between ELBO and $\mathrm{Valid}$, which makes model selection somewhat difficult.

\paragraph*{Implicit Node Probabilities.} \label{sec:implicit}
Our decoder assumes independence of node and edge probabilities, which allows for isolated nodes or edges. Making further use of the fact that molecules are connected graphs, here we investigate the effect of making node probabilities a function of edge probabilities. Specifically, we consider the probability for node $a$ as that of its most probable edge: $\widetilde{A}_{a,a} = \max_b{\widetilde{A}_{a,b}}$. 

The evaluation on QM9 in Table~\ref{tab:qm9impl} shows a clear improvement in $\mathrm{Valid}$, $\mathrm{Accurate}$, and $\mathrm{Novel}$ metrics in both the conditional and unconditional setting. However, this is paid for by lower variability and higher reconstruction loss. This indicates that while the new constraint is useful, the model cannot fully cope with it.

\begin{table*}[bt]
\centering
\begin{tabular}{cccccccc}
\toprule
& & $\log \dec$ & ELBO & $\mathrm{Valid}$ & $\mathrm{Accurate}$ & $\mathrm{Unique}$ & $\mathrm{Novel}$\tabularnewline
\midrule

\multirow{4}{*}{\rotatebox{90}{Cond.}}
& Ours/imp $c=20$ & -0.784 & -0.919 & \emph{0.572} & \emph{0.482} & 0.238 & \emph{0.718} \tabularnewline
& Ours/imp $c=40$ & -0.671 & -0.776 & \emph{0.611} & \emph{0.518} & 0.307 & \emph{0.665} \tabularnewline
& Ours/imp $c=60$ & -0.618 & -0.714 & \emph{0.566} & \emph{0.448} & 0.416 & \emph{0.710} \tabularnewline
& Ours/imp $c=80$ & -0.627 & -0.713 & \emph{0.583} & \emph{0.451} & 0.475 & \emph{0.681} \tabularnewline

\midrule
\multirow{4}{*}{\rotatebox{90}{Uncond.}}
& Ours/imp $c=20$ & -0.857 & -1.091 & \emph{0.533} & \emph{0.533} & 0.228 & \emph{0.610} \tabularnewline
& Ours/imp $c=40$ & -0.737 & -0.932 & \emph{0.562} & \emph{0.562} & 0.420 & \emph{0.758} \tabularnewline
& Ours/imp $c=60$ & -0.634 & -0.797 & \emph{0.587} & \emph{0.587} & 0.459 & \emph{0.730} \tabularnewline
& Ours/imp $c=80$ & -0.642 & -0.777 & \emph{0.571} & \emph{0.571} & 0.520 & \emph{0.719} \tabularnewline

\bottomrule
\end{tabular}
\vspace{1.5ex}
\caption{\label{tab:qm9impl}
Performance on conditional and unconditional QM9 models with implicit node probabilities. Improvement with respect to Table~\ref{tab:qm9} is emphasized in italics.}
\end{table*}

\paragraph*{Unregularized Autoencoder.} The regularization in VAE works against achieving perfect reconstruction of training data, especially for small embedding sizes. To understand the reconstruction ability of our architecture, we train it as unregularized in this section, \ie with a deterministic encoder and without KL-divergence term in Equation~\ref{eq:elbo}. 

Unconditional models for QM9 achieve mean test log-likelihood $\log \dec$ of roughly $-0.37$ (about $-0.50$ for the implicit node probability model) for all $c \in \{20,40,60,80\}$. While these log-likelihoods are significantly higher than in Tables~\ref{tab:qm9} and ~\ref{tab:qm9impl}, our architecture cannot achieve perfect reconstruction of inputs. We were successful to increase training negative log-likelihood to zero only on fixed small training sets of hundreds of examples, where the network could overfit. This indicates that the network has problems finding generally valid rules for assembly of output tensors.

\subsection{ZINC Dataset}

ZINC dataset \citep{zinc} contains about 250k drug-like organic molecules of up to 38 heavy atoms with 9 distinct atomic numbers and 4 bond types, we set $k=38$, $d_e=4$ and $d_n=9$ and use the same split strategy as with QM9. We investigate the degree of scalability of an unconditional generative model.

\paragraph*{Setup.} The setup is equivalent as for QM9 but with a wider encoder (64, 128, 256 channels).

\paragraph*{Decoder Quality Metrics.} Our best model with $c=40$ has archived $\mathrm{Valid}=0.135$, which is clearly worse than for QM9. Using implicit node probabilities brought no improvement. For comparison, CVAE failed to generated any valid sample, while GVAE achieved $\mathrm{Valid}=0.357$ (models provided by \citet{KusnerPH17}, $c=56$). 

We attribute such a low performance to a generally much higher chance of producing a chemically-relevant inconsistency (number of possible edges growing quadratically). To confirm the relationship between performance and graph size $k$, we kept only graphs not larger than $k=20$ nodes, corresponding to 21\% of ZINC, and obtained $\mathrm{Valid}=0.341$ (and $\mathrm{Valid}=0.185$ for $k=30$ nodes, 92\% of ZINC). To verify that the problem is likely not caused by our proposed graph matching loss, we synthetically evaluate it in the following.

\paragraph*{Matching Robustness.} Robust behavior of graph matching using our similarity function $S$ is important for good performance of GraphVAE. Here we study graph matching in isolation to investigate its scalability. To that end, we add Gaussian noise $N(0,\epsilon_A), N(0,\epsilon_E), N(0,\epsilon_F)$ to each tensor of input graph $G$, truncating and renormalizing to keep their probabilistic interpretation, to create its noisy version $G_N$. We are interested in the quality of matching between self, $P[G,G]$, using noisy assignment matrix $X$ between $G$ and $G_N$. The advantage to naive checking $X$ for identity is the invariance to permutation of equivalent nodes. 

In Table~\ref{tab:robustness} we vary $k$ and $\epsilon$ for each tensor separately and report mean accuracies (computed in the same fashion as losses in Equation~\ref{eq:logli}) over 100 random samples from ZINC with size up to $k$ nodes. While we observe an expected fall of accuracy with stronger noise, the behavior is fairly robust with respect to increasing $k$ at a fixed noise level, the most sensitive being the adjacency matrix. Note that accuracies are not comparable across tensors due to different dimensionalities of random variables. We may conclude that the quality of the matching process is not a major hurdle to scalability.

\begin{table}[bt]
\centering
\begin{tabular}{ccccccc}
\toprule
Noise & $k=15$ & $k=20$ & $k=25$ & $k=30$ & $k=35$ & $k=40$\tabularnewline
\midrule
$\epsilon_{A,E,F}=0$ & 99.55 & 99.52 & 99.45 & 99.4 &  99.47 & 99.46 \tabularnewline
\midrule
$\epsilon_A=0.4$ & 90.95 & 89.55 & 86.64 & 87.25 & 87.07 & 86.78 \tabularnewline
$\epsilon_A=0.8$ & 82.14 & 81.01 & 79.62 & 79.67 & 79.07 & 78.69 \tabularnewline
\midrule
$\epsilon_E=0.4$ & 97.11 & 96.42 & 95.65 & 95.90 &  95.69 & 95.69 \tabularnewline
$\epsilon_E=0.8$ & 92.03 & 90.76 & 89.76 & 89.70 &  88.34 & 89.40 \tabularnewline
\midrule
$\epsilon_F=0.4$ & 98.32 & 98.23 & 97.64 & 98.28 & 98.24 & 97.90 \tabularnewline
$\epsilon_F=0.8$ & 97.26 & 97.00 & 96.60 & 96.91 & 96.56 & 97.17 \tabularnewline
\bottomrule
\end{tabular}
\vspace{1.5ex}
\caption{\label{tab:robustness}
Mean accuracy of matching ZINC graphs to their noisy counterparts in a synthetic benchmark as a function of maximum graph size $k$.}
\end{table}

\section{Discussion}
		
Having presented one of the first works on graph generation, we are aware of range of limitations and possible extensions, many of them already addressed by the follow-up works reviewed in Section~\ref{sec:genrelated}.

\paragraph*{Size Restriction.} As each edge is predicted individually, the proposed model is expected to be useful only for generating small graphs. This is due to growth of GPU memory requirements and number of parameters in $O(k^2)$ as well as matching complexity in $O(k^4)$, with small decrease in quality for high values of $k$. In Section~\ref{sec:exper} we demonstrated results for up to $k=38$. Nevertheless, in many applications, such as drug design, even generation of small graphs is still very useful.

\paragraph*{Independence Assumption.} We model the existence and attribute of each node and edge with an independent random variable, which allows us to easily take a point-wise estimate of the decoded probabilistic graph at the end. However, this assumption has turned out as too brittle in the case of imperfect predictions, as we have not observed strong correlation between ELBO and the ratio of valid samples in the experimental section, which suggests a mismatch between the training objective (probabilistic graph) and the desired objective (discrete graph). We have presented and evaluated two amendments to partially mitigate this problem: post-processing with a maximum spanning tree and making node probabilities implicit. Subsequent work has adopted  sequential sampling approaches, which allows to explicitly control for validity at each step \citep{JinBJ18,samanta18,liu18}.

\paragraph*{Simplistic Prior.} We use isotropic Gaussian prior $\prior = N(0,I)$ for regularization. However, there is a large body of work suggesting more complex distributions, such as using normalization flows on prior \citep{DinhSB16} or posterior \citep{KingmaSW16}, or better formulations of VAEs, notably Wasserstein autoencoders \citep{wasservaes}. While we expect that building on more powerful frameworks would lead to better results, it would not solve the limitations discussed above.

\section{Conclusion}		
		
In this chapter we addressed the problem of generating graphs from a continuous embedding in the context of variational autoencoders. We evaluated our method on two molecular datasets of different maximum graph size. While we achieved to learn embedding of reasonable quality on small molecules, our decoder had a hard time capturing complex chemical interactions for larger molecules. Nevertheless, we believe our method has been an important initial step towards more powerful decoders and has sparked interest in the community.

\chapter{Conclusion} \label{chap:conclusion}

The goal of this thesis was to study and mutually relate deep learning-based architectures for processing graph-structured data and, primarily, to propose several novel contributions to this nascent but fast evolving field. Throughout the past pages, we have seen that many core concepts (\eg convolutions, superpixels, or autoencoders) can be taken over from the domain of regular grids and repurposed for irregular data. We have touched upon both discriminative and generative models. The field being so young and challenging, it has offered us many unexplored paths to tread; many of them having a dead end but others leading us to good experimental results and novel approaches to tasks such as point cloud processing, graph classification or graph generation. Nevertheless, the proposed methods are hardly perfect and we tried to remain critical in stating their limitations as well. 

Deep learning is a booming topic and with the justified popularity of preprint repositories, there are often multiple lines of work proposed independently. Specifically in deep learning on graphs, these papers often come from several, originally rather separate communities working in signal processing and vision, in biochemistry, in natural language processing, or in information retrieval. We have thus attempted at providing a somewhat fused but not exhaustive picture in our literature reviews and at helping the reader to better understand the context of our work. 

In the next paragraphs, we will recapitulate our contributions and conclude with an outlook on possible future directions of work.

The first main contribution was the introduction of a crucial primitive for processing data structured on variable attributed graphs, called Edge-Conditioned Convolutions (ECC). The major insight was learning how to generate filters for every edge attribute rather than learning such filters directly. We have shown that the standard discrete convolution on grids as well as many formulations proposed previously can be seen as a special case of ECC. Our other contributions rely on having such an operation available. In addition, this was the first time graph convolutions were applied to processing of (neighborhood graphs constructed on) point clouds.

The second main contribution was the formulation of a superpixel-inspired intermediate point cloud representation, called SuperPoint Graph (SPG). The major insight was that considering point clouds as a collection of interconnected simple shapes, rather than individual points, allows ECC to start off from more informative embeddings and scale up to massive point clouds consisting of millions of points without major sacrifice in fine details. Important ingredients were also the very efficient partitioning algorithm and the definition of rich contextual features captured in SPG. Having combined the parts together and performed an extensive ablation study, we were able to significantly improve on the state of the art methods on the two largest publicly available datasets covering outdoor and indoor scenes.

Finally, the third main contribution was the presentation of one the first works on deep-learning based graph generation. The major insight was to use approximate graph matching for aligning predictions of an autoencoder with its inputs, which allowed us to decode graphs with variable but upper-bounded number of nodes. Our motivation was to avoid several challenges associated with step-wise construction. While the method could outperform past text generation-based autoencoders on a small-sized dataset, the practical performance left space for improvement, quickly addressed by consecutive works. Nevertheless, we believe our work has helped to spark interest for graph generation in the community and has looked into the direction of generation of graph tensors at once.

In summary, we have ventured a journey from graphs to their embeddings and back. We hope that the research developed in the context of this thesis has provided the community with new insights about some of the challenges in the emergent field of deep learning on graphs and brings us one more step closer to solving real-world problems arising in the field's manifold applications, such as drug discovery, remote sensing, social network understanding, or autonomous systems, for the benefit of all of us.

\section{Future Directions}

Deep learning on graphs is an exciting field with large applicability, offering many future direction to follow. Here we expand just on some of them.

\paragraph*{Dynamic Graphs.} In practice, it is often the case that graph data is not static but rather evolves in time and, in fact, the precise timing of events can be important for the task at hand. This can include changes in node signal, edge attributes, or in the number of nodes and their connectivity. For example, users communicate on social networks, a moving car updates information about its surroundings, new facts are included in knowledge bases, or - as we have seen in concurrent work in Chapter~\ref{chap:iclr18} - graphs are sequentially constructed. How to update the representation efficiently and how incorporate timing information, on what level of granularity, and whether to maintain history or treat the system as Markovian are just a few questions, the answering of which is likely application-driven. Extending graph embedding techniques in this direction could open up a wide range of exciting application domains.

With a few exceptions \citep{YuanLWYG17}, past DL research has dealt only with time series on graphs with fixed structure, such as for traffic prediction \citep{li2018dcrnn_traffic} or for modeling the dynamics of physical systems represented as complete graphs \citep{BattagliaPLRK16,KipfFWWZ18}. Extending the latter works to larger-scale systems is an open question; one way could be to adapt the graph structure according to object distances or attention scores, which is an application of graph editing discussed below.

\paragraph*{Graph Editing.} Graph generation is a very recent topic, which has just started to get explored in DL. While the primary real-world application has been in molecule generation so far, we believe that many more are awaiting. Past work has already looked into generating graphs as an intermediate representation, see \citet{johnson2017learning} for reasoning tasks and \citet{BrockschmidtGCMG} for code generation. We are particularly excited about the applications of auto-regressive methods to graph editing, where one starts from an existing graph rather than from scratch. Note that only the case of adding edges (link prediction) has been well researched. Graph editing could find use in simulating evolution of phenomena in time, generating similar molecules and translating measurements into plausible molecules, predicting chemical reactions, or locally optimizing certain combinatorial properties of graphs. In drug discovery, finding the right ligands for a specific binding site on a protein (analogous to finding a key for a keyhole) could been seen as graph inpainting. Also, graph generative models could be used as samplers for various reinforcement learning-based approaches for automating the architecture design process of deep networks, as \eg in \citet{BakerGNR16}. Ultimately, the variety of tasks that can be achieved with generative models in images  can serve as an inspiration.

\paragraph*{Node-Conditioned Convolutions.} In ECC, we conditioned message function $m$ on edge attributes, which allowed specializing the communication between each pair of nodes. However, all nodes shared the same learned weight matrices. Thus, it could be interesting to condition update function $u$ (in the form of RNN) on node attributes as well. We suspect this might be useful in cases where there are distinct node types present in a graph, such as in biological networks (with nodes for proteins, diseases, chemical compounds or symptoms) or perhaps multi-agent systems.

\paragraph*{Packages and Benchmarks.} Open source implementations and public datasets are one of the driving forces of the momentum behind deep learning. While many graph convolution methods have been open sourced, including ours, their programming interfaces are fairly diverse and written in different DL frameworks, which makes benchmarking and selection of the right method for an application laborious. It would be thus beneficial to initiate a single, up-to-date package containing a range of published methods, implemented with reusable modules and accessible under a unified interface. This should also make exploring novel model variants easier.

A related point is the desire for better benchmarks for evaluating graph representation learning approaches. Many popular datasets currently used in the community are tiny and prone to overfitting or not providing enough variety. For example, not all graphs may exhibit the degree of locality and stationarity assumed by convolutional methods\footnote{The fitness of different architectures to different graph datasets could be potentially measured in a way similar to the work of \citet{UlyanovVL17}, who have shown that the structure of  networks is sufficient to capture a great deal of low-level image statistics.}. Evaluating models under different conditions may often lead to (empirical) research insights. For example, a recently introduced benchmark for molecular machine learning \citep{wu2018moleculenet} has shown that traditional methods are not yet outperformed by a selection of graph convolution models on several task.

\paragraph*{Theoretical Understanding.} The lack of theoretical understanding is a popular complaint against deep learning, and spatial graph convolution models are no exception. By leaving the graph spectral theory \citep{defferrard16} and dynamic system theory \citep{scarselli09} behind, we could learn feed-forward models working on arbitrary graphs but also forfeited theoretical insights, proposing innovations based mostly on intuition and experiments. The future work should look more into the properties and guarantees offered by various message passing and aggregation methods. Specifically, the convergence at training time and the stability and the way of spreading information at inference time should be of interest. A source of inspiration could be studies performed for RNNs as in \eg \citet{tallec2018can,chen2018dynamical}.

\begin{spacing}{0.9}

\bibliographystyle{apalike}
\cleardoublepage
\bibliography{References/references} %

\begin{thebibliography}{}

\bibitem[Abadi et~al., 2015]{tensorflow}
Abadi, M., Agarwal, A., Barham, P., Brevdo, E., Chen, Z., Citro, C., Corrado,
  G.~S., Davis, A., Dean, J., Devin, M., Ghemawat, S., Goodfellow, I., Harp,
  A., Irving, G., Isard, M., Jia, Y., Jozefowicz, R., Kaiser, L., Kudlur, M.,
  Levenberg, J., Man\'{e}, D., Monga, R., Moore, S., Murray, D., Olah, C.,
  Schuster, M., Shlens, J., Steiner, B., Sutskever, I., Talwar, K., Tucker, P.,
  Vanhoucke, V., Vasudevan, V., Vi\'{e}gas, F., Vinyals, O., Warden, P.,
  Wattenberg, M., Wicke, M., Yu, Y., and Zheng, X. (2015).
\newblock {TensorFlow}: Large-scale machine learning on heterogeneous systems.

\bibitem[Achanta et~al., 2012]{achanta2012slic}
Achanta, R., Shaji, A., Smith, K., Lucchi, A., Fua, P., and S{\"u}sstrunk, S.
  (2012).
\newblock {SLIC} superpixels compared to state-of-the-art superpixel methods.
\newblock {\em IEEE Transactions on Pattern Analysis and Machine Intelligence},
  34(11):2274--2282.

\bibitem[Agaskar and Lu, 2013]{AgaskarL13}
Agaskar, A. and Lu, Y.~M. (2013).
\newblock A spectral graph uncertainty principle.
\newblock {\em {IEEE} Trans. Information Theory}, 59(7):4338--4356.

\bibitem[Alvar et~al., 2016]{orion}
Alvar, N.~S., Zolfaghari, M., and Brox, T. (2016).
\newblock Orientation-boosted voxel nets for {3D} object recognition.
\newblock {\em CoRR}, abs/1604.03351.

\bibitem[Anand et~al., 2013]{anand2013contextually}
Anand, A., Koppula, H.~S., Joachims, T., and Saxena, A. (2013).
\newblock Contextually guided semantic labeling and search for
  three-dimensional point clouds.
\newblock {\em The International Journal of Robotics Research}, 32(1):19--34.

\bibitem[Armeni et~al., 2016]{armeni_cvpr16}
Armeni, I., Sener, O., Zamir, A.~R., Jiang, H., Brilakis, I., Fischer, M., and
  Savarese, S. (2016).
\newblock {3D} semantic parsing of large-scale indoor spaces.
\newblock In {\em IEEE Conference on Computer Vision and Pattern Recognition
  (CVPR)}.

\bibitem[Atwood and Towsley, 2016]{dcnn}
Atwood, J. and Towsley, D. (2016).
\newblock Diffusion-convolutional neural networks.
\newblock In {\em Advances in Neural Information Processing Systems (NIPS)}.

\bibitem[Ba et~al., 2016]{layernorm}
Ba, L.~J., Kiros, R., and Hinton, G.~E. (2016).
\newblock Layer normalization.
\newblock {\em CoRR}, abs/1607.06450.

\bibitem[Baker et~al., 2017]{BakerGNR16}
Baker, B., Gupta, O., Naik, N., and Raskar, R. (2017).
\newblock Designing neural network architectures using reinforcement learning.
\newblock In {\em International Conference on Learning Representations (ICLR)}.

\bibitem[Barab{\'a}si and Albert, 1999]{barabasi1999emergence}
Barab{\'a}si, A.-L. and Albert, R. (1999).
\newblock Emergence of scaling in random networks.
\newblock {\em Science}, 286(5439):509--512.

\bibitem[Battaglia et~al., 2016]{BattagliaPLRK16}
Battaglia, P.~W., Pascanu, R., Lai, M., Rezende, D.~J., and Kavukcuoglu, K.
  (2016).
\newblock Interaction networks for learning about objects, relations and
  physics.
\newblock In {\em Advances in Neural Information Processing Systems (NIPS)},
  pages 4502--4510.

\bibitem[Belkin and Niyogi, 2001]{BelkinN01}
Belkin, M. and Niyogi, P. (2001).
\newblock Laplacian eigenmaps and spectral techniques for embedding and
  clustering.
\newblock In {\em Advances in Neural Information Processing Systems (NIPS)},
  pages 585--591.

\bibitem[Bengio et~al., 2015]{bengio2015scheduled}
Bengio, S., Vinyals, O., Jaitly, N., and Shazeer, N. (2015).
\newblock Scheduled sampling for sequence prediction with recurrent neural
  networks.
\newblock In {\em Advances in Neural Information Processing Systems (NIPS)},
  pages 1171--1179.

\bibitem[Bengio et~al., 2013]{bengio2013representation}
Bengio, Y., Courville, A., and Vincent, P. (2013).
\newblock Representation learning: A review and new perspectives.
\newblock {\em IEEE transactions on pattern analysis and machine intelligence},
  35(8):1798--1828.

\bibitem[Bojchevski and G{\"{u}}nnemann, 2017]{BojchevskiG17}
Bojchevski, A. and G{\"{u}}nnemann, S. (2017).
\newblock Deep gaussian embedding of attributed graphs: Unsupervised inductive
  learning via ranking.
\newblock {\em CoRR}, abs/1707.03815.

\bibitem[Bojchevski et~al., 2018]{netgan2018}
Bojchevski, A., Shchur, O., Z{\"{u}}gner, D., and G{\"{u}}nnemann, S. (2018).
\newblock {NetGAN}: Generating graphs via random walks.
\newblock {\em CoRR}, abs/1803.00816.

\bibitem[Borgwardt and Kriegel, 2005]{borgwardtK05}
Borgwardt, K.~M. and Kriegel, H. (2005).
\newblock Shortest-path kernels on graphs.
\newblock In {\em {IEEE} International Conference on Data Mining {(ICDM})},
  pages 74--81.

\bibitem[Boscaini et~al., 2016]{BoscainiMRB16}
Boscaini, D., Masci, J., Rodol{\`{a}}, E., and Bronstein, M.~M. (2016).
\newblock Learning shape correspondence with anisotropic convolutional neural
  networks.
\newblock In {\em Advances in Neural Information Processing Systems (NIPS)},
  pages 3189--3197.

\bibitem[Boulch et~al., 2017]{boulch2017unstructured}
Boulch, A., Saux, B.~L., and Audebert, N. (2017).
\newblock Unstructured point cloud semantic labeling using deep segmentation
  networks.
\newblock In {\em Eurographics Workshop on 3D Object Retrieval}, volume~2.

\bibitem[Bowman et~al., 2016]{BowmanVVDJB16}
Bowman, S.~R., Vilnis, L., Vinyals, O., Dai, A.~M., J{\'{o}}zefowicz, R., and
  Bengio, S. (2016).
\newblock Generating sentences from a continuous space.
\newblock In {\em CoNLL}, pages 10--21.

\bibitem[Boykov et~al., 2001]{boykov2001fast}
Boykov, Y., Veksler, O., and Zabih, R. (2001).
\newblock Fast approximate energy minimization via graph cuts.
\newblock {\em IEEE Transactions on Pattern Analysis and Machine Intelligence},
  23(11):1222--1239.

\bibitem[Brabandere et~al., 2016]{dfn16}
Brabandere, B.~D., Jia, X., Tuytelaars, T., and Gool, L.~V. (2016).
\newblock Dynamic filter networks.
\newblock In {\em Advances in Neural Information Processing Systems (NIPS)}.

\bibitem[Brockschmidt et~al., 2018]{BrockschmidtGCMG}
Brockschmidt, M., Allamanis, M., Gaunt, A.~L., and Polozov, O. (2018).
\newblock Generative code modeling with graphs.
\newblock {\em CoRR}, abs/1805.08490.

\bibitem[Bronstein et~al., 2017]{bronstein2017geometric}
Bronstein, M.~M., Bruna, J., LeCun, Y., Szlam, A., and Vandergheynst, P.
  (2017).
\newblock Geometric deep learning: Going beyond {E}uclidean data.
\newblock {\em IEEE Signal Processing Magazine}, 34(4):18--42.

\bibitem[Bruna et~al., 2013]{bruna13}
Bruna, J., Zaremba, W., Szlam, A., and LeCun, Y. (2013).
\newblock Spectral networks and locally connected networks on graphs.
\newblock {\em CoRR}, abs/1312.6203.

\bibitem[Cao and Kipf, 2018]{molgan18}
Cao, N.~D. and Kipf, T. (2018).
\newblock Molgan: An implicit generative model for small molecular graphs.
\newblock {\em CoRR}, abs/1805.11973.

\bibitem[Chandra and Kokkinos, 2016]{ChandraK16}
Chandra, S. and Kokkinos, I. (2016).
\newblock Fast, exact and multi-scale inference for semantic image segmentation
  with deep {G}aussian {CRFs}.
\newblock In {\em {IEEE} European Conference on Computer Vision (ECCV)}.

\bibitem[Chen et~al., 2018]{chen2018dynamical}
Chen, M., Pennington, J., and Schoenholz, S.~S. (2018).
\newblock Dynamical isometry and a mean field theory of {RNNs}: Gating enables
  signal propagation in recurrent neural networks.
\newblock {\em arXiv preprint arXiv:1806.05394}.

\bibitem[Chen et~al., 2014]{gfhsvm}
Chen, T., Dai, B., Liu, D., and Song, J. (2014).
\newblock Performance of global descriptors for velodyne-based urban object
  recognition.
\newblock In {\em {IEEE} Intelligent Vehicles Symposium Proceedings}, pages
  667--673.

\bibitem[Cho et~al., 2014a]{cho-gru14}
Cho, K., van Merri{\"{e}}nboer, B., G{\"{u}}l{\c c}ehre, {\c C}., Bahdanau, D.,
  Bougares, F., Schwenk, H., and Bengio, Y. (2014a).
\newblock Learning phrase representations using {RNN} encoder--decoder for
  statistical machine translation.
\newblock In {\em Conference on Empirical Methods in Natural Language
  Processing ({EMNLP})}.

\bibitem[Cho et~al., 2014b]{Cho}
Cho, M., Sun, J., Duchenne, O., and Ponce, J. (2014b).
\newblock Finding matches in a haystack: {A} max-pooling strategy for graph
  matching in the presence of outliers.
\newblock In {\em IEEE Conference on Computer Vision and Pattern Recognition
  (CVPR)}, pages 2091--2098.

\bibitem[Dai et~al., 2016]{struct2vec}
Dai, H., Dai, B., and Song, L. (2016).
\newblock Discriminative embeddings of latent variable models for structured
  data.
\newblock In {\em International Conference on Machine Learning (ICML)}.

\bibitem[Dai et~al., 2017]{dai2017learning}
Dai, H., Khalil, E.~B., Zhang, Y., Dilkina, B., and Song, L. (2017).
\newblock Learning combinatorial optimization algorithms over graphs.
\newblock In {\em Advances in Neural Information Processing Systems (NIPS)}.

\bibitem[Dai et~al., 2018]{sd-vae}
Dai, H., Tian, Y., Dai, B., Skiena, S., and Song, L. (2018).
\newblock Syntax-directed variational autoencoder for structured data.
\newblock In {\em International Conference on Learning Representations (ICLR)}.

\bibitem[Date and Nagi, 2016]{DateN16}
Date, K. and Nagi, R. (2016).
\newblock {GPU}-accelerated {H}ungarian algorithms for the linear assignment
  problem.
\newblock {\em Parallel Computing}, 57:52--72.

\bibitem[De~Deuge et~al., 2013]{trianglesvm}
De~Deuge, M., Quadros, A., Hung, C., and Douillard, B. (2013).
\newblock Unsupervised feature learning for classification of outdoor {3D}
  scans.
\newblock In {\em Australasian Conference on Robotics and Automation},
  volume~2.

\bibitem[Debnath et~al., 1991]{debnath1991structure}
Debnath, A.~K., Lopez~de Compadre, R.~L., Debnath, G., Shusterman, A.~J., and
  Hansch, C. (1991).
\newblock Structure-activity relationship of mutagenic aromatic and
  heteroaromatic nitro compounds. correlation with molecular orbital energies
  and hydrophobicity.
\newblock {\em Journal of medicinal chemistry}, 34(2):786--797.

\bibitem[Defferrard et~al., 2016]{defferrard16}
Defferrard, M., Bresson, X., and Vandergheynst, P. (2016).
\newblock Convolutional neural networks on graphs with fast localized spectral
  filtering.
\newblock In {\em Advances in Neural Information Processing Systems (NIPS)}.

\bibitem[Demantké et~al., 2011]{demantke_dimensionality_2011}
Demantké, J., Mallet, C., David, N., and Vallet, B. (2011).
\newblock Dimensionality based scale selection in {3D} lidar point clouds.
\newblock {\em International Archives of the Photogrammetry, Remote Sensing and
  Spatial Information Sciences}, {XXXVIII}-5/W12:97--102.

\bibitem[Dhillon et~al., 2007]{dhillon2007weighted}
Dhillon, I.~S., Guan, Y., and Kulis, B. (2007).
\newblock Weighted graph cuts without eigenvectors a multilevel approach.
\newblock {\em IEEE transactions on pattern analysis and machine intelligence},
  29(11).

\bibitem[Dinh et~al., 2016]{DinhSB16}
Dinh, L., Sohl{-}Dickstein, J., and Bengio, S. (2016).
\newblock Density estimation using real {NVP}.
\newblock {\em CoRR}, abs/1605.08803.

\bibitem[Dobson and Doig, 2003]{dobson2003}
Dobson, P.~D. and Doig, A.~J. (2003).
\newblock Distinguishing enzyme structures from non-enzymes without alignments.
\newblock {\em Journal of molecular biology}, 330(4):771--783.

\bibitem[D{\"{o}}rfler and Bullo, 2013]{kron}
D{\"{o}}rfler, F. and Bullo, F. (2013).
\newblock Kron reduction of graphs with applications to electrical networks.
\newblock {\em {IEEE} Trans. on Circuits and Systems}, 60-I(1):150--163.

\bibitem[Duvenaud et~al., 2015]{duvenaud}
Duvenaud, D.~K., Maclaurin, D., Aguilera{-}Iparraguirre, J., Bombarell, R.,
  Hirzel, T., Aspuru{-}Guzik, A., and Adams, R.~P. (2015).
\newblock Convolutional networks on graphs for learning molecular fingerprints.
\newblock In {\em Advances in Neural Information Processing Systems (NIPS)}.

\bibitem[Edwards and Xie, 2016]{edwards16}
Edwards, M. and Xie, X. (2016).
\newblock Graph based convolutional neural network.
\newblock In {\em British Machine Vision Conference (BMVC)}.

\bibitem[Engelmann et~al., 2017]{Engelmann17_3dsemseg}
Engelmann, F., Kontogianni, T., Hermans, A., and Leibe, B. (2017).
\newblock Exploring spatial context for 3d semantic segmentation of point
  clouds.
\newblock In {\em {IEEE} International Conference on Computer Vision (ICCV),
  3DRMS Workshop}.

\bibitem[Erdos and R{\'e}nyi, 1960]{erdos1960evolution}
Erdos, P. and R{\'e}nyi, A. (1960).
\newblock On the evolution of random graphs.
\newblock {\em Publ. Math. Inst. Hung. Acad. Sci}, 5(1):17--60.

\bibitem[Fan et~al., 2017]{FanSG17}
Fan, H., Su, H., and Guibas, L.~J. (2017).
\newblock A point set generation network for {3D} object reconstruction from a
  single image.
\newblock In {\em IEEE Conference on Computer Vision and Pattern Recognition
  (CVPR)}, pages 2463--2471.

\bibitem[Gadde et~al., 2016]{gadde16bi}
Gadde, R., Jampani, V., Kiefel, M., Kappler, D., and Gehler, P. (2016).
\newblock Superpixel convolutional networks using bilateral inceptions.
\newblock In {\em {IEEE} European Conference on Computer Vision (ECCV)}.

\bibitem[Gilmer et~al., 2017]{GilmerSRVD17}
Gilmer, J., Schoenholz, S.~S., Riley, P.~F., Vinyals, O., and Dahl, G.~E.
  (2017).
\newblock Neural message passing for quantum chemistry.
\newblock In {\em International Conference on Machine Learning (ICML)}, pages
  1263--1272.

\bibitem[G{\'{o}}mez{-}Bombarelli et~al., 2016]{Bombarelli16}
G{\'{o}}mez{-}Bombarelli, R., Duvenaud, D.~K., Hern{\'{a}}ndez{-}Lobato, J.~M.,
  Aguilera{-}Iparraguirre, J., Hirzel, T.~D., Adams, R.~P., and Aspuru{-}Guzik,
  A. (2016).
\newblock Automatic chemical design using a data-driven continuous
  representation of molecules.
\newblock {\em CoRR}, abs/1610.02415.

\bibitem[Gong and Xiang, 2003]{GongX03}
Gong, S. and Xiang, T. (2003).
\newblock Recognition of group activities using dynamic probabilistic networks.
\newblock In {\em {IEEE} International Conference on Computer Vision (ICCV)},
  pages 742--749.

\bibitem[Goodfellow et~al., 2014]{GoodfellowGAN14}
Goodfellow, I.~J., Pouget{-}Abadie, J., Mirza, M., Xu, B., Warde{-}Farley, D.,
  Ozair, S., Courville, A.~C., and Bengio, Y. (2014).
\newblock Generative adversarial networks.
\newblock {\em CoRR}, abs/1406.2661.

\bibitem[Graham, 2014]{fractpool}
Graham, B. (2014).
\newblock Fractional max-pooling.
\newblock {\em CoRR}, abs/1412.6071.

\bibitem[Graham, 2015]{Graham15}
Graham, B. (2015).
\newblock Sparse 3d convolutional neural networks.
\newblock In {\em British Machine Vision Conference (BMVC)}, pages
  150.1--150.9.

\bibitem[Graham et~al., 2018]{SubmanifoldSparseConvNet}
Graham, B., Engelcke, M., and van~der Maaten, L. (2018).
\newblock 3d semantic segmentation with submanifold sparse convolutional
  networks.
\newblock In {\em IEEE Conference on Computer Vision and Pattern Recognition
  (CVPR)}.

\bibitem[Grover and Leskovec, 2016]{GroverL16}
Grover, A. and Leskovec, J. (2016).
\newblock node2vec: Scalable feature learning for networks.
\newblock In {\em {ACM} {SIGKDD} International Conference on Knowledge
  Discovery and Data Mining}, pages 855--864.

\bibitem[Guinard and Landrieu, 2017]{guinard_weakly_2017}
Guinard, S. and Landrieu, L. (2017).
\newblock Weakly supervised segmentation-aided classification of urban scenes
  from {3D} {LiDAR} point clouds.
\newblock In {\em {ISPRS} 2017}.

\bibitem[Hackel et~al., 2017]{hackel2017semantic3d}
Hackel, T., Savinov, N., Ladicky, L., Wegner, J.~D., Schindler, K., and
  Pollefeys, M. (2017).
\newblock Semantic3d. net: A new large-scale point cloud classification
  benchmark.
\newblock {\em arXiv preprint arXiv:1704.03847}.

\bibitem[Hackel et~al., 2016]{hackel2016fast}
Hackel, T., Wegner, J.~D., and Schindler, K. (2016).
\newblock Fast semantic segmentation of {3D} point clouds with strongly varying
  density.
\newblock {\em ISPRS Annals of Photogrammetry, Remote Sensing \& Spatial
  Information Sciences}, 3(3).

\bibitem[Hamilton et~al., 2017a]{hamilton2017representation}
Hamilton, W.~L., Ying, R., and Leskovec, J. (2017a).
\newblock Representation learning on graphs: Methods and applications.
\newblock {\em arXiv preprint arXiv:1709.05584}.

\bibitem[Hamilton et~al., 2017b]{HamiltonYL17}
Hamilton, W.~L., Ying, Z., and Leskovec, J. (2017b).
\newblock Inductive representation learning on large graphs.
\newblock In {\em Advances in Neural Information Processing Systems (NIPS)},
  pages 1025--1035.

\bibitem[He and Sun, 2015]{hesun14}
He, K. and Sun, J. (2015).
\newblock Convolutional neural networks at constrained time cost.
\newblock In {\em IEEE Conference on Computer Vision and Pattern Recognition
  (CVPR)}.

\bibitem[He et~al., 2016]{residuals}
He, K., Zhang, X., Ren, S., and Sun, J. (2016).
\newblock Deep residual learning for image recognition.
\newblock In {\em IEEE Conference on Computer Vision and Pattern Recognition
  (CVPR)}.

\bibitem[Hornik, 1991]{Hornik91}
Hornik, K. (1991).
\newblock Approximation capabilities of multilayer feedforward networks.
\newblock {\em Neural Networks}, 4(2):251--257.

\bibitem[Hu et~al., 2013]{hu2013efficient}
Hu, H., Munoz, D., Bagnell, J.~A., and Hebert, M. (2013).
\newblock Efficient 3-d scene analysis from streaming data.
\newblock In {\em IEEE International Conference on Robotics and Automation
  (ICRA)}.

\bibitem[Huang et~al., 2017]{densenet17}
Huang, G., Liu, Z., and Weinberger, K.~Q. (2017).
\newblock Densely connected convolutional networks.
\newblock In {\em IEEE Conference on Computer Vision and Pattern Recognition
  (CVPR)}.

\bibitem[Huang and You, 2016]{pclabeling16}
Huang, J. and You, S. (2016).
\newblock Point cloud labeling using {3D} convolutional neural network.
\newblock In {\em ICPR}.

\bibitem[Huang et~al., 2018]{huangslice18}
Huang, Q., Wang, W., and Neumann, U. (2018).
\newblock Recurrent slice networks for {3D} segmentation on point clouds.
\newblock {\em arXiv preprint arXiv:1802.04402}.

\bibitem[Ioffe and Szegedy, 2015]{batchnorm}
Ioffe, S. and Szegedy, C. (2015).
\newblock Batch normalization: Accelerating deep network training by reducing
  internal covariate shift.
\newblock In {\em International Conference on Machine Learning (ICML)}.

\bibitem[Irwin et~al., 2012]{zinc}
Irwin, J.~J., Sterling, T., Mysinger, M.~M., Bolstad, E.~S., and Coleman, R.~G.
  (2012).
\newblock {ZINC:} {A} free tool to discover chemistry for biology.
\newblock {\em Journal of Chemical Information and Modeling}, 52(7):1757--1768.

\bibitem[Isola et~al., 2017]{IsolaZZE17}
Isola, P., Zhu, J., Zhou, T., and Efros, A.~A. (2017).
\newblock Image-to-image translation with conditional adversarial networks.
\newblock In {\em IEEE Conference on Computer Vision and Pattern Recognition
  (CVPR)}, pages 5967--5976.

\bibitem[Jaderberg et~al., 2015]{JaderbergSTN}
Jaderberg, M., Simonyan, K., Zisserman, A., and Kavukcuoglu, K. (2015).
\newblock Spatial transformer networks.
\newblock In {\em Advances in Neural Information Processing Systems (NIPS)},
  pages 2017--2025.

\bibitem[Jang et~al., 2016]{JangGP16}
Jang, E., Gu, S., and Poole, B. (2016).
\newblock Categorical reparameterization with gumbel-softmax.
\newblock {\em CoRR}, abs/1611.01144.

\bibitem[Jaromczyk and Toussaint, 1992]{jaromczyk1992relative}
Jaromczyk, J.~W. and Toussaint, G.~T. (1992).
\newblock Relative neighborhood graphs and their relatives.
\newblock {\em Proceedings of the IEEE}, 80(9):1502--1517.

\bibitem[Jin et~al., 2018]{JinBJ18}
Jin, W., Barzilay, R., and Jaakkola, T.~S. (2018).
\newblock Junction tree variational autoencoder for molecular graph generation.
\newblock In {\em International Conference on Machine Learning (ICML)}, pages
  2328--2337.

\bibitem[Johnson, 2017]{johnson2017learning}
Johnson, D.~D. (2017).
\newblock Learning graphical state transitions.
\newblock In {\em International Conference on Learning Representations (ICLR)}.

\bibitem[Kearnes et~al., 2016]{KearnesMBPR16}
Kearnes, S.~M., McCloskey, K., Berndl, M., Pande, V.~S., and Riley, P. (2016).
\newblock Molecular graph convolutions: moving beyond fingerprints.
\newblock {\em Journal of Computer-Aided Molecular Design}, 30(8):595--608.

\bibitem[Kim et~al., 2013]{kim20133d}
Kim, B.-S., Kohli, P., and Savarese, S. (2013).
\newblock {3D} scene understanding by voxel-{CRF}.
\newblock In {\em {IEEE} International Conference on Computer Vision (ICCV)}.

\bibitem[Kingma and Ba, 2015]{KingmaB14_adam}
Kingma, D.~P. and Ba, J. (2015).
\newblock Adam: {A} method for stochastic optimization.
\newblock In {\em International Conference on Learning Representations (ICLR)}.

\bibitem[Kingma et~al., 2016]{KingmaSW16}
Kingma, D.~P., Salimans, T., and Welling, M. (2016).
\newblock Improving variational inference with inverse autoregressive flow.
\newblock {\em CoRR}, abs/1606.04934.

\bibitem[Kingma and Welling, 2013]{vae}
Kingma, D.~P. and Welling, M. (2013).
\newblock Auto-encoding variational bayes.
\newblock {\em CoRR}, abs/1312.6114.

\bibitem[Kipf et~al., 2018]{KipfFWWZ18}
Kipf, T.~N., Fetaya, E., Wang, K., Welling, M., and Zemel, R.~S. (2018).
\newblock Neural relational inference for interacting systems.
\newblock In {\em International Conference on Machine Learning (ICML)}, pages
  2693--2702.

\bibitem[Kipf and Welling, 2016a]{kipf}
Kipf, T.~N. and Welling, M. (2016a).
\newblock Semi-supervised classification with graph convolutional networks.
\newblock {\em CoRR}, abs/1609.02907.

\bibitem[Kipf and Welling, 2016b]{kipf2016variational}
Kipf, T.~N. and Welling, M. (2016b).
\newblock Variational graph auto-encoders.
\newblock {\em Advances in Neural Information Processing Systems (NIPS),
  Workshop on Bayesian Deep Learning}.

\bibitem[Klokov and Lempitsky, 2017]{KlokovL17KdNet}
Klokov, R. and Lempitsky, V.~S. (2017).
\newblock Escape from cells: Deep {Kd}-networks for the recognition of {3D}
  point cloud models.
\newblock {\em CoRR}, abs/1704.01222.

\bibitem[Kondor, 2018]{nBodyNets}
Kondor, R. (2018).
\newblock N-body networks: a covariant hierarchical neural network architecture
  for learning atomic potentials.
\newblock {\em CoRR}, abs/1803.01588.

\bibitem[Kondor et~al., 2018]{kondorCCN}
Kondor, R., Son, H.~T., Pan, H., Anderson, B., and Trivedi, S. (2018).
\newblock Covariant compositional networks for learning graphs.
\newblock {\em CoRR}, abs/1801.02144.

\bibitem[Koppula et~al., 2011]{koppula2011semantic}
Koppula, H.~S., Anand, A., Joachims, T., and Saxena, A. (2011).
\newblock Semantic labeling of {3D} point clouds for indoor scenes.
\newblock In {\em Advances in Neural Information Processing Systems (NIPS)},
  pages 244--252.

\bibitem[Ktena et~al., 2017]{ktena2017distance}
Ktena, S.~I., Parisot, S., Ferrante, E., Rajchl, M., Lee, M., Glocker, B., and
  Rueckert, D. (2017).
\newblock Distance metric learning using graph convolutional networks:
  Application to functional brain networks.
\newblock In {\em Medical Image Computing and Computer-Assisted Intervention
  (MICCAI)}.

\bibitem[Kullback and Leibler, 1951]{Kullback_and_Leibler_1951}
Kullback, S. and Leibler, R.~A. (1951).
\newblock On information and sufficiency.
\newblock {\em The Annals of Mathematical Statistics}, 22 (1):79--86.

\bibitem[Kusner and Hern{\'{a}}ndez{-}Lobato, 2016]{KusnerH16}
Kusner, M.~J. and Hern{\'{a}}ndez{-}Lobato, J.~M. (2016).
\newblock {GANS} for sequences of discrete elements with the gumbel-softmax
  distribution.
\newblock {\em CoRR}, abs/1611.04051.

\bibitem[Kusner et~al., 2017]{KusnerPH17}
Kusner, M.~J., Paige, B., and Hern{\'{a}}ndez{-}Lobato, J.~M. (2017).
\newblock Grammar variational autoencoder.
\newblock In {\em International Conference on Machine Learning (ICML)}, pages
  1945--1954.

\bibitem[Landrieu and Obozinski, 2017]{landrieu2017cut}
Landrieu, L. and Obozinski, G. (2017).
\newblock Cut pursuit: Fast algorithms to learn piecewise constant functions on
  general weighted graphs.
\newblock {\em SIAM Journal on Imaging Sciences}, 10(4):1724--1766.

\bibitem[Landrieu et~al., 2017]{LANDRIEU2017102}
Landrieu, L., Raguet, H., Vallet, B., Mallet, C., and Weinmann, M. (2017).
\newblock A structured regularization framework for spatially smoothing
  semantic labelings of {3D} point clouds.
\newblock {\em ISPRS Journal of Photogrammetry and Remote Sensing}, 132:102 --
  118.

\bibitem[Landrieu and Simonovsky, 2018]{superpointgraphs}
Landrieu, L. and Simonovsky, M. (2018).
\newblock Large-scale point cloud semantic segmentation with superpoint graphs.
\newblock In {\em IEEE Conference on Computer Vision and Pattern Recognition
  (CVPR)}.

\bibitem[Landrum, 2011]{rdkit}
Landrum, G. (2011).
\newblock {RDKit}: Open-source cheminformatics.

\bibitem[Larsson et~al., 2017]{LarssonK0ATH17}
Larsson, M., Kahl, F., Zheng, S., Arnab, A., Torr, P. H.~S., and Hartley, R.~I.
  (2017).
\newblock Learning arbitrary potentials in {CRF}s with gradient descent.
\newblock {\em CoRR}, abs/1701.06805.

\bibitem[Lawin et~al., 2017]{lawin2017deep}
Lawin, F.~J., Danelljan, M., Tosteberg, P., Bhat, G., Khan, F.~S., and
  Felsberg, M. (2017).
\newblock Deep projective {3D} semantic segmentation.
\newblock {\em arXiv preprint arXiv:1705.03428}.

\bibitem[LeCun et~al., 2015]{lecun2015deep}
LeCun, Y., Bengio, Y., and Hinton, G. (2015).
\newblock Deep learning.
\newblock {\em Nature}, 521(7553):436--444.

\bibitem[LeCun et~al., 1998]{mnist}
LeCun, Y., Bottou, L., Bengio, Y., and Haffner, P. (1998).
\newblock Gradient-based learning applied to document recognition.
\newblock {\em Proceedings of the IEEE}, 86(11):2278--2324.

\bibitem[Ledig et~al., 2017]{LedigTHCCAATTWS17}
Ledig, C., Theis, L., Huszar, F., Caballero, J., Cunningham, A., Acosta, A.,
  Aitken, A.~P., Tejani, A., Totz, J., Wang, Z., and Shi, W. (2017).
\newblock Photo-realistic single image super-resolution using a generative
  adversarial network.
\newblock In {\em IEEE Conference on Computer Vision and Pattern Recognition
  (CVPR)}, pages 105--114.

\bibitem[Lei et~al., 2017]{lei2017deriving}
Lei, T., Jin, W., Barzilay, R., and Jaakkola, T. (2017).
\newblock Deriving neural architectures from sequence and graph kernels.
\newblock {\em arXiv preprint arXiv:1705.09037}.

\bibitem[Li et~al., 2018a]{pointcnn18}
Li, Y., Bu, R., Sun, M., and Chen, B. (2018a).
\newblock {PointCNN}.
\newblock {\em CoRR}, abs/1801.07791.

\bibitem[Li et~al., 2016a]{LiPSQG16}
Li, Y., Pirk, S., Su, H., Qi, C.~R., and Guibas, L.~J. (2016a).
\newblock {FPNN:} field probing neural networks for 3d data.
\newblock In {\em Advances in Neural Information Processing Systems (NIPS)},
  pages 307--315.

\bibitem[Li et~al., 2015]{gmmn}
Li, Y., Swersky, K., and Zemel, R.~S. (2015).
\newblock Generative moment matching networks.
\newblock In {\em International Conference on Machine Learning (ICML)}, pages
  1718--1727.

\bibitem[Li et~al., 2016b]{yujia16}
Li, Y., Tarlow, D., Brockschmidt, M., and Zemel, R.~S. (2016b).
\newblock Gated graph sequence neural networks.
\newblock In {\em International Conference on Learning Representations (ICLR)}.

\bibitem[Li et~al., 2018b]{LiDGM18}
Li, Y., Vinyals, O., Dyer, C., Pascanu, R., and Battaglia, P. (2018b).
\newblock Learning deep generative models of graphs.
\newblock In {\em International Conference on Machine Learning (ICML)}.

\bibitem[Li et~al., 2018c]{li2018dcrnn_traffic}
Li, Y., Yu, R., Shahabi, C., and Liu, Y. (2018c).
\newblock Diffusion convolutional recurrent neural network: Data-driven traffic
  forecasting.
\newblock In {\em International Conference on Learning Representations (ICLR)}.

\bibitem[Li et~al., 2018d]{LiZL18}
Li, Y., Zhang, L., and Liu, Z. (2018d).
\newblock Multi-objective de novo drug design with conditional graph generative
  model.
\newblock {\em J. Cheminformatics}, 10(1):33:1--33:24.

\bibitem[Liang et~al., 2017]{LiangLSFYX17}
Liang, X., Lin, L., Shen, X., Feng, J., Yan, S., and Xing, E.~P. (2017).
\newblock Interpretable structure-evolving {LSTM}.
\newblock In {\em IEEE Conference on Computer Vision and Pattern Recognition
  (CVPR)}, pages 2175--2184.

\bibitem[Liang et~al., 2016]{LiangSFLY16}
Liang, X., Shen, X., Feng, J., Lin, L., and Yan, S. (2016).
\newblock Semantic object parsing with graph {LSTM}.
\newblock In {\em {IEEE} European Conference on Computer Vision (ECCV)}, pages
  125--143.

\bibitem[Lin et~al., 2016]{LinSHR16}
Lin, G., Shen, C., van~den Hengel, A., and Reid, I.~D. (2016).
\newblock Efficient piecewise training of deep structured models for semantic
  segmentation.
\newblock In {\em IEEE Conference on Computer Vision and Pattern Recognition
  (CVPR)}.

\bibitem[Liu et~al., 2018]{liu18}
Liu, Q., Allamanis, M., Brockschmidt, M., and Gaunt, A.~L. (2018).
\newblock Constrained graph variational autoencoders for molecule design.
\newblock {\em CoRR}, abs/1805.09076.

\bibitem[Long et~al., 2015]{Long2015fnc}
Long, J., Shelhamer, E., and Darrell, T. (2015).
\newblock Fully convolutional networks for semantic segmentation.
\newblock In {\em IEEE Conference on Computer Vision and Pattern Recognition
  (CVPR)}, pages 3431--3440.

\bibitem[Lu and Rasmussen, 2012]{lu2012simplified}
Lu, Y. and Rasmussen, C. (2012).
\newblock Simplified {M}arkov random fields for efficient semantic labeling of
  3d point clouds.
\newblock In {\em {IEEE/RSJ} International Conference on Intelligent Robots and
  Systems ({IROS})}, pages 2690--2697.

\bibitem[Makhzani et~al., 2015]{aae}
Makhzani, A., Shlens, J., Jaitly, N., and Goodfellow, I.~J. (2015).
\newblock Adversarial autoencoders.
\newblock {\em CoRR}, abs/1511.05644.

\bibitem[Martinovic et~al., 2015]{martinovic20153d}
Martinovic, A., Knopp, J., Riemenschneider, H., and Van~Gool, L. (2015).
\newblock {3D} all the way: Semantic segmentation of urban scenes from start to
  end in {3D}.
\newblock In {\em IEEE Conference on Computer Vision and Pattern Recognition
  (CVPR)}.

\bibitem[Masci et~al., 2015]{masci15}
Masci, J., Boscaini, D., Bronstein, M.~M., and Vandergheynst, P. (2015).
\newblock Geodesic convolutional neural networks on {Riemannian} manifolds.
\newblock In {\em {IEEE} International Conference on Computer Vision (ICCV),
  Workshop}, pages 37--45.

\bibitem[Masoumi and Hamza, 2017]{MasoumiH17}
Masoumi, M. and Hamza, A.~B. (2017).
\newblock Shape classification using spectral graph wavelets.
\newblock {\em Appl. Intell.}, 47(4):1256--1269.

\bibitem[Maturana and Scherer, 2015]{voxnet}
Maturana, D. and Scherer, S. (2015).
\newblock Voxnet: {A} {3D} convolutional neural network for real-time object
  recognition.
\newblock In {\em {IEEE/RSJ} International Conference on Intelligent Robots and
  Systems ({IROS})}.

\bibitem[McKay and Piperno, 2014]{McKay2014nauty}
McKay, B.~D. and Piperno, A. (2014).
\newblock Practical graph isomorphism, {II}.
\newblock {\em Journal of Symbolic Computation}, 60(0):94 -- 112.

\bibitem[Mikolov et~al., 2013]{mikolov2013efficient}
Mikolov, T., Chen, K., Corrado, G., and Dean, J. (2013).
\newblock Efficient estimation of word representations in vector space.
\newblock {\em arXiv preprint arXiv:1301.3781}.

\bibitem[Mishkin and Matas, 2016]{goodinit}
Mishkin, D. and Matas, J. (2016).
\newblock All you need is a good init.
\newblock In {\em International Conference on Learning Representations (ICLR)}.

\bibitem[Monti et~al., 2017]{MontiBMRSB17}
Monti, F., Boscaini, D., Masci, J., Rodol{\`{a}}, E., Svoboda, J., and
  Bronstein, M.~M. (2017).
\newblock Geometric deep learning on graphs and manifolds using mixture model
  {CNNs}.
\newblock In {\em IEEE Conference on Computer Vision and Pattern Recognition
  (CVPR)}, pages 5425--5434.

\bibitem[Monti et~al., 2018]{MotifNet}
Monti, F., Otness, K., and Bronstein, M.~M. (2018).
\newblock Motifnet: a motif-based graph convolutional network for directed
  graphs.
\newblock {\em CoRR}, abs/1802.01572.

\bibitem[Mou et~al., 2016]{MouLZWJ16}
Mou, L., Li, G., Zhang, L., Wang, T., and Jin, Z. (2016).
\newblock Convolutional neural networks over tree structures for programming
  language processing.
\newblock In {\em {AAAI} Conference on Artificial Intelligence}, pages
  1287--1293.

\bibitem[Munoz et~al., 2009]{munoz2009contextual}
Munoz, D., Bagnell, J.~A., Vandapel, N., and Hebert, M. (2009).
\newblock Contextual classification with functional max-margin {M}arkov
  networks.
\newblock In {\em IEEE Conference on Computer Vision and Pattern Recognition
  (CVPR)}.

\bibitem[Narayanan et~al., 2016]{narayanan2016subgraph2vec}
Narayanan, A., Chandramohan, M., Chen, L., Liu, Y., and Saminathan, S. (2016).
\newblock subgraph2vec: Learning distributed representations of rooted
  sub-graphs from large graphs.
\newblock {\em arXiv preprint arXiv:1606.08928}.

\bibitem[Niemeyer et~al., 2014]{niemeyer2014contextual}
Niemeyer, J., Rottensteiner, F., and Soergel, U. (2014).
\newblock Contextual classification of lidar data and building object detection
  in urban areas.
\newblock {\em ISPRS Journal of Photogrammetry and Remote Sensing},
  87:152--165.

\bibitem[Niepert et~al., 2016]{niepert}
Niepert, M., Ahmed, M., and Kutzkov, K. (2016).
\newblock Learning convolutional neural networks for graphs.
\newblock In {\em International Conference on Machine Learning (ICML)}.

\bibitem[Olivecrona et~al., 2017]{OlivecronaBEC17}
Olivecrona, M., Blaschke, T., Engkvist, O., and Chen, H. (2017).
\newblock Molecular de novo design through deep reinforcement learning.
\newblock {\em CoRR}, abs/1704.07555.

\bibitem[Paszke et~al., 2017]{pytorch}
Paszke, A., Gross, S., Chintala, S., Chanan, G., Yang, E., DeVito, Z., Lin, Z.,
  Desmaison, A., Antiga, L., and Lerer, A. (2017).
\newblock Automatic differentiation in {PyTorch}.
\newblock {\em Advances in Neural Information Processing Systems (NIPS),
  Autodiff Workshop}.

\bibitem[Pearl, 1988]{Pearl:1988}
Pearl, J. (1988).
\newblock {\em Probabilistic Reasoning in Intelligent Systems: Networks of
  Plausible Inference}.
\newblock Morgan Kaufmann Publishers Inc., San Francisco, CA, USA.

\bibitem[Perozzi et~al., 2014]{PerozziAS14}
Perozzi, B., Al{-}Rfou, R., and Skiena, S. (2014).
\newblock Deepwalk: online learning of social representations.
\newblock In {\em {ACM} {SIGKDD} International Conference on Knowledge
  Discovery and Data Mining}, pages 701--710.

\bibitem[Perraudin et~al., 2014]{gspbox}
Perraudin, N., Paratte, J., Shuman, D.~I., Kalofolias, V., Vandergheynst, P.,
  and Hammond, D.~K. (2014).
\newblock {GSPBOX:} {A} toolbox for signal processing on graphs.
\newblock {\em CoRR}, abs/1408.5781.

\bibitem[Qi et~al., 2017a]{qi2016pointnet}
Qi, C.~R., Su, H., Mo, K., and Guibas, L.~J. (2017a).
\newblock {PointNet}: Deep learning on point sets for {3D} classification and
  segmentation.
\newblock In {\em IEEE Conference on Computer Vision and Pattern Recognition
  (CVPR)}.

\bibitem[Qi et~al., 2016]{qi16}
Qi, C.~R., Su, H., Nie{\ss}ner, M., Dai, A., Yan, M., and Guibas, L.~J. (2016).
\newblock Volumetric and multi-view {CNNs} for object classification on {3D}
  data.
\newblock In {\em IEEE Conference on Computer Vision and Pattern Recognition
  (CVPR)}.

\bibitem[Qi et~al., 2017b]{QiYSG17PointNetPP}
Qi, C.~R., Yi, L., Su, H., and Guibas, L.~J. (2017b).
\newblock {PointNet++}: Deep hierarchical feature learning on point sets in a
  metric space.
\newblock In {\em Advances in Neural Information Processing Systems (NIPS)}.

\bibitem[Qi et~al., 2017c]{QiLJFU17}
Qi, X., Liao, R., Jia, J., Fidler, S., and Urtasun, R. (2017c).
\newblock {3D} graph neural networks for {RGBD} semantic segmentation.
\newblock In {\em {IEEE} International Conference on Computer Vision (ICCV)},
  pages 5209--5218.

\bibitem[Radford et~al., 2015]{radford2015unsupervised}
Radford, A., Metz, L., and Chintala, S. (2015).
\newblock Unsupervised representation learning with deep convolutional
  generative adversarial networks.
\newblock {\em arXiv preprint arXiv:1511.06434}.

\bibitem[Ramakrishnan et~al., 2014]{qm9}
Ramakrishnan, R., Dral, P.~O., Rupp, M., and von Lilienfeld, O.~A. (2014).
\newblock Quantum chemistry structures and properties of 134 kilo molecules.
\newblock {\em Scientific Data}, 1.

\bibitem[Reed et~al., 2016]{ReedAYLSL16}
Reed, S.~E., Akata, Z., Yan, X., Logeswaran, L., Schiele, B., and Lee, H.
  (2016).
\newblock Generative adversarial text to image synthesis.
\newblock In {\em International Conference on Machine Learning (ICML)}, pages
  1060--1069.

\bibitem[Riegler et~al., 2017a]{Riegler2017OctNetFusion}
Riegler, G., Ulusoy, A.~O., Bischof, H., and Geiger, A. (2017a).
\newblock {OctNetFusion}: Learning depth fusion from data.
\newblock In {\em Proceedings of the International Conference on 3D Vision}.

\bibitem[Riegler et~al., 2017b]{Riegler2017OctNet}
Riegler, G., Ulusoy, A.~O., and Geiger, A. (2017b).
\newblock {OctNet}: Learning deep {3D} representations at high resolutions.
\newblock In {\em IEEE Conference on Computer Vision and Pattern Recognition
  (CVPR)}.

\bibitem[Rusu and Cousins, 2011]{pclrusu}
Rusu, R.~B. and Cousins, S. (2011).
\newblock {3D} is here: Point cloud library (pcl).
\newblock In {\em IEEE International Conference on Robotics and Automation
  (ICRA)}, pages 1--4. IEEE.

\bibitem[Samanta et~al., 2018]{samanta18}
Samanta, B., De, A., Ganguly, N., and Gomez{-}Rodriguez, M. (2018).
\newblock Designing random graph models using variational autoencoders with
  applications to chemical design.
\newblock {\em CoRR}, abs/1802.05283.

\bibitem[Sardellitti et~al., 2017]{SardellittiBL17}
Sardellitti, S., Barbarossa, S., and Lorenzo, P.~D. (2017).
\newblock On the graph fourier transform for directed graphs.
\newblock {\em J. Sel. Topics Signal Processing}, 11(6):796--811.

\bibitem[Saxe et~al., 2014]{orthoinit}
Saxe, A.~M., McClelland, J.~L., and Ganguli, S. (2014).
\newblock Exact solutions to the nonlinear dynamics of learning in deep linear
  neural networks.
\newblock In {\em International Conference on Learning Representations (ICLR)}.

\bibitem[Scarselli et~al., 2009]{scarselli09}
Scarselli, F., Gori, M., Tsoi, A.~C., Hagenbuchner, M., and Monfardini, G.
  (2009).
\newblock The graph neural network model.
\newblock {\em {IEEE} Trans. Neural Networks}, 20(1):61--80.

\bibitem[Schlegl et~al., 2017]{SchleglSWSL17}
Schlegl, T., Seeb{\"{o}}ck, P., Waldstein, S.~M., Schmidt{-}Erfurth, U., and
  Langs, G. (2017).
\newblock Unsupervised anomaly detection with generative adversarial networks
  to guide marker discovery.
\newblock In {\em Information Processing in Medical Imaging ({IPMI})}, pages
  146--157.

\bibitem[Schlichtkrull et~al., 2017]{schlichtkrull2017modeling}
Schlichtkrull, M., Kipf, T.~N., Bloem, P., Berg, R. v.~d., Titov, I., and
  Welling, M. (2017).
\newblock Modeling relational data with graph convolutional networks.
\newblock {\em arXiv preprint arXiv:1703.06103}.

\bibitem[Sch{\"o}lkopf and Smola, 2002]{scholkopf2002kernels}
Sch{\"o}lkopf, B. and Smola, A.~J. (2002).
\newblock {\em Learning with kernels: support vector machines, regularization,
  optimization, and beyond}.
\newblock MIT press.

\bibitem[Sch{\"u}tt et~al., 2017]{Schtt2017Quantum}
Sch{\"u}tt, K.~T., Arbabzadah, F., Chmiela, S., M{\"u}ller, K.~R., and
  Tkatchenko, A. (2017).
\newblock Quantum-chemical insights from deep tensor neural networks.
\newblock In {\em Nature communications}.

\bibitem[Sch{\"{u}}tt et~al., 2017]{SchuttKFCTM17}
Sch{\"{u}}tt, K.~T., Kindermans, P., Felix, H. E.~S., Chmiela, S., Tkatchenko,
  A., and M{\"{u}}ller, K. (2017).
\newblock Schnet: {A} continuous-filter convolutional neural network for
  modeling quantum interactions.
\newblock In {\em Advances in Neural Information Processing Systems (NIPS)},
  pages 992--1002.

\bibitem[Schwing and Urtasun, 2015]{SchwingU15}
Schwing, A.~G. and Urtasun, R. (2015).
\newblock Fully connected deep structured networks.
\newblock {\em CoRR}, abs/1503.02351.

\bibitem[Segler et~al., 2018]{segler17}
Segler, M. H.~S., Kogej, T., Tyrchan, C., and Waller, M.~P. (2018).
\newblock Generating focused molecule libraries for drug discovery with
  recurrent neural networks.
\newblock {\em ACS Central Science}, 4(1):120--131.

\bibitem[Shapovalov et~al., 2013]{shapovalov2013spatial}
Shapovalov, R., Vetrov, D., and Kohli, P. (2013).
\newblock Spatial inference machines.
\newblock In {\em IEEE Conference on Computer Vision and Pattern Recognition
  (CVPR)}.

\bibitem[Shen et~al., 2017]{shen2017deep}
Shen, D., Wu, G., and Suk, H.-I. (2017).
\newblock Deep learning in medical image analysis.
\newblock {\em Annual review of biomedical engineering}, 19:221--248.

\bibitem[Shervashidze et~al., 2011]{shervashidze}
Shervashidze, N., Schweitzer, P., van Leeuwen, E.~J., Mehlhorn, K., and
  Borgwardt, K.~M. (2011).
\newblock Weisfeiler-lehman graph kernels.
\newblock {\em Journal of Machine Learning Research}, 12:2539--2561.

\bibitem[Shuman et~al., 2016]{shumanFV16}
Shuman, D.~I., Faraji, M.~J., and Vandergheynst, P. (2016).
\newblock A multiscale pyramid transform for graph signals.
\newblock {\em {IEEE} Trans. Signal Processing}, 64(8):2119--2134.

\bibitem[Shuman et~al., 2013]{shuman2013emerging}
Shuman, D.~I., Narang, S.~K., Frossard, P., Ortega, A., and Vandergheynst, P.
  (2013).
\newblock The emerging field of signal processing on graphs: Extending
  high-dimensional data analysis to networks and other irregular domains.
\newblock {\em IEEE Signal Processing Magazine}, 30(3):83--98.

\bibitem[Shuman et~al., 2011]{ShumanVF11}
Shuman, D.~I., Vandergheynst, P., and Frossard, P. (2011).
\newblock Chebyshev polynomial approximation for distributed signal processing.
\newblock In {\em Distributed Computing in Sensor Systems ({DCOSS})}, pages
  1--8.

\bibitem[Simonovsky et~al., 2016]{simonovsky2016deep}
Simonovsky, M., Guti{\'{e}}rrez{-}Becker, B., Mateus, D., Navab, N., and
  Komodakis, N. (2016).
\newblock A deep metric for multimodal registration.
\newblock In {\em Medical Image Computing and Computer-Assisted Intervention
  (MICCAI)}, pages 10--18.

\bibitem[Simonovsky and Komodakis, 2016]{simonovsky2016onionnet}
Simonovsky, M. and Komodakis, N. (2016).
\newblock {OnionNet}: Sharing features in cascaded deep classifiers.
\newblock In {\em British Machine Vision Conference (BMVC)}.

\bibitem[Simonovsky and Komodakis, 2017]{simonovsky2017dynamic}
Simonovsky, M. and Komodakis, N. (2017).
\newblock Dynamic edge-conditioned filters in convolutional neural networks on
  graphs.
\newblock In {\em IEEE Conference on Computer Vision and Pattern Recognition
  (CVPR)}.

\bibitem[Simonovsky and Komodakis, 2018a]{simonovsky2018graphvae}
Simonovsky, M. and Komodakis, N. (2018a).
\newblock {GraphVAE}: {T}owards generation of small graphs using variational
  autoencoders.
\newblock In {\em International Conference on Artificial Neural Networks
  (ICANN)}.

\bibitem[Simonovsky and Komodakis, 2018b]{simonovsky2018towards}
Simonovsky, M. and Komodakis, N. (2018b).
\newblock Towards variational generation of small graphs.
\newblock In {\em International Conference on Learning Representations (ICLR),
  Workshop track}.

\bibitem[Sinkhorn and Knopp, 1967]{sinkhorn1967concerning}
Sinkhorn, R. and Knopp, P. (1967).
\newblock Concerning nonnegative matrices and doubly stochastic matrices.
\newblock {\em Pacific Journal of Mathematics}, 21(2):343--348.

\bibitem[Snijders and Nowicki, 1997]{Snijders1997}
Snijders, T.~A. and Nowicki, K. (1997).
\newblock Estimation and prediction for stochastic blockmodels for graphs with
  latent block structure.
\newblock {\em Journal of Classification}, 14(1):75--100.

\bibitem[Sohn et~al., 2015]{cvae}
Sohn, K., Lee, H., and Yan, X. (2015).
\newblock Learning structured output representation using deep conditional
  generative models.
\newblock In {\em Advances in Neural Information Processing Systems (NIPS)},
  pages 3483--3491.

\bibitem[Spielman and Srivastava, 2011]{spielman2011graph}
Spielman, D.~A. and Srivastava, N. (2011).
\newblock Graph sparsification by effective resistances.
\newblock {\em SIAM Journal on Computing}, 40(6):1913--1926.

\bibitem[Stewart et~al., 2016]{stewart2016end}
Stewart, R., Andriluka, M., and Ng, A.~Y. (2016).
\newblock End-to-end people detection in crowded scenes.
\newblock In {\em IEEE Conference on Computer Vision and Pattern Recognition
  (CVPR)}, pages 2325--2333.

\bibitem[Su et~al., 2015]{Su15}
Su, H., Maji, S., Kalogerakis, E., and Learned{-}Miller, E.~G. (2015).
\newblock Multi-view convolutional neural networks for {3D} shape recognition.
\newblock In {\em {IEEE} International Conference on Computer Vision (ICCV)}.

\bibitem[Sutskever et~al., 2011]{SutskeverMH11}
Sutskever, I., Martens, J., and Hinton, G.~E. (2011).
\newblock Generating text with recurrent neural networks.
\newblock In {\em International Conference on Machine Learning (ICML)}, pages
  1017--1024.

\bibitem[Tallec and Ollivier, 2018]{tallec2018can}
Tallec, C. and Ollivier, Y. (2018).
\newblock Can recurrent neural networks warp time?
\newblock In {\em International Conference on Learning Representations (ICLR)}.

\bibitem[Tatarchenko et~al., 2017]{ogn2017}
Tatarchenko, M., Dosovitskiy, A., and Brox, T. (2017).
\newblock Octree generating networks: Efficient convolutional architectures for
  high-resolution 3d outputs.
\newblock In {\em {IEEE} International Conference on Computer Vision (ICCV)}.

\bibitem[Tatarchenko et~al., 2018]{tanconv18}
Tatarchenko, M., Park, J., Koltun, V., and Zhou., Q.-Y. (2018).
\newblock Tangent convolutions for dense prediction in {3D}.
\newblock {\em IEEE Conference on Computer Vision and Pattern Recognition
  (CVPR)}.

\bibitem[Tchapmi et~al., 2017]{tchapmi2017segcloud}
Tchapmi, L.~P., Choy, C.~B., Armeni, I., Gwak, J., and Savarese, S. (2017).
\newblock {SEGCloud}: Semantic segmentation of {3D} point clouds.
\newblock {\em arXiv preprint arXiv:1710.07563}.

\bibitem[Theis et~al., 2015]{TheisOB15}
Theis, L., van~den Oord, A., and Bethge, M. (2015).
\newblock A note on the evaluation of generative models.
\newblock {\em CoRR}, abs/1511.01844.

\bibitem[Thomas et~al., 2018]{tensorFieldNets}
Thomas, N., Smidt, T., Kearnes, S.~M., Yang, L., Li, L., Kohlhoff, K., and
  Riley, P. (2018).
\newblock Tensor field networks: Rotation- and translation-equivariant neural
  networks for 3d point clouds.
\newblock {\em CoRR}, abs/1802.08219.

\bibitem[Tolstikhin et~al., 2018]{wasservaes}
Tolstikhin, I.~O., Bousquet, O., Gelly, S., and Sch{\"{o}}lkopf, B. (2018).
\newblock Wasserstein auto-encoders.
\newblock In {\em International Conference on Learning Representations (ICLR)}.

\bibitem[Tu et~al., 2018]{Tu-CVPR-2018}
Tu, W.-C., Liu, M.-Y., Jampani, V., Sun, D., Chien, S.-Y., Yang, M.-H., and
  Kautz, J. (2018).
\newblock Learning superpixels with segmentation-aware affinity loss.
\newblock In {\em IEEE Conference on Computer Vision and Pattern Recognition
  (CVPR)}.

\bibitem[Ulyanov et~al., 2017]{UlyanovVL17}
Ulyanov, D., Vedaldi, A., and Lempitsky, V. (2017).
\newblock Deep image prior.
\newblock {\em arXiv:1711.10925}.

\bibitem[van~den Oord et~al., 2016]{oord2016pixelrnn}
van~den Oord, A., Kalchbrenner, N., and Kavukcuoglu, K. (2016).
\newblock Pixel recurrent neural networks.
\newblock In {\em International Conference on Machine Learning (ICML)}, pages
  1747--1756.

\bibitem[Veli{\v{c}}kovi{\'{c}} et~al., 2018]{GraphAtt}
Veli{\v{c}}kovi{\'{c}}, P., Cucurull, G., Casanova, A., Romero, A., Li{\`{o}},
  P., and Bengio, Y. (2018).
\newblock {Graph Attention Networks}.
\newblock {\em International Conference on Learning Representations (ICLR)}.

\bibitem[Verelst et~al., 2018]{Verelst18}
Verelst, T., Berman, M., and Blaschko, M.~B. (2018).
\newblock Generating superpixels with deep representations.
\newblock In {\em IEEE Conference on Computer Vision and Pattern Recognition
  (CVPR), Deep Vision Workshop}.

\bibitem[Verma et~al., 2018]{verma2018feastnet}
Verma, N., Boyer, E., and Verbeek, J. (2018).
\newblock {FeaStNet}: Feature-steered graph convolutions for {3D} shape
  analysis.
\newblock In {\em IEEE Conference on Computer Vision and Pattern Recognition
  (CVPR)}.

\bibitem[Vinyals et~al., 2015]{vinyals2015order}
Vinyals, O., Bengio, S., and Kudlur, M. (2015).
\newblock Order matters: Sequence to sequence for sets.
\newblock {\em arXiv preprint arXiv:1511.06391}.

\bibitem[Wale et~al., 2008]{nci1db}
Wale, N., Watson, I.~A., and Karypis, G. (2008).
\newblock Comparison of descriptor spaces for chemical compound retrieval and
  classification.
\newblock {\em Knowledge and Information Systems}, 14(3):347--375.

\bibitem[Wang et~al., 2017]{Wang-2017-ocnn}
Wang, P.-S., Liu, Y., Guo, Y.-X., Sun, C.-Y., and Tong, X. (2017).
\newblock {O-CNN}: Octree-based convolutional neural networks for 3d shape
  analysis.
\newblock {\em ACM Transactions on Graphics (SIGGRAPH)}, 36(4).

\bibitem[Wang et~al., 2018a]{wang2018non}
Wang, X., Girshick, R., Gupta, A., and He, K. (2018a).
\newblock Non-local neural networks.
\newblock In {\em IEEE Conference on Computer Vision and Pattern Recognition
  (CVPR)}.

\bibitem[Wang et~al., 2018b]{dgcnn}
Wang, Y., Sun, Y., Liu, Z., Sarma, S.~E., Bronstein, M.~M., and Solomon, J.~M.
  (2018b).
\newblock Dynamic graph {CNN} for learning on point clouds.
\newblock {\em arXiv preprint arXiv:1801.07829}.

\bibitem[Weininger, 1988]{Weininger88}
Weininger, D. (1988).
\newblock {SMILES}, a chemical language and information system. 1.
  {I}ntroduction to methodology and encoding rules.
\newblock {\em Journal of Chemical Information and Computer Sciences},
  28(1):31--36.

\bibitem[Weinmann et~al., 2017]{weinmann2017hybrid}
Weinmann, M., Hinz, S., and Weinmann, M. (2017).
\newblock A hybrid semantic point cloud classification-segmentation framework
  based on geometric features and semantic rules.
\newblock {\em PFG--Journal of Photogrammetry, Remote Sensing and
  Geoinformation Science}, 85(3):183--194.

\bibitem[Weinmann et~al., 2015]{weinmann_contextual_2015}
Weinmann, M., Schmidt, A., Mallet, C., Hinz, S., Rottensteiner, F., and Jutzi,
  B. (2015).
\newblock Contextual classification of point cloud data by exploiting
  individual {3D} neighborhoods.
\newblock {\em {ISPRS} Annals of the Photogrammetry, Remote Sensing and Spatial
  Information Sciences}, {II}-3/W4:271--278.

\bibitem[Williams and Zipser, 1989]{WilliamsZ89}
Williams, R.~J. and Zipser, D. (1989).
\newblock A learning algorithm for continually running fully recurrent neural
  networks.
\newblock {\em Neural Computation}, 1(2):270--280.

\bibitem[Wolf et~al., 2015]{wolf2015fast}
Wolf, D., Prankl, J., and Vincze, M. (2015).
\newblock Fast semantic segmentation of {3D} point clouds using a dense {CRF}
  with learned parameters.
\newblock In {\em IEEE International Conference on Robotics and Automation
  (ICRA)}.

\bibitem[Wolf et~al., 2017]{WolfSKH17}
Wolf, S., Schott, L., K{\"{o}}the, U., and Hamprecht, F.~A. (2017).
\newblock Learned watershed: End-to-end learning of seeded segmentation.
\newblock In {\em {IEEE} International Conference on Computer Vision (ICCV)},
  pages 2030--2038.

\bibitem[Wu et~al., 2018]{wu2018moleculenet}
Wu, Z., Ramsundar, B., Feinberg, E.~N., Gomes, J., Geniesse, C., Pappu, A.~S.,
  Leswing, K., and Pande, V. (2018).
\newblock Moleculenet: a benchmark for molecular machine learning.
\newblock {\em Chemical science}, 9(2):513--530.

\bibitem[Wu et~al., 2015a]{modelnet}
Wu, Z., Song, S., Khosla, A., Tang, X., and Xiao, J. (2015a).
\newblock {3D ShapeNets} for {2.5D} object recognition and next-best-view
  prediction.
\newblock In {\em IEEE Conference on Computer Vision and Pattern Recognition
  (CVPR)}.

\bibitem[Wu et~al., 2015b]{WuSKYZTX15}
Wu, Z., Song, S., Khosla, A., Yu, F., Zhang, L., Tang, X., and Xiao, J.
  (2015b).
\newblock {3D ShapeNets}: {A} deep representation for volumetric shapes.
\newblock In {\em IEEE Conference on Computer Vision and Pattern Recognition
  (CVPR)}, pages 1912--1920.

\bibitem[Xu et~al., 2017]{Xu17GraphGen}
Xu, D., Zhu, Y., Choy, C.~B., and Fei{-}Fei, L. (2017).
\newblock Scene graph generation by iterative message passing.
\newblock In {\em IEEE Conference on Computer Vision and Pattern Recognition
  (CVPR)}, pages 3097--3106.

\bibitem[Yanardag and Vishwanathan, 2015]{deepkern}
Yanardag, P. and Vishwanathan, S. V.~N. (2015).
\newblock Deep graph kernels.
\newblock In {\em {ACM} {SIGKDD} International Conference on Knowledge
  Discovery and Data Mining}.

\bibitem[Yi et~al., 2017]{SyncSpecCNN}
Yi, L., Su, H., Guo, X., and Guibas, L.~J. (2017).
\newblock {SyncSpecCNN}: Synchronized spectral {CNN} for {3D} shape
  segmentation.
\newblock In {\em IEEE Conference on Computer Vision and Pattern Recognition
  (CVPR)}, pages 6584--6592.

\bibitem[You et~al., 2018a]{YouPolicy18}
You, J., Liu, B., Ying, R., Pande, V.~S., and Leskovec, J. (2018a).
\newblock Graph convolutional policy network for goal-directed molecular graph
  generation.
\newblock {\em CoRR}, abs/1806.02473.

\bibitem[You et~al., 2018b]{graphrnn}
You, J., Ying, R., Ren, X., Hamilton, W.~L., and Leskovec, J. (2018b).
\newblock {GraphRNN}: {A} deep generative model for graphs.
\newblock In {\em International Conference on Machine Learning (ICML)}.

\bibitem[Yu et~al., 2017]{SeqGAN}
Yu, L., Zhang, W., Wang, J., and Yu, Y. (2017).
\newblock {SeqGAN}: Sequence generative adversarial nets with policy gradient.
\newblock In {\em {AAAI} Conference on Artificial Intelligence}.

\bibitem[Yuan et~al., 2017]{YuanLWYG17}
Yuan, Y., Liang, X., Wang, X., Yeung, D., and Gupta, A. (2017).
\newblock Temporal dynamic graph {LSTM} for action-driven video object
  detection.
\newblock In {\em {IEEE} International Conference on Computer Vision (ICCV)},
  pages 1819--1828.

\bibitem[Zhang et~al., 2018]{ZhangCNC18}
Zhang, M., Cui, Z., Neumann, M., and Chen, Y. (2018).
\newblock An end-to-end deep learning architecture for graph classification.
\newblock In {\em {AAAI} Conference on Artificial Intelligence}.

\bibitem[Zheng et~al., 2015]{Zheng15crf}
Zheng, S., Jayasumana, S., Romera{-}Paredes, B., Vineet, V., Su, Z., Du, D.,
  Huang, C., and Torr, P. H.~S. (2015).
\newblock Conditional random fields as recurrent neural networks.
\newblock In {\em {IEEE} International Conference on Computer Vision (ICCV)}.

\end{thebibliography}

\end{spacing}

\begin{appendices} %

\end{appendices}

\printthesisindex %

\end{document}